\documentclass[10pt,twocolumn,letterpaper]{article}

\usepackage{titling}
\usepackage{iccv}
\usepackage{times}
\usepackage{epsfig}
\usepackage{graphicx}
\usepackage{amsmath}
\usepackage{amssymb}
\usepackage{booktabs}
\usepackage{gensymb}
\usepackage{xspace}

\usepackage{physics}
\usepackage[obeyDraft]{todonotes}
\usepackage{nicefrac}

\usepackage{multirow}
\usepackage[export]{adjustbox}

\usepackage{siunitx}
\usepackage{enumitem}
\ifdefined\unit\else
  \ifdefined\NewCommandCopy
    \NewCommandCopy\unit\si
  \else
    \NewDocumentCommand\unit{O{}m}{\si[#1]{#2}}
  \fi
\fi

\def\valeur{\textcolor{black}}

\renewcommand{\etal}{et al.\xspace}

\usepackage[pagebackref=true,breaklinks=true,letterpaper=true,colorlinks,bookmarks=false]{hyperref}

\usepackage[capitalize]{cleveref}
\crefname{section}{sec.}{secs.}
\Crefname{section}{Sec.}{Secs.}
\crefname{table}{tab.}{tabs.}
\Crefname{table}{Tab.}{Tabs.}
\crefname{figure}{fig.}{figs.}
\Crefname{figure}{Fig.}{Figs.}
\crefname{equation}{eq.}{eqs.}
\Crefname{equation}{Eq.}{Eqs.}

\iccvfinalcopy 

\def\iccvPaperID{6978} 

\ificcvfinal\fi

\begin{document}

\title{Beyond the Pixel: a Photometrically Calibrated HDR Dataset \\ for Luminance and Color Prediction}

\author{Christophe Bolduc, Justine Giroux, Marc Hébert, Claude Demers, and Jean-François Lalonde\\
Université Laval\\
}

\maketitle
\ificcvfinal\thispagestyle{empty}\fi

\begin{abstract}
Light plays an important role in human well-being. However, most computer vision tasks treat pixels without considering their relationship to physical luminance. To address this shortcoming, we introduce the Laval Photometric Indoor HDR Dataset, the first large-scale photometrically calibrated dataset of high dynamic range \ang{360} panoramas. Our key contribution is the calibration of an existing, uncalibrated HDR Dataset. We do so by accurately capturing RAW bracketed exposures simultaneously with a professional photometric measurement device (chroma meter) for multiple scenes across a variety of lighting conditions. Using the resulting measurements, we establish the calibration coefficients to be applied to the HDR images. The resulting dataset is a rich representation of indoor scenes which displays a wide range of illuminance and color, and varied types of light sources. We exploit the dataset to introduce three novel tasks, where: per-pixel luminance, per-pixel color and planar illuminance can be predicted from a single input image. Finally, we also capture another smaller photometric dataset with a commercial \ang{360} camera, to experiment on generalization across cameras. We are optimistic that the release of our datasets and associated code will spark interest in physically accurate light estimation within the community. Dataset and code are available at \url{https://lvsn.github.io/beyondthepixel/}.
\end{abstract}
\section{Introduction}
\label{sec:intro}


Natural light has shaped the way our human visual system evolved~\cite{cronin2014visual}, plays a key role in driving our circadian rhythm~\cite{duffy2009effect}, and affects our mental health~\cite{magnusson2003seasonal} and social organization~\cite{brox2010brilliant}. It has also been shown~\cite{murray2019visual} that human vision relies on stable properties of light, measured in terms of luminance (in \unit{\candela\per\meter\squared}), in order to perceive object features such as shape and color. 

Natural light is also at the heart of photography and computer vision. However, most if not all computer vision approaches consider pixel values as a 3-channel input to be processed without considering the relationship between pixel intensity and luminance. This is understandable since modern digital cameras pursue a goal different from measuring physically accurate perceived brightness: they strive to create visually pleasing photographs. In doing so, their internal image signal processors (ISP) perform a series of operations on the measured light (denoising, contrast enhancement, tonemapping, etc.~\cite{karaimer2016isp}) in order to produce pixel values which, while visually appealing, no longer correspond to the physical properties of the environment.

Modeling camera ISPs and inverting their image formation process has been the subject of many works (e.g. \cite{xiong2012pixels,heide2014flexisp,nam2022learning}). Here, most approaches aim at recovering an image where pixel values are linearly proportional to the scene radiance (or luminance). Closely related are approaches for capturing high dynamic range (HDR) images~\cite{debevec2008recovering}, or predicting HDR from low dynamic range (LDR) photographs~\cite{endo2017reversetone,lee2018deepchain,marnerides2018expandnet,li2019hdrnet,liu2020single}. While linear pixel values can be extremely useful for physics-based vision applications (e.g. \cite{Haefner2021}), the scale factor to absolute luminance is still unknown. Can we go beyond (linear) pixel values and recover per-pixel luminance from a single image?



In this paper, we propose the Laval Photometric Indoor HDR Dataset: what we believe to be the first large-scale dataset to help the community answer this question. The novel dataset of physically accurate luminance and colors acquired in a wide variety of indoor scenes. Our key idea is to leverage the camera and RAW captures of an existing dataset of HDR indoor \ang{360} panoramas that was previously captured~\cite{gardner-tog-17}. We contribute by carefully calibrating the camera with a chroma meter to determine the per-channel correction factors to be applied to each panorama. Our analysis shows that the Laval Photometric Indoor HDR Dataset contains a wide range of illuminance (e.g. $[\valeur{\SI{0}{\lux}}, \valeur{\SI{7000}{\lux}}]$) and color, expressed in correlated color temperature (CCT) (e.g. $[\valeur{\SI{2000}{\K}}, \valeur{\SI{8000}{\K}}]$), capturing the diversity of indoor environments. We also explore the luminance and color distributions of individual light sources in the dataset, which span several orders of magnitude of average luminance. 

We present three novel learning tasks that are enabled by our calibrated dataset. Given a single image as input, we explore how per-pixel luminance, per-pixel color, and planar illuminance can be estimated. More importantly, we also consider what information must be available for accurate light prediction. Indeed, democratizing the process of capturing physical luminance begs several important questions: can luminance be accurately estimated using conventional, uncalibrated cameras? If so, is HDR imagery needed or is a single, well-exposed shot sufficient? Is a generic approach appropriate or do methods need to be finetuned to specific cameras? We provide initial answers to these challenging questions by presenting learning experiments on our novel dataset, as well as on another, smaller photometric dataset captured with an off-the-shelf \ang{360} camera. By publicly releasing the calibrated datasets and associated code, we hope to spur interest in the community and help it consider the physical light measurements that lie beyond the pixels.




\section{Related work}
\label{sec:related_work}



\paragraph{Radiometric camera calibration} A large body of work has tackled the recovery of the camera response function, the (usually proprietary) non-linear tonemapping curve applied by cameras~\cite{grossberg2004modeling}. This can be done for example from multiple exposures~\cite{debevec2008recovering}, an image sequence~\cite{kim2008radiometric}, or even from a single image~\cite{lin2004radiometric,lin2005determining,lin2011revisiting}. This can also be done jointly with other tasks, e.g. vignetting correction~\cite{kim2008robust}. 

\paragraph{HDR reconstruction}
Fusing multiple low dynamic range (LDR) images at different exposures into one high dynamic range (HDR) image has been studied extensively~\cite{robertson1999dynamic,debevec2008recovering}. The overlapping exposures can be leveraged to simultaneously linearize the input images and reconstruct the radiance. Images must be aligned if the camera moves~\cite{ward2003fast}, and more complex treatment must be given to avoid ghosting if the scene is in movement~\cite{granados2013automatic,kalantari2017deep,niu2021hdr}. HDR images can also be reconstructed from image bursts~\cite{lecouat2022}, specialized optics~\cite{sun2020learning,metzler2020deep}, or even during NeRF~\cite{mildenhall2021nerf} training: from RAW inputs~\cite{mildenhall2022nerfinthedark}, non-overlapping exposures~\cite{huang2022hdrnerf}, or by using a dedicated network~\cite{gera2022casual}. While we leverage HDR reconstruction techniques to build the Laval Photometric Indoor HDR Dataset, we wish to recover absolute luminance values: we must therefore use specialized tools for acquiring these measurements.

\paragraph{Inverse tonemapping}
Algorithms for recovering HDR from LDR images (known as inverse tone mapping) have been proposed. This was done by inverting tonemapping operators~\cite{reinhard2002photographic,banterle2006itmo,banterle2007framework}, expanding the dynamic range via edge stopping functions~\cite{rempel2007ldr2hdr} or by employing scene-specific iTMO~\cite{kuo2012content}. Recently, deep learning methods have been proposed, for example by predicting the exposure stack that is then fused with HDR methods~\cite{endo2017reversetone,lee2018deepchain} or by directly predicting the HDR image~\cite{zhang2017learning,li2019hdrnet,marnerides2018expandnet,santos2020single,yu2021luminance}. Of note, ~\cite{liu2020single} models each stage of the camera pipeline using individual networks. In this work, we employ a Unet (similar to \cite{liu2020single}) to explore novel tasks enabled by our photometric dataset.

\paragraph{Photometric calibration} 
HDR images store relative luminance values. Multiple techniques can be used to identify the (linear) photometric calibration coefficients of an imaging system to retrieve absolute luminance: measure the illuminance of a scene using a chroma meter and compare it to the integration of a calibrated \ang{180} fisheye lens~\cite{jung2019measuring}, use a luminance meter to measure the luminance of an incoming direction and compare directly with the corresponding pixel of the camera~\cite{pierson2021}, or employ a calibrated display~\cite{macpherson2022360}. Here, we follow ~\cite{jung2019measuring} to calibrate our dataset.




\paragraph{Color prediction and post-capture white balance}
To our knowledge, no current method allows for the prediction of photometric color from a single LDR image. Closely related, automatic white balance (or illuminant estimation, or color constancy) has been thoroughly explored in computer vision~\cite{barron2015convolutional,barron2017fast,bianco2019quasi,finlayson2015color,cheng2014illuminant,gehler2008bayesian,shi2016deep}. For example, correcting the white balance based on presets~\cite{mahmoud2020deep} allows the network to understand the color temperature of a scene. While most of these works assume a single illuminant, correcting for multiple illuminants has also been explored~\cite{hsu2008light,gijsenij2011color,afifi2022auto}. In these works, no absolute color values are obtained.

\paragraph{Luminance prediction}
Prediction of HDR at physical luminance is a relatively new task. Of note, Wei~\etal~\cite{Wei2021BeyondVA} tackle the problem, but their dataset is limited to stores, only luminance without its chromaticity is available, and the exposure is given as input in the experiments, hence the network does not learn to predict illuminance from the visual features in the scene. We believe that ours is the first large-scale HDR dataset containing absolute colorimetric information of the luminance maps.

\section{The Laval Photometric Indoor HDR Dataset}
\label{sec:dataset}

\subsection{Base dataset}

We rely on the existing Laval Indoor HDR dataset~\cite{gardner-tog-17} (herein referred to as ``Laval dataset'') which comprises over \num{2300} HDR panoramas captured in a large variety of indoor scenes. The data was captured with a Canon 5D Mark III camera and a Sigma \SI{8}{\mm} fisheye lens mounted on a tripod equipped with a robotic panoramic tripod head, and programmed to shoot 7 bracketed exposures at \ang{60} increments along the azimuth. The resulting 42 photos were shot in RAW mode, and automatically stitched into a 22 f-stop HDR \ang{360} panorama using the PTGui Pro commercial software. As with other HDR datasets (e.g. \cite{debevec2008recovering,polyhaven,kalantari2017deep,liu2020single}), the Laval dataset captures up-to-scale luminance values. In this work, we explicitly calibrate for the unknown, per-channel scale factors in order to recover calibrated luminance and color values at every pixel. 

\subsection{Capturing a calibration dataset}

We can photometrically calibrate the Laval dataset by first capturing a ``calibration dataset'' to determine the per-channel scale factors for the camera. The estimated scale factors can then be applied to the panoramas to obtain accurate luminance and color values. 

\begin{figure}[t]
    \includegraphics[width=\linewidth]{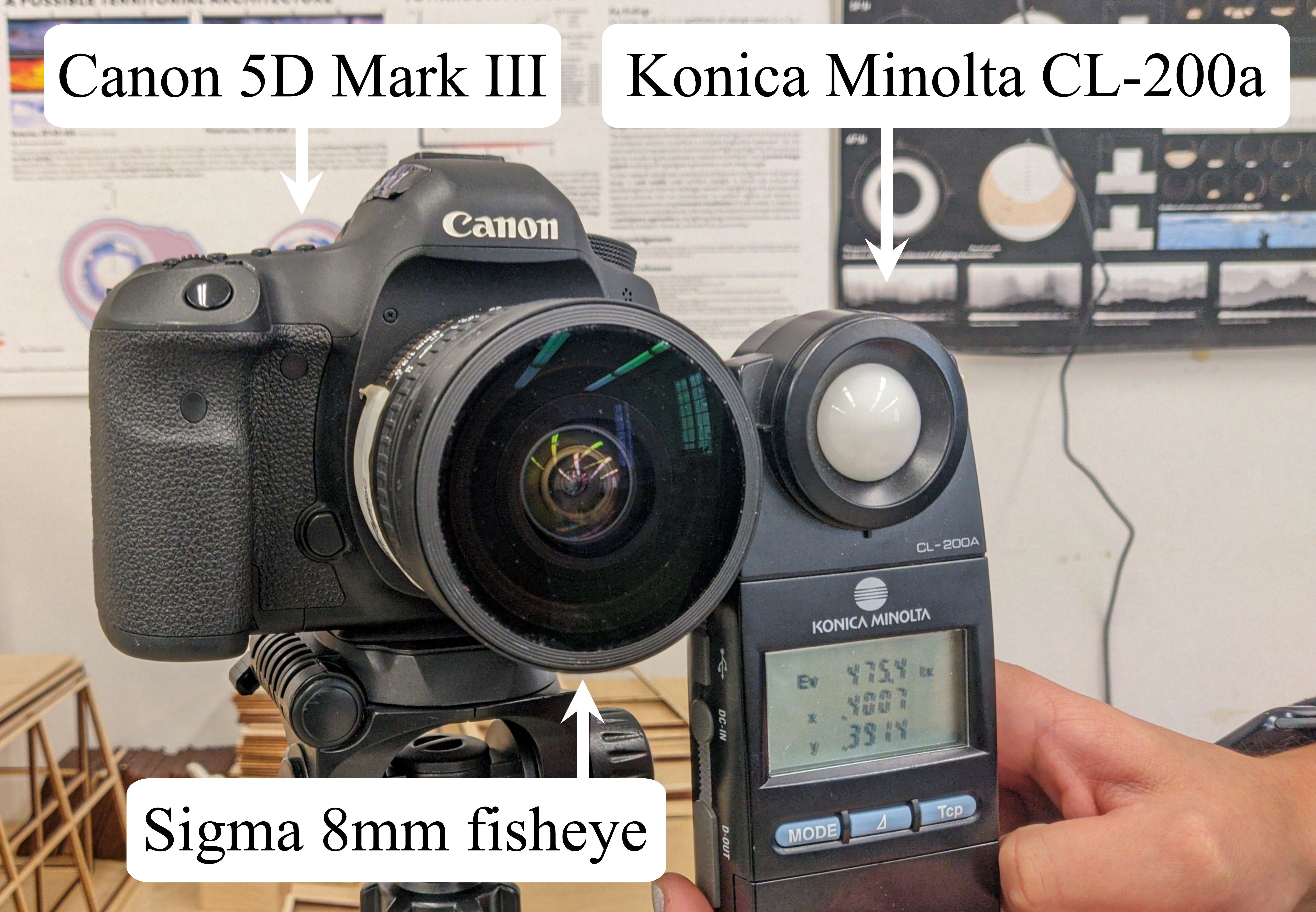}
    \caption{Setup used for the capture of the calibration dataset. The original Canon 5D Mark III camera used for the Laval dataset (left, graciously provided by the original authors) captures a bracket of images at different exposures while the chroma meter CL-200a (right) measures the illuminance and chromaticity of the scene.}
    \label{fig:calibration_setup}
\end{figure}

To capture the calibration dataset, we place the camera side-by-side with a Konica Minolta CL-200a chroma meter, as shown in \cref{fig:calibration_setup}. We then simultaneously acquire a bracket of 7 RAW exposures from the camera and a reading from the chroma meter. To ensure a match to the images from the Laval dataset, the exact same camera parameters (retrieved from the EXIF header in the RAW files) were used. We found that a total of 5 exposure configurations (differing mainly by the aperture used\footnote{f/4, f/11, f/13, f/14, f/18, with f/4 and f/14 representing the vast majority (\SI{98}{\percent}) of the Laval dataset.}) were used to capture the Laval dataset --- we therefore captured brackets for each of the 5 camera configurations at each scene to ensure proper calibration across all panoramas. The chroma meter measures both the scene illuminance (in \unit{\candela\steradian\per\m\squared}, or \unit{\lux}) and its chromaticity (in the CIE xy color space).

The PTGui Pro software was then used to merge the different exposure images into one HDR image. Vignetting correction is applied by PTGui using a polynomial model optimized during a previous panorama stitching from overlapping images. Since vignetting varies with the aperture, the model's parameters are computed for each of the 5 exposure configurations. This process was repeated in a variety of scenes (different from the scenes in the original dataset) with diverse illumination conditions. In all, \num{135} scenes were captured to establish the calibration dataset.



\subsection{Illuminance computation}

To compute the illuminance from the captured HDR image, we first geometrically calibrate the camera using \cite{li2013multiple} with the projection model from \cite{mei2007single} (implemented in OpenCV). Using the recovered lens parameters, the captured fisheye images are re-projected to an orthographic projection. The (uncalibrated) illuminance $E$ can then be computed for each color channel using
\begin{equation}
    E = \frac{\pi}{N} \sum_i^N L(i) \,,
    \label{eq:illuminance}
\end{equation}
where $L(i)$ is the value of pixel $i$, and $N$ the number of pixels in the (circular) image. \Cref{eq:illuminance} derives from the CIE illuminance equation and is explained in the supplementary.

%

\subsection{Calibration coefficients}

The uncalibrated illuminance from \cref{eq:illuminance} can be compared to the absolute illuminance measured with the chroma meter. To obtain per-channel illuminance, the xyY color value captured by the chroma meter is converted to RGB (see supplementary). A linear regression identifies the coefficients to be applied to the Laval dataset to obtain the photometrically accurate HDR panoramas. For example, the regression for the f/14 capture configuration for each RGB channel has coefficients of determination ($R^2$) of (\num{0.985}, \num{0.987}, \num{0.989}) for (R, G, B) respectively, indicating the high reliability of the fits. The uncertainty on the calibration is discussed in the supplementary. 

\subsection{Photometric correction}

All panoramas from the Laval dataset were regenerated with PTGui Pro, and each one corrected according to the linear coefficients of its capture configuration. In total, after filtering out the few panoramas that were oversaturated, we have \valeur{2362} HDR photometric panoramas of physically accurate luminance and color at \valeur{$3884 \times 7768$} pixel resolution.

\section{Visualizing the dataset}
\label{sec:viz_dataset}

In this section, physical properties of the entire scene and of individual light sources are derived for each panorama in the Laval Photometric Indoor HDR Dataset. For visualization of photometric colors, each pixel is expressed as the correlated color temperature (CCT) of its luminance (cf. supplementary). We now explore the diversity in the dataset by computing statistics of the different physical parameters of scenes and light sources present therein. 

\subsection{Entire scene}
\label{subsec:viz_dataset_whole}


\begin{figure}
    \includegraphics[width=\linewidth]{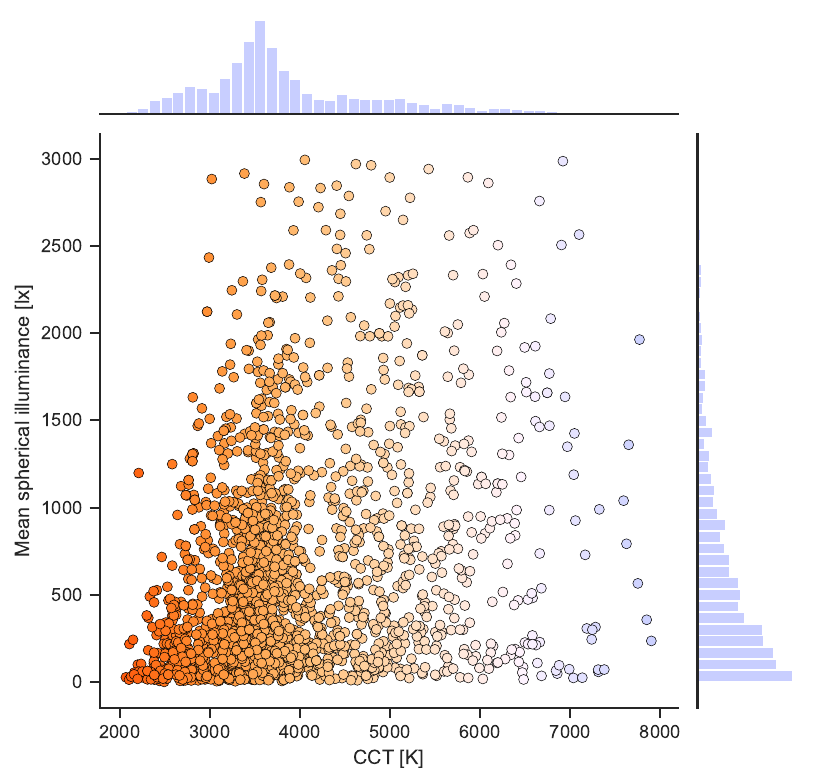}
    \caption{Correlation between the CCT (\unit{\kelvin}) and the mean spherical illuminance (\unit{\lux}) of the photometric panoramas of the dataset.  The distributions of CCT (top) and mean spherical illuminance (right) of the data are also displayed. Only the data with a CCT in $[\valeur{\SI{2000}{\K}}, \valeur{\SI{8000}{\K}}]$ and a mean spherical illuminance in $[\valeur{\SI{0}{\lux}}, \valeur{\SI{3000}{\lux}}]$ are included to better see the trends (\valeur{\num{2187}} out of \valeur{\num{2362}} panoramas). Points are color-coded according to their CCT.}
    \label{fig:jointplot_temperature_illuminance_whole_pano}
\end{figure}

\begin{figure*}
   \centering
   \footnotesize
   \setlength{\tabcolsep}{0.5pt} 
   \newlength{\tmplength}
   \setlength{\tmplength}{0.155\linewidth}
    \begin{tabular}{ccccccc}
    \valeur{\SI{1}{\%}} (\valeur{\SI{13}{\lux}}) & 
    \valeur{\SI{10}{\%}} (\valeur{\SI{65}{\lux}}) & 
    \valeur{\SI{20}{\%}} (\valeur{\SI{142}{\lux}}) & 
    \valeur{\SI{30}{\%}} (\valeur{\SI{226}{\lux}}) & 
    \valeur{\SI{40}{\%}} (\valeur{\SI{320}{\lux}}) & 
    \valeur{\SI{50}{\%}} (\valeur{\SI{462}{\lux}}) &  \multirow{5}{*}{\includegraphics[trim={{.013\linewidth} 0 0 {.005\linewidth}},clip, height=0.38\linewidth]{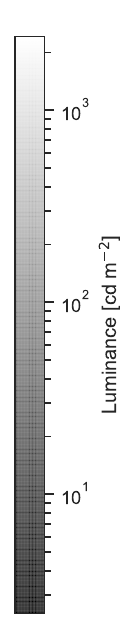}} \\
    \includegraphics[width=\tmplength]{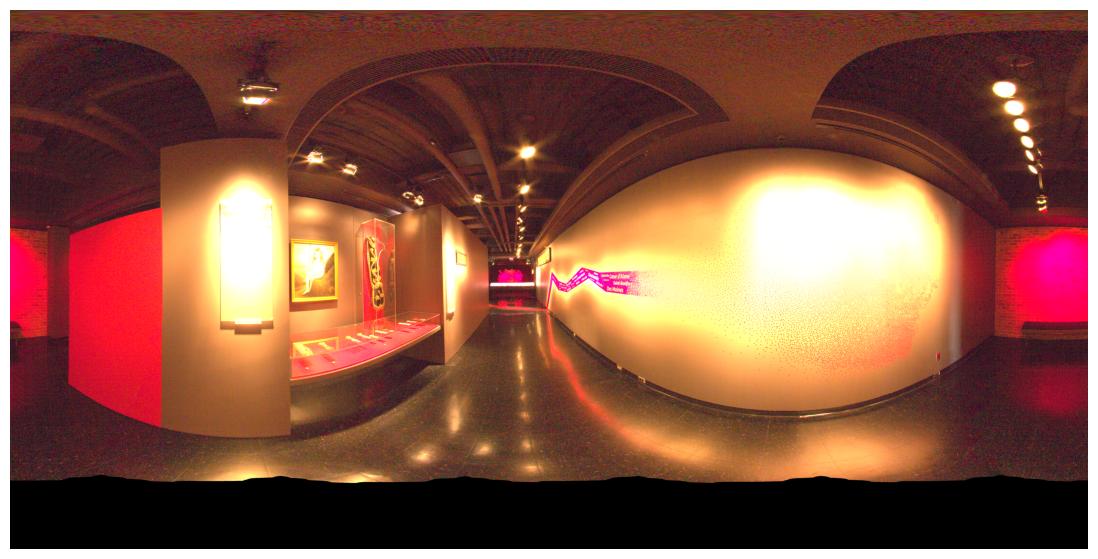}&
    \includegraphics[width=\tmplength]{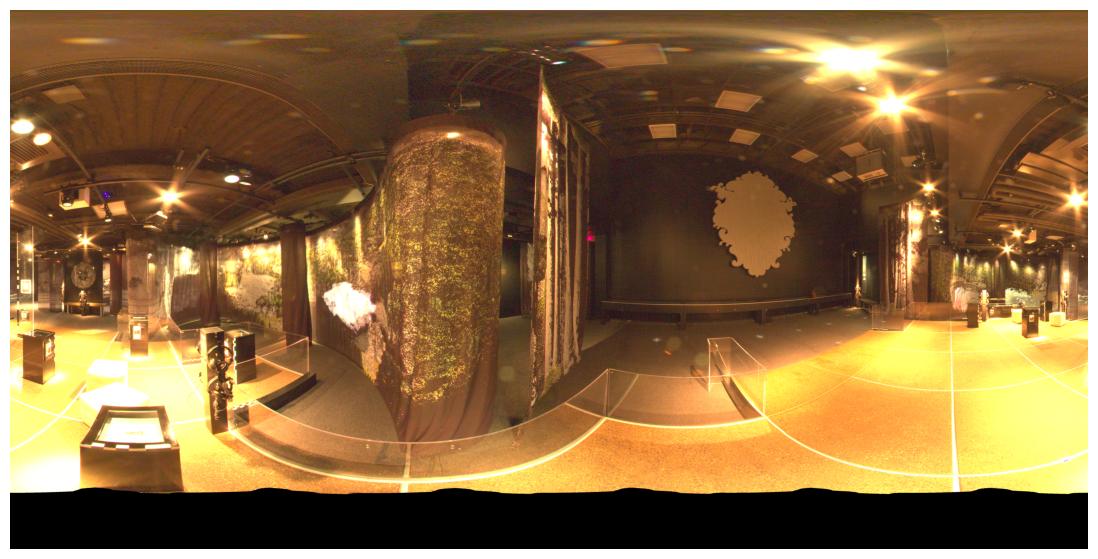}&
    \includegraphics[width=\tmplength]{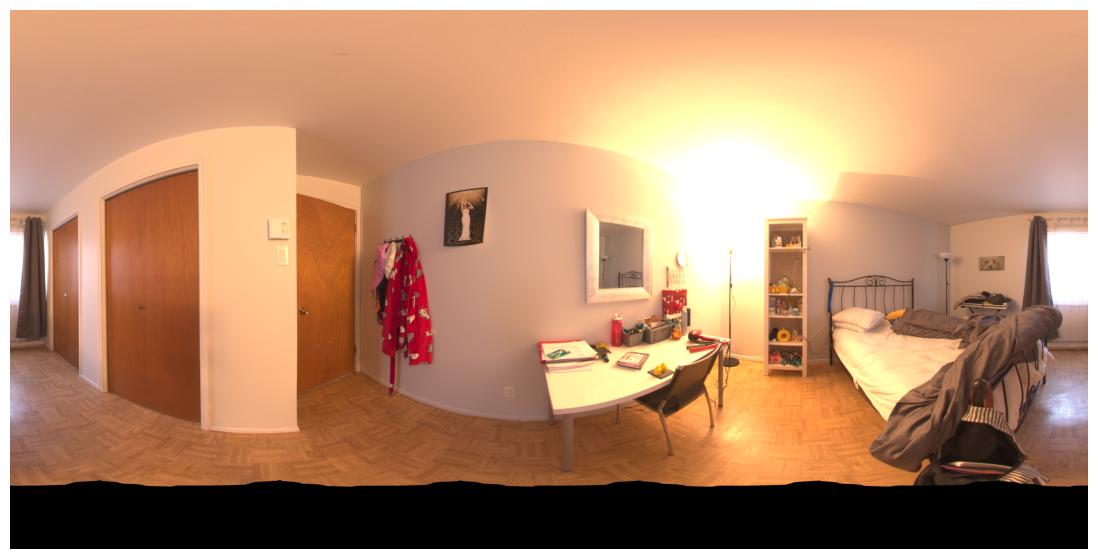}&
    \includegraphics[width=\tmplength]{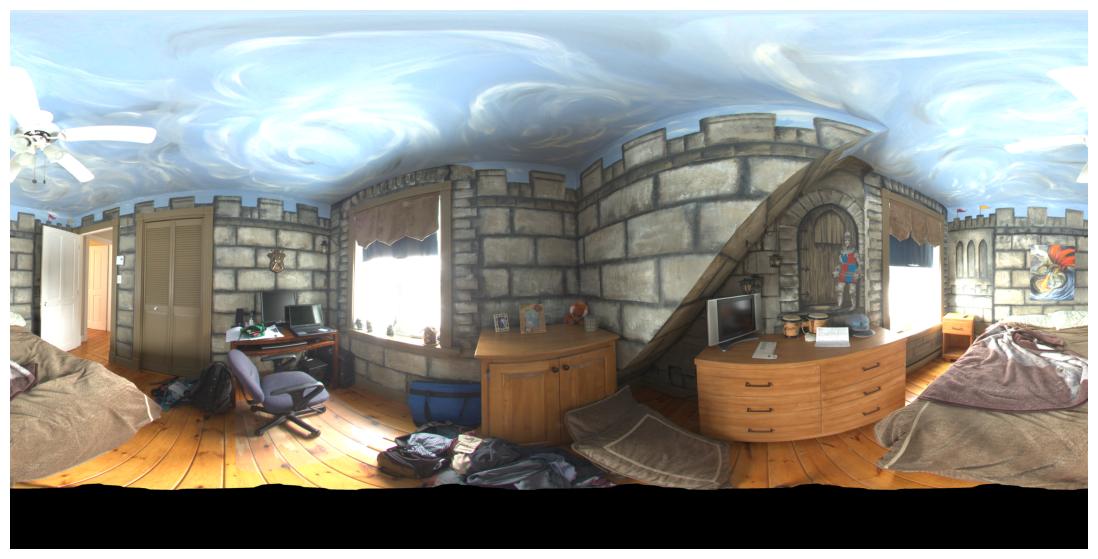}&
    \includegraphics[width=\tmplength]{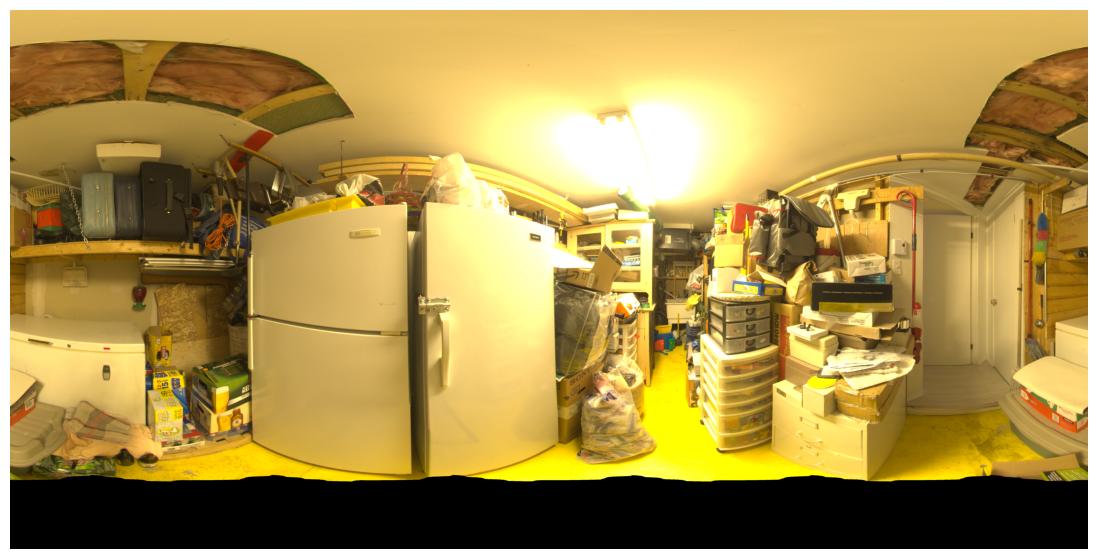}&
    \includegraphics[width=\tmplength]{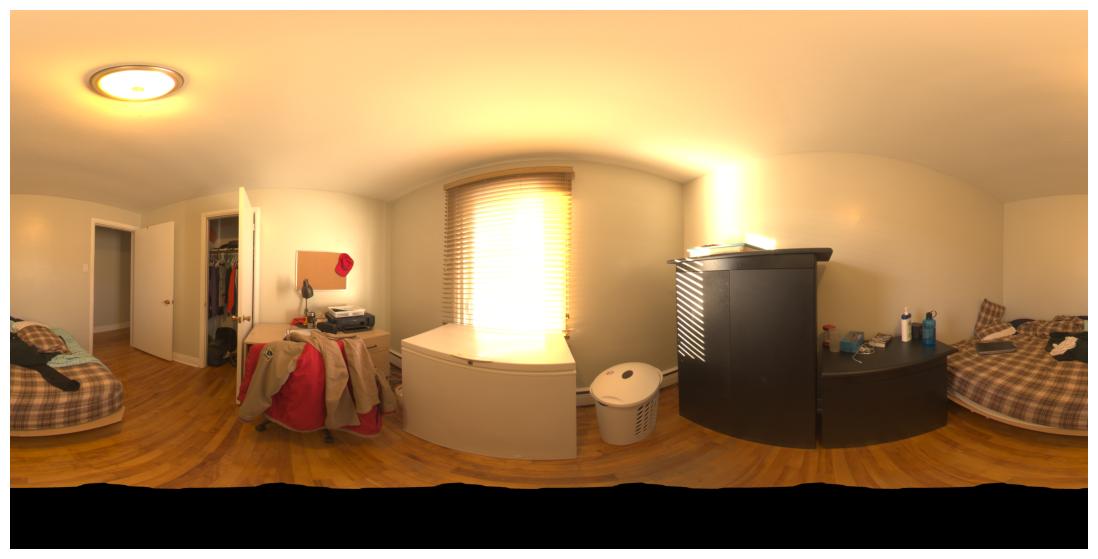}  \\
    \includegraphics[width=\tmplength]{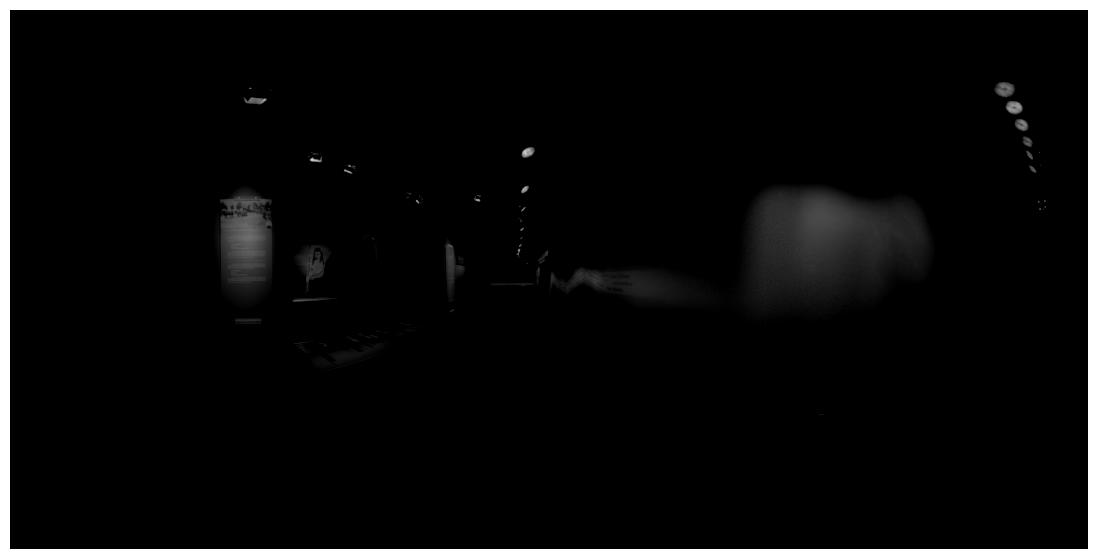}&
    \includegraphics[width=\tmplength]{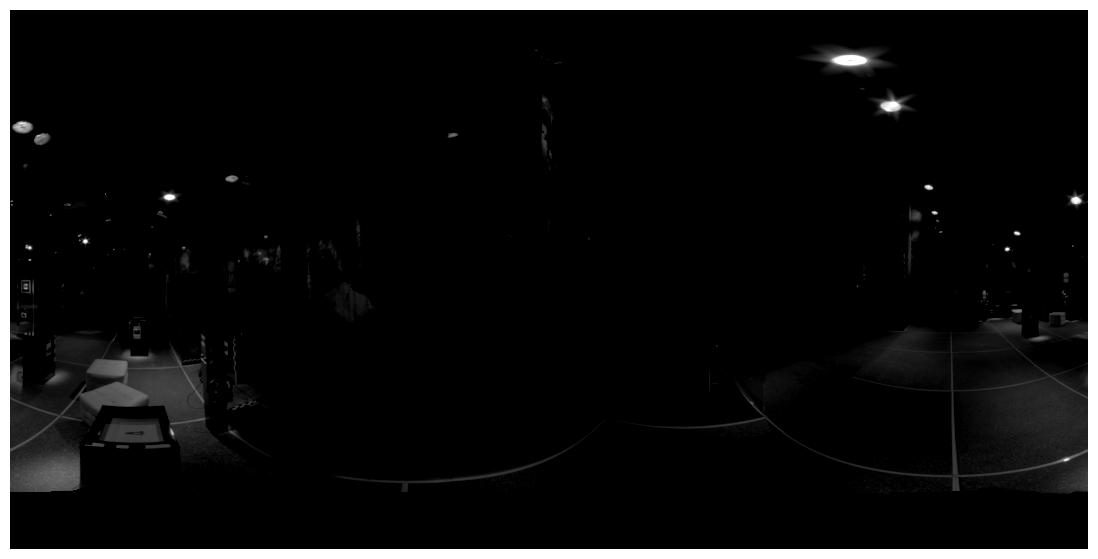}&
    \includegraphics[width=\tmplength]{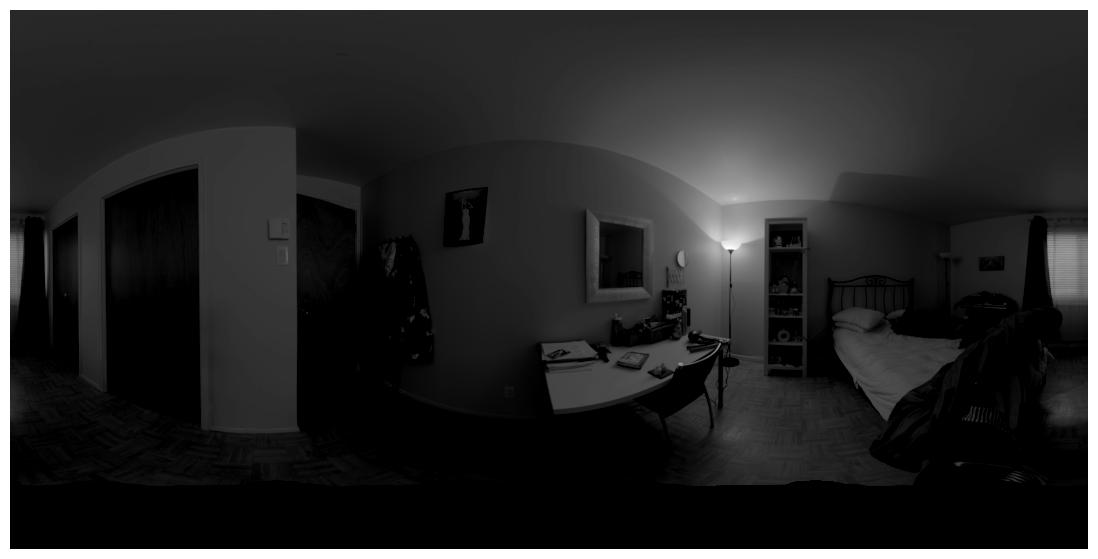}&
    \includegraphics[width=\tmplength]{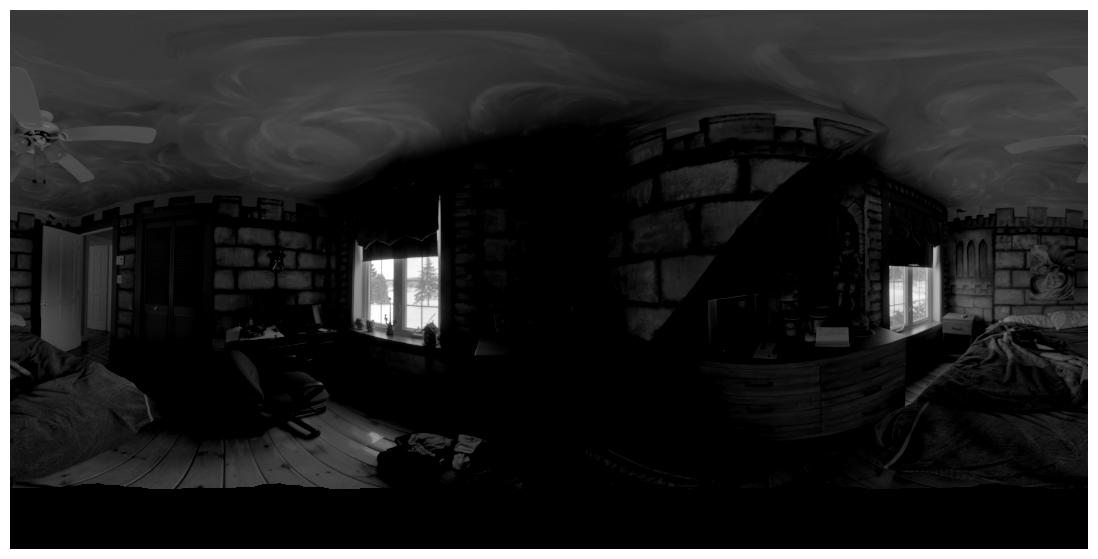}&
    \includegraphics[width=\tmplength]{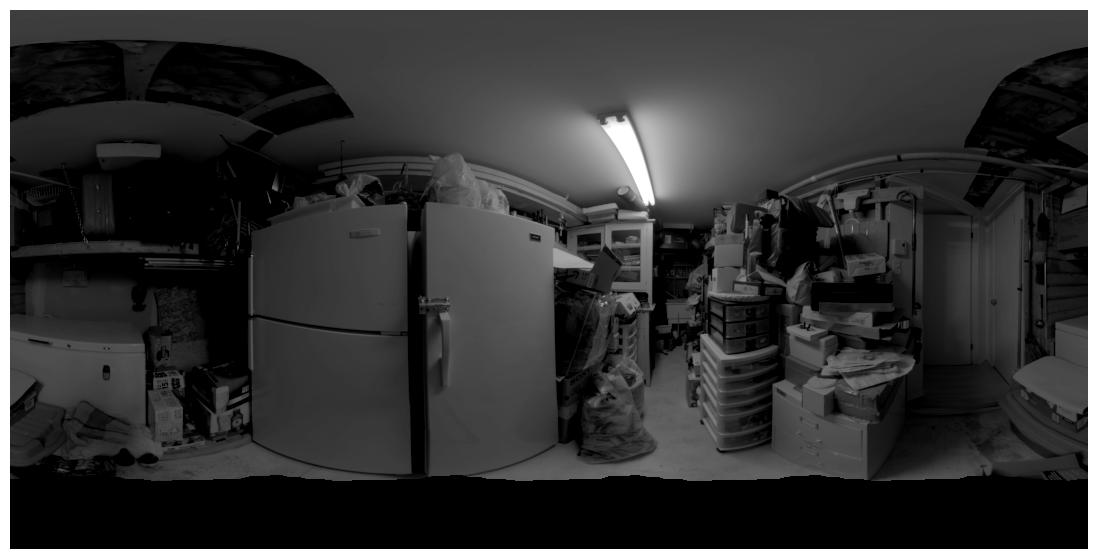}&
    \includegraphics[width=\tmplength]{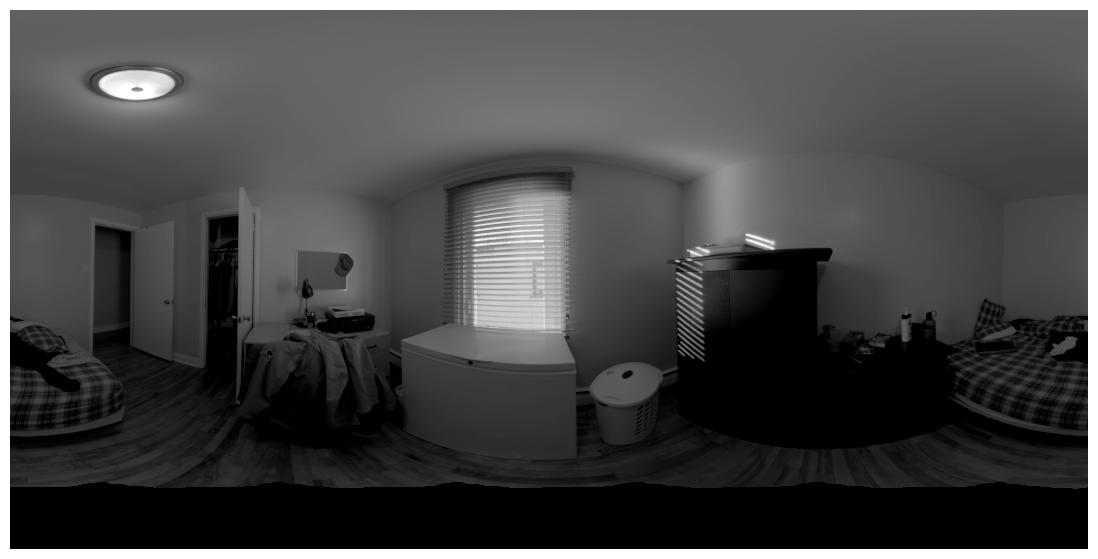}\\
    \valeur{\SI{60}{\%}} (\valeur{\SI{616}{\lux}}) & 
    \valeur{\SI{70}{\%}} (\valeur{\SI{835}{\lux}}) & 
    \valeur{\SI{80}{\%}} (\valeur{\SI{1138}{\lux}}) & 
    \valeur{\SI{90}{\%}} (\valeur{\SI{1771}{\lux}}) & 
    \valeur{\SI{95}{\%}} (\valeur{\SI{2723}{\lux}}) & 
    \valeur{\SI{99}{\%}} (\valeur{\SI{7000}{\lux}}) \\
    \includegraphics[width=\tmplength]{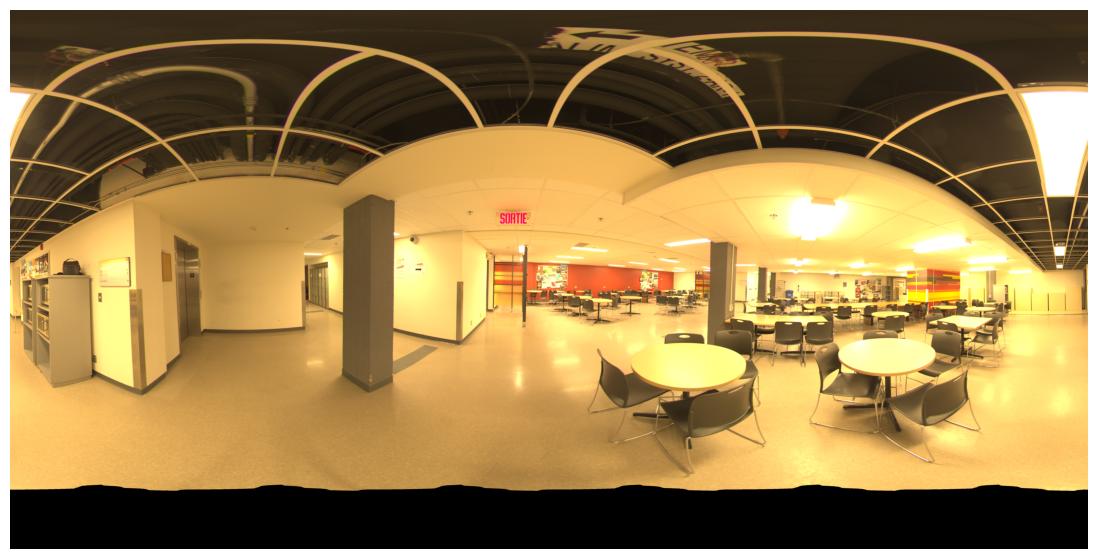}&
    \includegraphics[width=\tmplength]{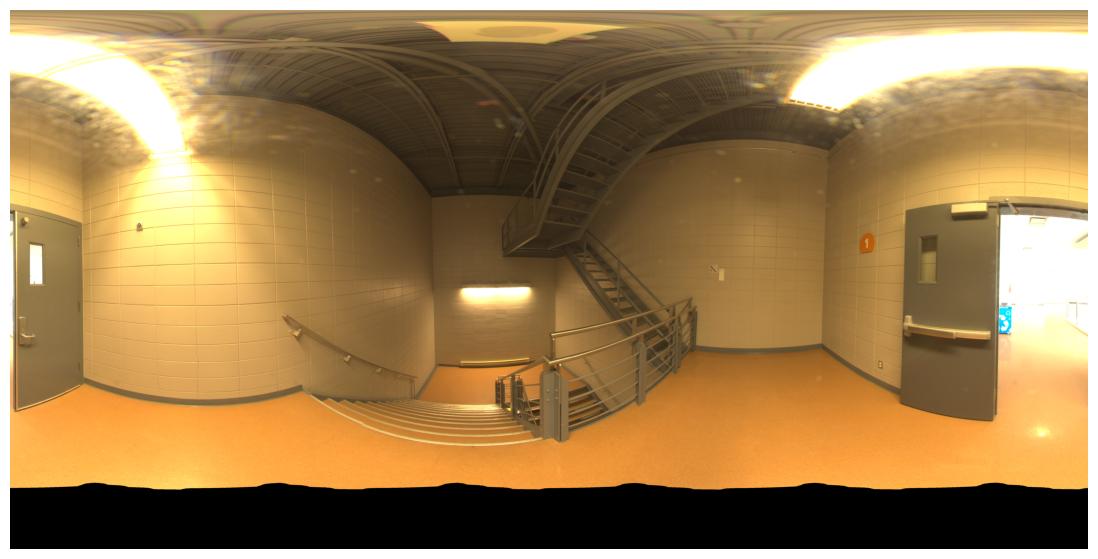}&
    \includegraphics[width=\tmplength]{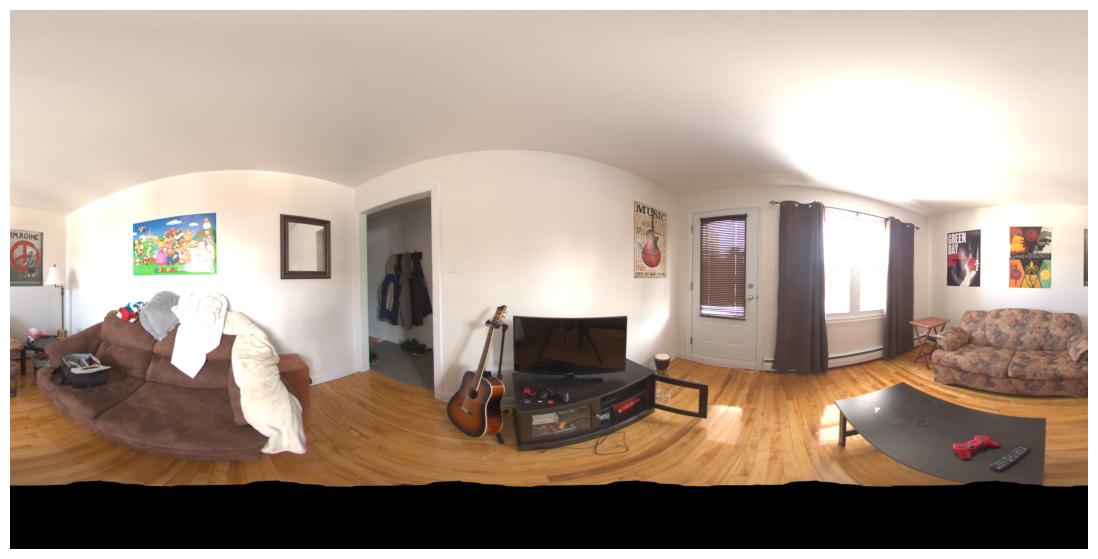}&
    \includegraphics[width=\tmplength]{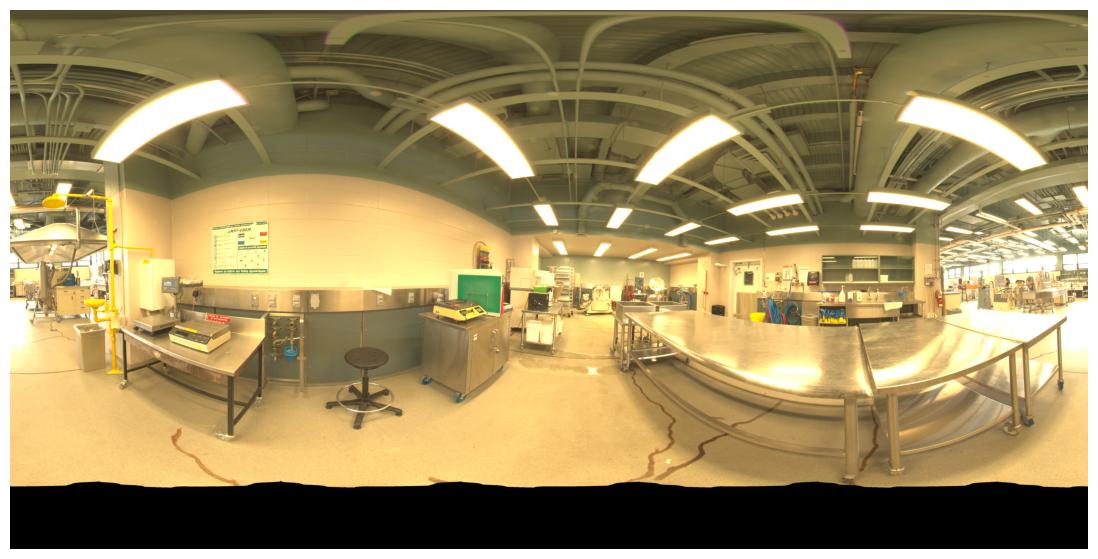}&
    \includegraphics[width=\tmplength]{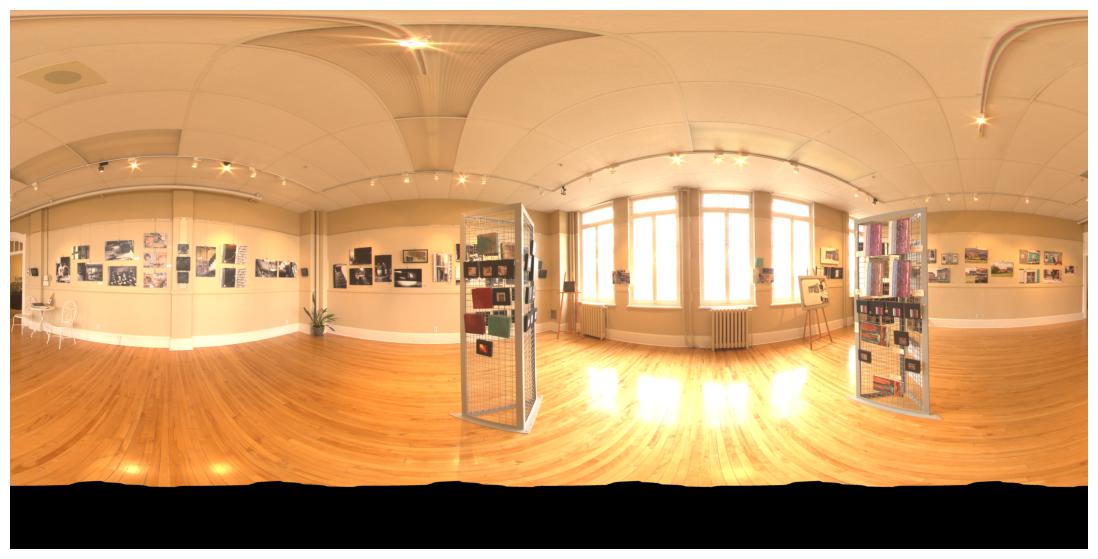}&
    \includegraphics[width=\tmplength]{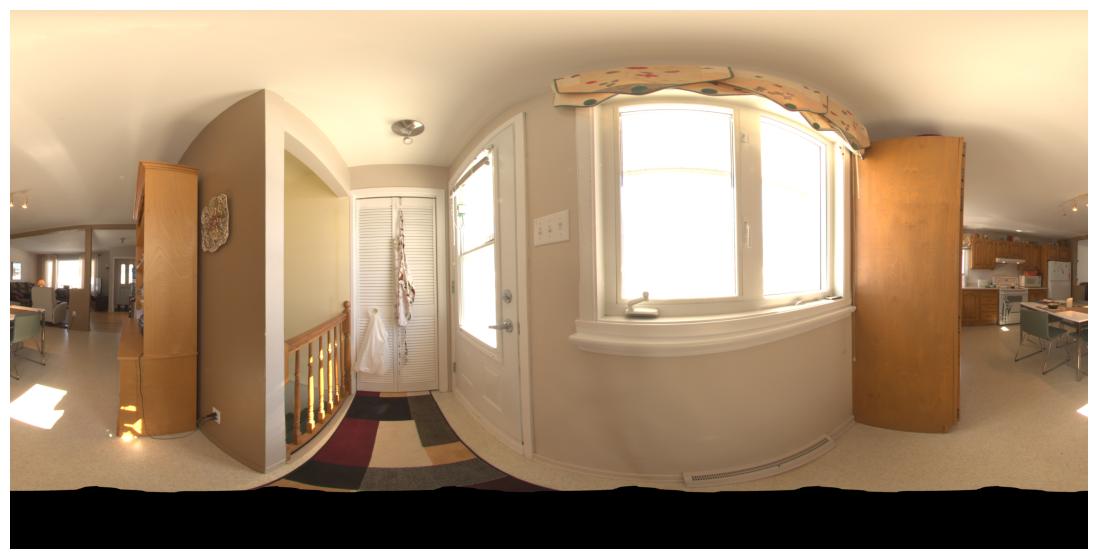}\\
    \includegraphics[width=\tmplength]{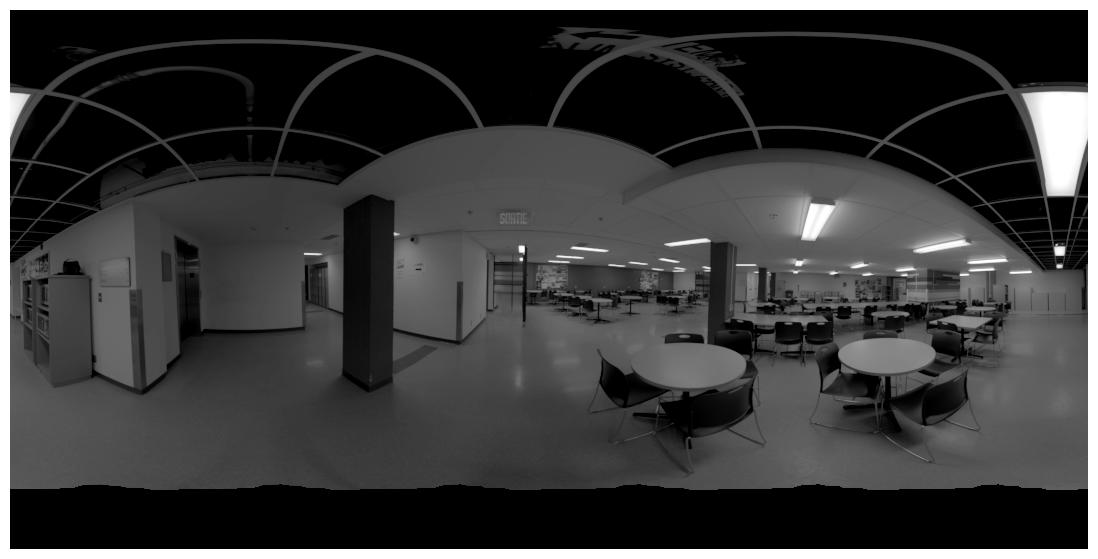}&
    \includegraphics[width=\tmplength]{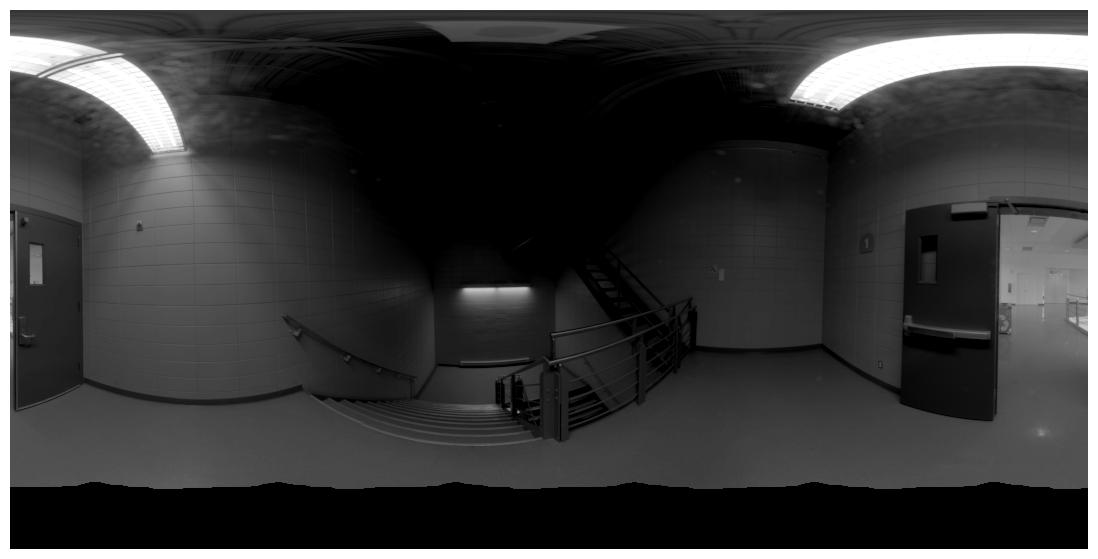}&
    \includegraphics[width=\tmplength]{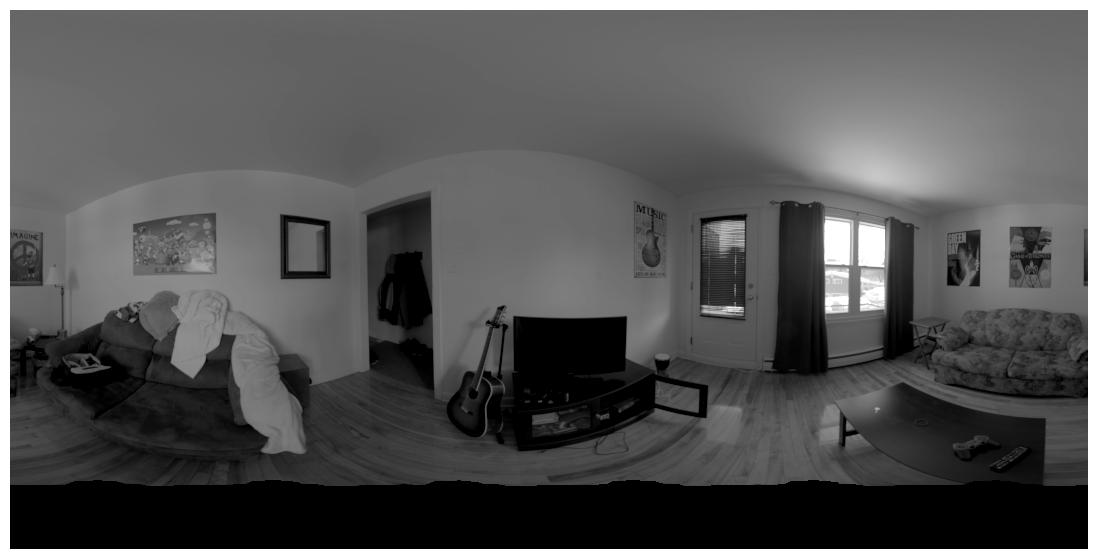}&
    \includegraphics[width=\tmplength]{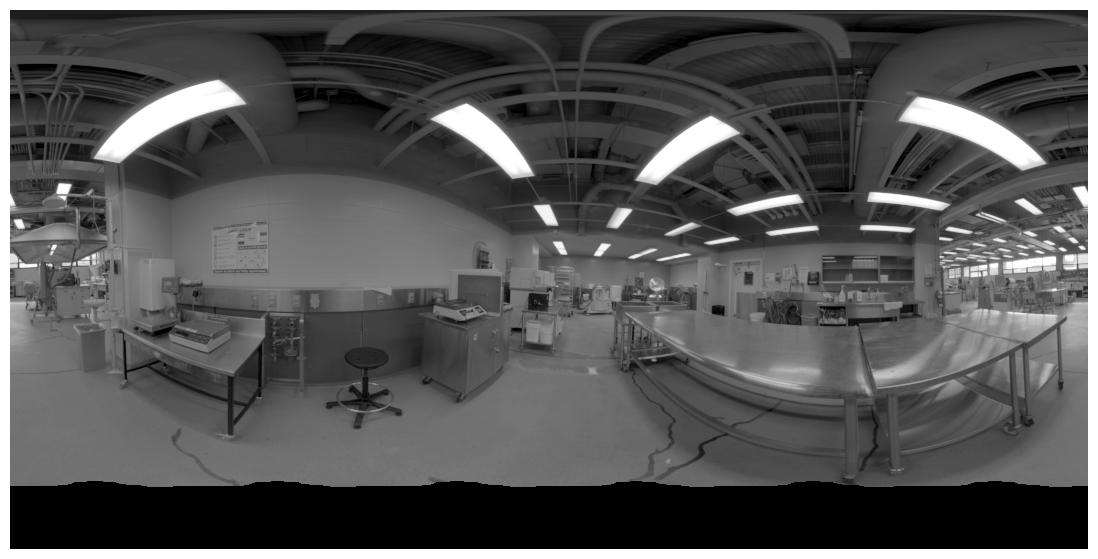}&
    \includegraphics[width=\tmplength]{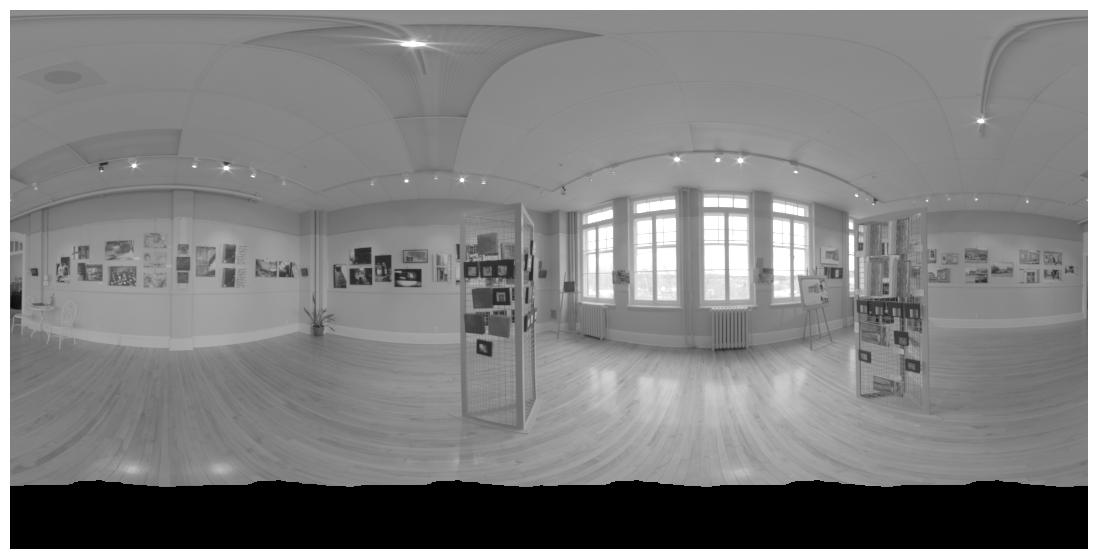}&
    \includegraphics[width=\tmplength]{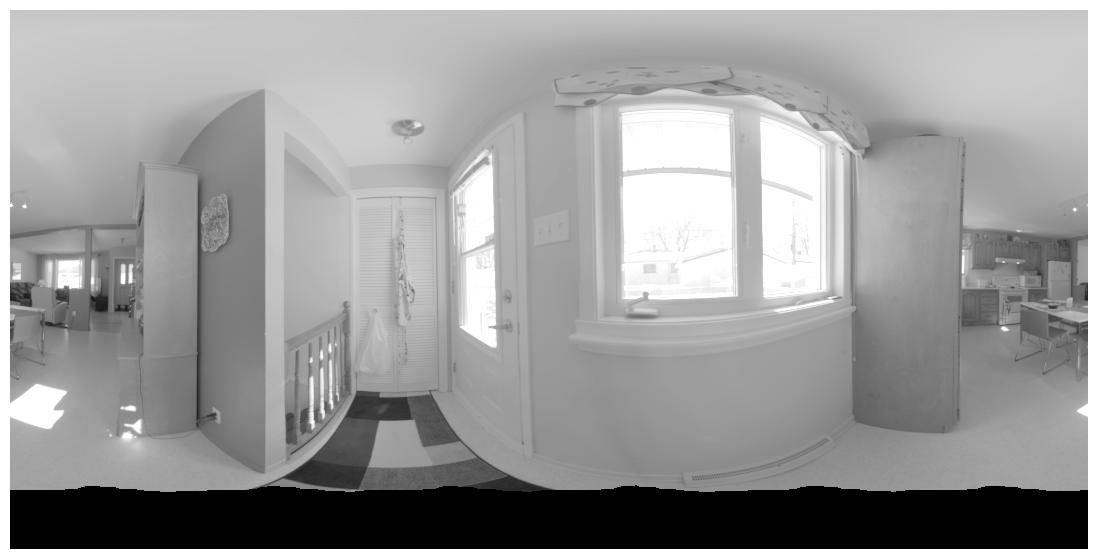}\\
    \end{tabular}
    \caption{Example scenes with mean spherical illuminance (MSI) close to the quantile values.  Greyscale images below show the corresponding log-luminance maps (color scale shown on the right).  The percentiles and corresponding measured MSI are indicated above the images.  Images are reexposed and tonemapped ($\gamma = \valeur{\num{2.2}}$) for display.} 
    \label{fig:distribution_illuminance_examples}
\end{figure*}
\begin{figure*}
   \centering
   \footnotesize
   \setlength{\tabcolsep}{0.5pt} 
   \setlength{\tmplength}{0.155\linewidth}
    \begin{tabular}{ccccccc}
    \valeur{\SI{1}{\%}} (\valeur{\SI{2199}{\K}}) & 
    \valeur{\SI{10}{\%}} (\valeur{\SI{2805}{\K}}) & 
    \valeur{\SI{20}{\%}} (\valeur{\SI{3200}{\K}}) & 
    \valeur{\SI{30}{\%}} (\valeur{\SI{3398}{\K}}) & 
    \valeur{\SI{40}{\%}} (\valeur{\SI{3532}{\K}}) & 
    \valeur{\SI{50}{\%}} (\valeur{\SI{3654}{\K}}) &  \multirow{5}{*}{\includegraphics[trim={{.010\linewidth} 0 0 {.005\linewidth}},clip, height=0.38\linewidth]{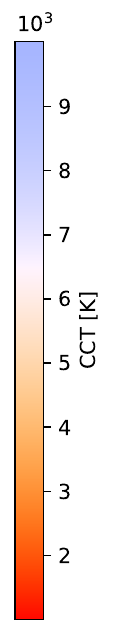}} \\
    \includegraphics[width=\tmplength]{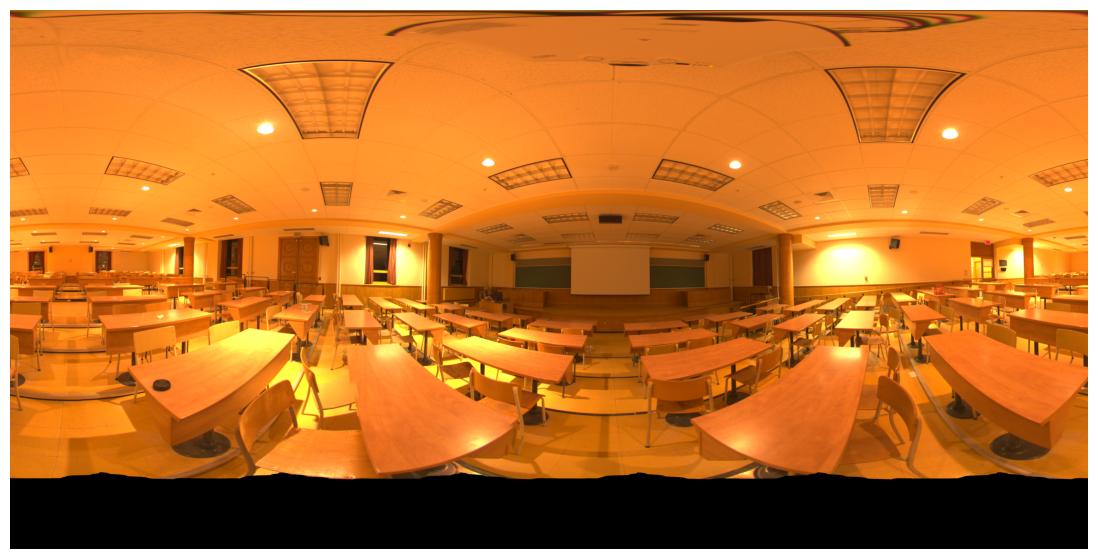}&
    \includegraphics[width=\tmplength]{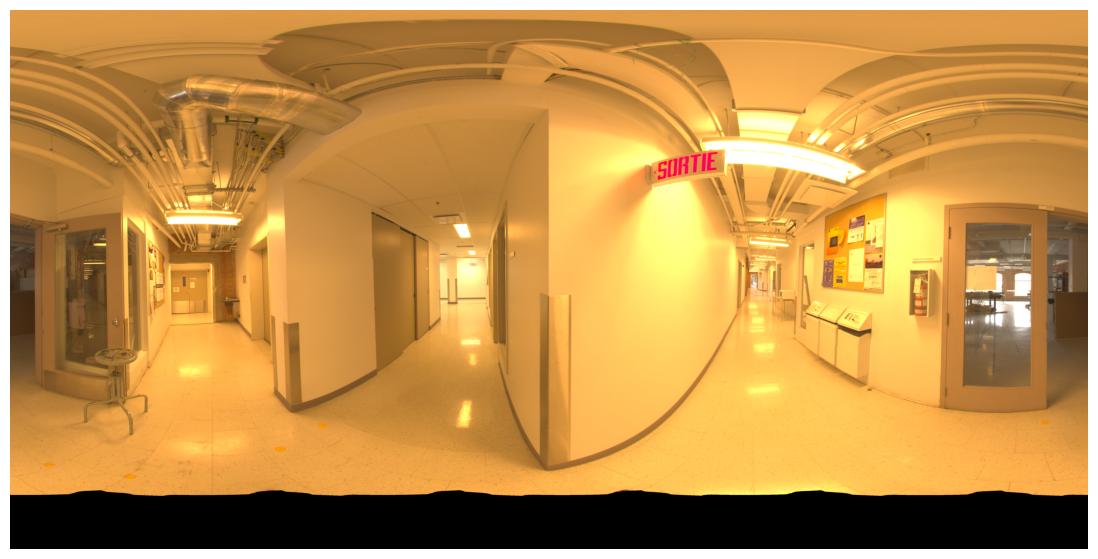}&
    \includegraphics[width=\tmplength]{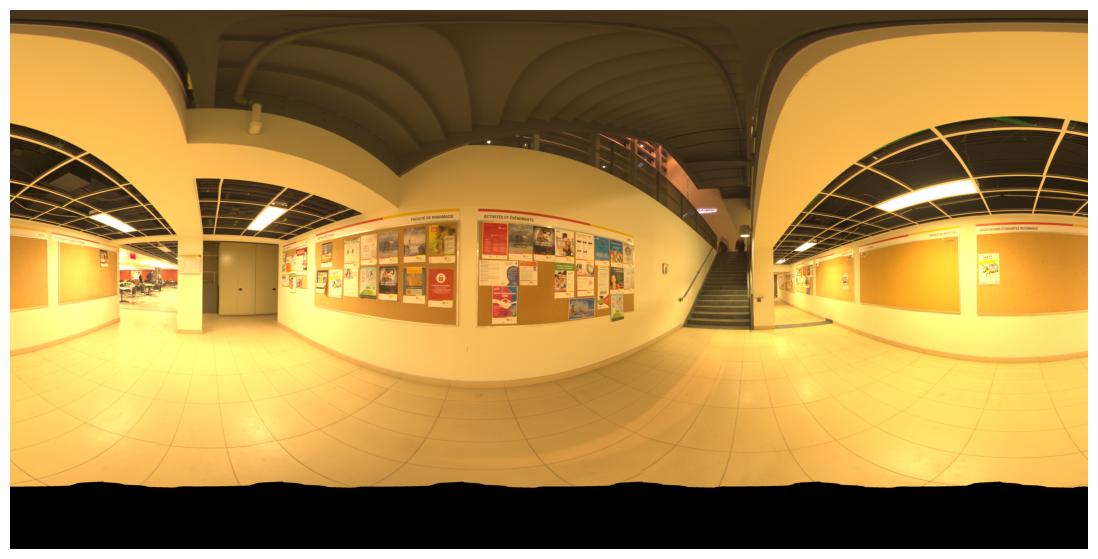}&
    \includegraphics[width=\tmplength]{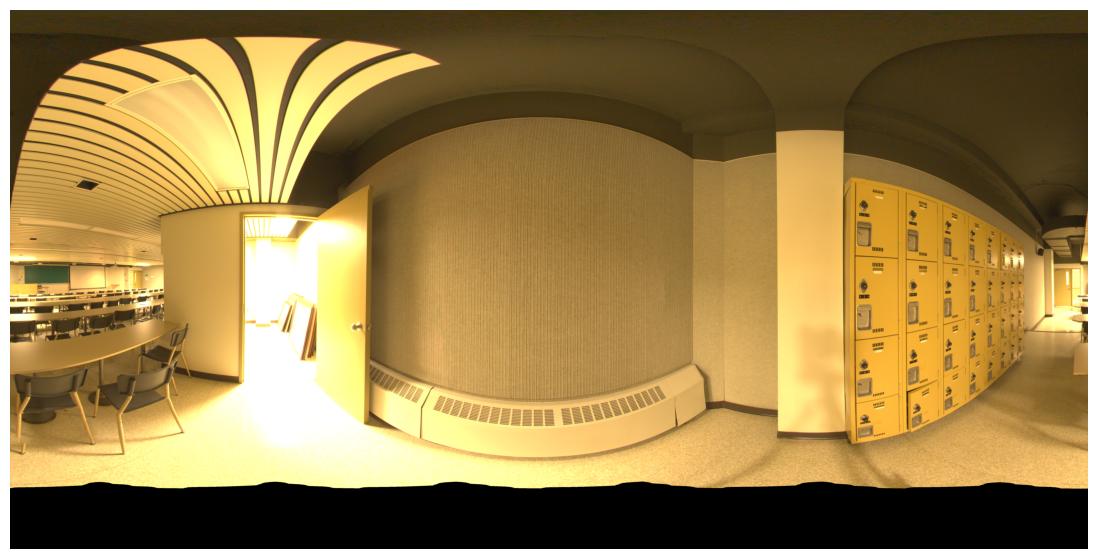}&
    \includegraphics[width=\tmplength]{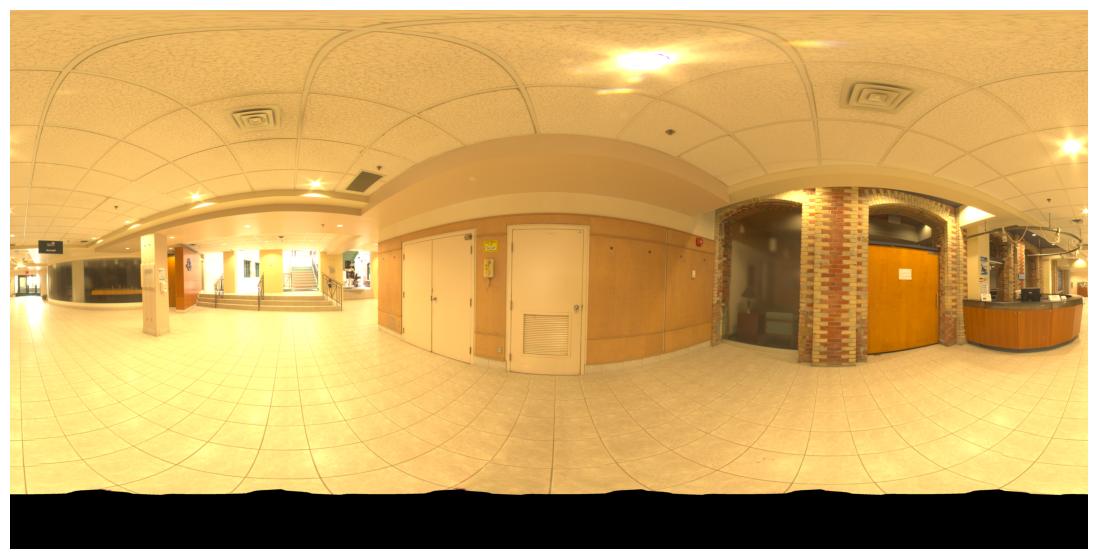}&
    \includegraphics[width=\tmplength]{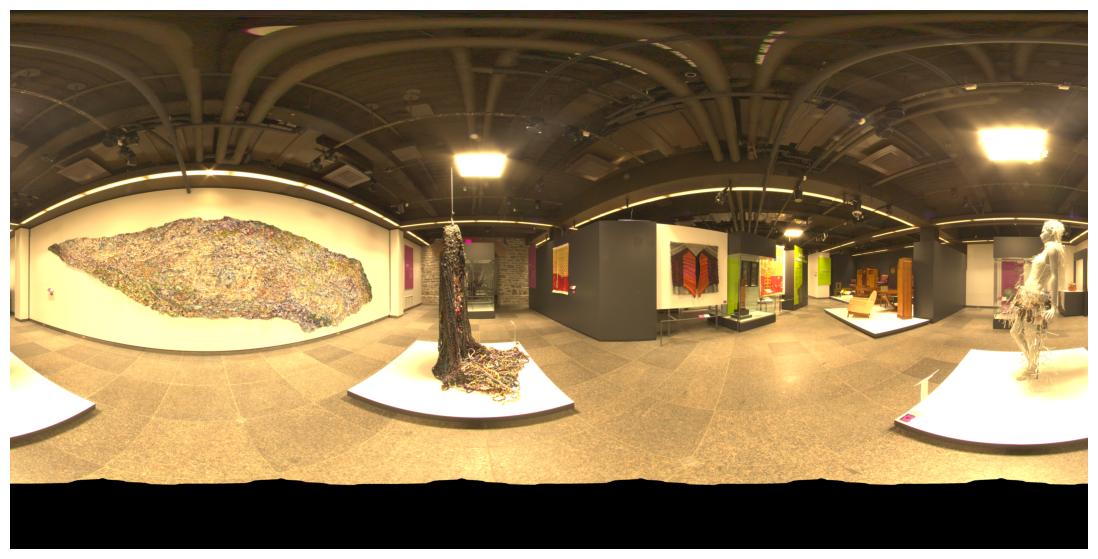}\\
    \includegraphics[width=\tmplength]{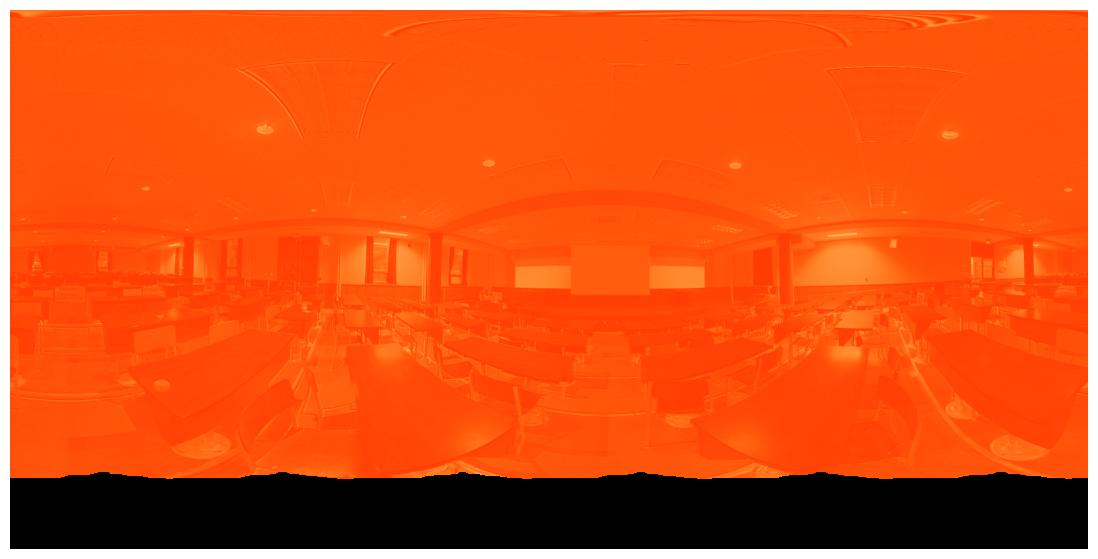}&
    \includegraphics[width=\tmplength]{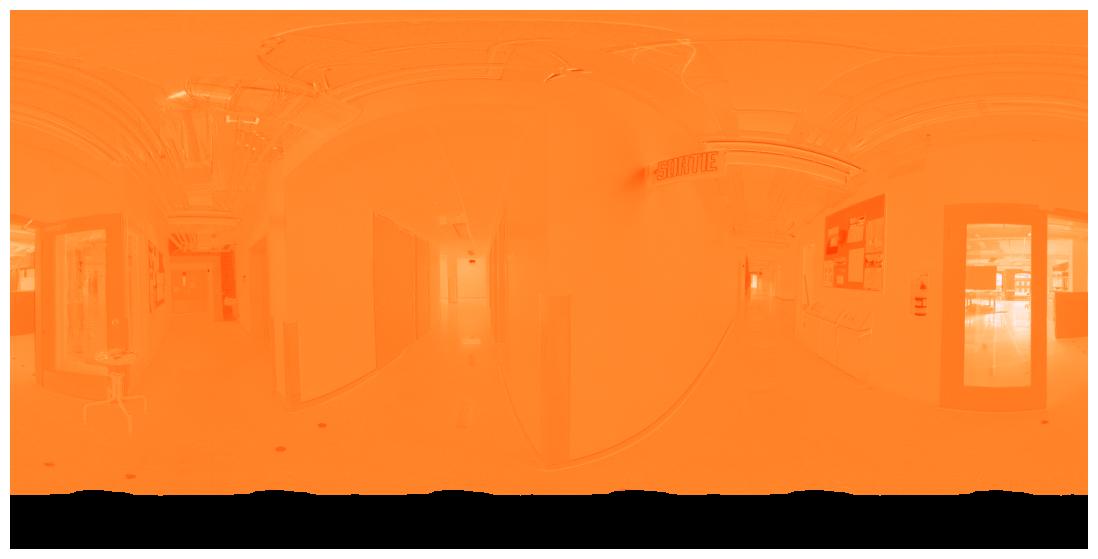}&
    \includegraphics[width=\tmplength]{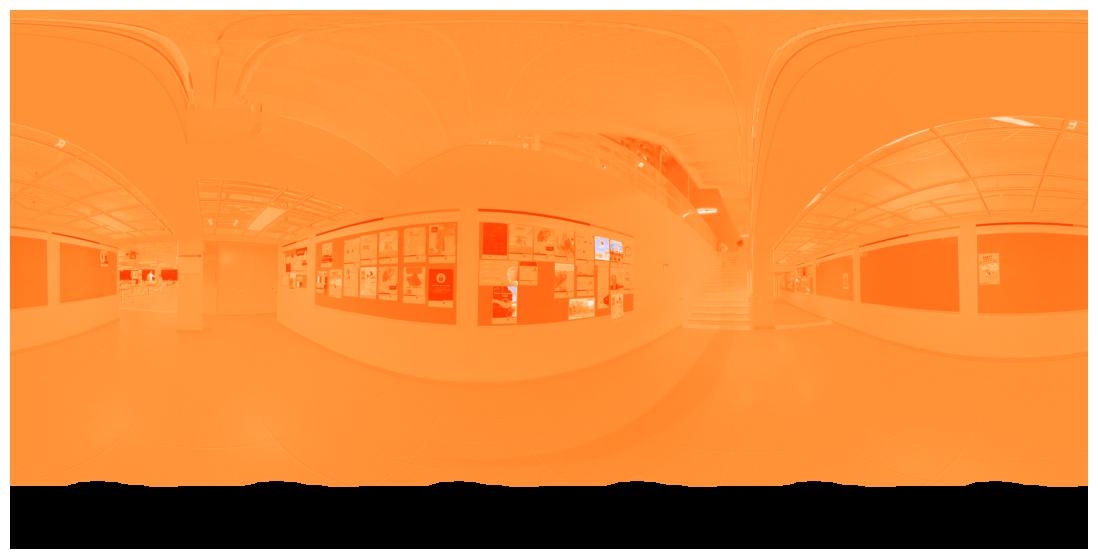}&
    \includegraphics[width=\tmplength]{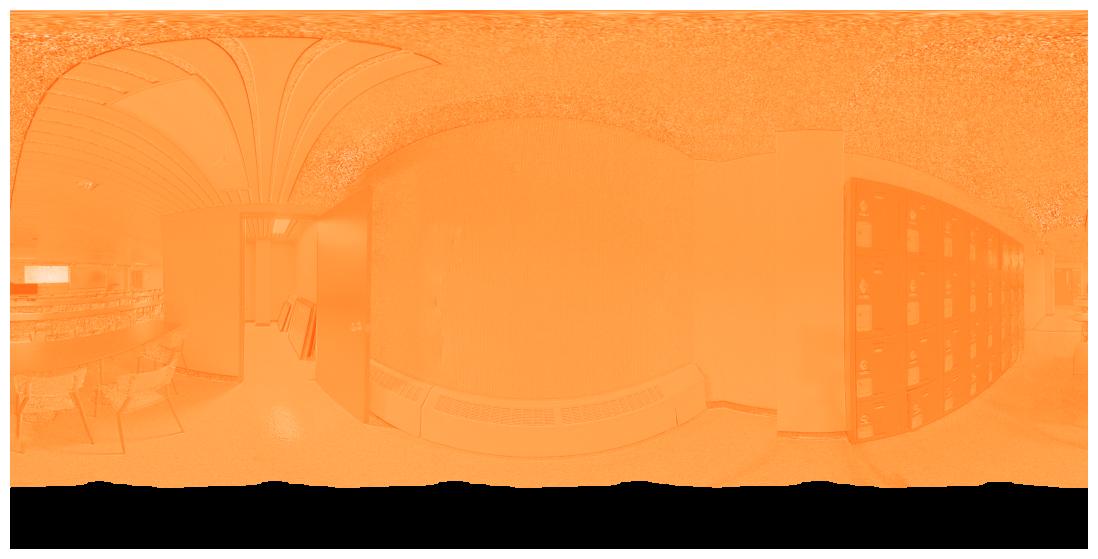}&
    \includegraphics[width=\tmplength]{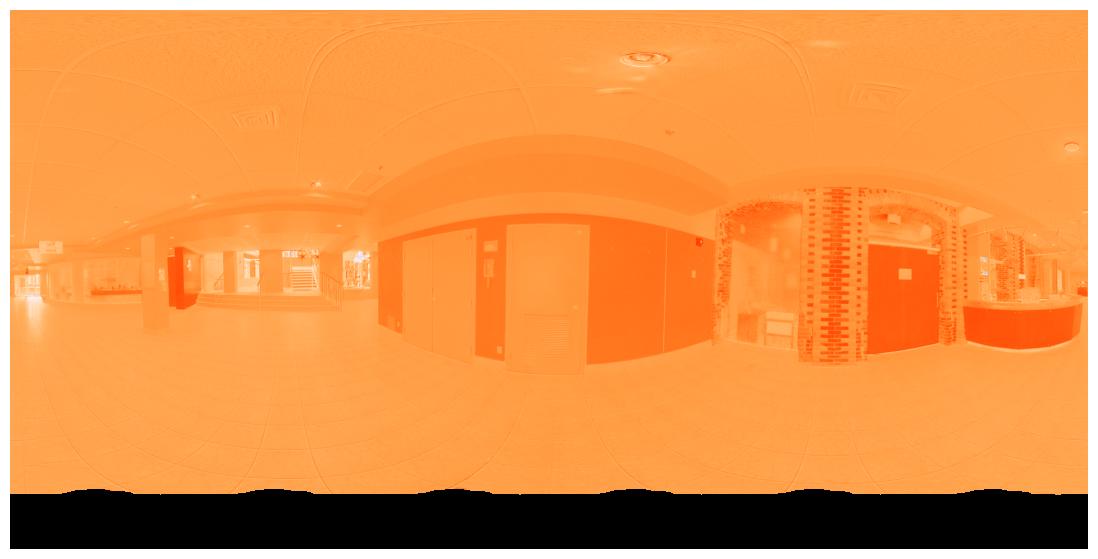}&
    \includegraphics[width=\tmplength]{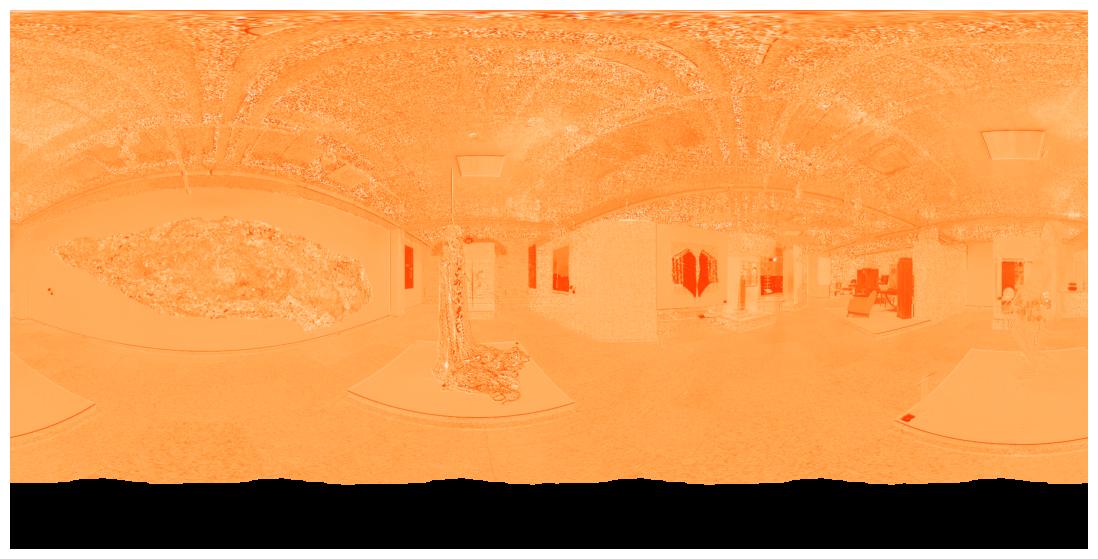}\\
    \valeur{\SI{60}{\%}} (\valeur{\SI{3847}{\K}}) & 
    \valeur{\SI{70}{\%}} (\valeur{\SI{4261}{\K}}) & 
    \valeur{\SI{80}{\%}} (\valeur{\SI{4876}{\K}}) & 
    \valeur{\SI{90}{\%}} (\valeur{\SI{5656}{\K}}) & 
    \valeur{\SI{95}{\%}} (\valeur{\SI{6304}{\K}}) & 
    \valeur{\SI{99}{\%}} (\valeur{\SI{8103}{\K}}) \\
    \includegraphics[width=\tmplength]{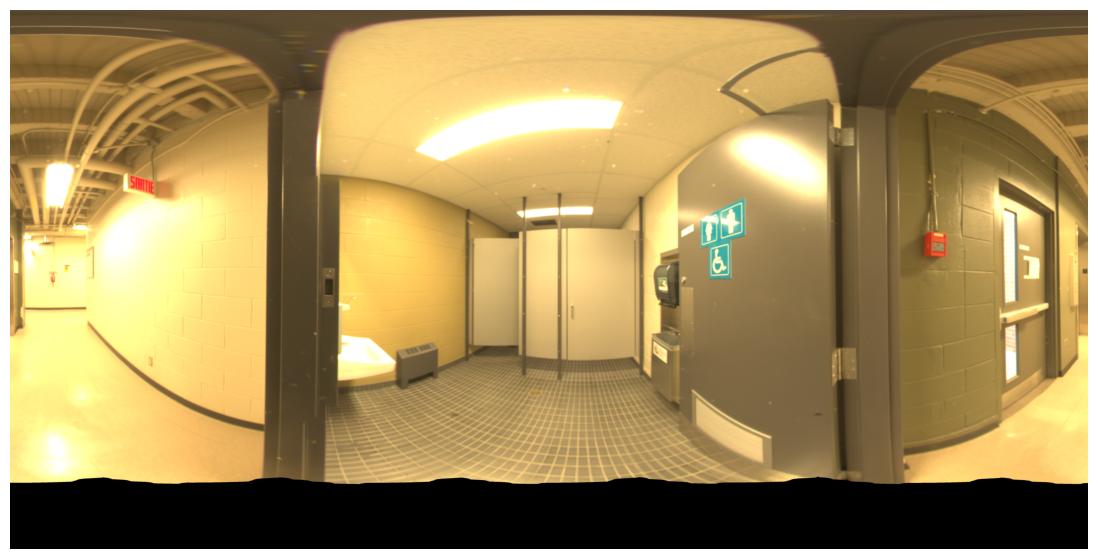}&
    \includegraphics[width=\tmplength]{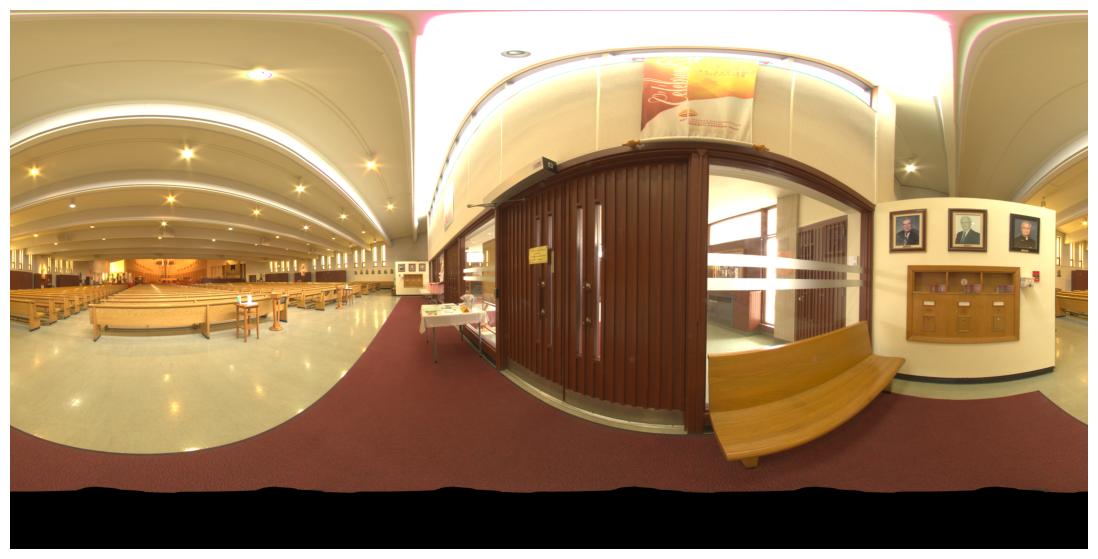}&
    \includegraphics[width=\tmplength]{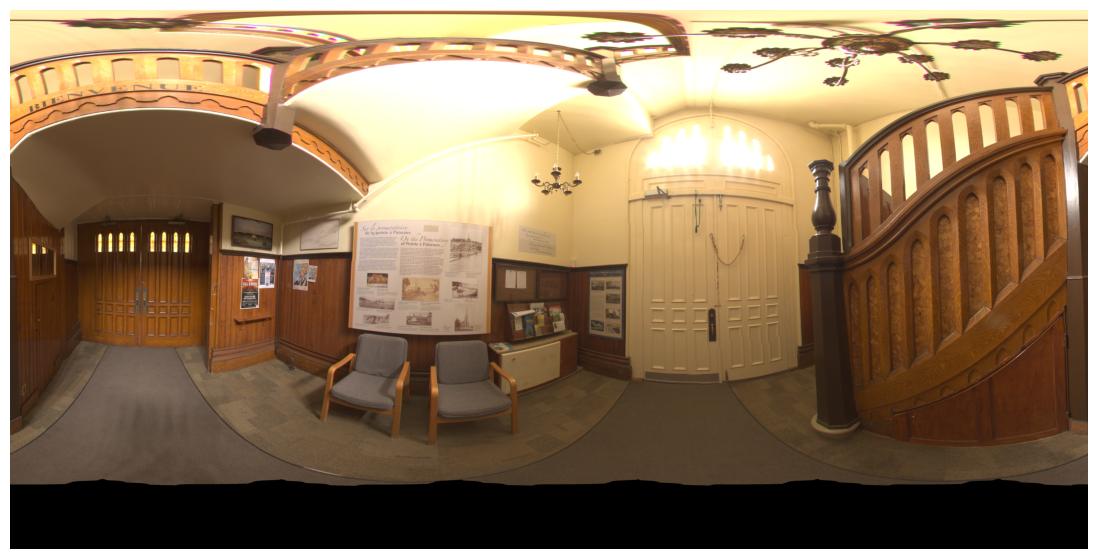}&
    \includegraphics[width=\tmplength]{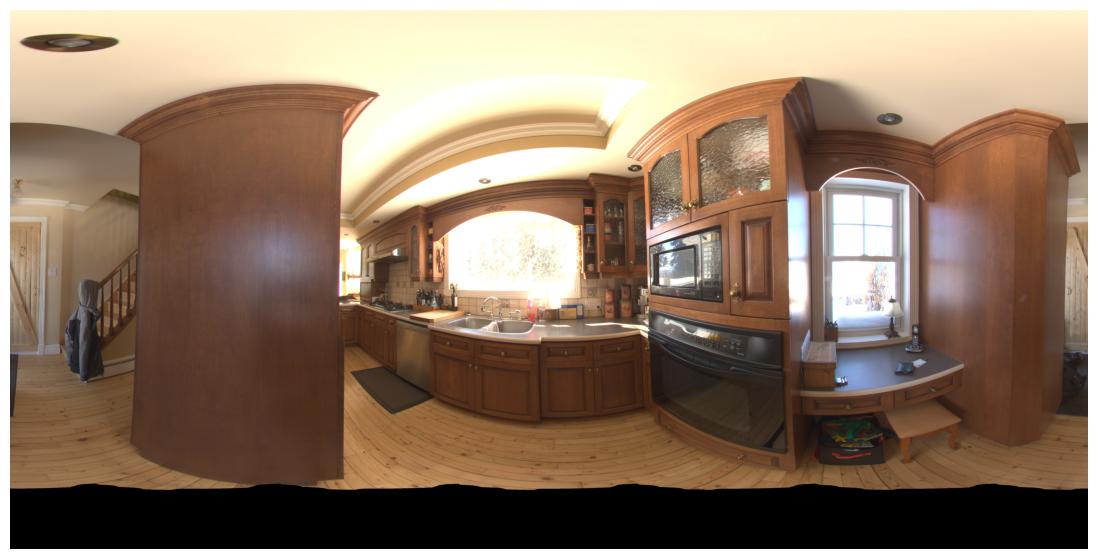}&
    \includegraphics[width=\tmplength]{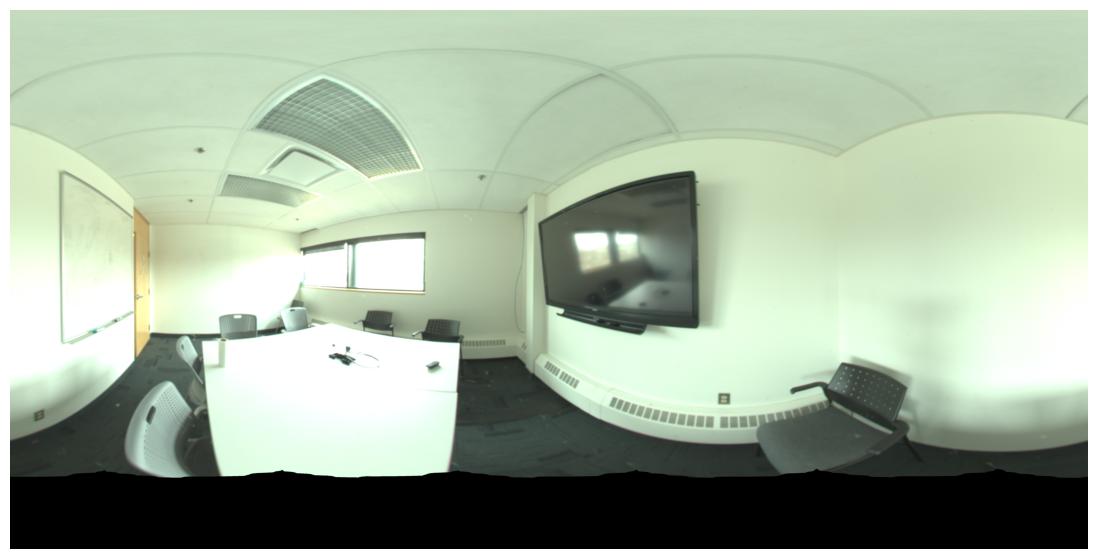}&
    \includegraphics[width=\tmplength]{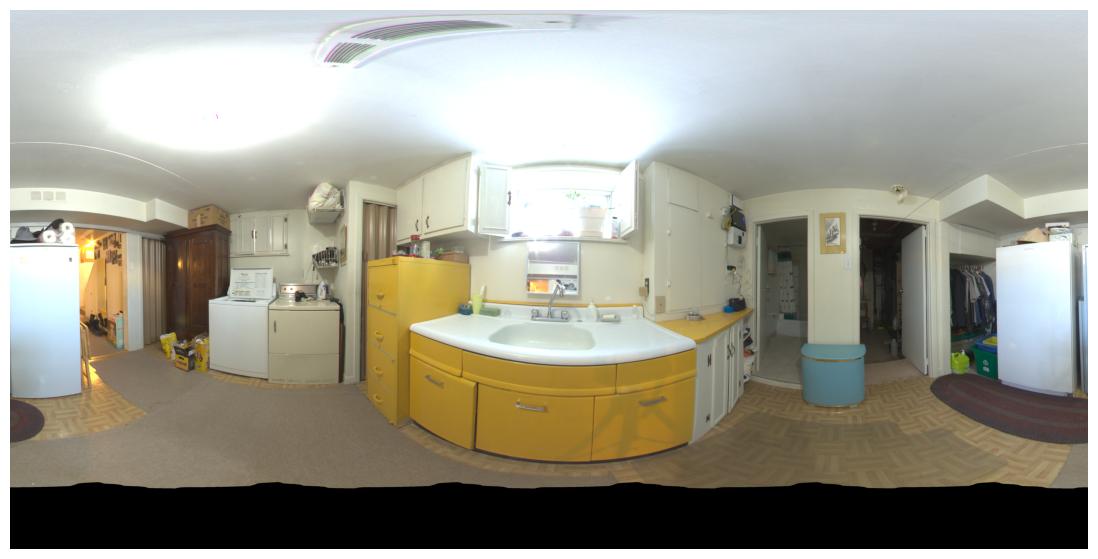}\\
    \includegraphics[width=\tmplength]{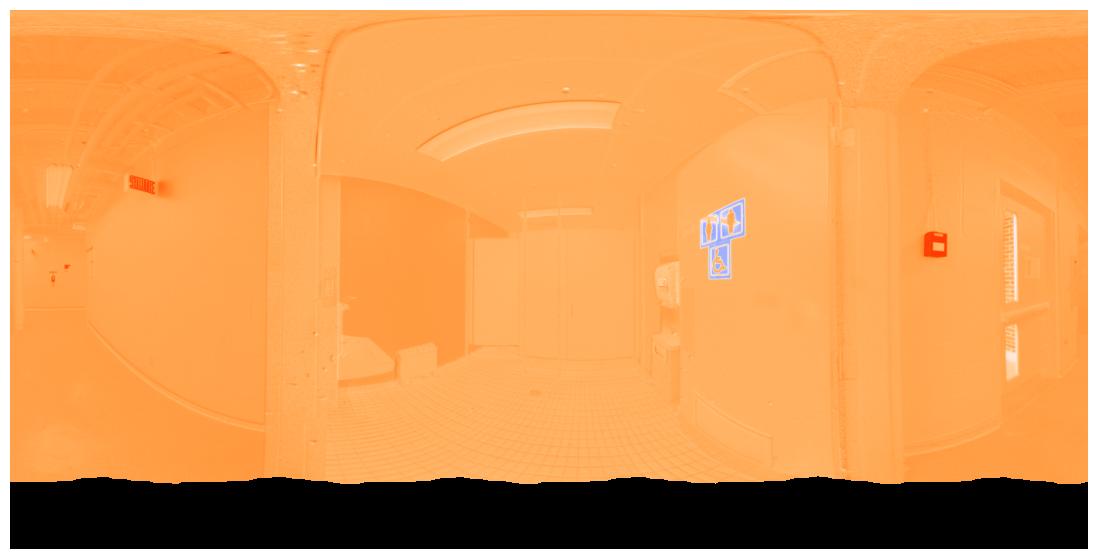}&
    \includegraphics[width=\tmplength]{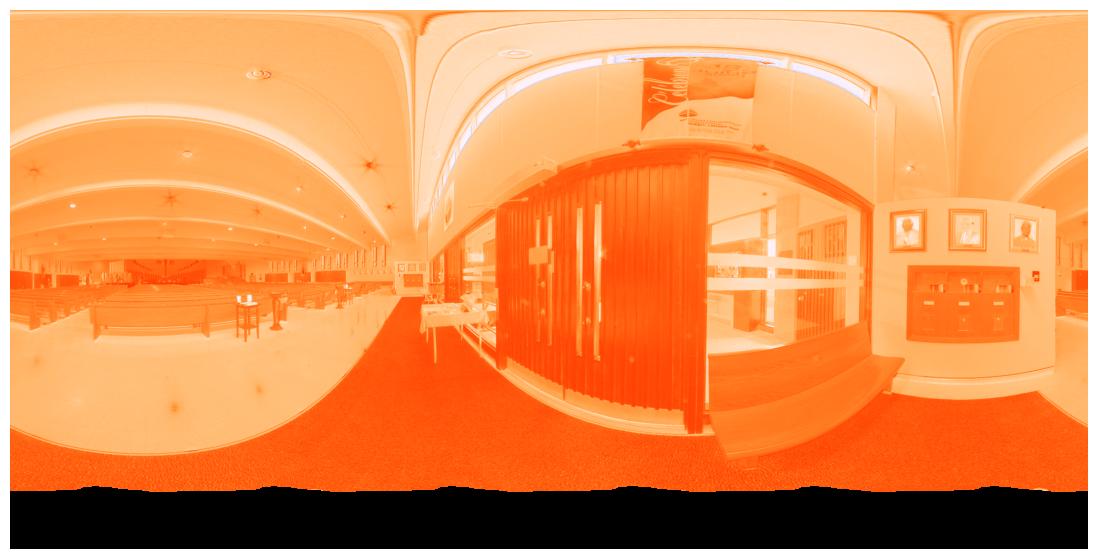}&
    \includegraphics[width=\tmplength]{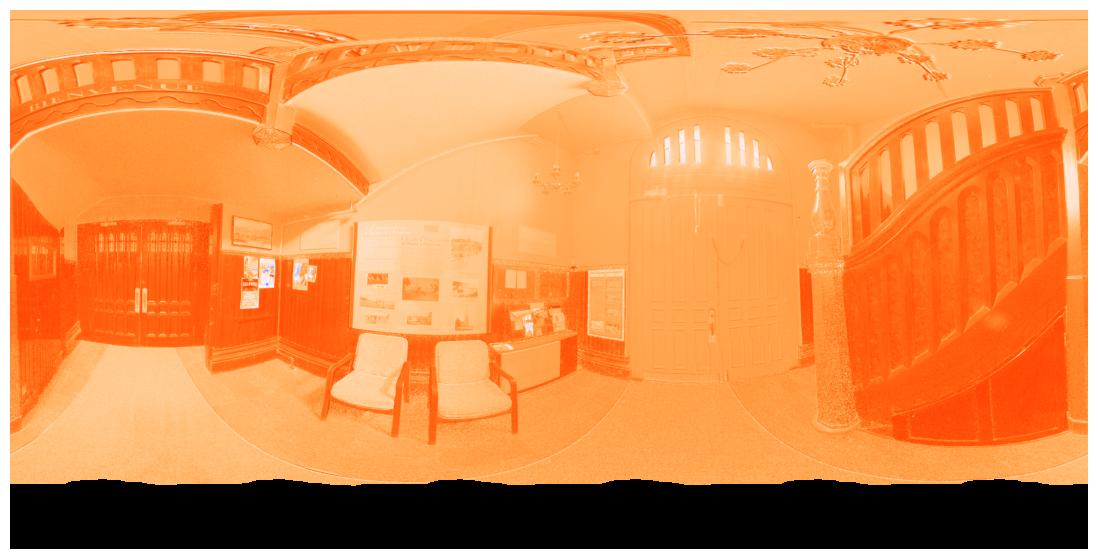}&
    \includegraphics[width=\tmplength]{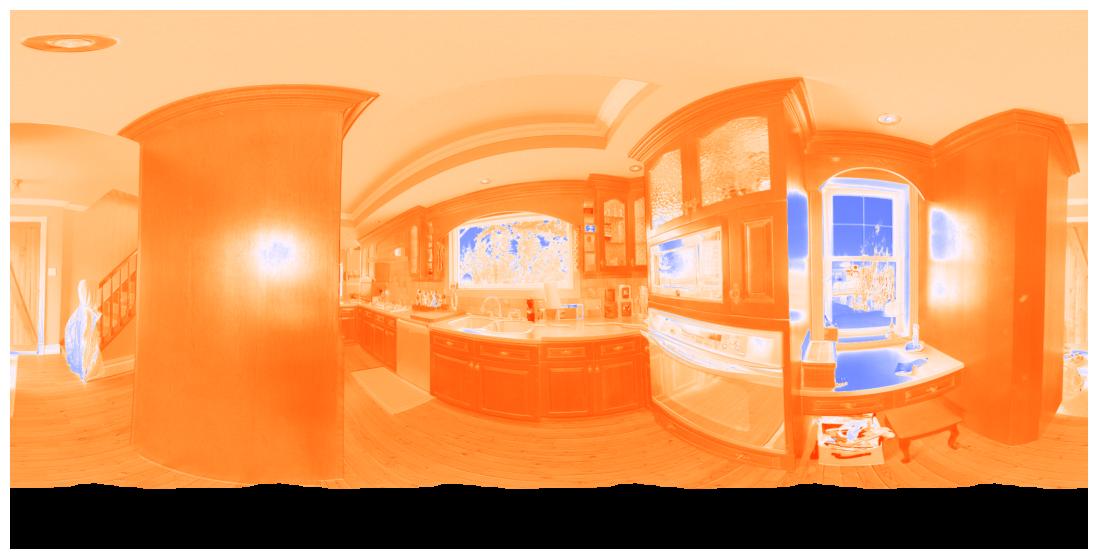}&
    \includegraphics[width=\tmplength]{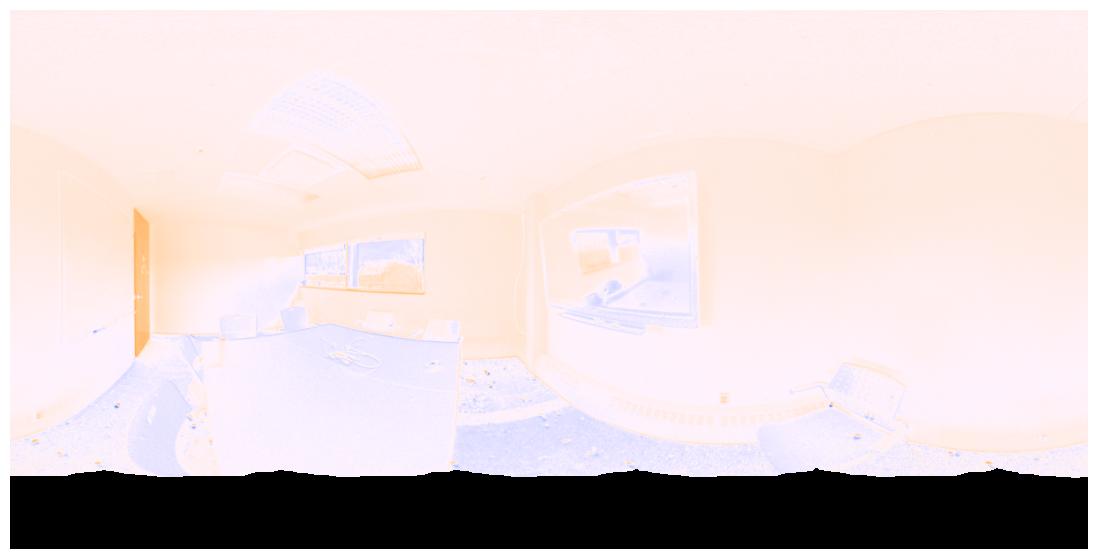}&
    \includegraphics[width=\tmplength]{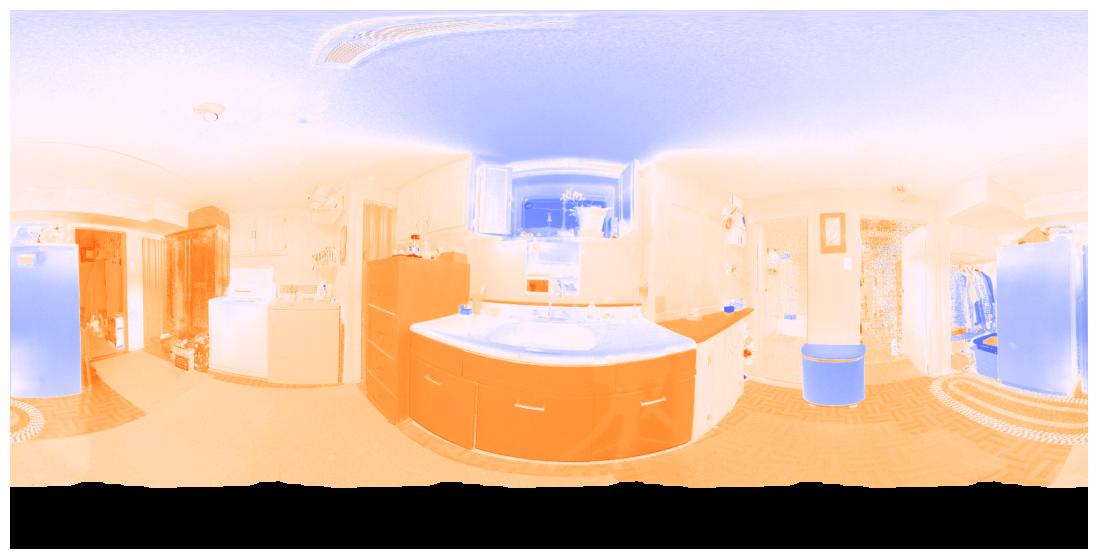}\\
    \end{tabular}
    \caption{Example scenes with CCT close to the quantile values.  Colored images below show the CCT map of the scenes, with corresponding scale shown on the right.  The percentiles and corresponding measured scene CCT are indicated above the images.  Images are reexposed and tonemapped ($\gamma = \valeur{\num{2.2}}$) for display.} 
    \label{fig:distribution_temp_examples}
\end{figure*}

We represent an entire \ang{360} panorama as its mean spherical illuminance (MSI) ~\cite{ILV2020} and CCT (cf. supplementary). The correlation between both quantities and their distributions over the entire dataset is presented in \cref{fig:jointplot_temperature_illuminance_whole_pano}. 
As can be observed in \cref{fig:jointplot_temperature_illuminance_whole_pano}, the majority of panoramas in the dataset possess relatively low MSI and CCT (bottom-left quadrant in the plot). Indeed, the median MSI and CCT of the dataset are \valeur{\SI{461}{\lux}} and \valeur{\SI{3654}{\K}} respectively.  \Cref{fig:distribution_illuminance_examples,fig:distribution_temp_examples} illustrate the dataset, showing visual examples sorted by MSI and CCT respectively.
For example, \cref{fig:distribution_illuminance_examples} shows that scenes with lower MSI correspond to basements and bedrooms with curtains and artificial lighting (incandescent, compact fluorescent and/or ``soft white'' LED), whereas scenes with higher MSI correspond to well-lit public spaces, often containing very bright ceiling lights or large windows on a sunny day. Similarly, \cref{fig:distribution_temp_examples} illustrates that scenes with lower CCT have predominantly artificial lighting, whereas higher CCT can be due to strong outdoor lighting from windows and/or more neutral surface reflectance.

\subsection{Individual light sources}
\label{subsec:viz_dataset_sources}

To provide a more fine-grained analysis of the dataset, we detect and segment light sources in panoramas using the approach by Gardner~\etal~\cite{gardner-iccv-19}. In total, \valeur{\num{11060}} light sources are detected, for which the mean CCT and luminance are computed. Overall, the average mean luminance for all the light sources included in the dataset is \valeur{\SI{27874}{\candela\per\m\squared}} (median of \valeur{\SI{3854}{\candela\per\m\squared}}), and the average mean CCT is \valeur{\SI{3648}{\K}} (median of \valeur{\SI{3380}{\K}}).

\begin{figure*}
   \centering
   \scriptsize
   \setlength{\tabcolsep}{0.5pt} 
   \newlength{\mywidth}
   \setlength{\mywidth}{0.14\linewidth}
    \begin{tabular}{lccccccc}
    & & & & & & & 
    \multirow{3}*{\includegraphics[width=0.2\linewidth]{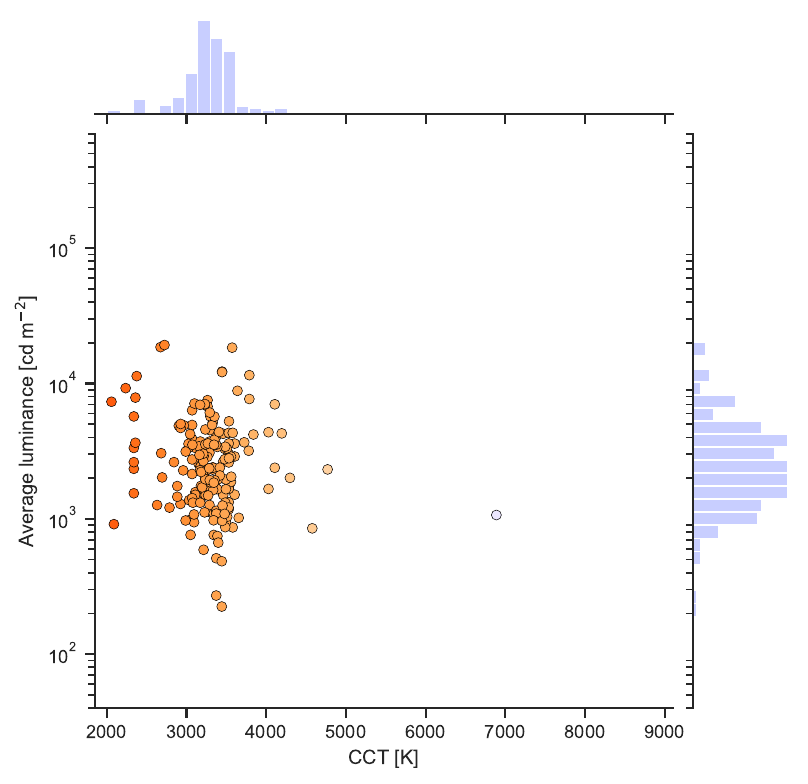}}
    \\
    & 
    \valeur{\SI{7330}{\candela\per\m\squared}} / \valeur{\SI{2059}{\K}} & 
    \valeur{\SI{3053}{\candela\per\m\squared}} / \valeur{\SI{2683}{\K}} & 
    \valeur{\SI{6095}{\candela\per\m\squared}} / \valeur{\SI{3292}{\K}} & 
    \valeur{\SI{3183}{\candela\per\m\squared}} / \valeur{\SI{3781}{\K}} & 
    \valeur{\SI{2318}{\candela\per\m\squared}} / \valeur{\SI{4771}{\K}} & & 
    \\
    \rotatebox{90}{\hspace{1cm} Tubes} & 
    \includegraphics[width=\mywidth]{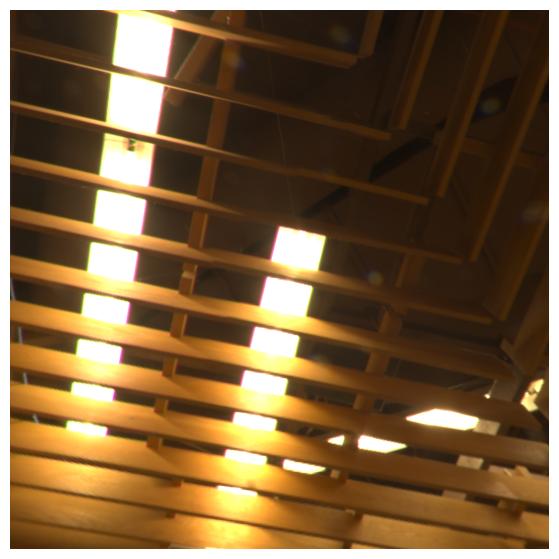}&
    \includegraphics[width=\mywidth]{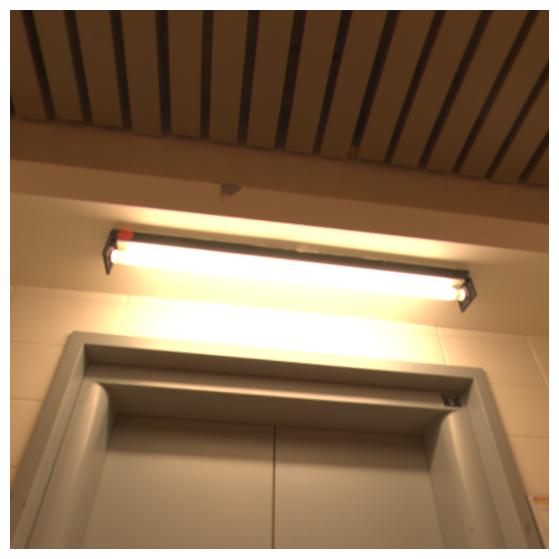}&
    \includegraphics[width=\mywidth]{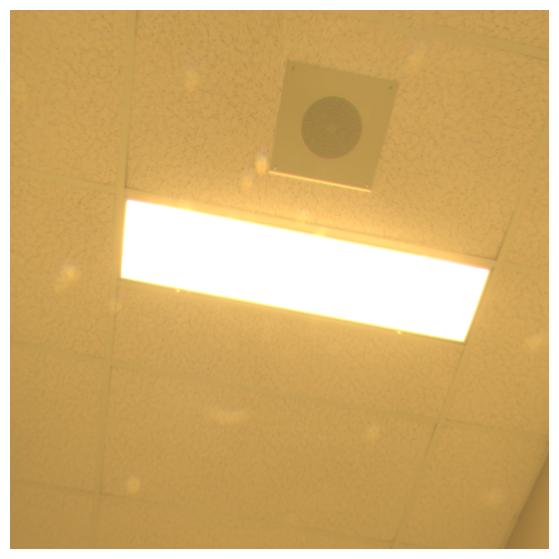}&
    \includegraphics[width=\mywidth]{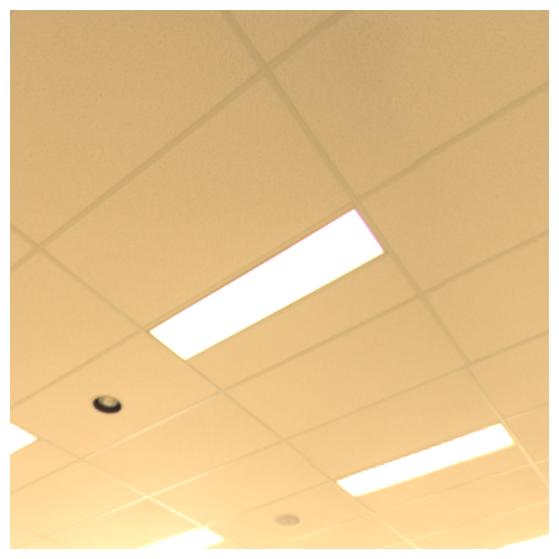}&
    \includegraphics[width=\mywidth]{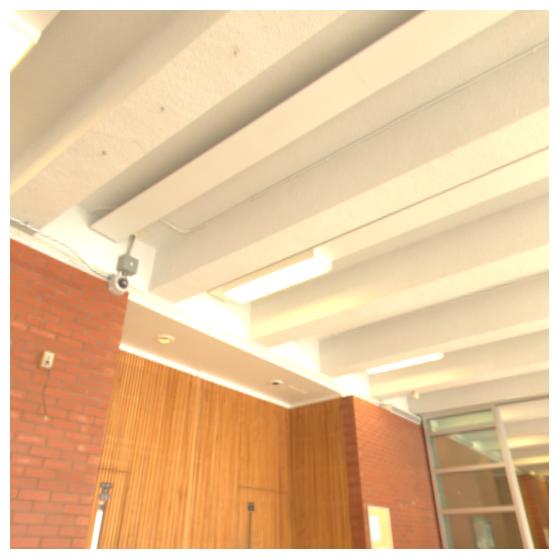}&
    & 
    \\*[0.5em]
    & & & & & & & 
    \multirow{3}*{\includegraphics[width=0.2\linewidth]{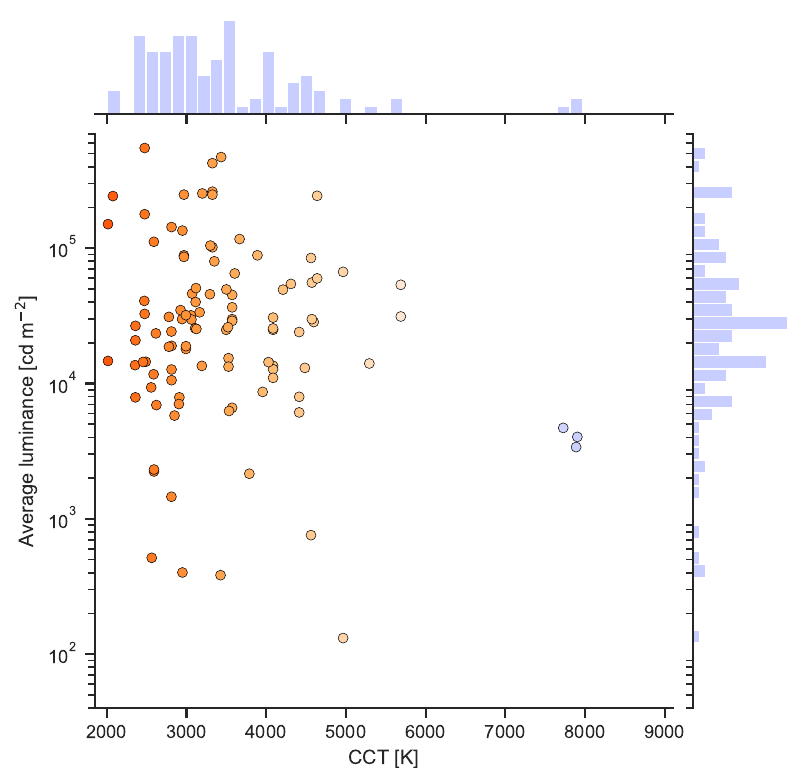}}
    \\ 
    & 
    \valeur{\SI{14476}{\candela\per\m\squared}} / \valeur{\SI{2489}{\K}} & 
    \valeur{\SI{23458}{\candela\per\m\squared}} / \valeur{\SI{2617}{\K}} & 
    \valeur{\SI{31972}{\candela\per\m\squared}} / \valeur{\SI{2993}{\K}} & 
    \valeur{\SI{28460}{\candela\per\m\squared}} / \valeur{\SI{4597}{\K}} & 
    \valeur{\SI{4698}{\candela\per\m\squared}} / \valeur{\SI{7727}{\K}} & & 
    \\
    \rotatebox{90}{\hspace{1cm} Bulbs} & 
    \includegraphics[width=\mywidth]{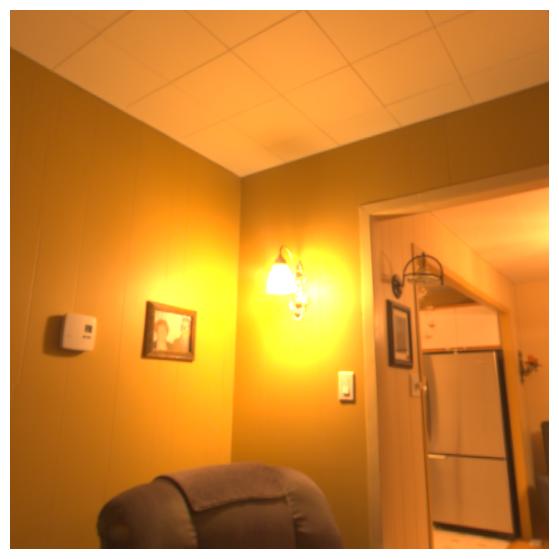}&
    \includegraphics[width=\mywidth]{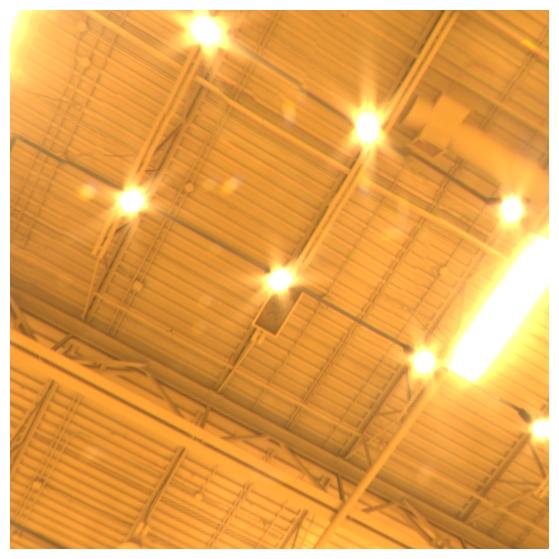}&
    \includegraphics[width=\mywidth]{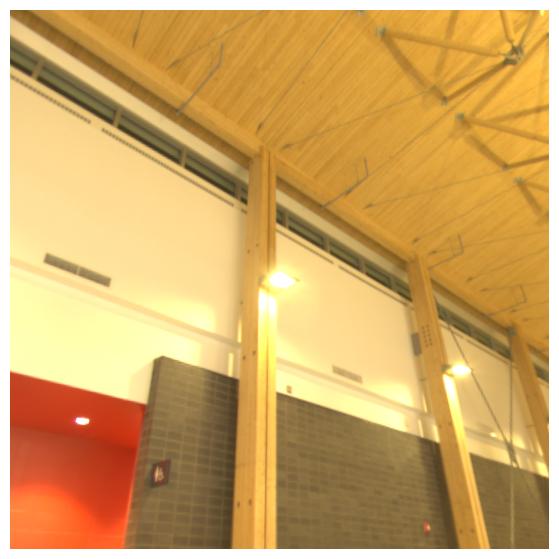}&
    \includegraphics[width=\mywidth]{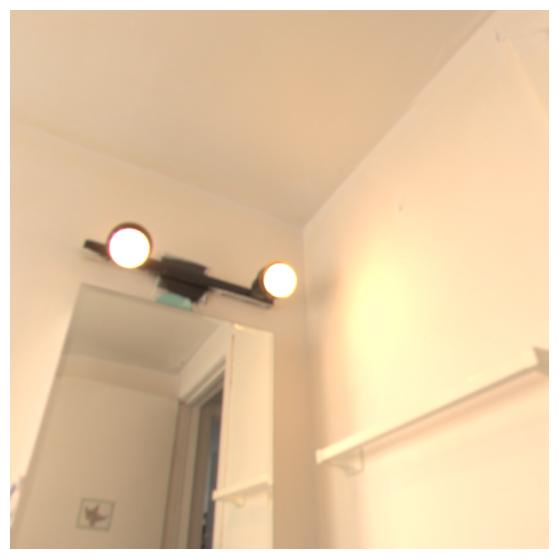}&
    \includegraphics[width=\mywidth]{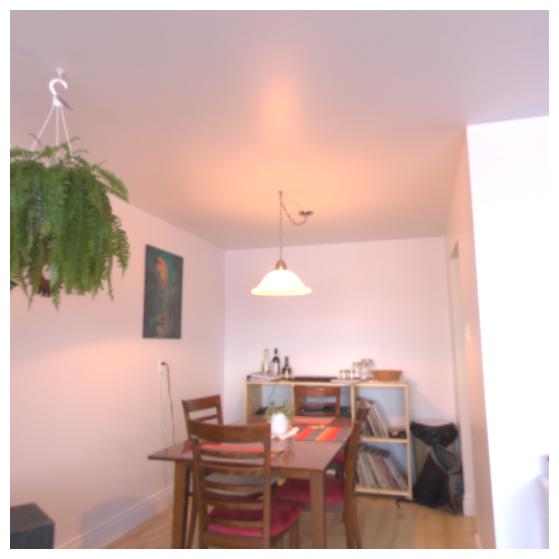}& 
    &
    \\*[0.5em]
    & & & & & & & 
    \multirow{3}*{\includegraphics[width=0.2\linewidth]{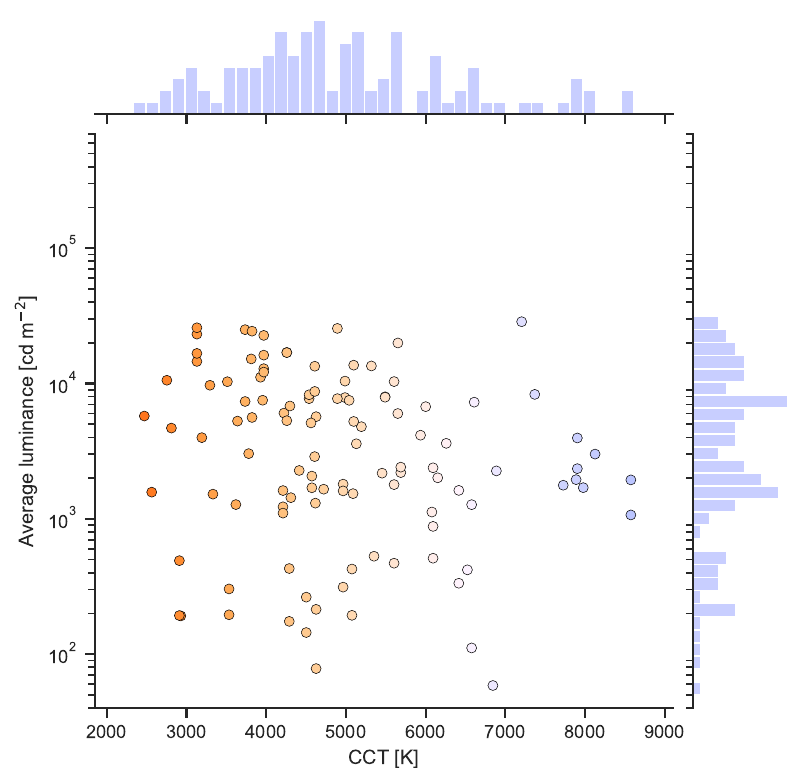}}
    \\ 
    & 
    \valeur{\SI{491}{\candela\per\m\squared}} / \valeur{\SI{2911}{\K}} & 
    \valeur{\SI{1274}{\candela\per\m\squared}} / \valeur{\SI{3620}{\K}} & 
    \valeur{\SI{1433}{\candela\per\m\squared}} / \valeur{\SI{4310}{\K}} & 
    \valeur{\SI{10333}{\candela\per\m\squared}} / \valeur{\SI{5604}{\K}} & 
    \valeur{\SI{1981}{\candela\per\m\squared}} / \valeur{\SI{9025}{\K}} & &
    \\
    \rotatebox{90}{\hspace{0.8cm} Windows} & 
    \includegraphics[width=\mywidth]{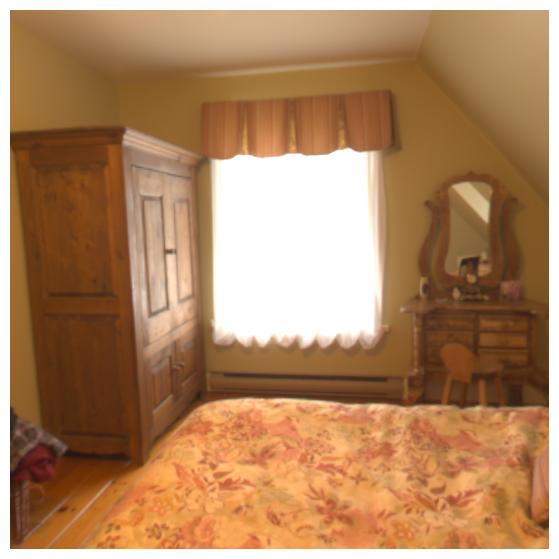}&
    \includegraphics[width=\mywidth]{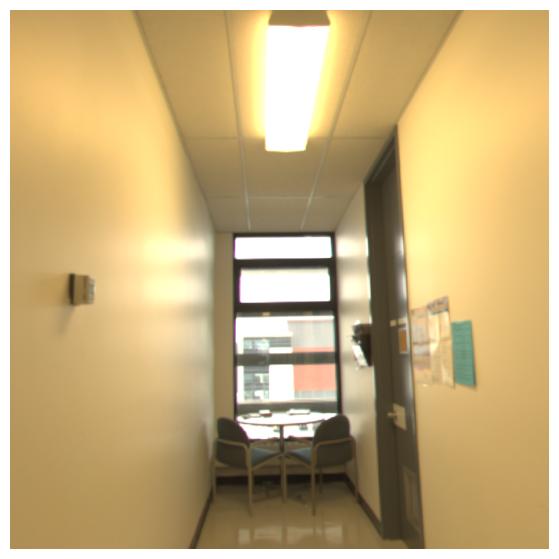}&
    \includegraphics[width=\mywidth]{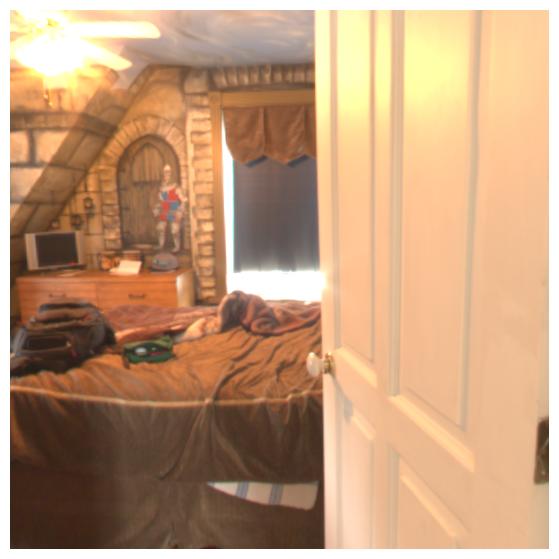}&
    \includegraphics[width=\mywidth]{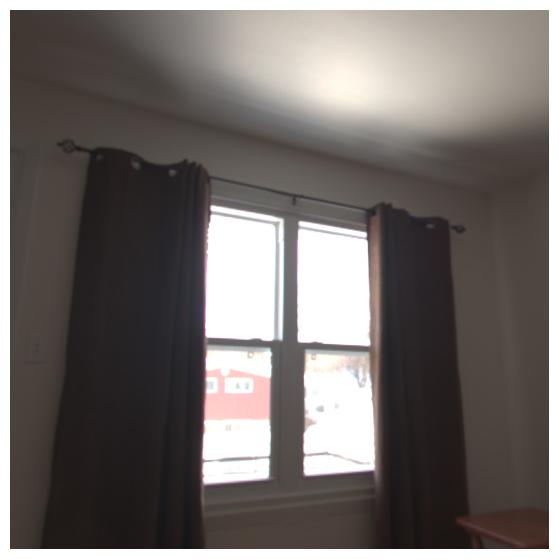}&
    \includegraphics[width=\mywidth]{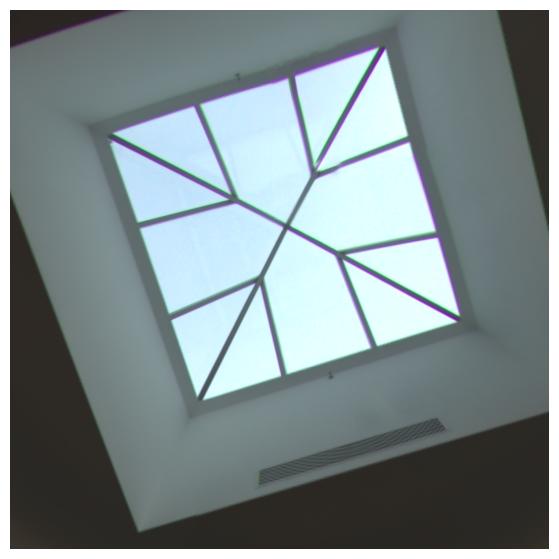}&
    & 
    \\*[1em]
    \end{tabular}
    \caption{From top to bottom: examples of light sources labeled as tubes, bulbs, and windows (left), and correlation between the CCT and the average luminance of the light sources (right).  Each light source is centered in the frame, with its average luminance/CCT indicated above. The mean/median values of each category are: ``window''  \valeur{\SI{4946}{\K}}/\valeur{\SI{4626}{\K}}; ``tubes'' \valeur{\SI{3293}{\K}}/\valeur{\SI{3290}{\K}}; ``bulb''  \valeur{\SI{3506}{\K}}/\valeur{\SI{3294}{\K}}. Images are reexposed and tonemapped ($\gamma = \valeur{\num{2.2}}$) for display.} 
    \label{fig:categorie_sources_examples_lum_moyenne}
\end{figure*}

A total of \valeur{406} randomly selected light sources were manually labeled in order to study the physical properties and diversity of three types of light sources differentiated by their visual appearance: elongated ``tubes'' (\valeur{\num{190}} samples), point-type ``bulbs'' (\valeur{\num{105}} samples), and large ``windows'' (\valeur{\num{111}} samples), including skylights and windows with curtains. 
\Cref{fig:categorie_sources_examples_lum_moyenne} shows examples of detected sources and the correlation between their CCT and average luminance values for each category.

Of all three categories, ``tubes'' have the most compact distributions, with standard deviations of \valeur{\SI{465}{\K}} and \valeur{\SI{2957}{\candela\per\m\squared}} for CCT and luminance respectively, as they all tend to use the same fluorescent lighting technology.  These types of lights are mostly used in public spaces, thus their properties are expected to be similar.
In comparison, the standard deviations of the CCT and luminance for the sources labeled as ``bulbs'' are \valeur{\SI{1090}{\K}}/\valeur{\SI{95381}{\candela\per\m\squared}} and \valeur{\SI{1395}{\K}}/\valeur{\SI{6821}{\candela\per\m\squared}} for the ``windows''.
As these numbers suggest, the average luminance of the ``bulbs'' varies considerably more than the ``tube'' category, as their area (size) and purpose differ and can be found both in public and private spaces.
The ``bulb'' category also shows a wide diversity in CCT, as it contains different types of lighting technologies (tungsten, compact fluorescent, or LED) that are difficult to separate visually.  They are the category with the highest average luminance, with mean/median values of \valeur{\num{59701}}/\valeur{\SI{26058}{\candela\per\m\squared}}, compared to ``tubes'' (\valeur{\num{3212}}/\valeur{\SI{2330}{\candela\per\m\squared}}) and ``windows'' (\valeur{\num{6087}}/\valeur{\SI{3032}{\candela\per\m\squared}}), as they are small but very bright light sources that are usually scattered by diffusers or reflectors. 

The ``windows'' category is the most diverse group of light sources, especially regarding the CCT, as the spectral properties of the incoming light depends on multiple factors, such as the time of day, weather, geographical position and orientation of the room, the scene outside the window, human-made light modifiers (curtains and blinds), etc.  That category also contains the hottest light sources. 
The total intensity of the light also varies greatly, as panoramas were taken during the day and at night, during sunny and cloudy days. 
The first and fourth image of the bottom row of \cref{fig:categorie_sources_examples_lum_moyenne} show the impact of curtains on the average luminance of the light penetrating in the scene, compared to the windows (other examples on the third row) without curtains.  Human-made light modifiers also have an impact on the ``bulb'' category due to the lampshade and type of fixture used, reducing the average luminance perceived, as it can be seen in the first and fifth examples of the second row of \cref{fig:categorie_sources_examples_lum_moyenne}.
The skylight shown in the fifth example of the third row of \cref{fig:categorie_sources_examples_lum_moyenne} has a very high CCT, due to the blue color of the sky, compared to the other lateral windows, indicating that the orientation in space of the window (and the view accessible from it) has an impact on the spectral properties of the light in the scene.  

\section{Learning to predict photometric values}
\label{sec:learning}


Our main goal is to develop algorithms that perform physically accurate lighting predictions from real-world photographs captured ``in the wild.'' It is our hope that the proposed Laval Photometric Indoor HDR Dataset helps the community make strides towards this goal. Here, we introduce new tasks that are enabled by our dataset, and analyze the conditions necessary for accurate light prediction. 

\subsection{Prediction tasks}
\label{sec:tasks}

We present three novel learning tasks that are enabled by our dataset. Given a single image as input, each task aims to predict the following values.
\begin{enumerate}[nosep,wide]
\item \textbf{Per-pixel luminance}: we wish to recover the luminance (in \unit{\candela\per\meter\squared}) at each pixel in the input. For clarity, losses are attributed independently to two subtasks: extrapolating HDR values from LDR inputs (similar to \cite{liu2020single}, see \cref{sec:related_work}); and predicting the scalar exposure to appropriately scale the HDR values to luminance. Here, we wonder how different degradations (e.g. noise, quantization, tonemapping) on the input affect the prediction.
\item \textbf{Per-pixel color}: we wish to estimate the color at each pixel in the input by predicting its CCT. We augment the white balance (WB) of the input using~\cite{Afifi2019} so that the colorimetry of the scene is unknown and wish to see if the network is able to correctly identify the true CCT, as well as predicting the CCT for saturated pixels.
\item \textbf{Planar illuminance}: we wish to predict the (scalar) planar illuminance. This is computed using \cref{eq:illuminance} from a \ang{180} photometric HDR image, but can it also be done from a narrower field of view (FOV)? We also explore the impact of the information provided in the input: can a single LDR image, at arbitrary exposure, be sufficient? Is HDR necessary, or alternatively, is the ground truth exposure needed? 
\end{enumerate}
The input to each of these tasks are adapted to measure if a deep learning architecture can understand the photometry of a scene from limited data.

\begin{figure}
    \includegraphics[width=\linewidth]{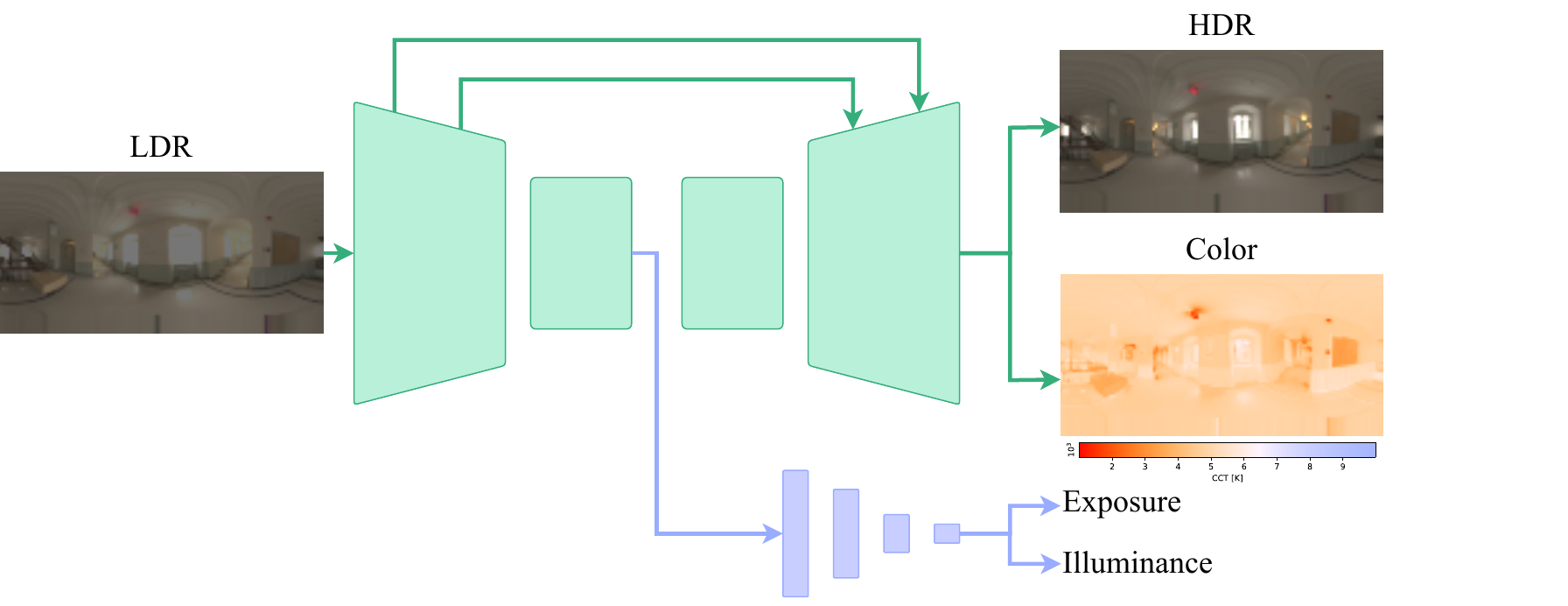}
    \caption{Architecture used for learning tasks. The learning architecture is based on a Unet with fixup initialization (green), with an added subnetwork for scalar prediction tasks (blue). The input is an LDR image of varying degradations and field of view (FOV). The outputs depend on the task: per-pixel luminance prediction outputs an HDR image and its (scalar) exposure; per-pixel color prediction outputs a CCT map; and illuminance prediction outputs an illuminance scalar only.}
    \label{fig:learning_pipeline}
\end{figure}

\subsection{Learning architecture and per-task data}

\Cref{fig:learning_pipeline} summarizes the architecture used for the experiments. A UNet~\cite{ronneberger2015unet} with fixup initialization~\cite{zhang2019fixup} (implementation from \cite{griffiths2022outcast}) takes an image as input and outputs an image of the same resolution. It consists of 5 down/up sampling with skip connections, with 6 residual blocks~\cite{he2016resnet} at each level and 6 bottleneck layers. The decoder part of the UNet is used for pixel prediction tasks. Additionally, a subnetwork is added at the center of the bottleneck layers of the UNet, consisting of a 4-layer MLP, and outputting a single scalar. This subnetwork is used for scalar prediction tasks. Every inner layer uses ReLU activation, and the last layers of the decoder and subnetwork have tanh activation function. The outputs are normalized in logarithmic scale accordingly. Each network is trained with the Adam optimizer with learning rates of $\num{e-6}$ for luminance and illuminance tasks, and of $\num{e-5}$ for color.

The architecture is adapted to each task. For the per-pixel luminance task, an HDR image in the same exposure as the given input is predicted. The subnetwork predicts the exposure needed to scale the predicted HDR to absolute luminance. Here, different degradations are applied to the input: clipping, reexposing, gamma tonemapping, 8 bits quantization and additive gaussian noise. For the per-pixel color task, the subnetwork is omitted and the decoder outputs a CCT map from a WB-augmented input. For the planar illuminance task, the subnetwork outputs the illuminance and the decoder is discarded. Since the planar illuminance is not defined for \ang{360}, the input is a hemisphere, in equirectangular projection or rectangular projection with a given FOV. The photometric dataset is randomly split \SI{80}{\percent}-\SI{10}{\percent}-\SI{10}{\percent} for train-val-test respectively for all experiments below. For computational efficiency, all HDR panoramas are rescaled to \valeur{$64 \times 128$} resolution with an energy preserving scaling function, except for illuminance prediction where we extract perspective projection at \valeur{$160 \times 120$}.

\begin{table}
\centering
\footnotesize
\begin{tabular}{lccc}
\toprule
 Input      & RMSE$_\downarrow$ & siRMSE$_\downarrow$ & HV3$_\uparrow$ \\
 \midrule
 Linear     & \num{116.2} & \num{83.4} & \num{96.4} \\
 Gamma      & \num{121.9} & \num{84.1} & \num{96.4} \\
 Quantized  & \num{125.4} & \num{83.8} & \num{95.9} \\
 Noise      & \num{116.3} & \num{84.2} & \num{96.6} \\
 All        & \num{121.0} & \num{84.7} & \num{96.0} \\
 \bottomrule
\end{tabular}
\caption{The effect of input degradations on the prediction of the per-pixel luminance. The rows indicate the degradation applied to the input of the network. Each input is reexposed and clipped in the range [0,1], and different transformations are applied: none ("Linear"), tonemapping (with $\gamma=2.2$) ("Gamma"), 8-bit quantization ("Quantized"), additive gaussian noise with variance drawn uniformly in the $[0,0.03]$ interval ("Noise") and all 3 degradations compounded ("All").``HV3'' is the HDR-VDP-3 metric~\cite{mantiuk2011hdrvdp}.}
\label{tab:degradations}
\end{table}

\begin{table}
\centering
\footnotesize
\setlength{\tabcolsep}{1pt}
\begin{tabular}{lcccccccc}
\toprule
 & \multicolumn{2}{c}{\ang{180}} & \multicolumn{2}{c}{\ang{120}} & \multicolumn{2}{c}{\ang{60}} & \multicolumn{2}{c}{\ang{60}--\ang{120}}  \\
 \midrule
 Input     & RMSE$_\downarrow$ & $R^2$$_\uparrow$ & RMSE$_\downarrow$ & $R^2$$_\uparrow$ & RMSE$_\downarrow$ & $R^2$$_\uparrow$ & RMSE$_\downarrow$ & $R^2$$_\uparrow$ \\
 \midrule
 HDR       & \num{22.0}     & \num{0.969} & \num{20.6} & \num{0.969} & \num{51.8} & \num{0.861} & \num{49.7} & \num{0.839} \\
 LDR+scale & \num{47.8}    & \num{0.830} & \num{50.9} & \num{0.834} & \num{57.4} & \num{0.819} & \num{52.8} & \num{0.876} \\
 LDR       & \num{123.7}   & \num{0.406} & \num{124.1} & \num{0.374} & \num{122.0} & \num{0.402} & \num{126.5} & \num{0.423} \\
 \bottomrule
\end{tabular}
\caption{Illuminance prediction at different FOV with different levels of information as input. The network is trained and evaluated on: the photometric HDR ("HDR"), the reexposed photometric HDR clipped in the range $[0,1]$ with ("LDR+scale") and without ("LDR") knowledge of the exposure. Additionally, the network is trained on multiple FOVs: full hemispherical \ang{180} (in equirectangular projection), \ang{120}, \ang{60}, and randomly varying in the $[60^\circ, 120^\circ]$ interval (using perspective projection).}
\label{tab:illuminance}
\end{table}

\subsection{Experimental results}

\paragraph{Per-pixel luminance} Here, the input is reexposed so that its \num{90}th percentile corresponds to \num{0.8} and is clipped in the $[0,1]$ range. We then explore the effect of degrading the input on the prediction. We experiment by tonemapping the input ($\gamma=2.2$), by quantizing it to 8 bits, by adding gaussian noise (with variance drawn uniformly in the $[0,0.03]$ interval), and by combining all three degradations. Here, the decoder predicts an HDR image in the same scale as the input LDR, and the subnetwork predicts the scalar which multiplies the HDR to obtain the luminance map. The results in \cref{tab:degradations} show the RMSE and its scale-invariant version (siRMSE)~\cite{barron2014shape} (both weighted by solid angles), and HDR-VDP-3~\cite{mantiuk2011hdrvdp} to measure the visual quality of the prediction. We observe that the network is robust to the degradations.

\paragraph{Per-pixel color} Here, we explore the capacity to predict the per-pixel CCT, having as input a LDR image with random WB augmentation. The HDR input is first reexposed so that its 90th percentile maps to 0.8, clipped to $[0, 1]$, and the WB is augmented to a random preset using~\cite{Afifi2019}. The predictions are scored using RMSE as well as the relative error between the prediction and the ground truth. \Cref{fig:learning_temperature} shows qualitative results of color prediction. The mean relative error and RMSE are \SI{4.25}{\percent} and \num{173.0} on the entire test set. We observe that the network struggles with larger color variations across the image. However, CCT is accurately predicted despite color changes in the input.

\iftrue
\begin{figure*}
   \centering
   \footnotesize
   \setlength{\tabcolsep}{0.5pt} 
   \setlength{\tmplength}{0.155\linewidth}
    \begin{tabular}{cccccccc}
    & 
    1st: \num{48.9} (\SI{1.56}{\percent}) & 
    20th: \num{115.3} (\SI{3.29}{\percent}) & 
    40th: \num{166.6} (\SI{10.1}{\percent}) & 
    60th: \num{235.7} (\SI{6.29}{\percent}) & 
    80th: \num{376.2} (\SI{5.66}{\percent}) & 
    99th: \num{1286.9} (\SI{19.2}{\percent}) & 
    \multirow{3}{*}{\includegraphics[trim={{.01\linewidth} 0 0 {-.0\linewidth}},clip, height=0.186\linewidth]{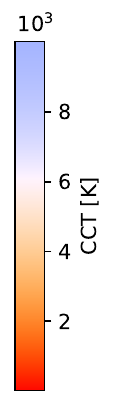}} \\
    \rotatebox{90}{\hspace{.3cm} Input} & \includegraphics[width=\tmplength]{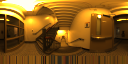}&
    \includegraphics[width=\tmplength]{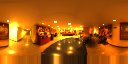}&
    \includegraphics[width=\tmplength]{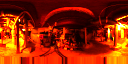}&
    \includegraphics[width=\tmplength]{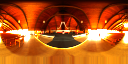}&
    \includegraphics[width=\tmplength]{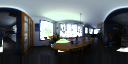}&
    \includegraphics[width=\tmplength]{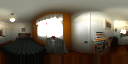}\\
    \rotatebox{90}{\hspace{.4cm} GT} & \includegraphics[width=\tmplength]{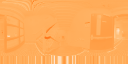}&
    \includegraphics[width=\tmplength]{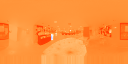}&
    \includegraphics[width=\tmplength]{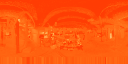}&
    \includegraphics[width=\tmplength]{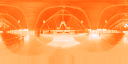}&
    \includegraphics[width=\tmplength]{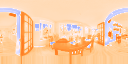}&
    \includegraphics[width=\tmplength]{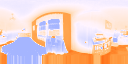} &  \multirow{3}{*}{\includegraphics[trim={{-.003\linewidth} 0 0 {-.025\linewidth}},clip, height=0.1865\linewidth]{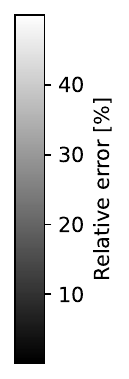}}\\
    \rotatebox{90}{\hspace{.025cm} Prediction} & \includegraphics[width=\tmplength]{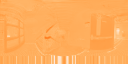}&
    \includegraphics[width=\tmplength]{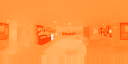}&
    \includegraphics[width=\tmplength]{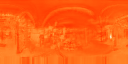}&
    \includegraphics[width=\tmplength]{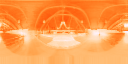}&
    \includegraphics[width=\tmplength]{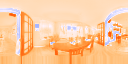}&
    \includegraphics[width=\tmplength]{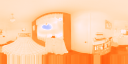}\\
    \rotatebox{90}{\hspace{.015cm} Rel $\varepsilon$ map} & \includegraphics[width=\tmplength]{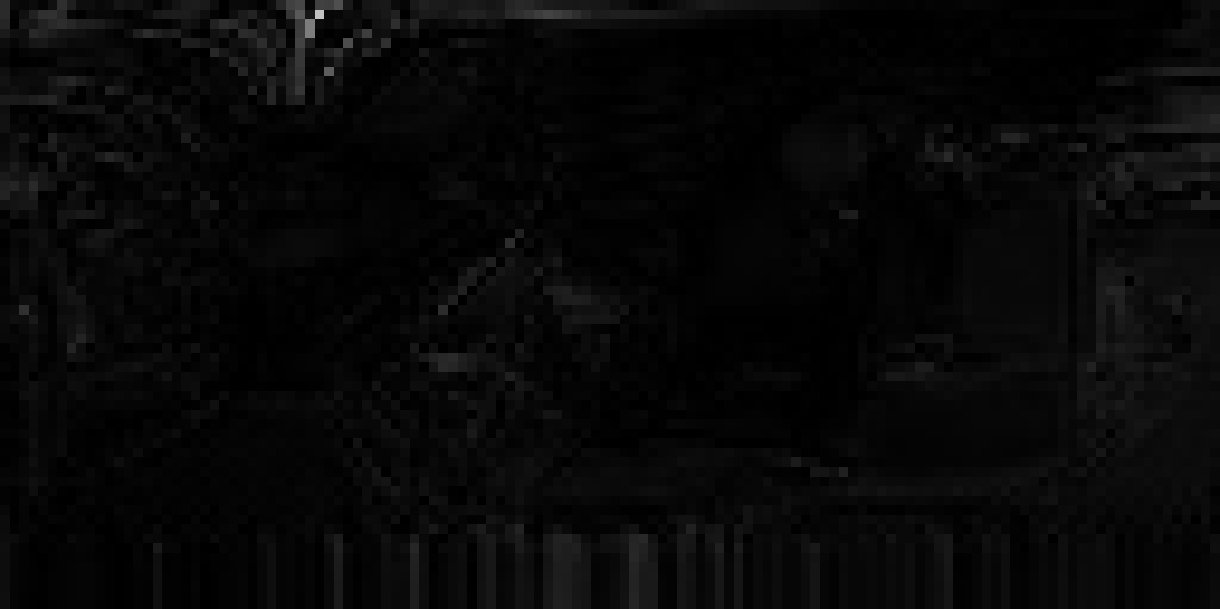}&
    \includegraphics[width=\tmplength]{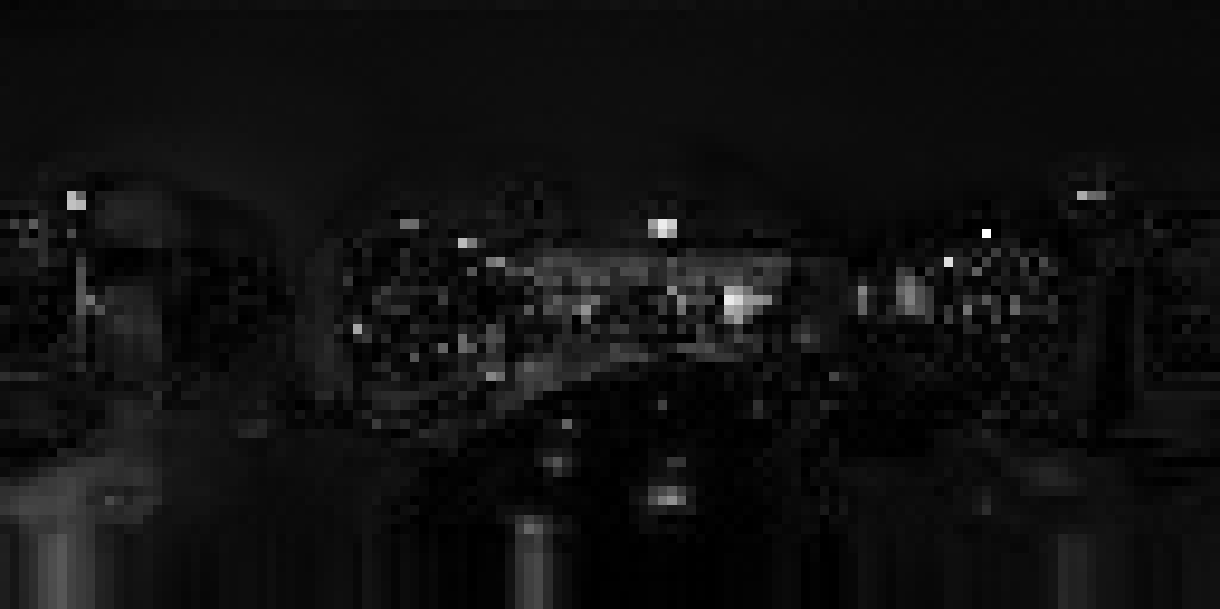}&
    \includegraphics[width=\tmplength]{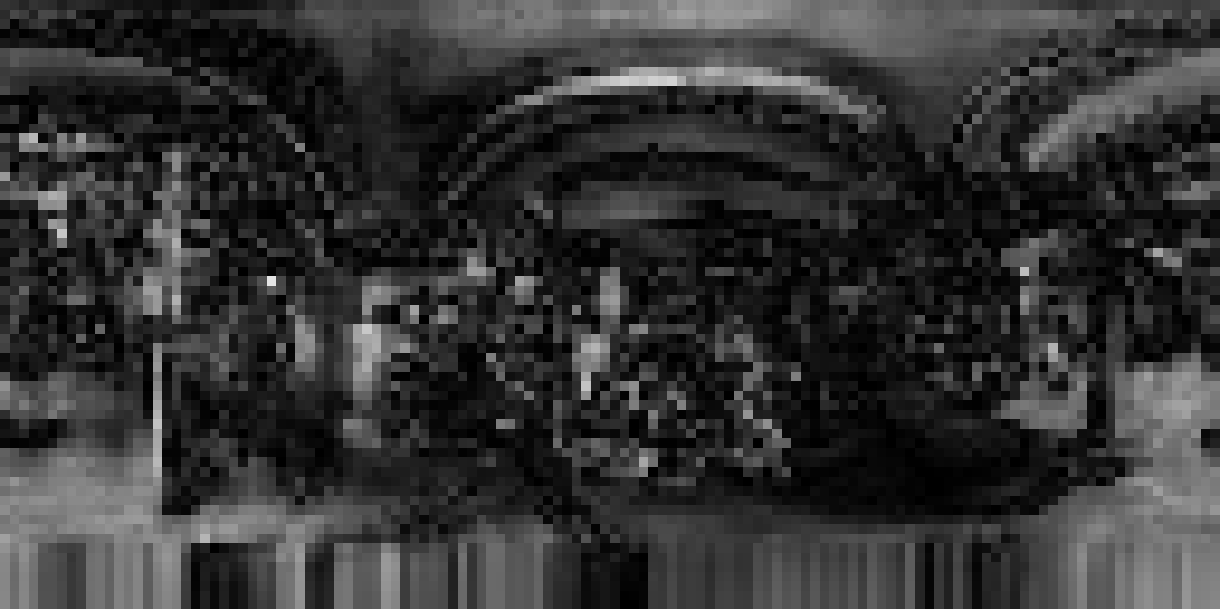}&
    \includegraphics[width=\tmplength]{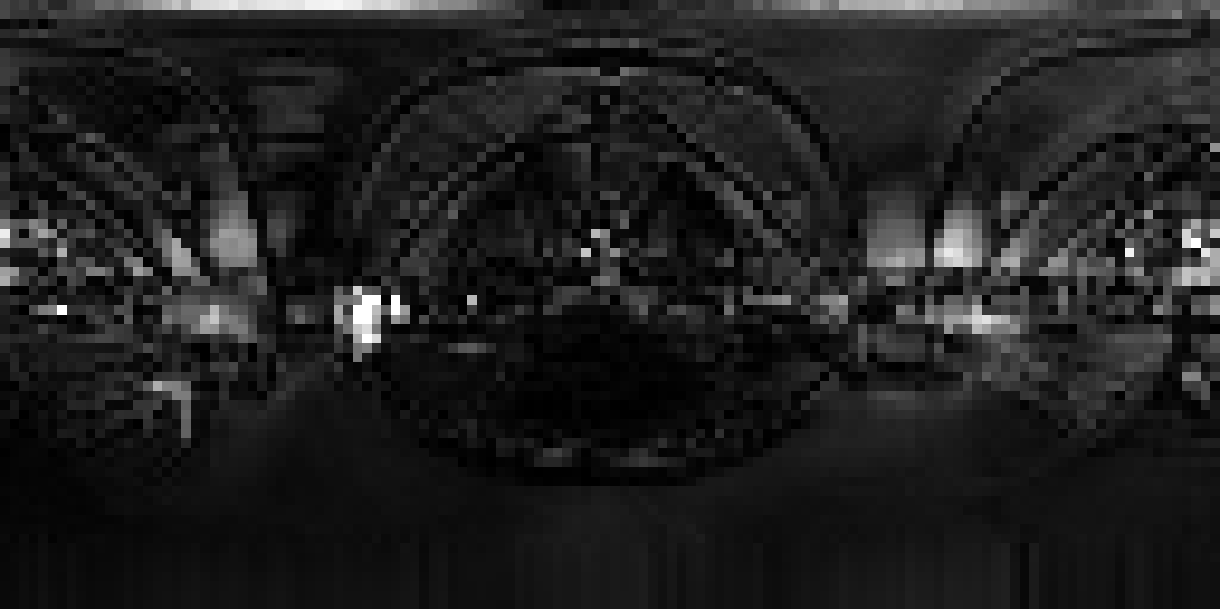}&
    \includegraphics[width=\tmplength]{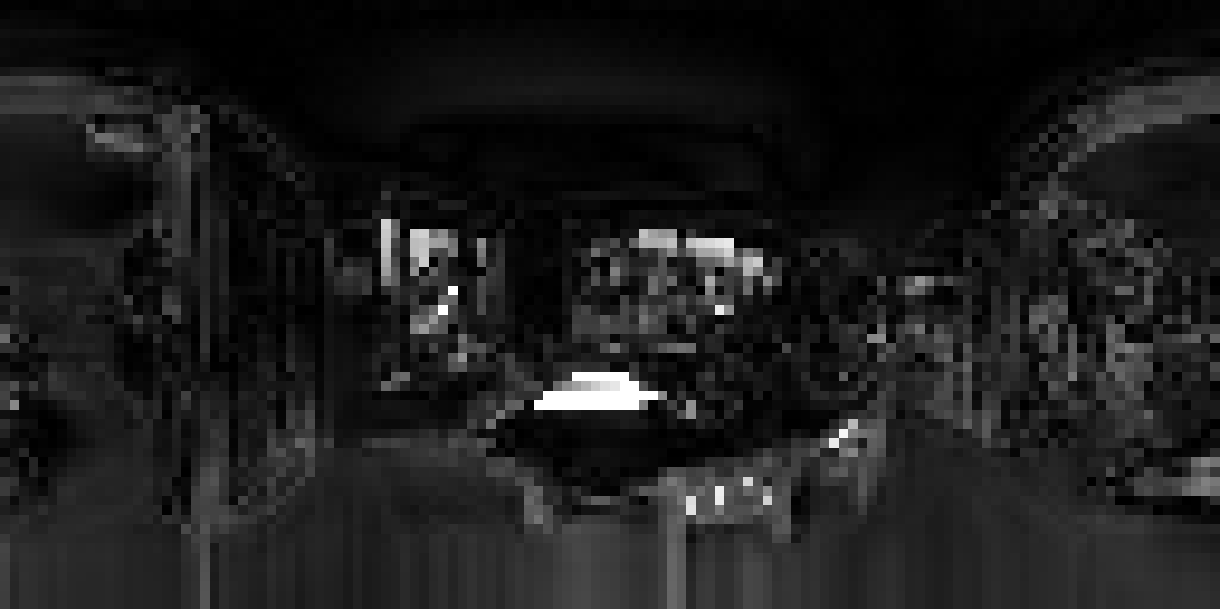}&
    \includegraphics[width=\tmplength]{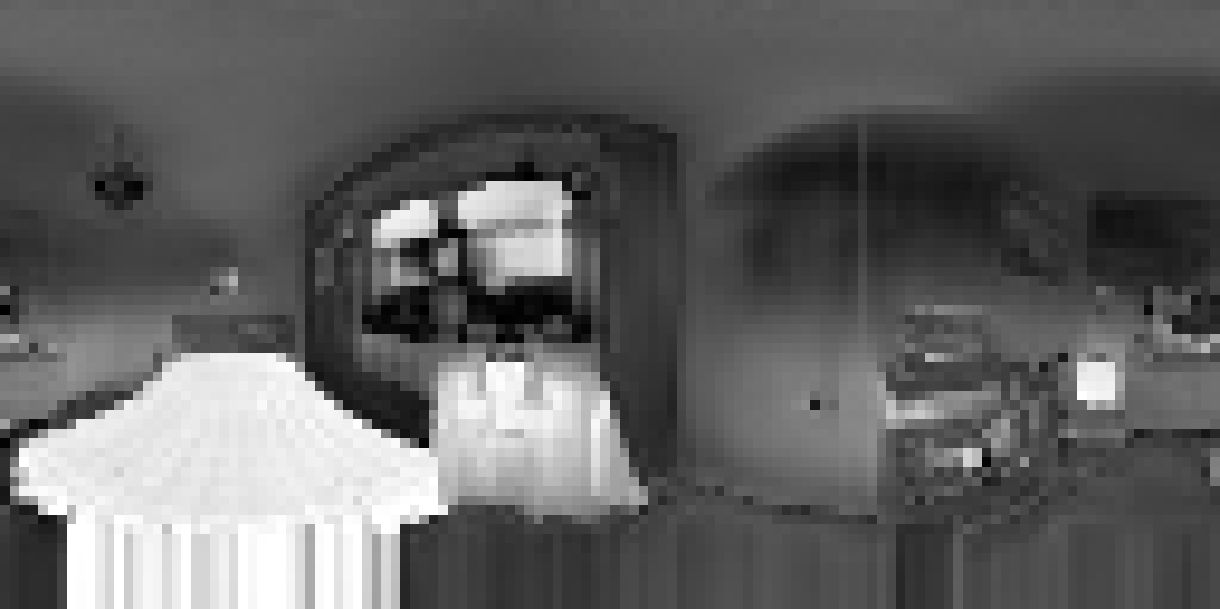}\\
    \end{tabular}
    \caption{Examples of color prediction. The first row indicates the RMSE percentile: the RMSE (relative error). The ``input'' is the calibrated HDR reexposed, clipped, with a random WB augmenter~\cite{Afifi2019}. Other rows show the ground truth and predicted CCT maps, and the relative error map. Colormaps for both the CCT and the relative error are shown at the right.}
    \label{fig:learning_temperature}
\end{figure*}
\fi

\paragraph{Planar illuminance} Here, we experiment with three types of inputs: a photometric HDR image ("HDR"), a linear LDR image (reexposed HDR clipped to the $[0,1]$ interval) with ("LDR+scale") and without ("LDR") knowledge of the exposure. In addition, we also evaluate the impact of the FOV of the input. We experiment with FOVs of \ang{180}, \ang{120}, \ang{60}, and uniformly random in the $[60, 120]^\circ$ interval. The image is stored in an equirectangular representation for \ang{180}, and perspective projection for the other, lower FOVs. 

\Cref{tab:illuminance} shows the results of these series of experiments. We report the RMSE and $R^2$ for each combination of input type and FOV. First, observe that the experiment with a FOV of \ang{180} with the HDR image (top-left in \cref{tab:illuminance}) amounts to learning the illuminance integration (\cref{eq:illuminance}). Unsurprisingly, narrowing the FOV results in decreased performance, due to the hidden lights beyond the FOV which may directly affect the planar illuminance.
Limiting the photometric input information by clipping the HDR but keeping the correct scale (LDR+scale), reduces the scores moderately. The network now has to predict the amount of light beyond the clipped pixels.
Discarding the scale from the input significantly worsens the results, indicating that full dynamic range (HDR) and/or knowing the absolute exposure are both necessary conditions for accurate illuminance inference.

\section{Generalization to another camera}
\label{sec:generalization}

We now present experiments to evaluate the usefulness of our proposed dataset for predicting luminance and color values from images captured with another camera. To this end, we rely on the Ricoh Theta Z1, an off-the-shelf \ang{360} camera, and captured a small dataset of 74 calibrated HDR panoramas (referred to as ``Theta dataset'') using the same process as in \cref{sec:dataset}. In addition to the bracketed RAW images, we also capture well-exposed LDR images (in jpeg format) produced by the camera. This small experimental dataset will also be released publicly. As opposed to \cite{macpherson2022360} which calibrated this camera for photometric measurements, we calibrate for color as well as luminance.

\paragraph{Architecture and data}
The network architecture for each task is kept the same as in \cref{sec:learning}. The networks are first trained on the calibrated Laval dataset with synthetically degraded LDR inputs (all degradations). The Theta dataset is split 40\%-10\%-50\% for train-val-test respectively. The pretrained networks are then fine-tuned on the Theta dataset (with jpeg images as input) with the same learning rate.

\paragraph{Experimental results}
\Cref{tab:theta} shows the results of the experiments for each task (cf. \cref{sec:tasks}). We experiment with a degraded LDR as input to the models pretrained on our dataset and the jpeg image of the camera as input to the pretrained and fine-tuned models. The input images have \ang{120} FOV for the planar illuminance prediction task, and all jpeg images are captured with the same white balance setting.

Directly providing the jpeg images as inputs to the pretrained model (2nd row in \cref{tab:theta}) results in significantly degraded performance across most tasks as compared to using degraded LDR images. This shows that the domain gap between produced jpeg images and simulated LDR images is still wide. Fine-tuning the networks on jpeg inputs is therefore necessary to obtain performance similar or sometimes slightly better than those obtained on the synthetic LDR.

\begin{table}
\centering
\footnotesize
\setlength{\tabcolsep}{2.4pt}
\begin{tabular}{lccccccc}
\toprule
& \multicolumn{3}{c}{Luminance} & \multicolumn{2}{c}{Color} & \multicolumn{2}{c}{Illuminance}  \\
\midrule
Input& RMSE$_\downarrow$ & siRMSE$_\downarrow$ & HV3$_\uparrow$ & RMSE$_\downarrow$ & rel $\varepsilon$$_\downarrow$ & RMSE$_\downarrow$ & $R^2$$_\uparrow$\\
\midrule
PT LDR   & \num{130.6} & \num{86.9} & \num{95.5} & \num{334.0} & \num{11.97} & \num{153.2} & \num{0.469} \\
PT Jpeg   & \num{170.0} & \num{100.6} & \num{90.3} & \num{676.4} &  \num{25.06} & \num{141.9} & \num{0.314}\\
FT Jpeg   & \num{156.7} & \num{96.4} & \num{91.4} & \num{177.9} & \num{5.52} & \num{143.3} & \num{0.385}\\
\bottomrule
\end{tabular}
\caption[]{Domain adaptation on a real-world dataset for which a jpeg image of a scene,  as well as the calibrated luminance map, are captured with a Ricoh Theta Z1. Here, ``HV3'' refers to the HDR-VDP-3 metric~\cite{mantiuk2011hdrvdp}. We report performance on all three tasks from \cref{sec:tasks}. Each row corresponds to degraded LDR and jpeg as input to the pretrained model (``PT LDR'' and ``PT Jpeg'' resp.), and jpeg as input to the fine-tuned model (``FT Jpeg'').}
\label{tab:theta}
\end{table}
\section{Conclusion}
\label{sec:conc}

We present the Laval Photometric Indoor HDR Dataset, the first photometrically accurate, large-scale dataset of HDR panoramic images. Our calibration method relies on a carefully curated calibration dataset of RAW exposure brackets captured with the original camera and a chroma meter. We also capture another small calibrated dataset with a Ricoh Theta Z1 for experiments on jpeg inputs. We present baselines for three novel tasks: per-pixel luminance, per-pixel color and planar illuminance predictions. We hope this new dataset will spur and catalyze research by empowering others to explore novel photometric and colorimetric tasks in computer vision, such as white balance prediction under multiple illuminations, physically based inverse rendering and ''in the wild'' image relighting.

%
{\small
\noindent \textbf{Acknowledgements} \quad
This research was supported by Sentinel North, NSERC grant RGPIN 2020-04799, and the Digital Research Alliance Canada. The authors thank Mojtaba Parsaee and Anthony Gagnon for their help with the chroma meter and the Theta Z1 calibration.
}

{\small
\bibliographystyle{ieee_fullname}
\bibliography{bibliography}
}




\title{Beyond the Pixel: a Photometrically Calibrated HDR Dataset \\ for Luminance and Color Prediction \\*[0.5em] Supplementary Materials}

\author{Christophe Bolduc, Justine Giroux, Marc Hébert, Claude Demers, and Jean-François Lalonde\\
Université Laval\\
}

\maketitle
\ificcvfinal\thispagestyle{empty}\fi

In this document, we present the following additional information to complement the main paper: 

\begin{itemize}[noitemsep]
\item A description of how to acquire the dataset;
\item A description of the photometric and colorimetric quantities in \cref{sec:photometric} to accompany sec. 3 to sec. 6 of the paper;
\item Additional information on the calibration, including more details on the capture configurations and the calibration uncertainty in \cref{sec:calibration} to complement sec. 3 of the paper;
\item More visualisations to explore the calibrated dataset in \cref{sec:viz} to augment sec. 4 of the paper;
\item More results for the learning tasks, including visual examples in \cref{sec:learning} to add to sec. 5 of the paper;
\end{itemize}

\section{Acquiring the dataset}
\label{sec:examples}
The dataset, released along with the paper, is available at \url{http://www.hdrdb.com/indoor_hdr_photometric/}. Access to the complete dataset for non-profit or educational organization is provided after a license agreement is signed. Additionally, a sample of the photometric dataset is directly available (100 samples at $2048 \times 1024$ pixel resolution). The HDR data, stored in the ``.exr'' file format, can be visualised using an HDR viewer such as TEV\footnote{\url{https://github.com/Tom94/tev}}.

\section{Photometric and colorimetric quantities}
\label{sec:photometric}

\subsection{Illuminance and luminance computations}

\paragraph{Planar illuminance}
The equation used to compute the illuminance on a plan from the luminance of the hemisphere ~\cite{ILV2020} is
\begin{equation} \label{equation:illuminance}
E_{p} = \int_{\Omega} L(p, \omega) \cos(\theta) \dd\omega \,,
\end{equation}
where $L(p, \omega)$ is the luminance of an area (subtended by solid angle $\omega$) of the hemisphere $\Omega$, and $\theta$ is the angle the surface normal of the plane.

\begin{figure}[t]
    \includegraphics[width=\linewidth]{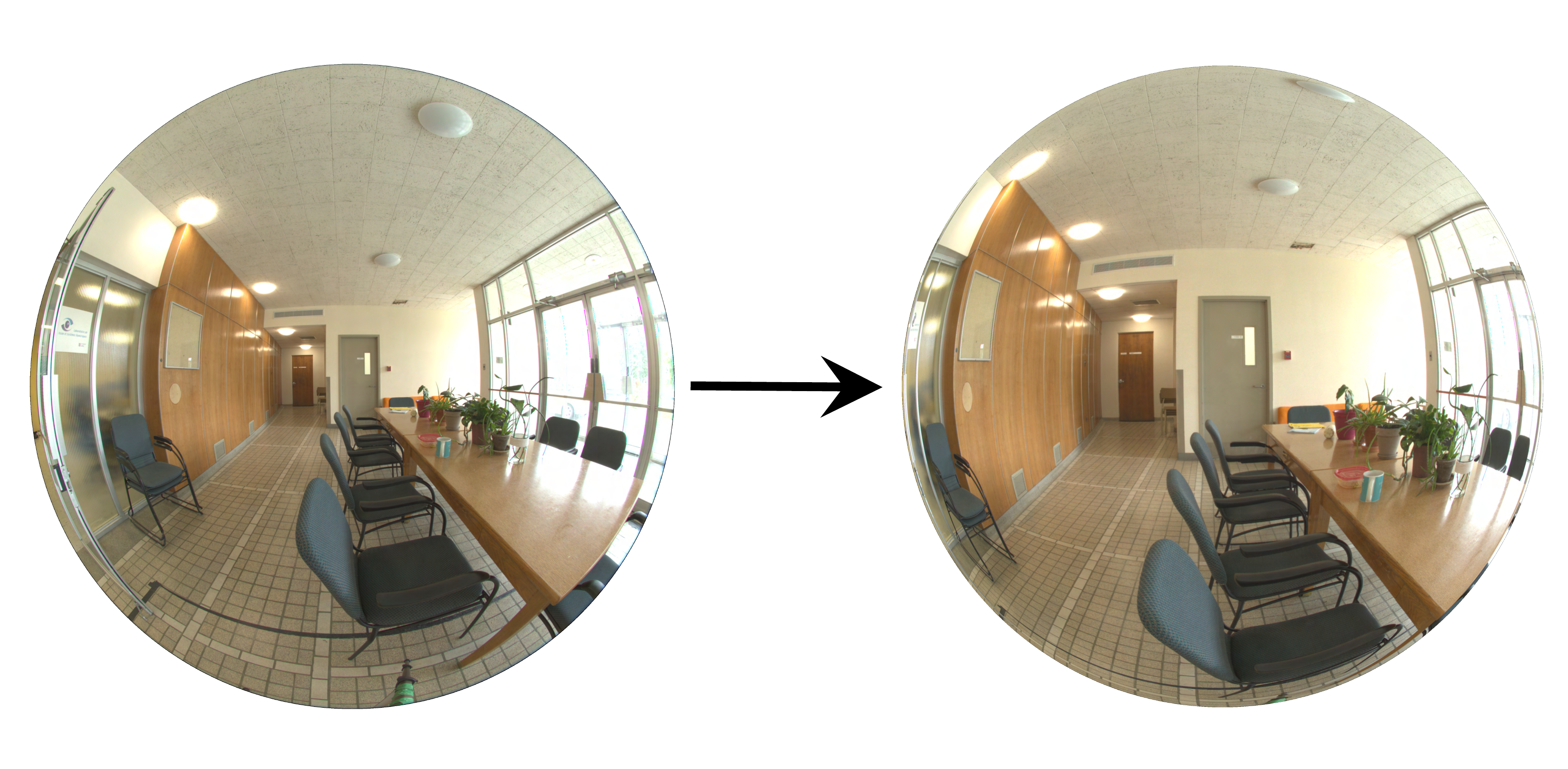}
    \caption{To compute the illuminance of the image, the geometric calibration of the camera is used to project the captured HDR (left) to an orthographic projection (right).}
    \label{fig:calibration_projection}
\end{figure}

When projecting the hemisphere on the plane with an orthographic projection (as is shown in \cref{fig:calibration_projection}), the projected solid angle with relation to the hemispherical solid angle corresponds to
\begin{equation}
\dd\omega_\perp = \cos(\theta) \dd\omega \,.
\end{equation}
\Cref{equation:illuminance} then becomes
\begin{equation}
    E_p = \int_{H} L(p, \omega) \dd\omega_\perp \,.
\end{equation}
Discretizing this equation and integrating on a planar pixel grid of $N$ pixels, the illuminance becomes
\begin{equation}
    E_p = \frac{\pi}{N} \sum_{i \in \Omega} L(i) \,.
    \label{equation:illuminance_disc}
\end{equation}
This is the equation used in sec. 3.1 of the main paper for the dataset calibration as well as sec. 5 and sec. 6 for computing the ground truth of illuminance prediction.

\paragraph{Mean spherical illuminance}
The mean spherical illuminance (MSI)~\cite{ILV2020} is used to measure the quantity of light received at a single point in the scene in the analysis in sec.~4.1 and is defined as
\begin{equation}
    E_\mathrm{ms} = \int_S L(p, \omega) \dd{\omega} \,,
    \label{eq:mean_spherical_illuminance}
\end{equation}
where $L(p, \omega)$ is the luminance of an area (subtended by solid angle $\omega$) of the sphere $S$.

Discretizing this equation over a planar pixel grid (in equirectangular format) of $N$ pixels gives
\begin{equation}
    E_\mathrm{ms} = 4 \pi \frac{\sum_{(i) \in S^\prime} L(i) \dd{\omega(i)}}{\sum_{(i) \in S^\prime} \dd{\omega(i)}} \,,
    \label{equation:mean_illuminance_disc}
\end{equation}
where $\dd{\omega(i)}$ is the solid angle subtended by pixel $i$, and $S^\prime$ represents the subset of valid pixels in the image\footnote{In practice, not all pixels are valid in the panoramas and a region at the nadir, corresponding to where the tripod was at the time of capture, is all-black.}.



\paragraph{Average luminance}
For each individual light source analyzed in sec.~4.2, the average luminance is computed as 
\begin{equation}
    \Bar{L} = \frac{\int_{A} L(p, \omega) \dd{\omega}}{\int_{A} \dd{\omega}} \,,
\end{equation}
where $L$ is the luminance of the pixel, $\dd{\omega}$ is its solid angle, and $A$ is the region which corresponds to the segmented light source.  

Its discretized version is defined as
\begin{equation}
\Bar{L} =  \frac{\sum_{(i) \in A} L(i) \dd{\omega(i)}}{\sum_{(i) \in A} \dd{\omega(i)}} \,.
\label{eq:mean_luminance_source}
\end{equation}

\subsection{Photopic values}
The previous photometric quantities are defined independently of any color space. In our work, we apply the planar illuminance (\cref{equation:illuminance_disc}) for the dataset calibration in sec. 3 and mean spherical illuminance (\cref{equation:mean_illuminance_disc}) in sec. 4.1 directly to each of the RGB channel. 

However, we also work with photopic luminance and illuminance, where the equations are applied to the photopic luminance, defined as:
\begin{equation}
     L = 0.212671 L_R + 0.715160 L_G + 0.072169 L_B \,.
     \label{eq:illuminance_photopic}
 \end{equation}
This is the case for the average source luminance (\cref{eq:mean_luminance_source}), luminance vizualisation (sec. 4) and planar illuminance prediction (sec. 5 and sec. 6).

\subsection{Color spaces conversions}

\paragraph{CIE Yxy to CIE XYZ}
The equations allowing the transformation from Yxy to XYZ color spaces \cite{Poynton2012} are
\begin{equation}
X = \frac{xY}{y} , \qq{} Y = Y , \qq{and} Z = \frac{(1-x-y)Y}{y} \,.
\label{eq:xyY2XYZ}
\end{equation}

\paragraph{CIE XYZ to CIE Yxy}
The inverse transformation corresponds to \cite{Poynton2012}
\begin{equation}
x = \frac{X}{X+Y+Z} , \qq{} y = \frac{Y}{X+Y+Z} , \qq{and} Y = Y \,.
\label{eq:XYZ2xyY}
\end{equation}

\paragraph{RGB to CIE XYZ}
The relation between linear sRGB under reference white D65 and XYZ is given by the following matrix multiplication \cite{Lindbloom2017XYZRGB}
\begin{equation}
\mqty[X \\ Y \\ Z] = \mqty[0.4124564 & 0.3575761 & 0.1804375 \\ 0.2126729 & 0.7151522 & 0.0721750 \\ 0.0193339 & 0.1191920 & 0.9503041] \mqty[R \\ G \\ B] .
\label{eq:XYZ2RGB}
\end{equation}

\paragraph{CIE XYZ to RGB}
The inverse transformation of \cref{eq:XYZ2RGB} is approximated as
\begin{equation}
\mqty[R \\ G \\ B] = \mqty[3.2404542 & -1.5371385 & -0.4985314 \\ -0.9692660 & 1.8760108 & 0.0415560 \\ 0.0556434 & -0.2040259 & 1.0572252] \mqty[X \\ Y \\ Z] .
\end{equation}

\paragraph{Chroma meter RGB conversion}
To convert the xyY color value captured by the chroma meter to RGB as is done in sec. 3.4 of the paper, the equations \cref{eq:xyY2XYZ} and \cref{eq:XYZ2RGB} are applied subsequently.

\subsection{Color temperature from photometric HDR}
We use McCamy's approximation to compute the correlated color temperature (CCT) from the chromaticity in CIE $xy$ format \cite{mccamy1992} in secs.~4.1, 5 and 6 of the main paper, defined as 
\begin{equation}
    T = 449 n^3 + 3525 n^2 + 6823.3 n + 5518.87 \,,
\label{eq:xy2CCT}
\end{equation}
where
\begin{equation}
n = \frac{x - 0.3320}{0.1858 - y} \,.
\end{equation}
The CCT is applied per-pixel to the photometric HDR by first using \cref{eq:XYZ2RGB} to convert from RGB to CIE XYZ, then using \cref{eq:XYZ2xyY} to convert to CIE xy and finally using \cref{eq:xy2CCT} to obtain the CCT.

\paragraph{Average CCT}
The average CCT of a source used in sec. 4.2 is computed on the per-pixel CCT image as
\begin{equation}
    \Bar{T} = \frac{\int_{A} T \dd{\omega}}{\int_{A} \dd{\omega}} \,,
    \label{eq:mean_temperature_source}
\end{equation}
and its discretized version
\begin{equation}
\Bar{T} = \frac{\sum_{(i) \in A} T(i) \dd{\omega(i)}}{\sum_{(i) \in A} \dd{\omega(i)}} \,.
\end{equation}

\section{Calibration}
\label{sec:calibration}

\begin{figure}[t]
\centering
    \includegraphics[width=\linewidth]{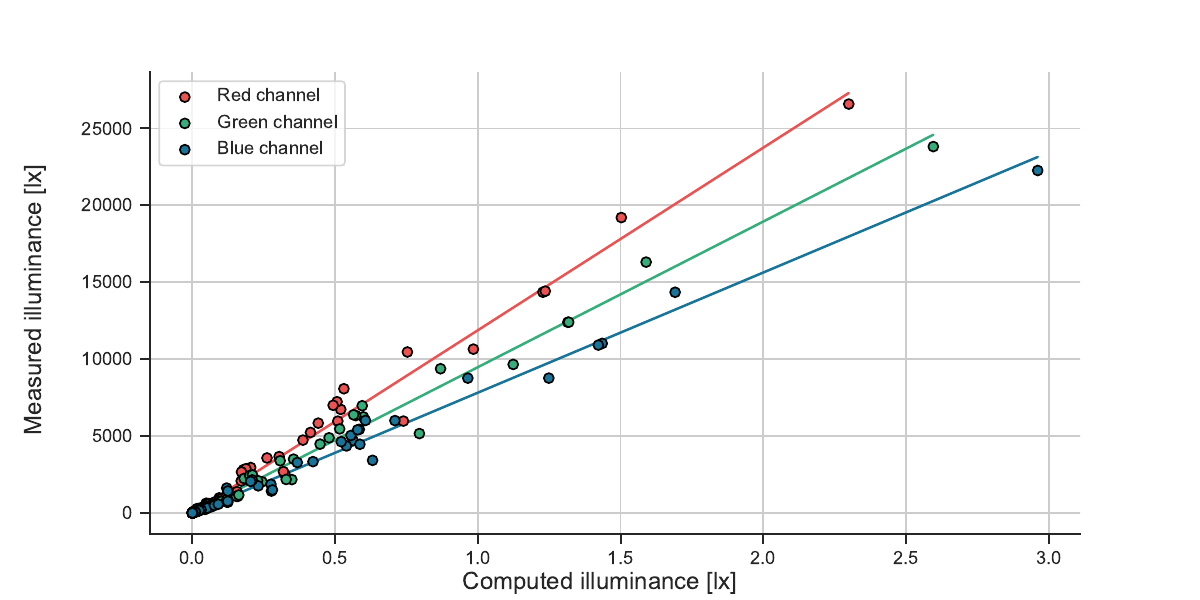} \\
    (a) f/14 \\
    \includegraphics[width=\linewidth]{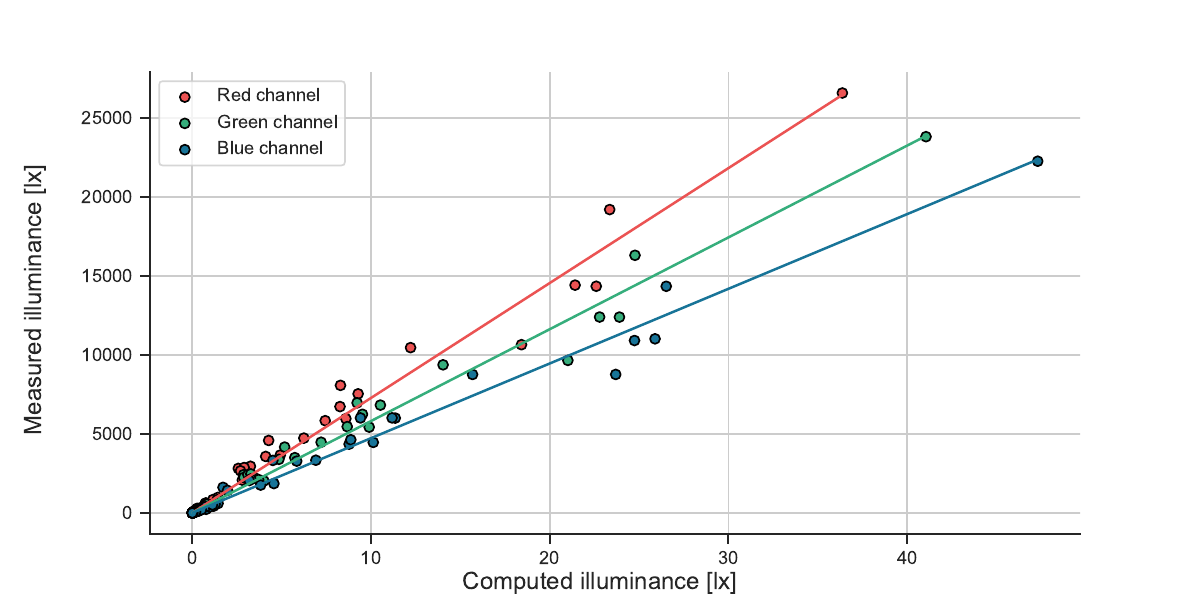} \\
    (b) f/4
    \caption{The resulting regression for the measured illuminance with the chroma meter over the integrated illuminance from the HDR images for the aperture of (a) f/14, and (b) f/4. (a) The resulting correction factors (slopes) are (\num{11872.8}, \num{9472.0}, \num{7814.3}) for (R, G, B), with $R^2$ regression coefficients of determination of (\num{0.985}, \num{0.987}, \num{0.989}) respectively. (b) The resulting correction factors (slopes) are (\num{727.5}, \num{581.3}, \num{472.8}) for (R, G, B), with $R^2$ regression coefficients of determination of (\num{0.982}, \num{0.984}, \num{0.982}) respectively.}
    \label{fig:calibration_regression14}
\end{figure}

\begin{figure}[t]
\centering
    \includegraphics[width=\linewidth]{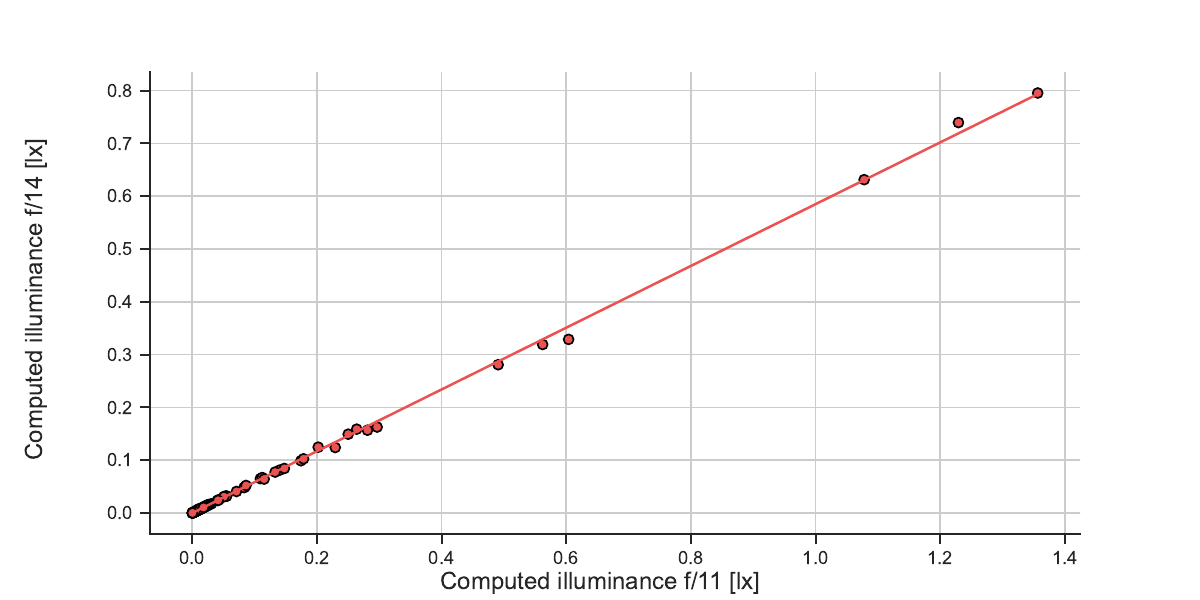} \\
    (a) f/14 over f/11 \\
    \includegraphics[width=\linewidth]{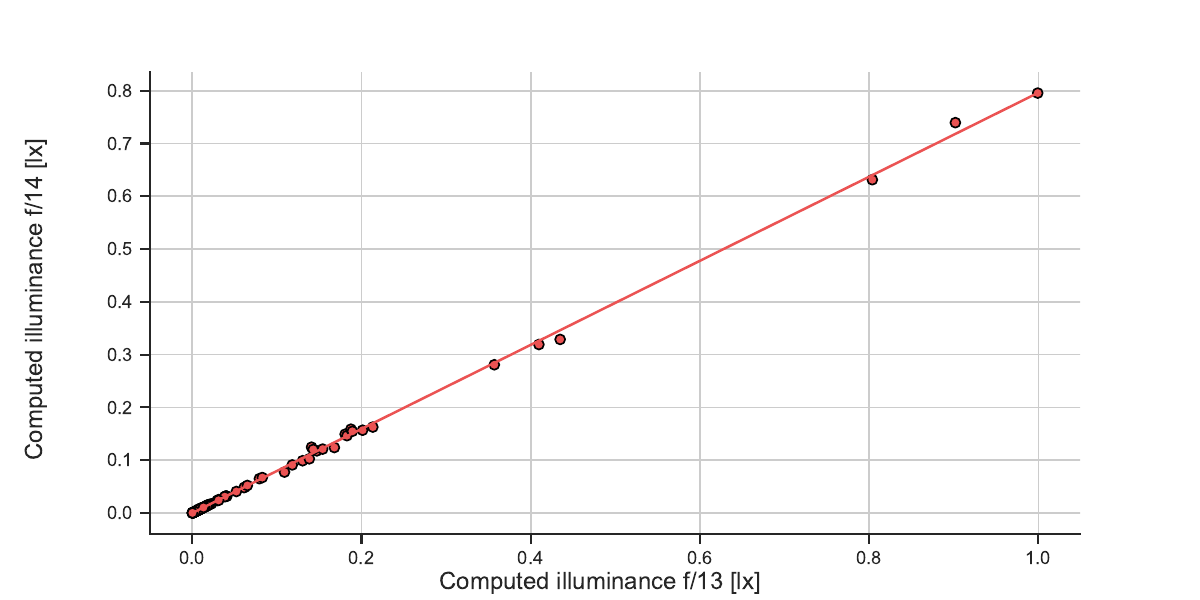} \\
    (b) f/14 over f/13 \\
    \includegraphics[width=\linewidth]{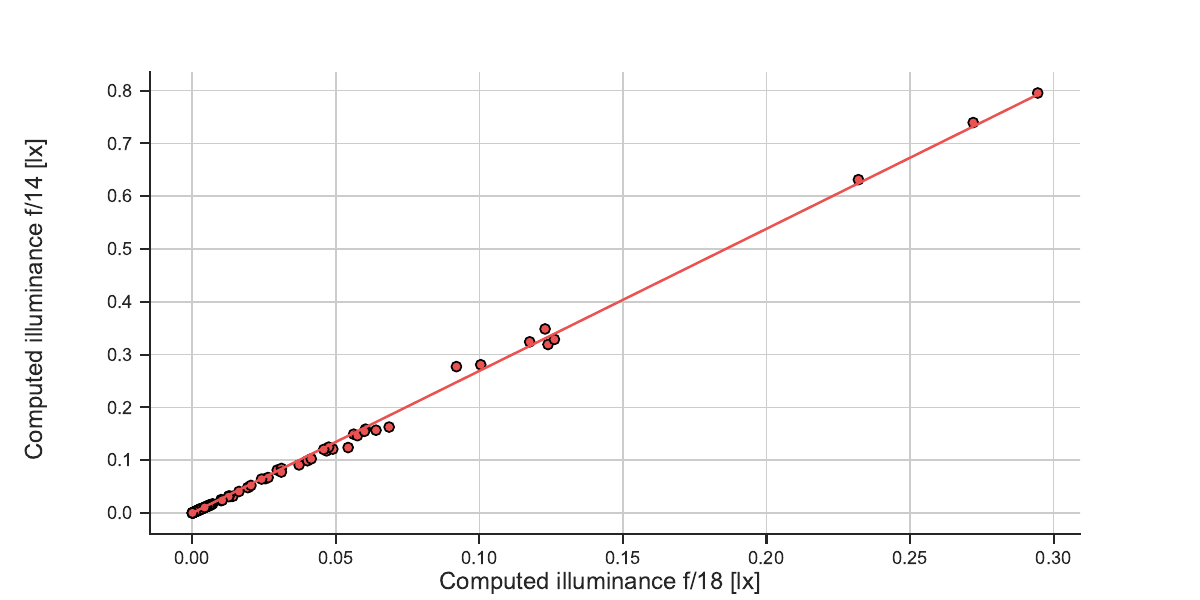} \\
    (c) f/14 over f/18
    \caption{The resulting regression for the integrated illuminance from the HDR images for the aperture of (a) f/14 over the integrated illuminance from the HDR images for the aperture of f/11, (b) f/14 over f/13, and (c) f/14 over f/18. (a) The resulting correction factors (slope) is 0.585, with $R^2$ regression coefficients of determination of 0.999. (b) The resulting correction factors (slope) is 0.796, with $R^2$ regression coefficients of determination of 0.999. (c) The resulting correction factors (slope) is 2.69, with $R^2$ regression coefficients of determination of 0.998.}
    \label{fig:calibration_regression11}
\end{figure}

\subsection{Coefficients regressions}
The calibration coefficients identified in sec. 3.4 of the paper are computed for each capture configuration.
The regressions for each channel of the two main configurations (f/14 and f/4) are shown in \cref{fig:calibration_regression14}. Since the other 3 configurations represent but a small minority (2\%) of the total number of images (\cref{tab:calibration_uncertainty}), they were not captured for all of the 135 calibration dataset scenes. Instead, they were only captured on a subset (43) of the scenes, and directly compared with the f/14 configuration instead of the chroma meter to compute the relationship with the RGB coefficients at aperture f/14. Since the change in aperture affects all three channels simultaneously, a single coefficient is computed from the three channels. The coefficients regression presented in \cref{fig:calibration_regression11} bring the f/11, f/13 and f/18 configurations respectively to their f/14 equivalent.

To calibrate the corresponding panoramas, the HDR is first multiplied by the factor correcting to obtain the f/14 equivalent, and the coefficients for each channel of the f/14 configuration are then applied afterwards.

\subsection{Uncertainty on calibration}
The uncertainty on the calibrated dataset of sec. 3 of the main paper depends on the configuration of the capture for a given panorama. \Cref{tab:calibration_uncertainty} lists the different configurations along with the standard deviation of the linear regression. In all, we achieve very low (less than \SI{1.5}{\percent}) uncertainty in the recovered luminance values across all configurations and three color channels.

\begin{table}
\centering
\footnotesize
\begin{tabular}{ccccccc}
\toprule
 \#panos & Aperture & Shutter speed & R STD & G STD & B STD  \\
  & & [\unit{\second}$\pm$\unit{\ev stop}] & [\%] & [\%] & [\%] \\
 \midrule
 540 & f/4  & 1/30 $\pm$ {2 2/3}  &1.43 &1.35&1.43\\
 7 & f/11 & 1/30 $\pm$ {2 2/3} & 1.24&1.18&1.10\\
 3 & f/13 & 1/30 $\pm$ {2 2/3}&  1.23&1.17&1.08\\
 1759 & f/14 &   1/30 $\pm$ {2 2/3} &1.18&1.12&1.03\\
 53 & f/18 & 1/60 $\pm$ {2 2/3}  &1.27&1.22&1.13\\
 \bottomrule
\end{tabular}
\caption{Uncertainty on the calibration process for each of the 5 capture configurations (aperture and shutter speed) in the dataset. The ISO for each configuration is 100.}
\label{tab:calibration_uncertainty}
\end{table}




\section{Visualisation}
\label{sec:viz}

\begin{figure*}
   \centering
   \scriptsize
   \setlength{\tabcolsep}{1pt} 
   \setlength{\tmplength}{0.15\linewidth}
    \begin{tabular}{cccccc}
    \valeur{Min} (\valeur{\SI{1}{\lux}}) & 
    \valeur{0.01th} (\valeur{\SI{5}{\lux}}) & 
    \valeur{1st} (\valeur{\SI{13}{\lux}}) & 
    \valeur{10th} (\valeur{\SI{65}{\lux}}) & 
    \valeur{12.5th} (\valeur{\SI{81}{\lux}}) & 
    \valeur{15th} (\valeur{\SI{103}{\lux}})  \\
    \includegraphics[width=\tmplength]{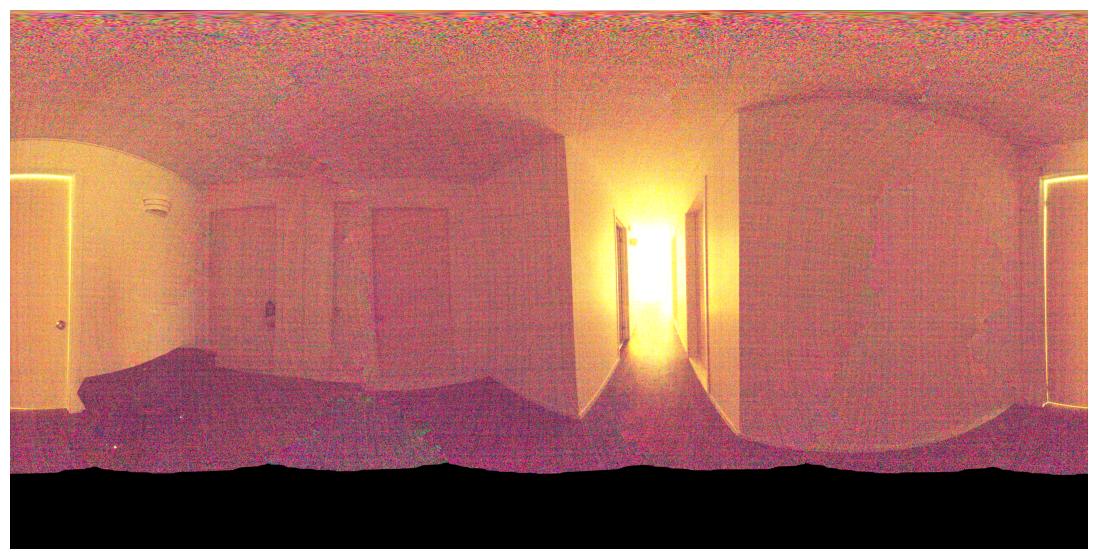}&
    \includegraphics[width=\tmplength]{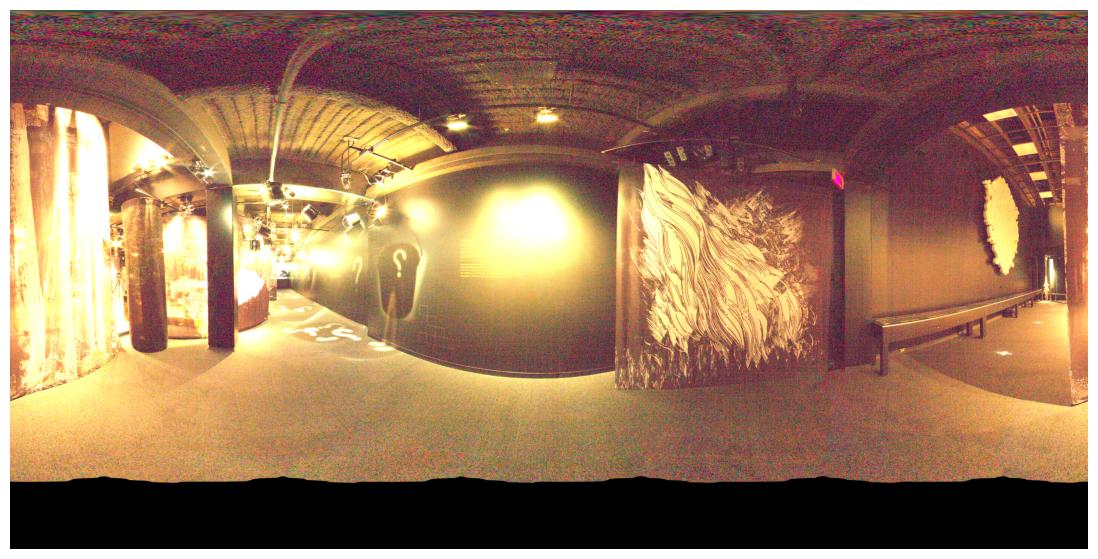}&
    \includegraphics[width=\tmplength]{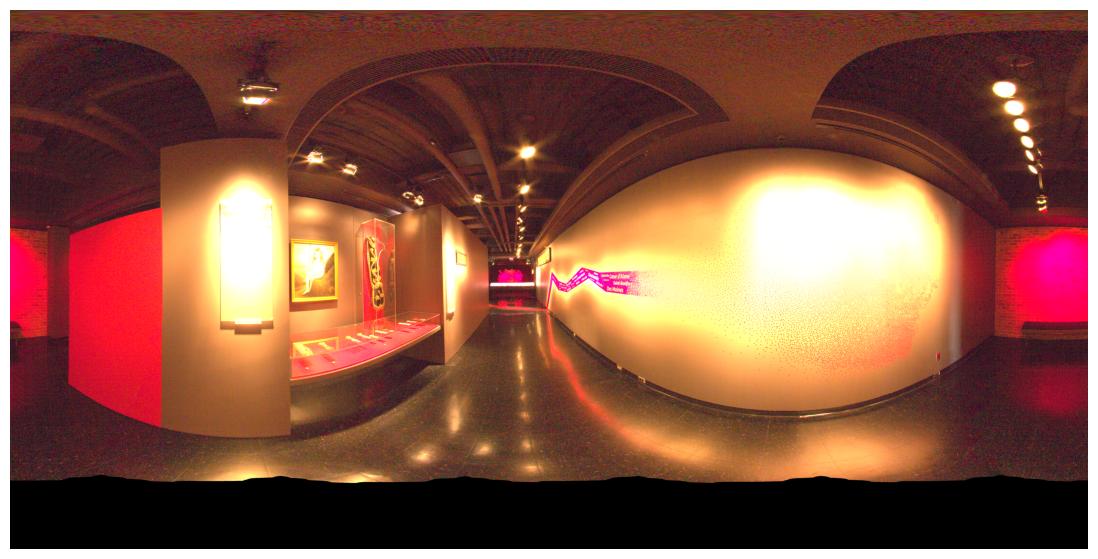}&
    \includegraphics[width=\tmplength]{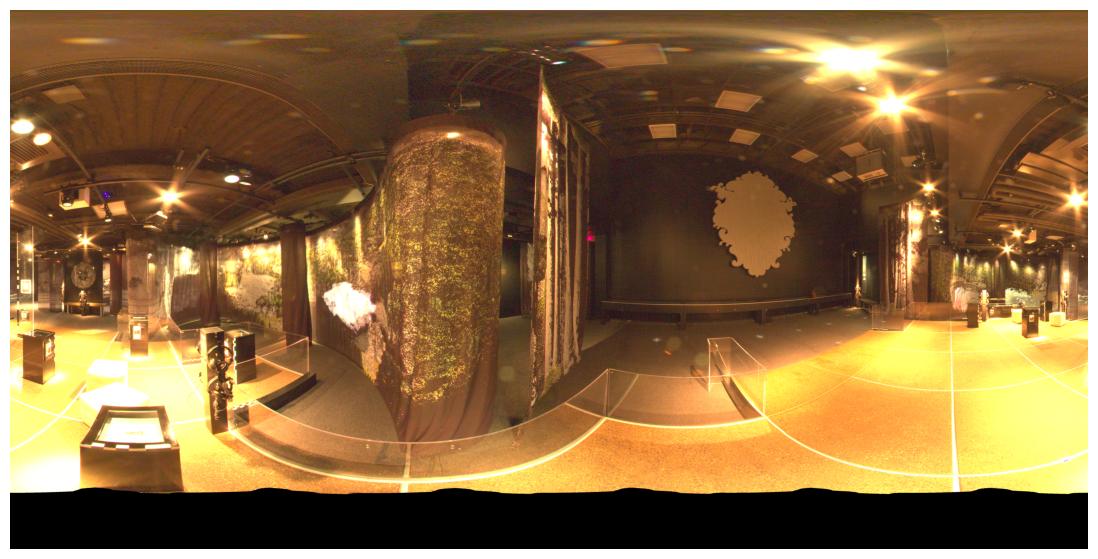}&
    \includegraphics[width=\tmplength]{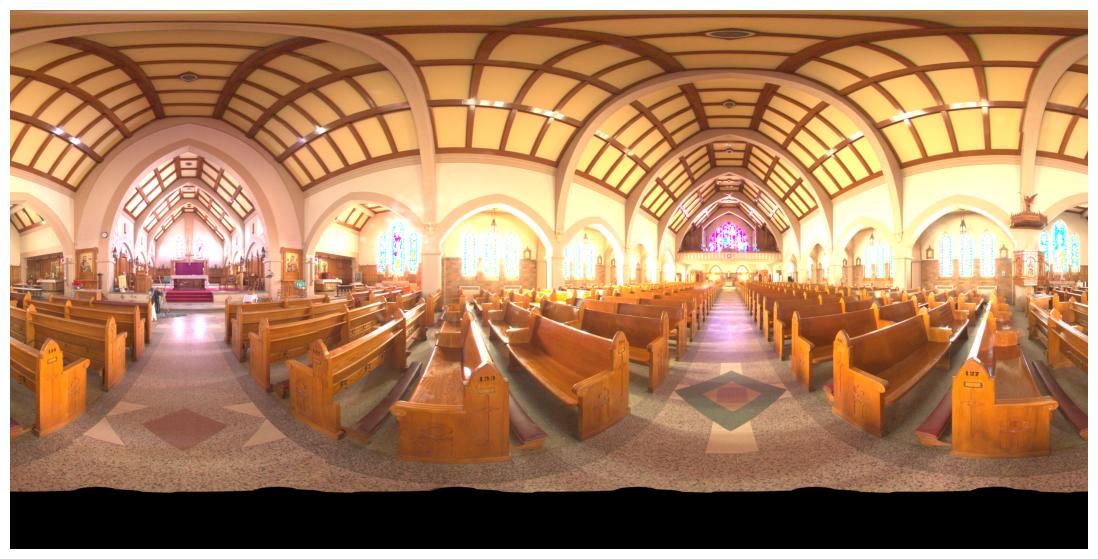}&
    \includegraphics[width=\tmplength]{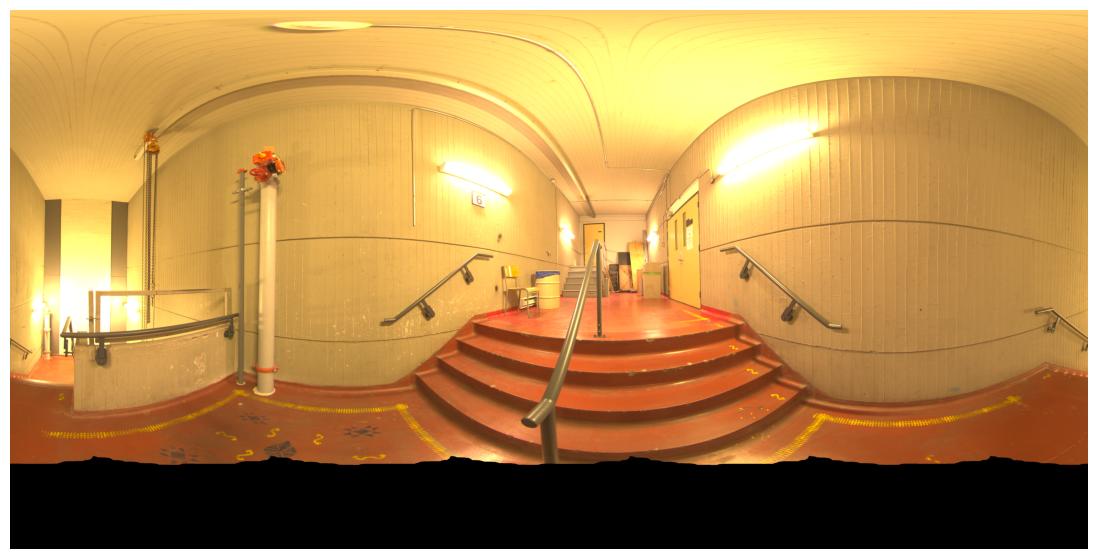}  \\
    \includegraphics[width=\tmplength]{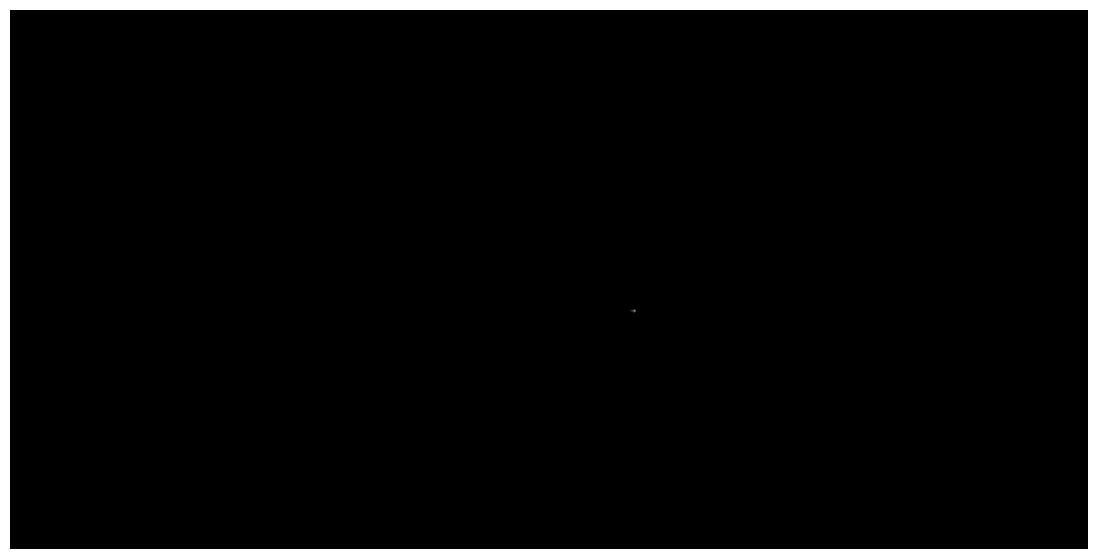}&
    \includegraphics[width=\tmplength]{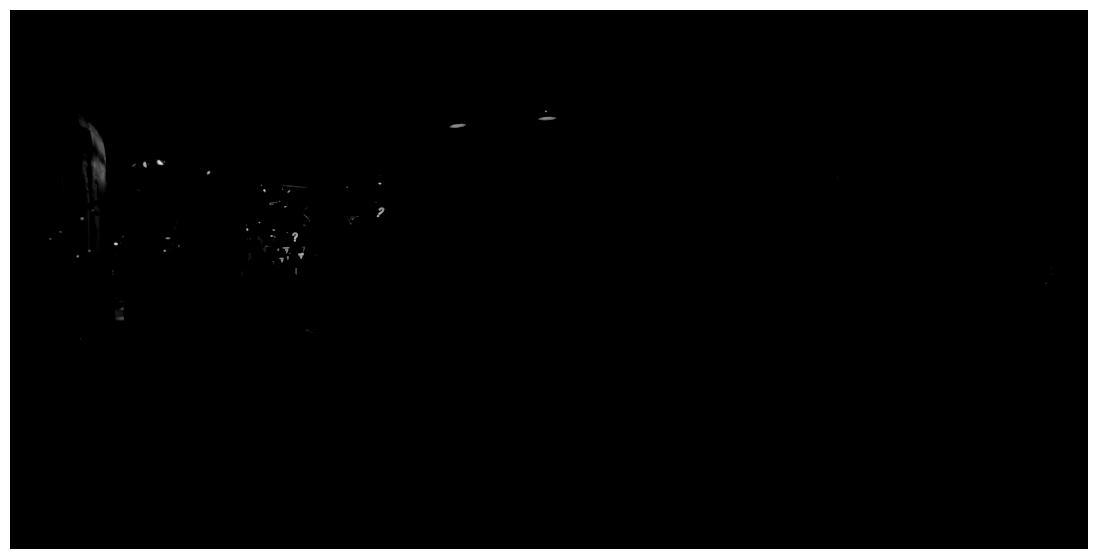}&
    \includegraphics[width=\tmplength]{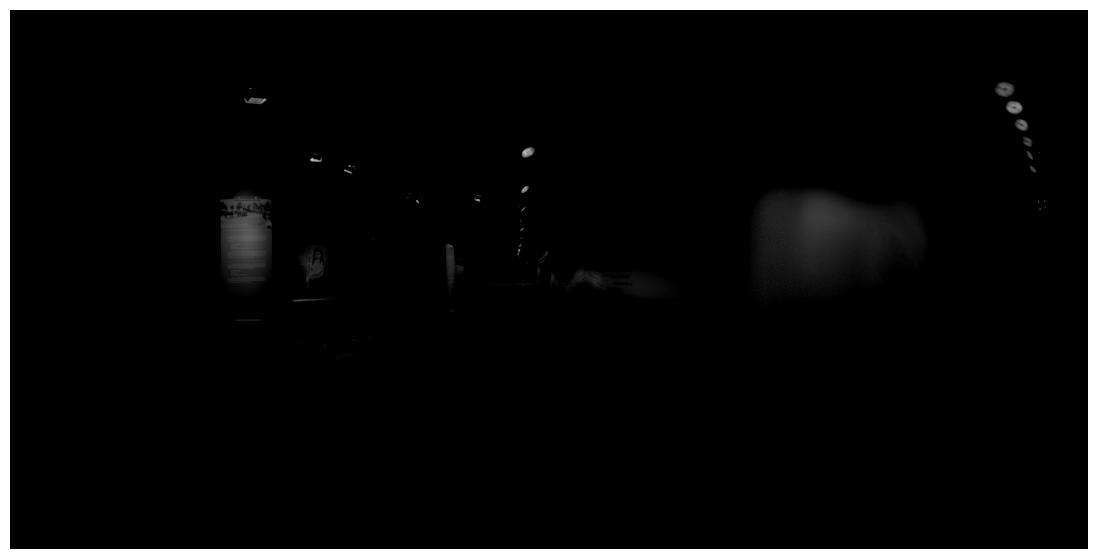}&
    \includegraphics[width=\tmplength]{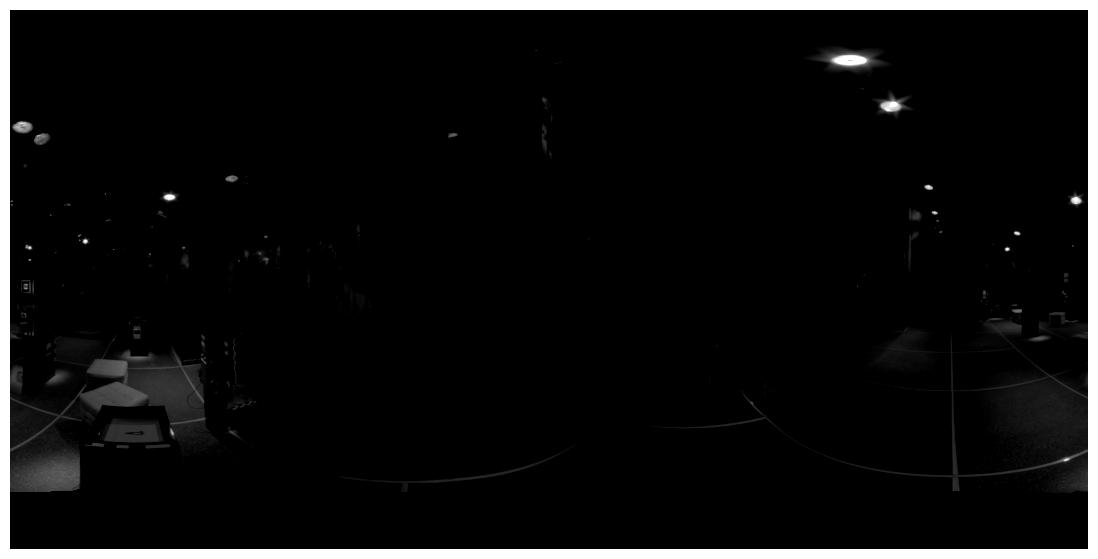}&
    \includegraphics[width=\tmplength]{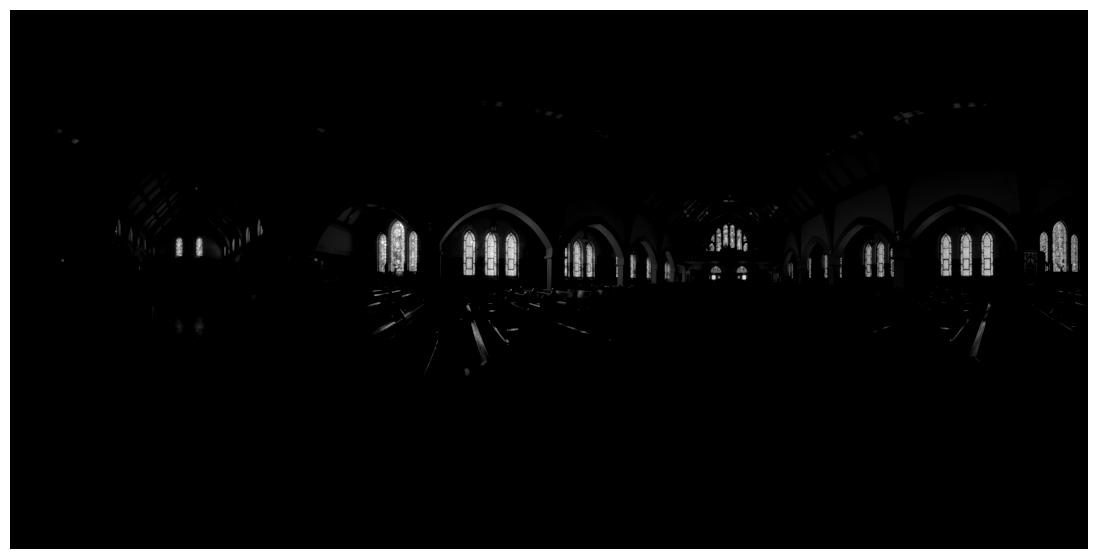}&
    \includegraphics[width=\tmplength]{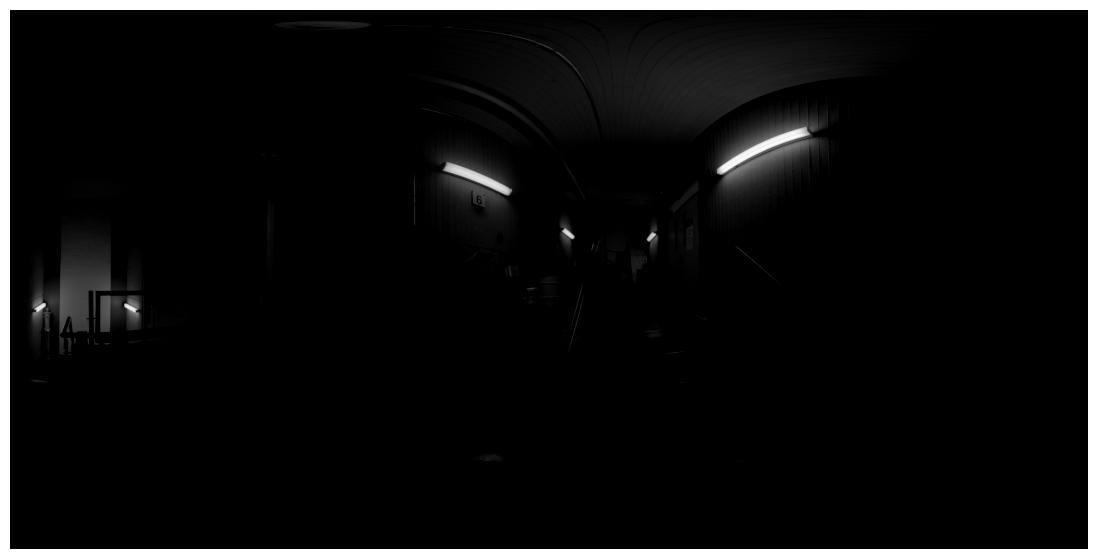}\\
    
    \valeur{17.5th} (\valeur{\SI{121}{\lux}}) & 
    \valeur{20th} (\valeur{\SI{142}{\lux}}) & 
    \valeur{22.5th} (\valeur{\SI{163}{\lux}}) & 
    \valeur{25th} (\valeur{\SI{182}{\lux}}) & 
    \valeur{27.5th} (\valeur{\SI{204}{\lux}}) & 
    \valeur{30th} (\valeur{\SI{226}{\lux}}) \\
    \includegraphics[width=\tmplength]{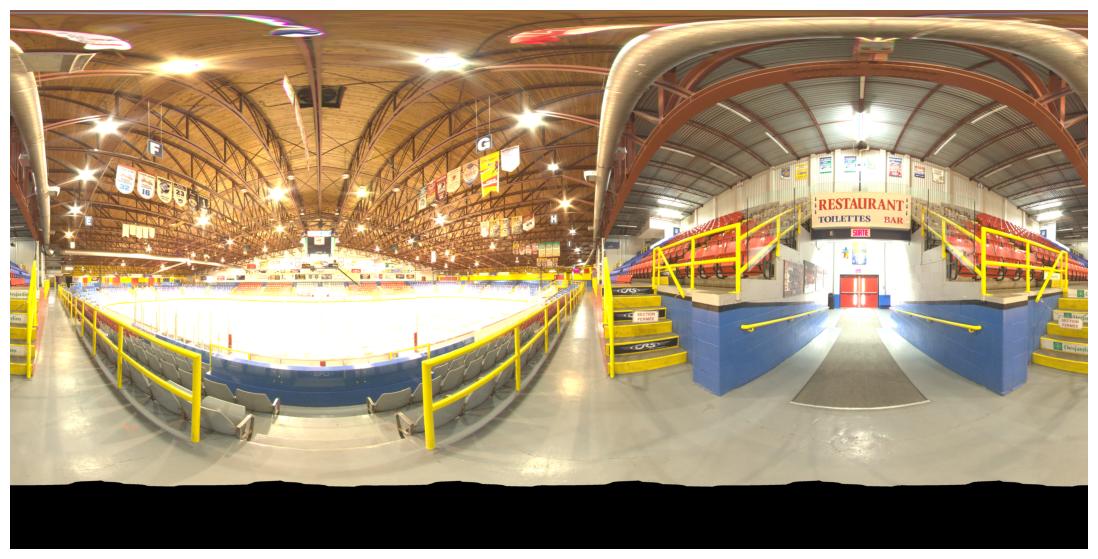}&
    \includegraphics[width=\tmplength]{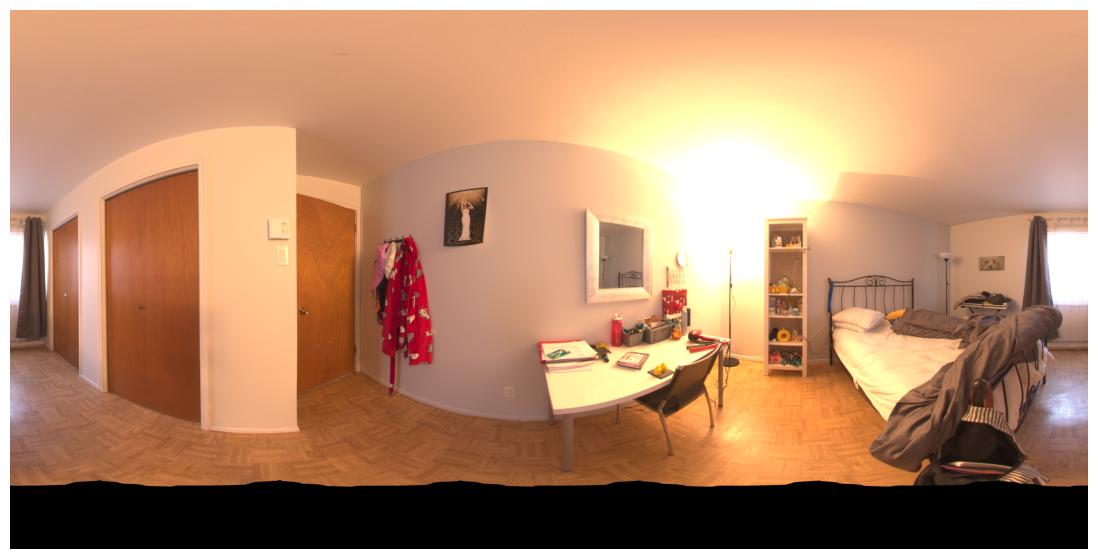}&
    \includegraphics[width=\tmplength]{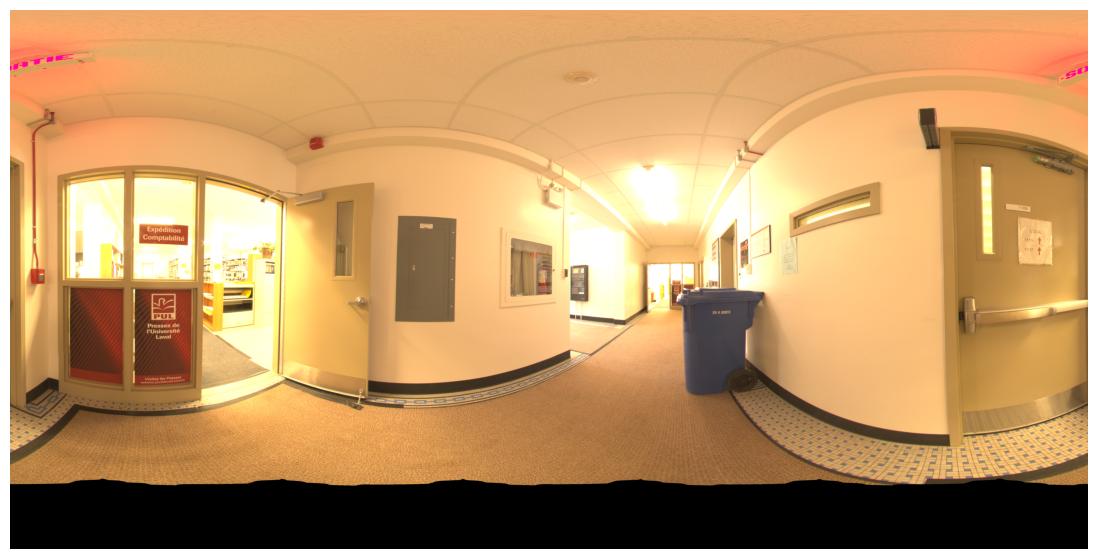}&
    \includegraphics[width=\tmplength]{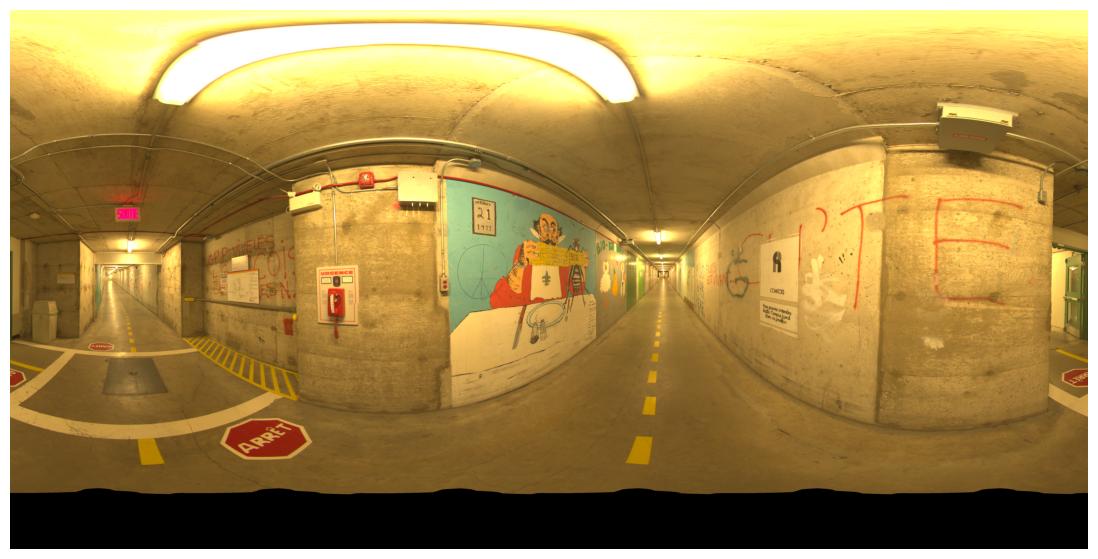}&
    \includegraphics[width=\tmplength]{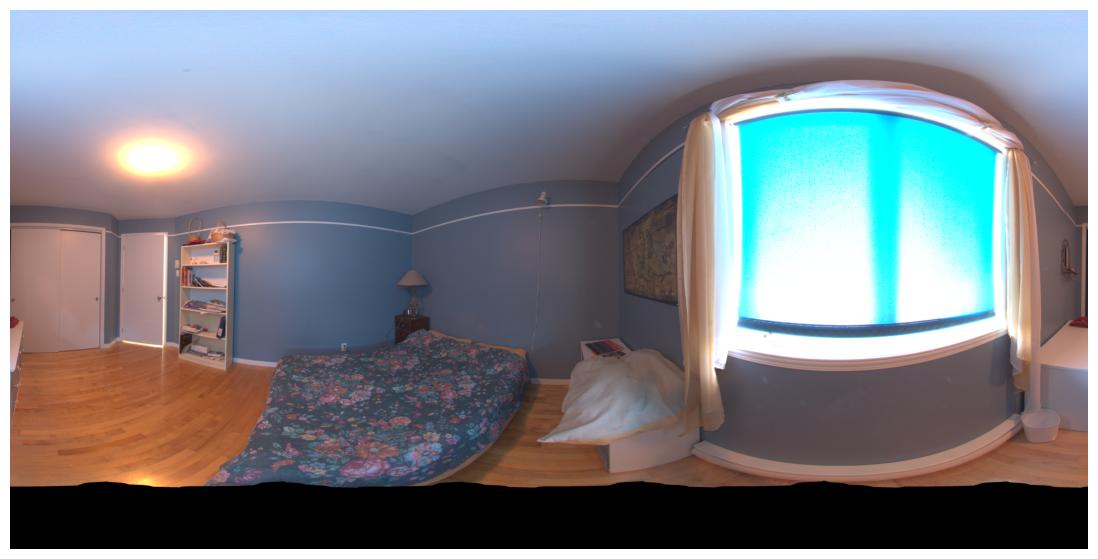}&
    \includegraphics[width=\tmplength]{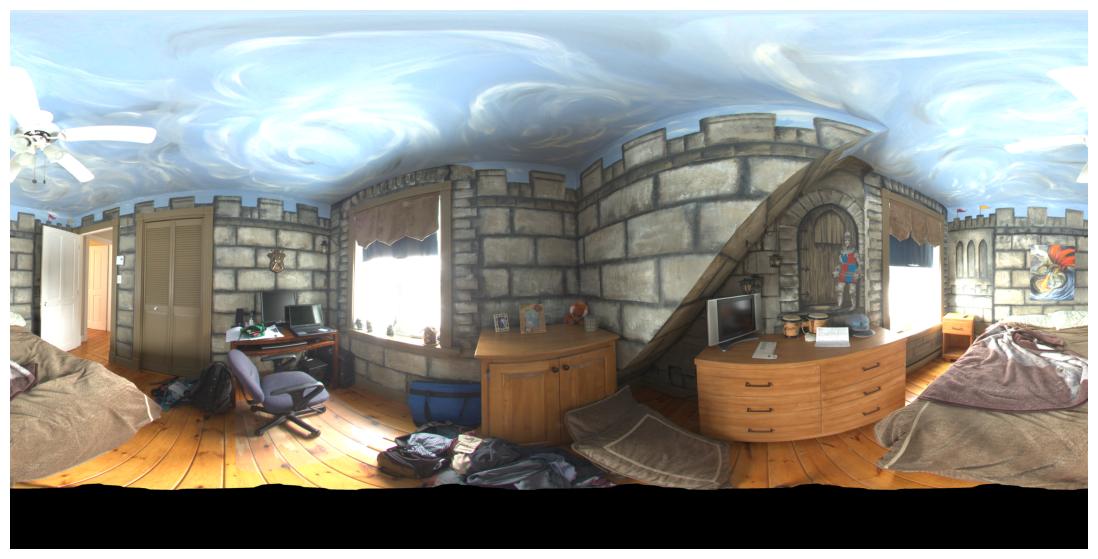}\\
    \includegraphics[width=\tmplength]{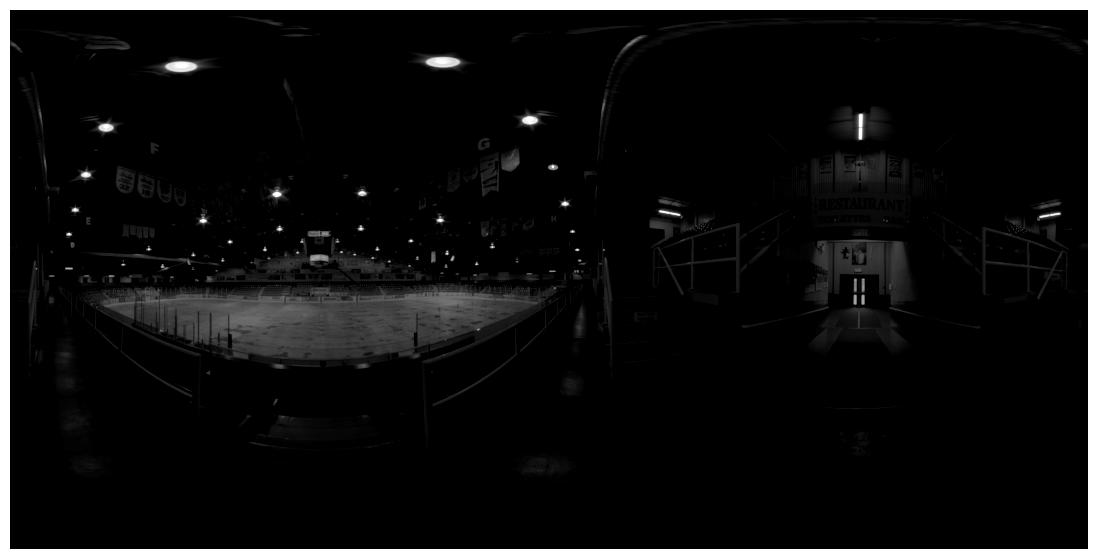}&
    \includegraphics[width=\tmplength]{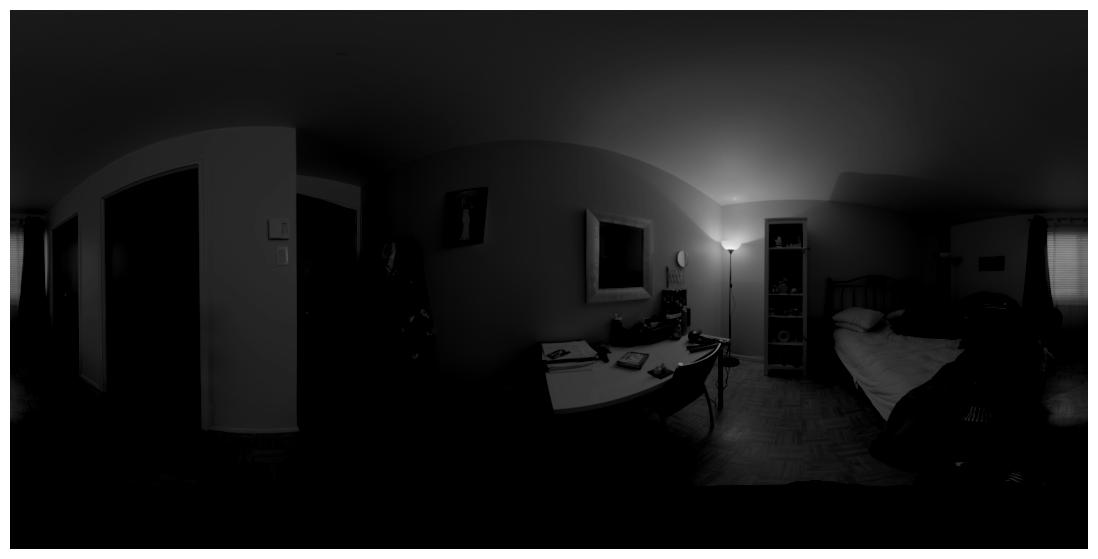}&
    \includegraphics[width=\tmplength]{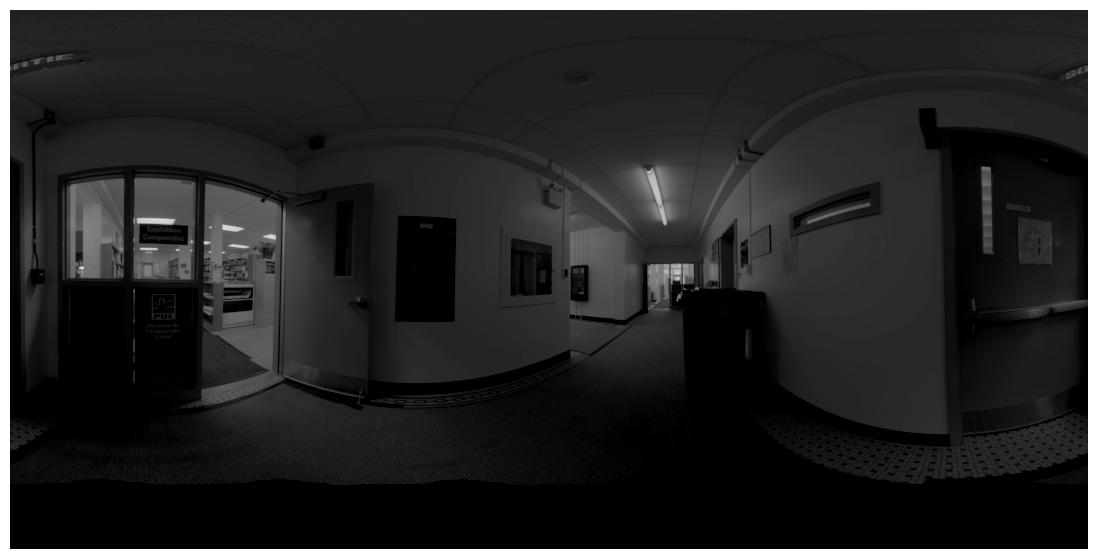}&
    \includegraphics[width=\tmplength]{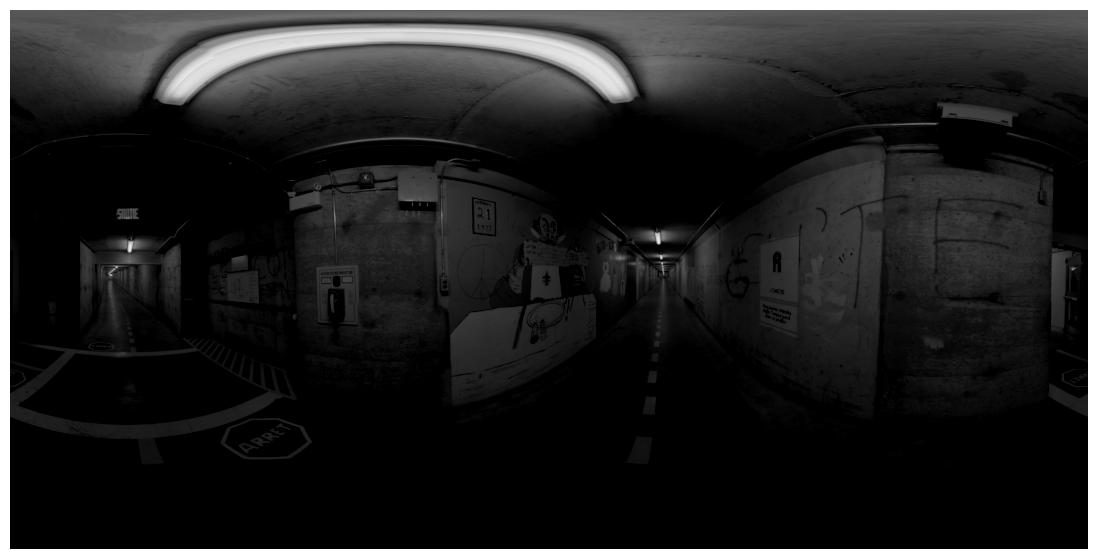}&
    \includegraphics[width=\tmplength]{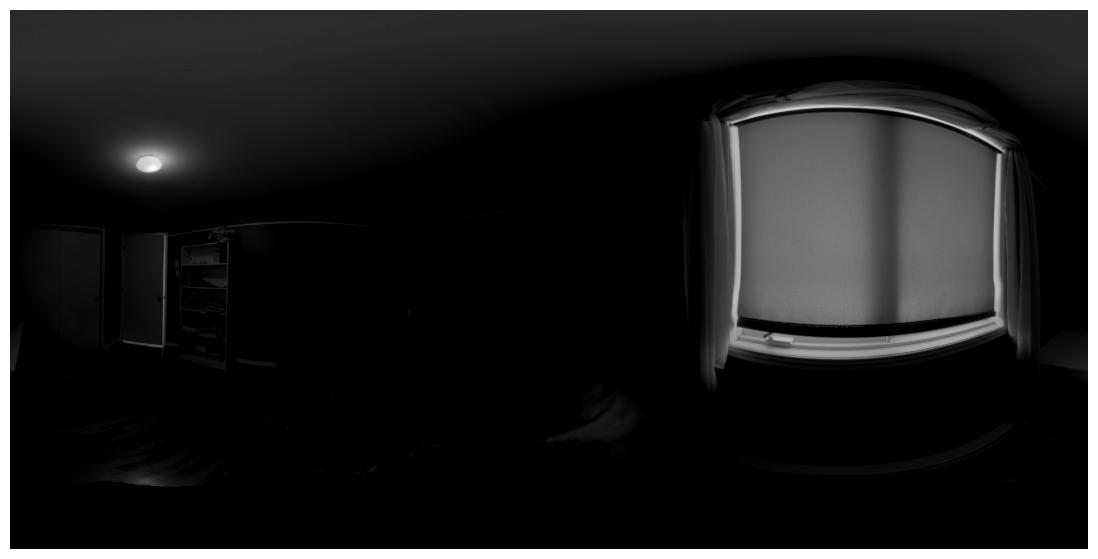}&
    \includegraphics[width=\tmplength]{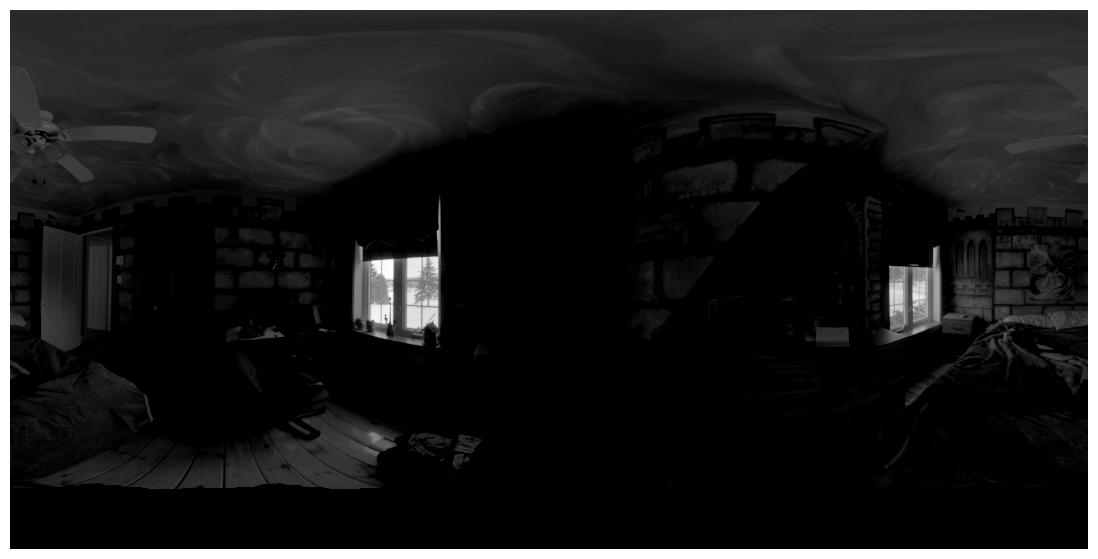}\\
    
    \valeur{32.5th} (\valeur{\SI{253}{\lux}}) & 
    \valeur{35th} (\valeur{\SI{275}{\lux}}) & 
    \valeur{37.5th} (\valeur{\SI{295}{\lux}}) & 
    \valeur{40th} (\valeur{\SI{320}{\lux}}) & 
    \valeur{42.5th} (\valeur{\SI{355}{\lux}}) & 
    \valeur{45th} (\valeur{\SI{390}{\lux}}) \\
    \includegraphics[width=\tmplength]{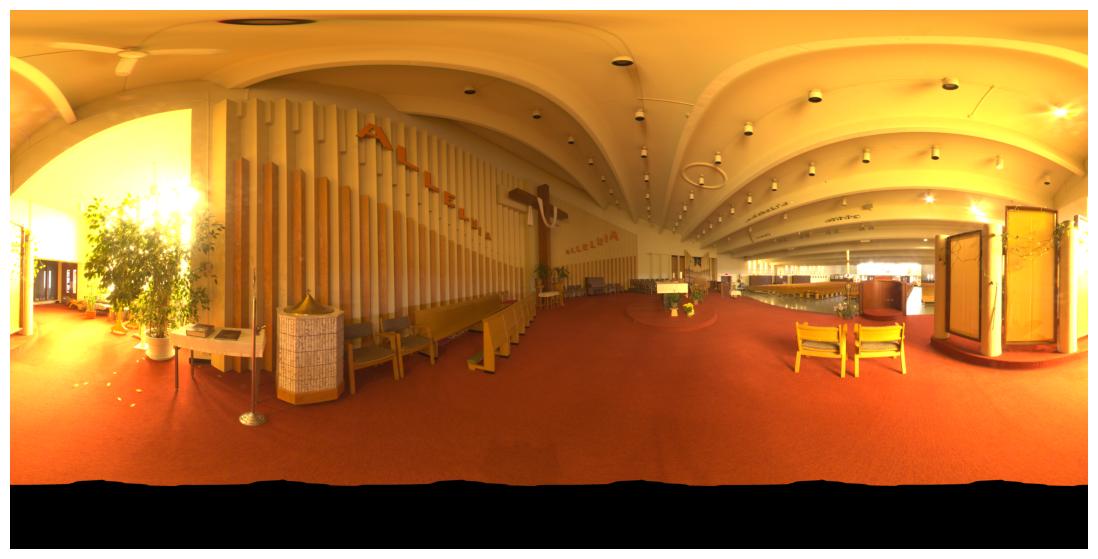}&
    \includegraphics[width=\tmplength]{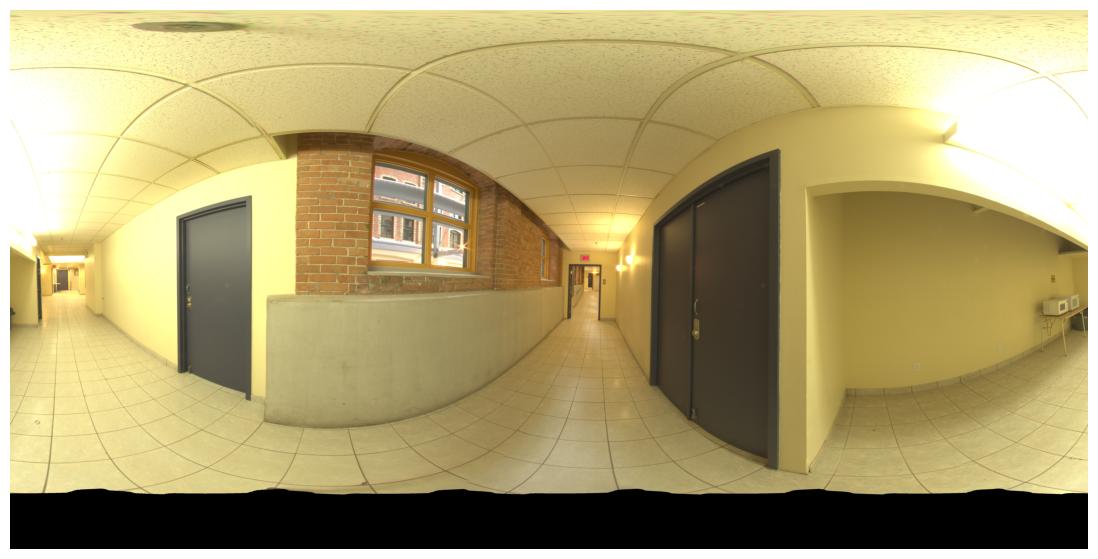}&
    \includegraphics[width=\tmplength]{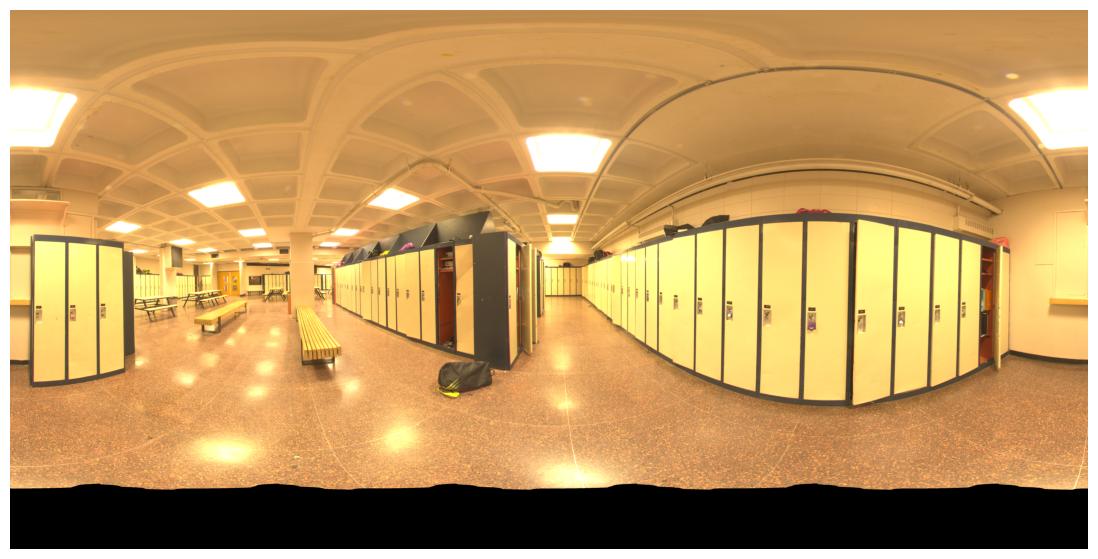}&
    \includegraphics[width=\tmplength]{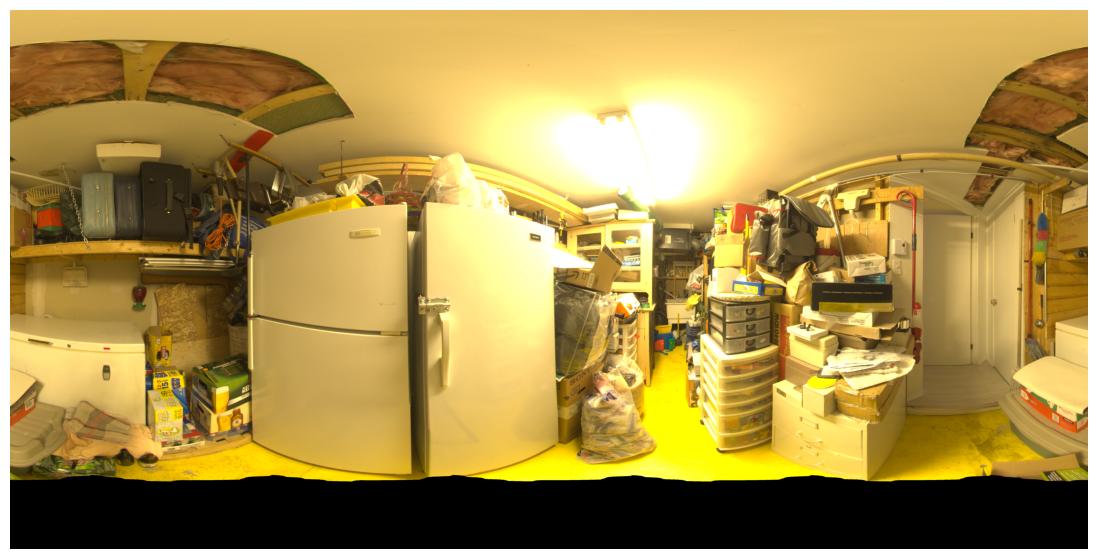}&
    \includegraphics[width=\tmplength]{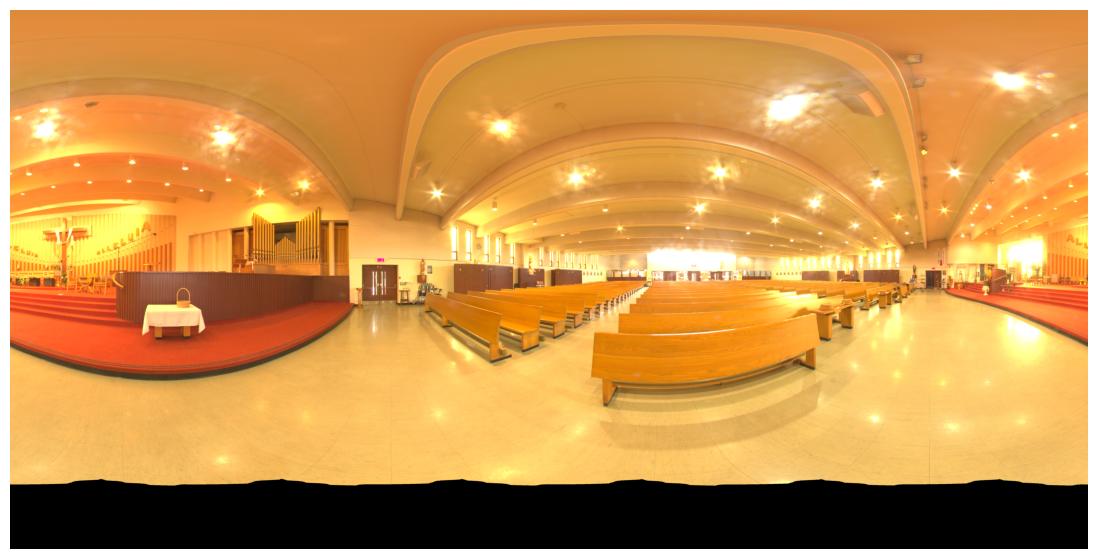}&
    \includegraphics[width=\tmplength]{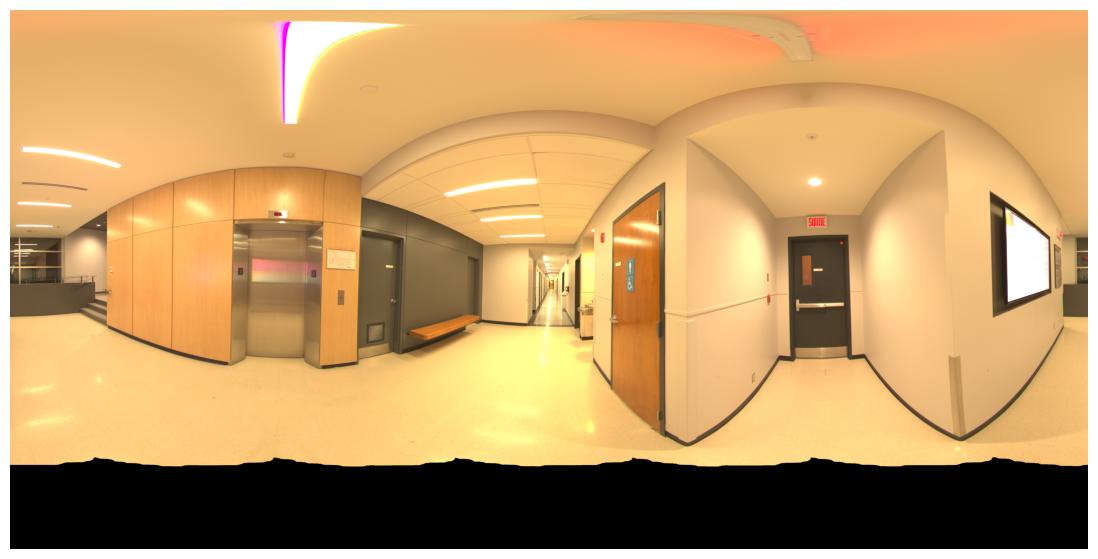}\\
    \includegraphics[width=\tmplength]{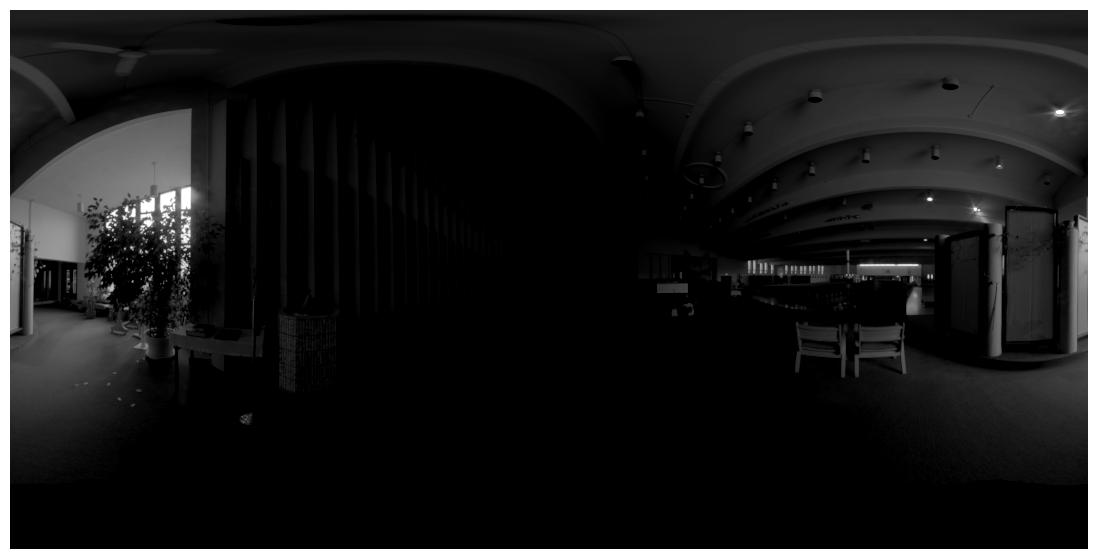}&
    \includegraphics[width=\tmplength]{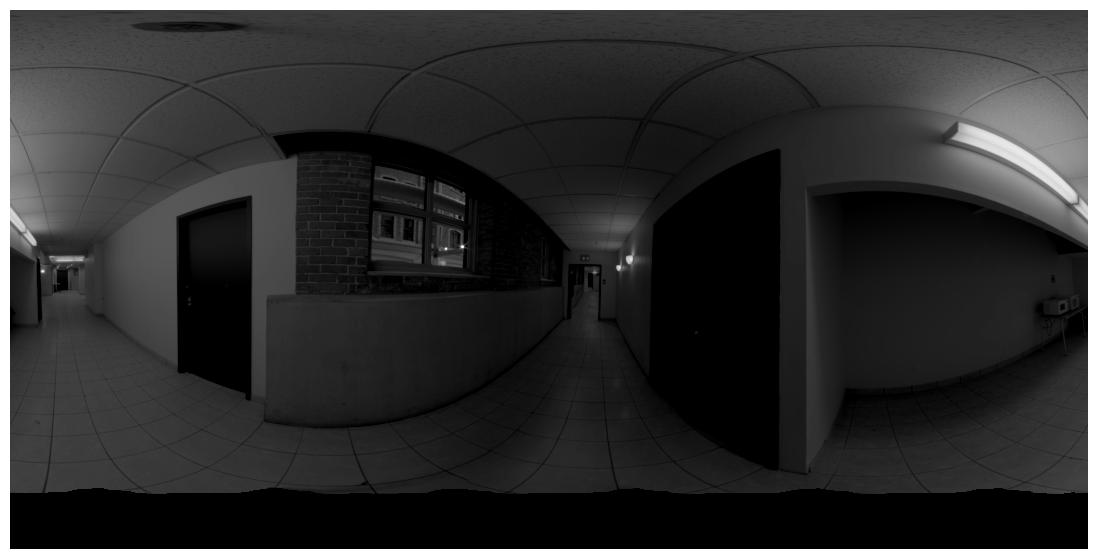}&
    \includegraphics[width=\tmplength]{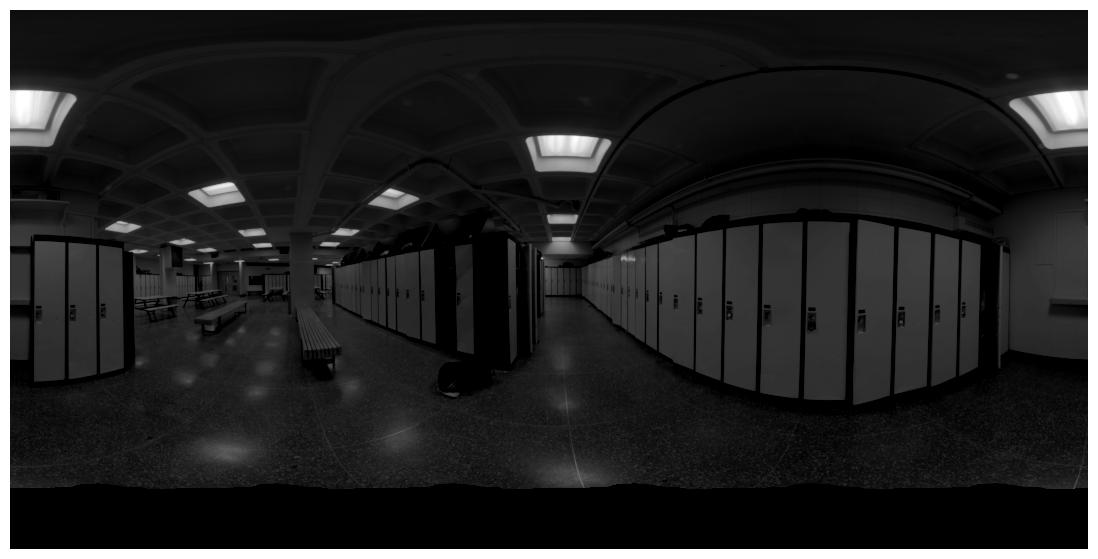}&
    \includegraphics[width=\tmplength]{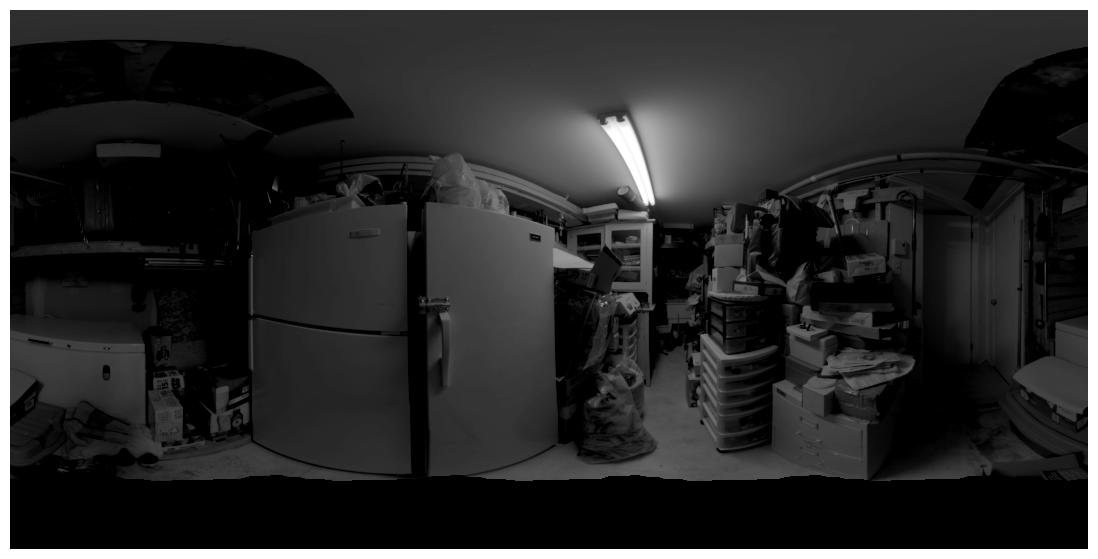}&
    \includegraphics[width=\tmplength]{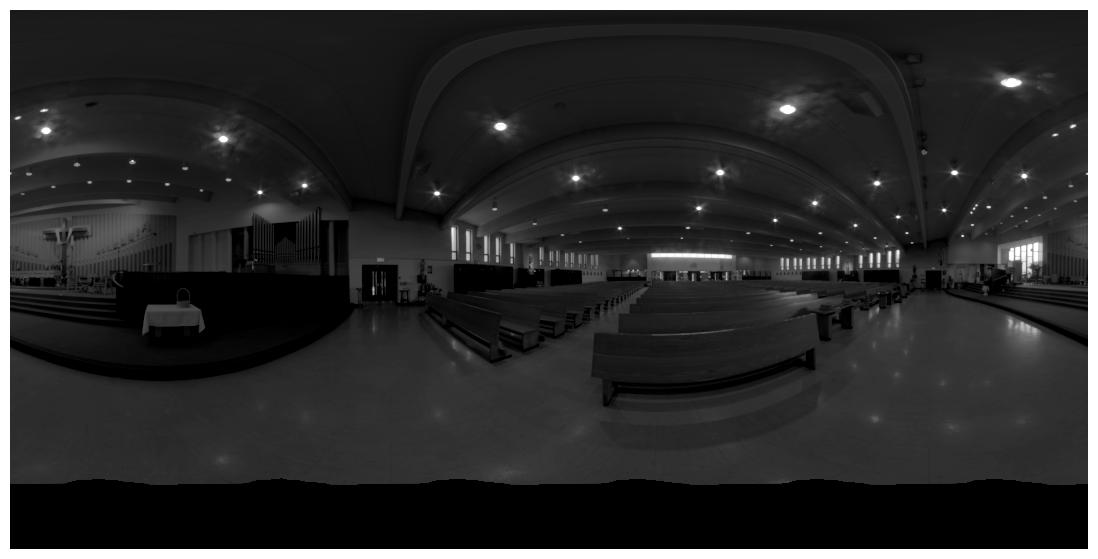}&
    \includegraphics[width=\tmplength]{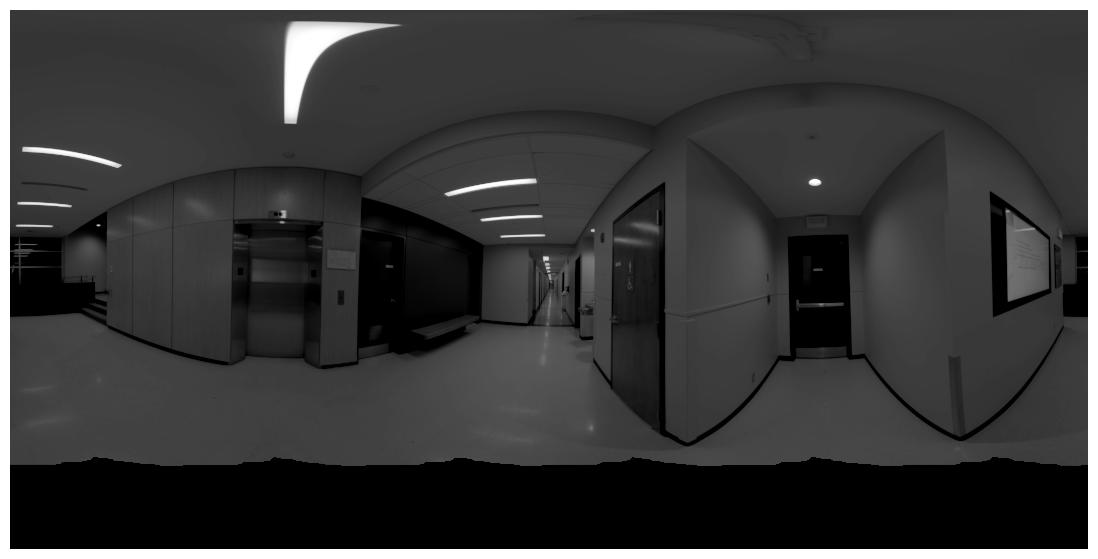}\\
    
    \valeur{47.5th} (\valeur{\SI{422}{\lux}}) & 
    \valeur{50th} (\valeur{\SI{460}{\lux}}) & 
    \valeur{52.5th} (\valeur{\SI{501}{\lux}}) & 
    \valeur{55th} (\valeur{\SI{539}{\lux}}) & 
    \valeur{57.5th} (\valeur{\SI{577}{\lux}}) & 
    \valeur{60th} (\valeur{\SI{616}{\lux}}) \\
    \includegraphics[width=\tmplength]{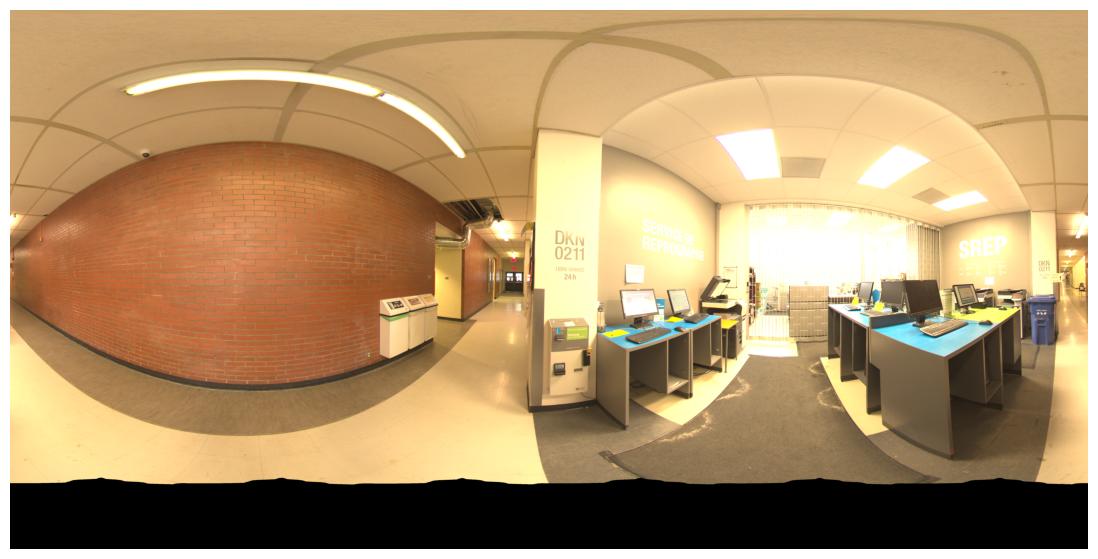}&
    \includegraphics[width=\tmplength]{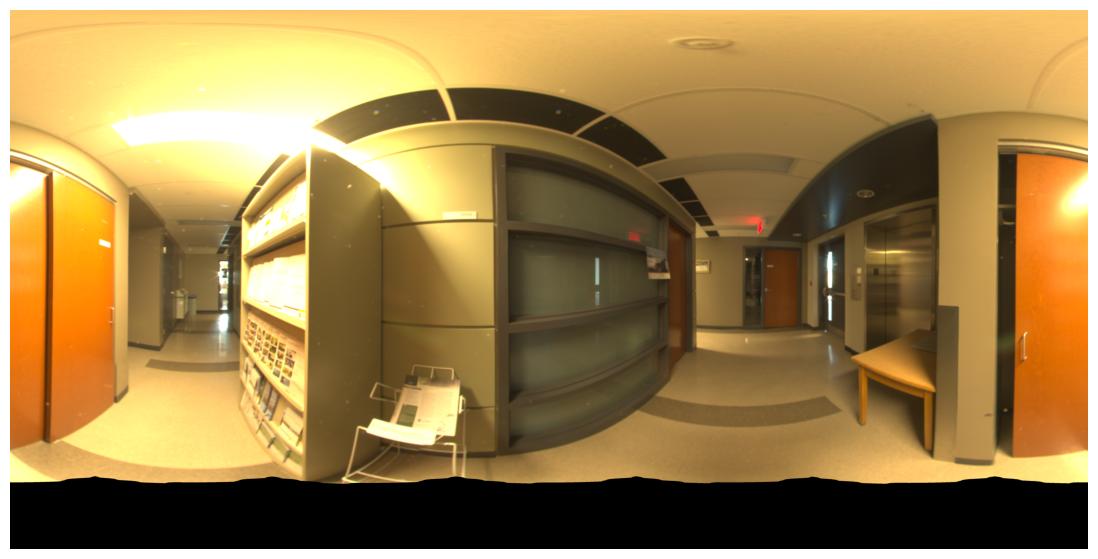}&
    \includegraphics[width=\tmplength]{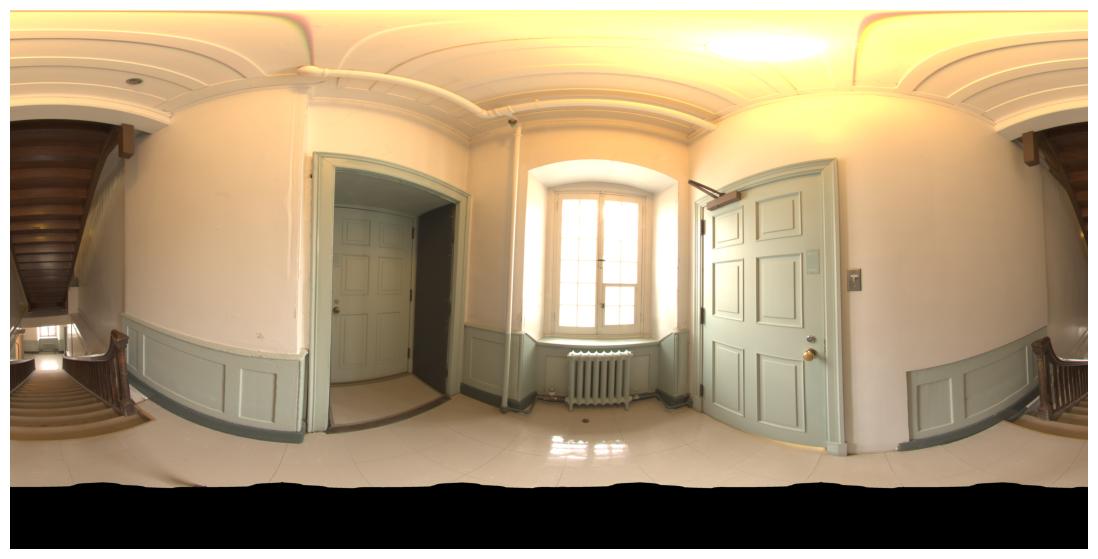}&
    \includegraphics[width=\tmplength]{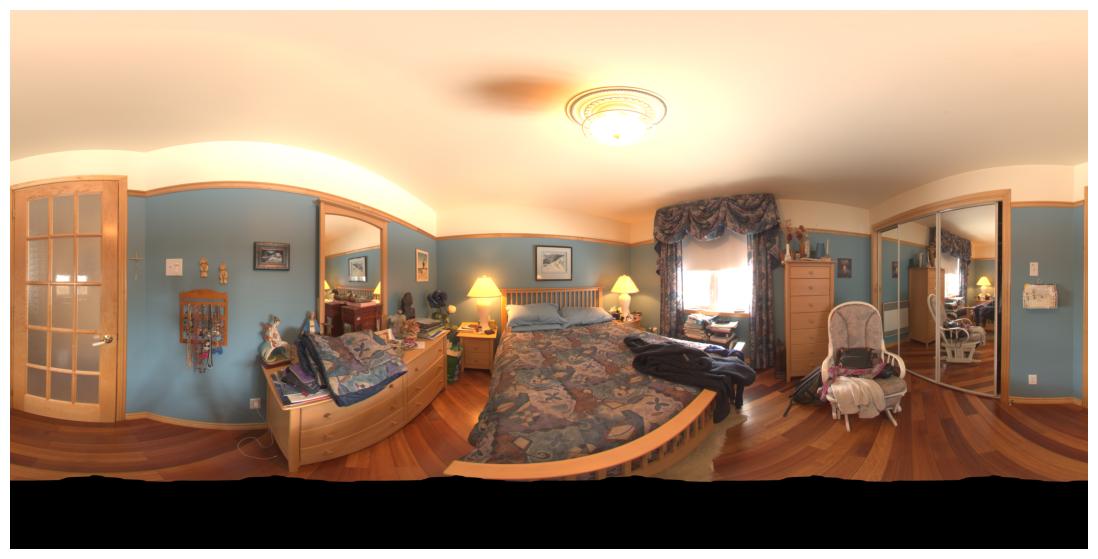}&
    \includegraphics[width=\tmplength]{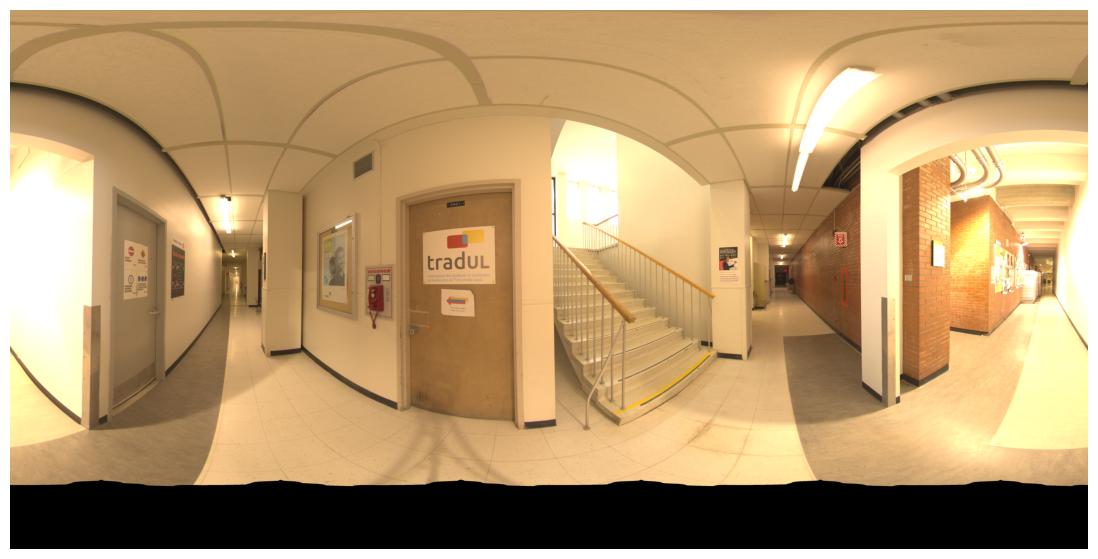}&
    \includegraphics[width=\tmplength]{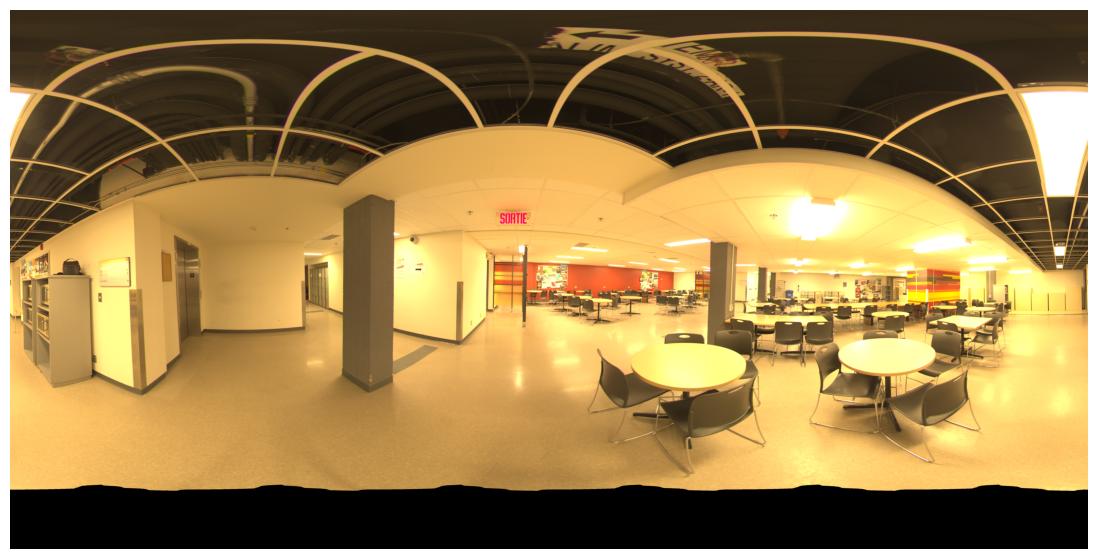}\\
    \includegraphics[width=\tmplength]{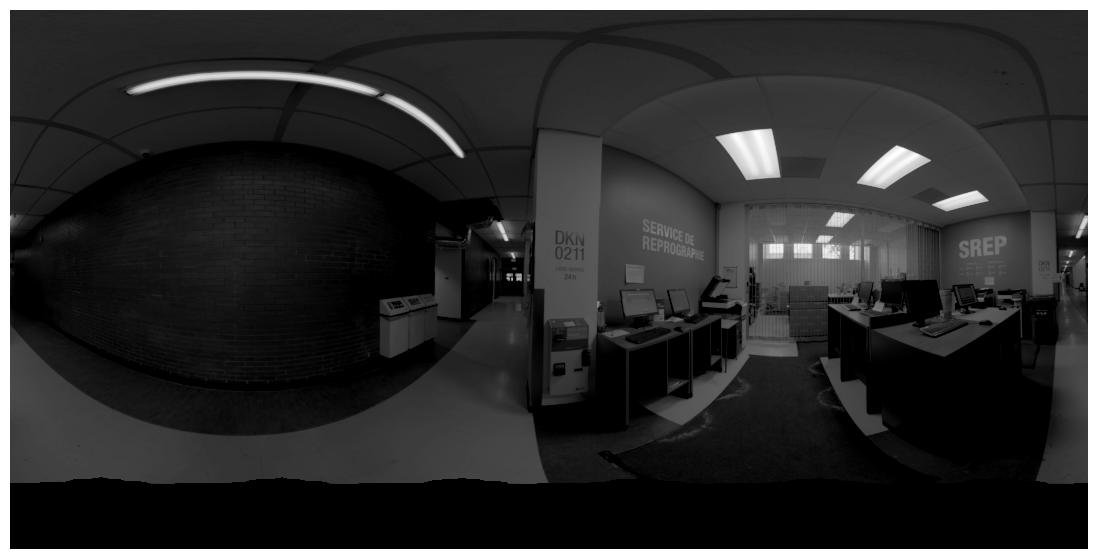}&
    \includegraphics[width=\tmplength]{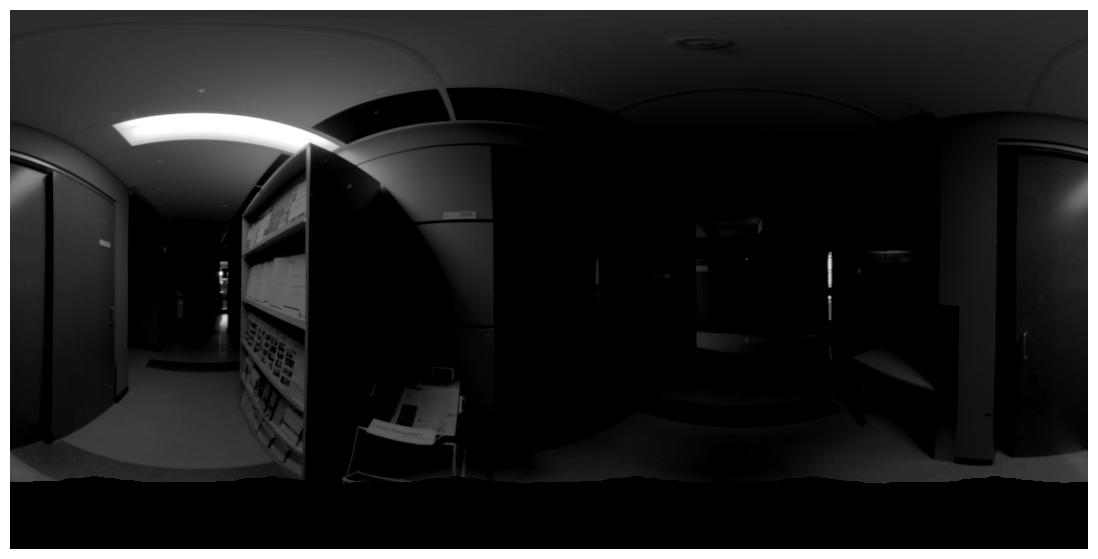}&
    \includegraphics[width=\tmplength]{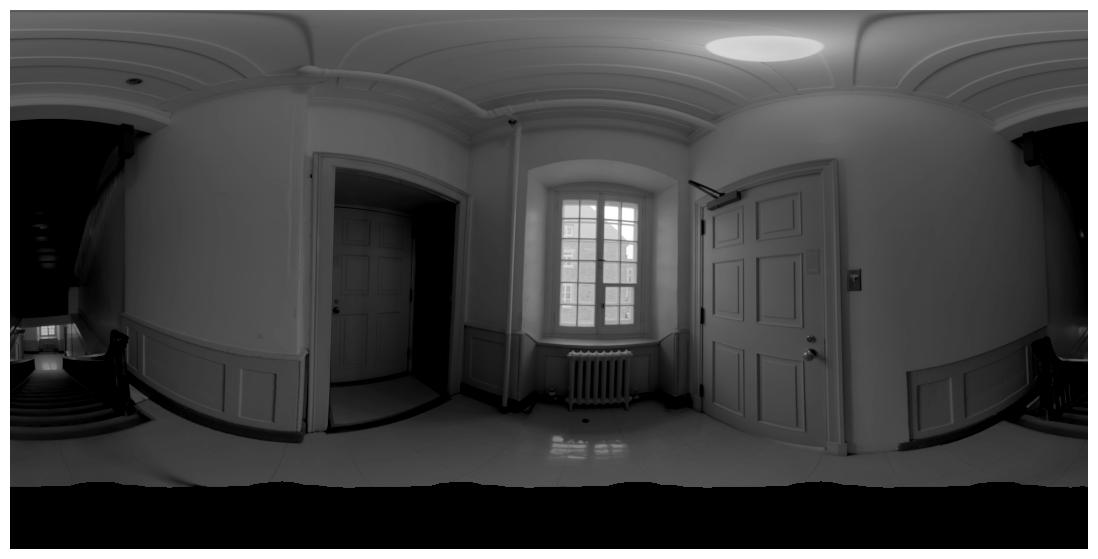}&
    \includegraphics[width=\tmplength]{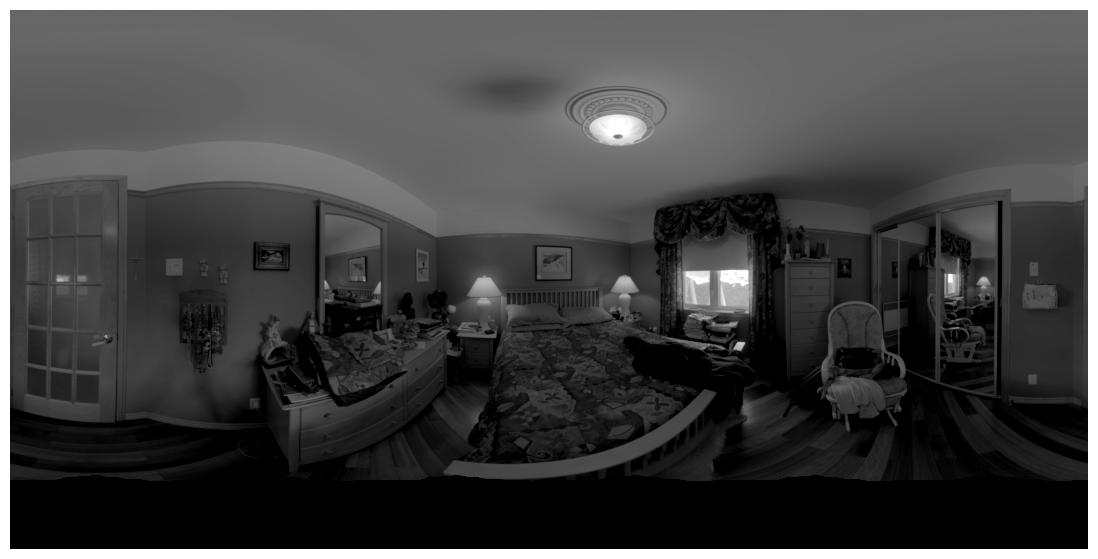}&
    \includegraphics[width=\tmplength]{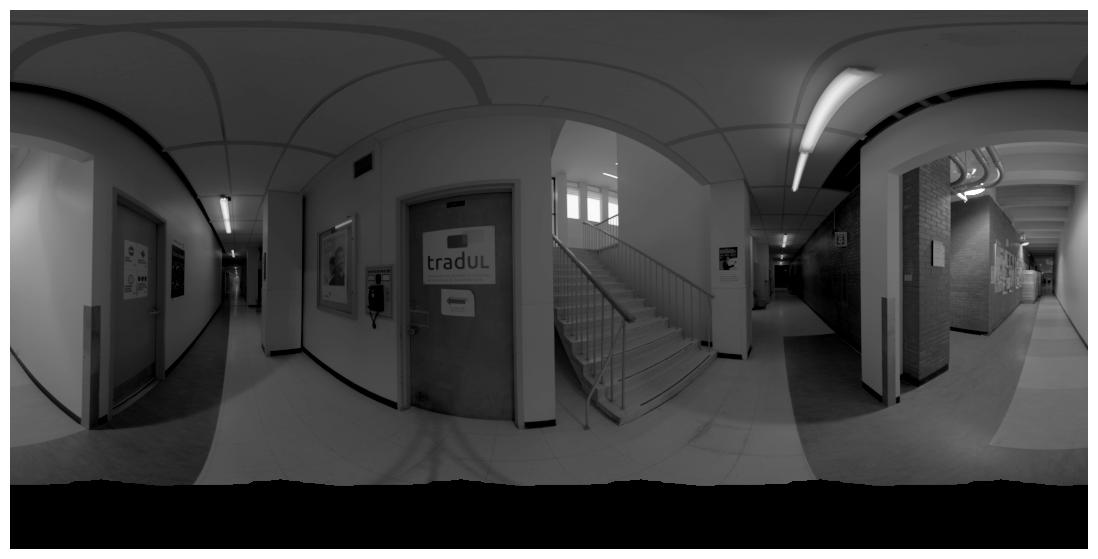}&
    \includegraphics[width=\tmplength]{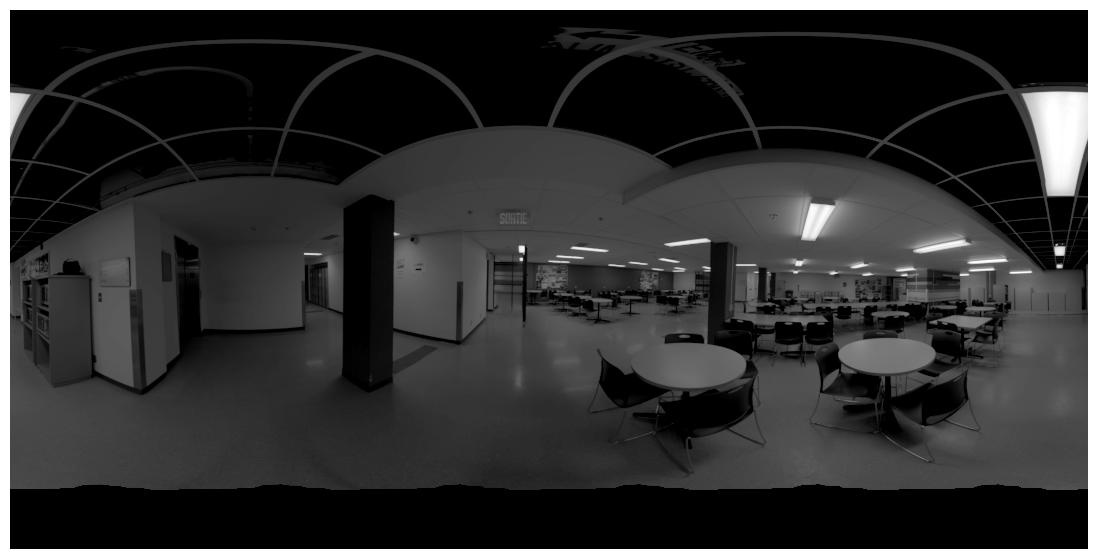}\\
    
    \valeur{62.5th} (\valeur{\SI{666}{\lux}}) & 
    \valeur{65th} (\valeur{\SI{717}{\lux}}) & 
    \valeur{67.5th} (\valeur{\SI{767}{\lux}}) & 
    \valeur{70th} (\valeur{\SI{835}{\lux}}) & 
    \valeur{72.5th} (\valeur{\SI{892}{\lux}}) & 
    \valeur{75th} (\valeur{\SI{955}{\lux}}) \\
    \includegraphics[width=\tmplength]{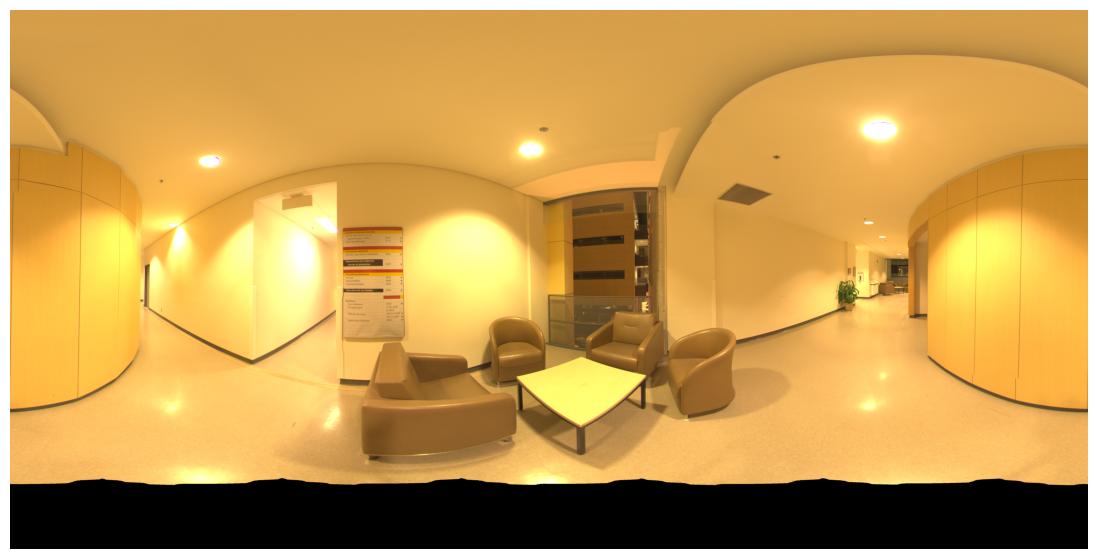}&
    \includegraphics[width=\tmplength]{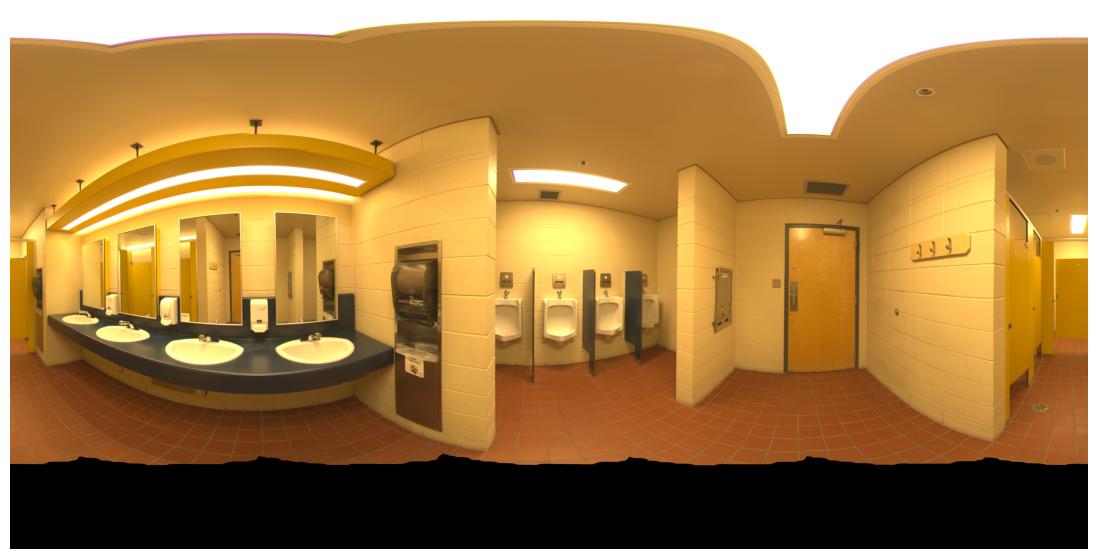}&
    \includegraphics[width=\tmplength]{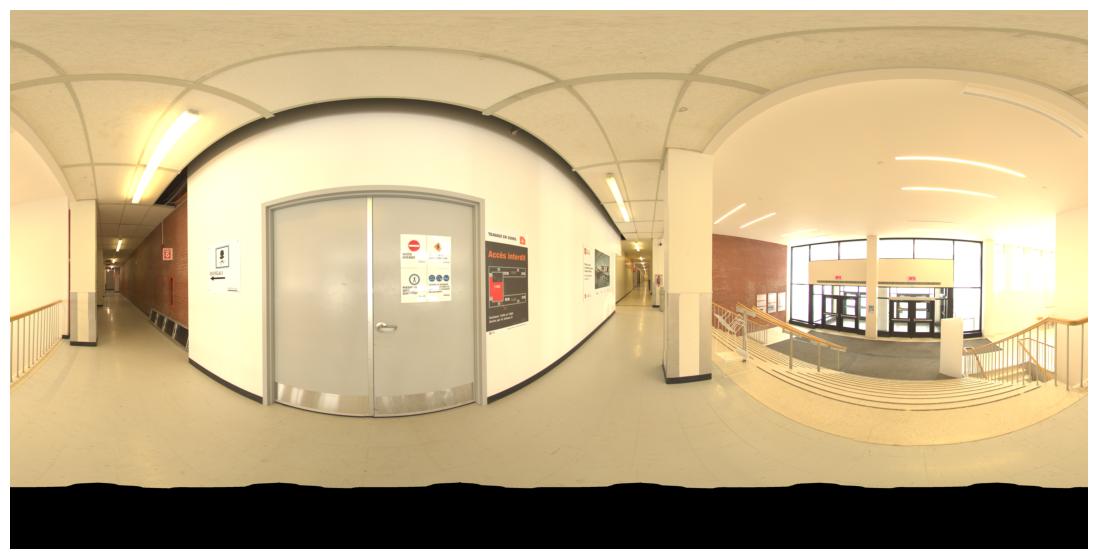}&
    \includegraphics[width=\tmplength]{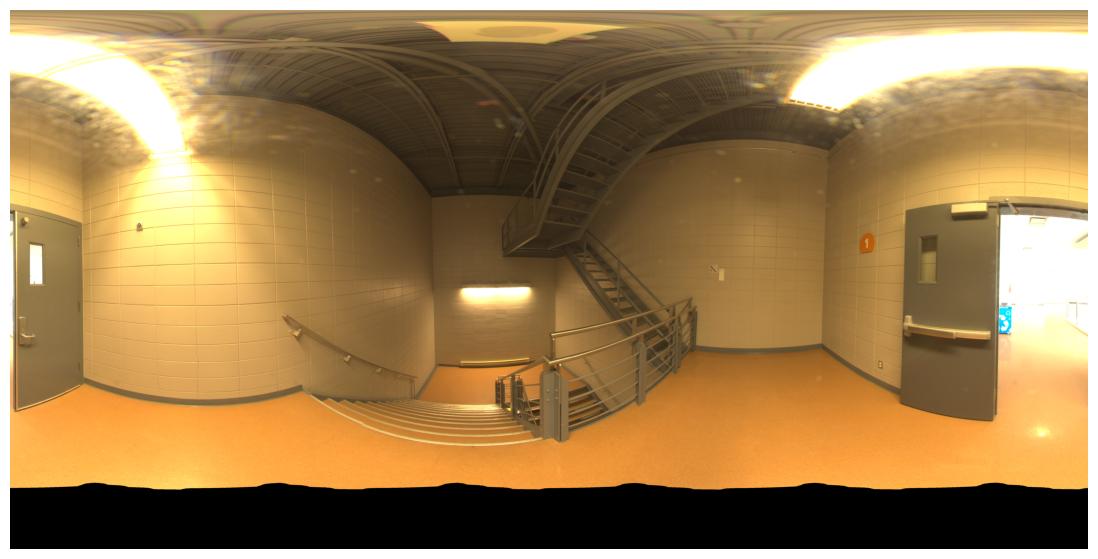}&
    \includegraphics[width=\tmplength]{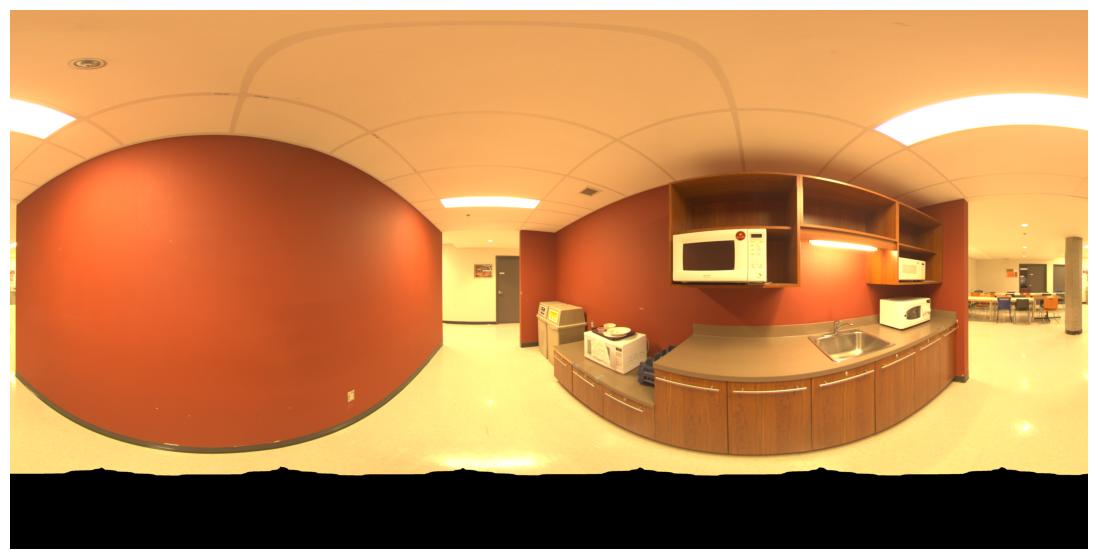}&
    \includegraphics[width=\tmplength]{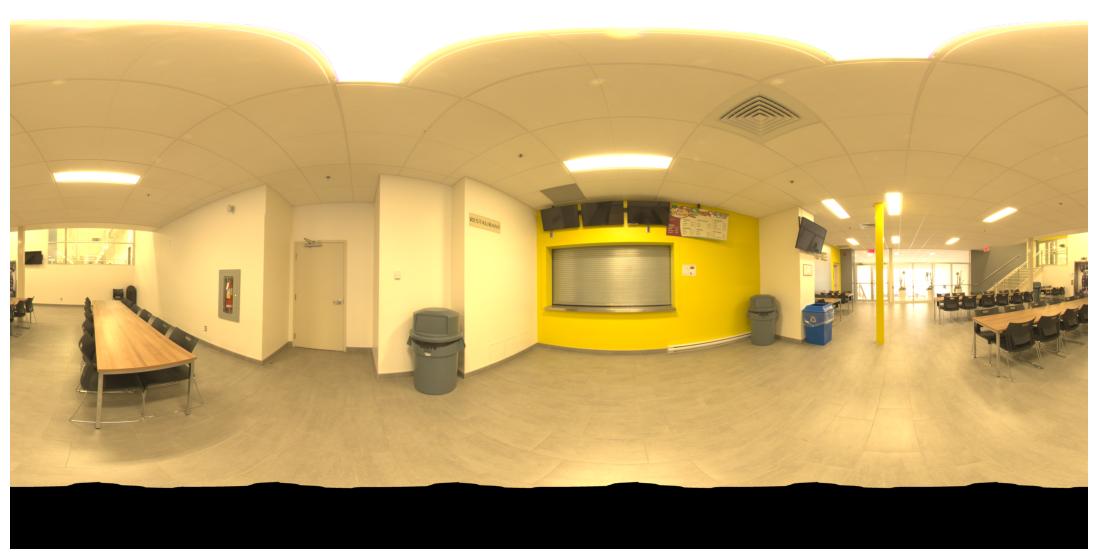}\\
    \includegraphics[width=\tmplength]{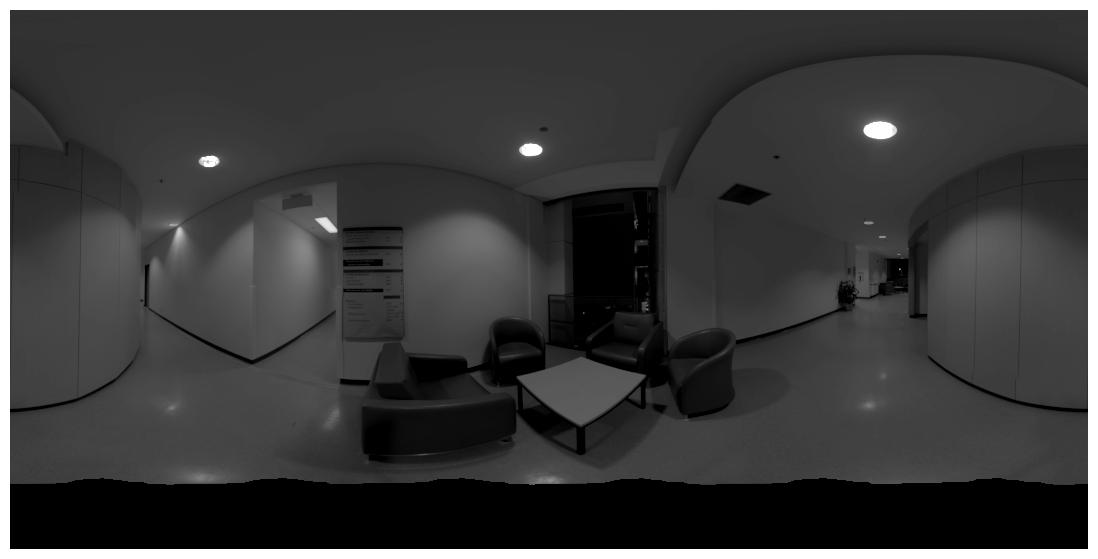}&
    \includegraphics[width=\tmplength]{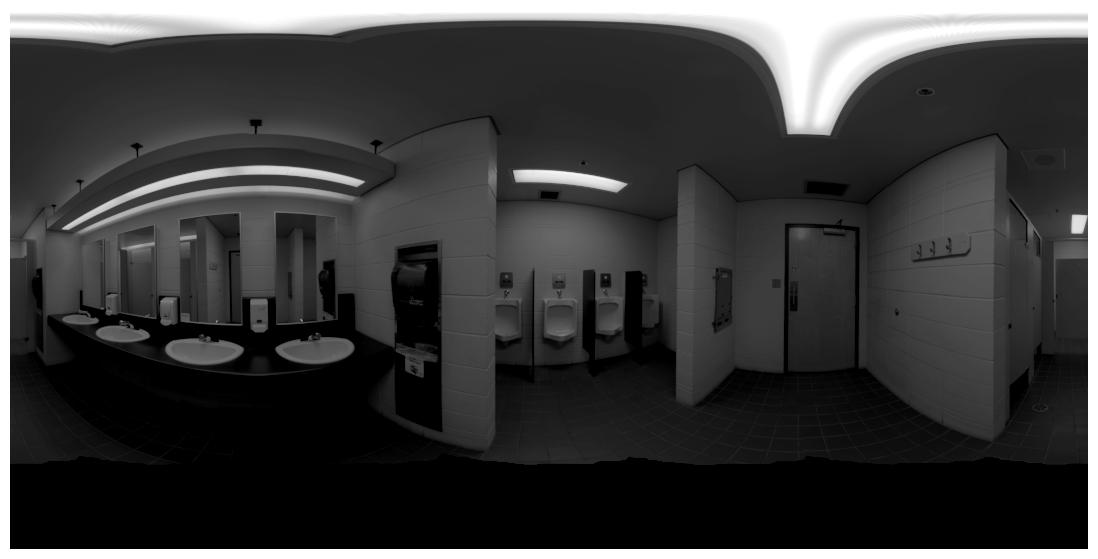}&
    \includegraphics[width=\tmplength]{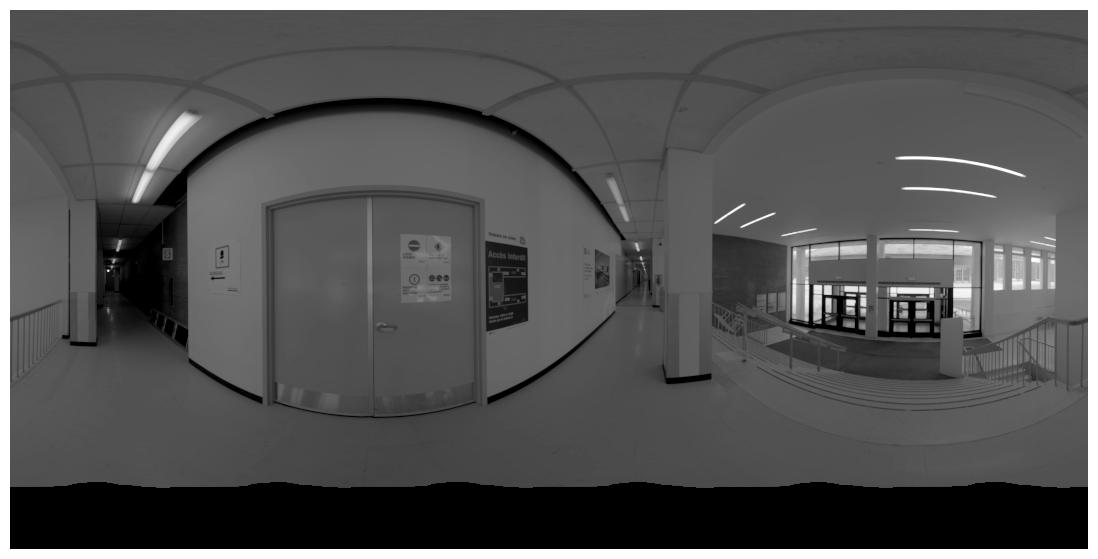}&
    \includegraphics[width=\tmplength]{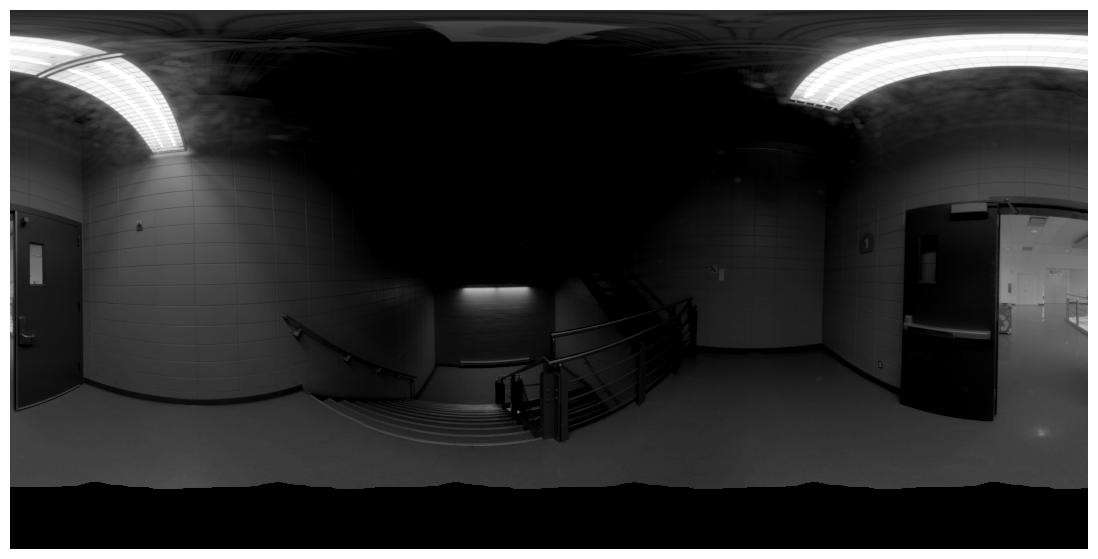}&
    \includegraphics[width=\tmplength]{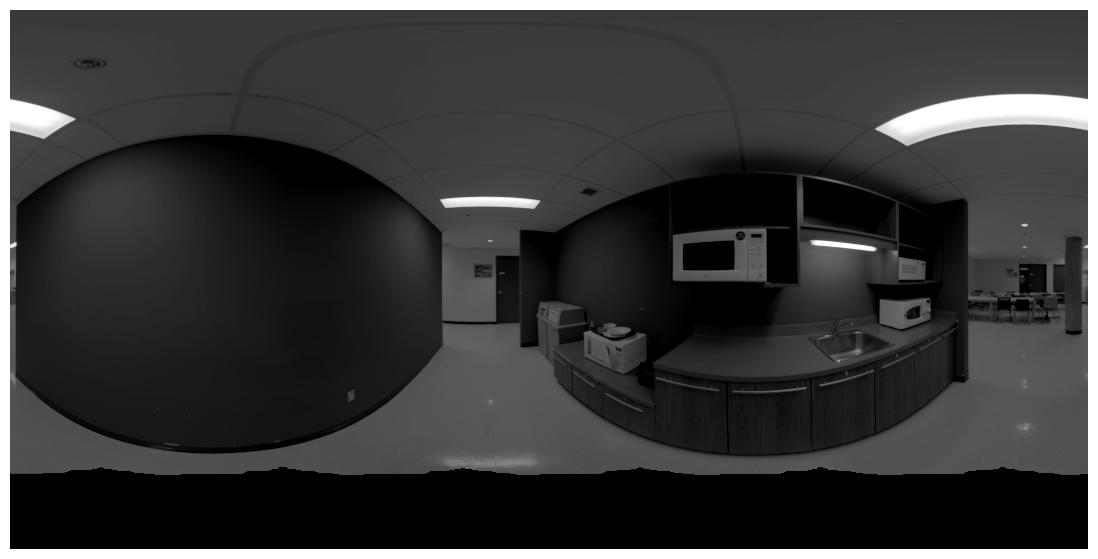}&
    \includegraphics[width=\tmplength]{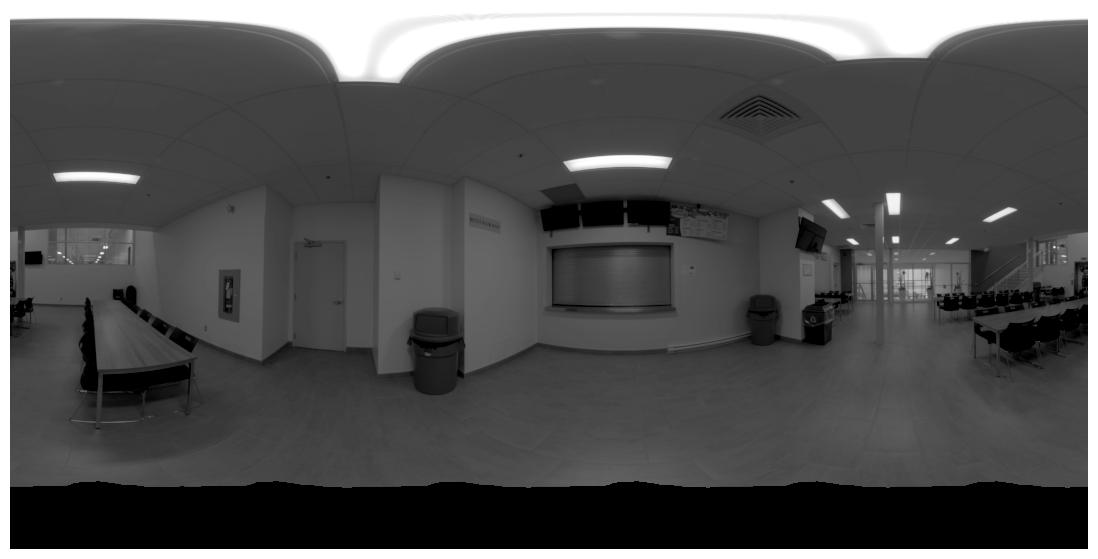}\\
    
    \valeur{77.5th} (\valeur{\SI{1037}{\lux}}) & 
    \valeur{80th} (\valeur{\SI{1138}{\lux}}) & 
    \valeur{82.5th} (\valeur{\SI{1257}{\lux}}) & 
    \valeur{85th} (\valeur{\SI{1414}{\lux}}) & 
    \valeur{87.5th} (\valeur{\SI{1541}{\lux}}) & 
    \valeur{90th} (\valeur{\SI{1771}{\lux}}) \\
    \includegraphics[width=\tmplength]{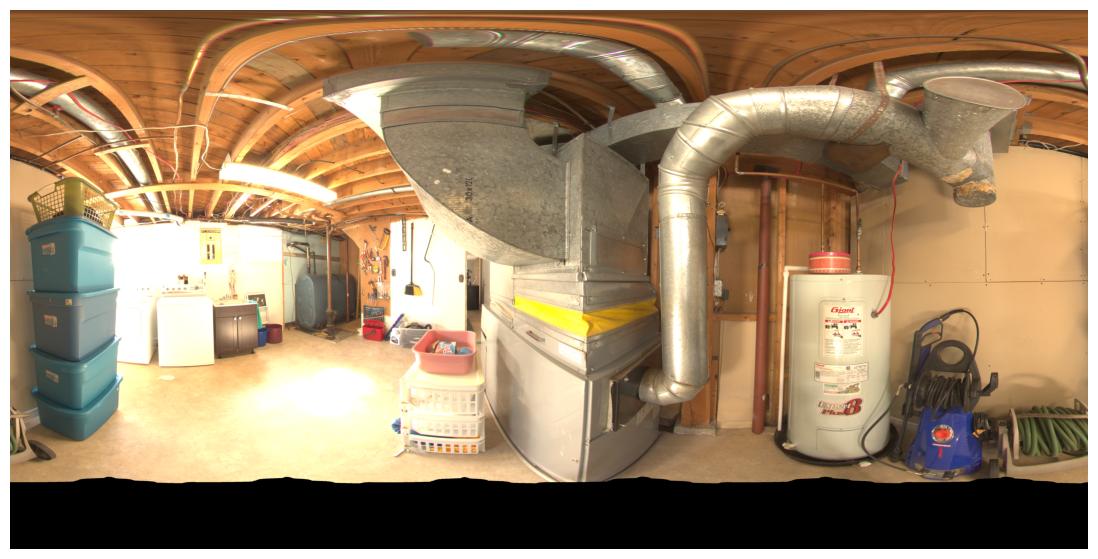}&
    \includegraphics[width=\tmplength]{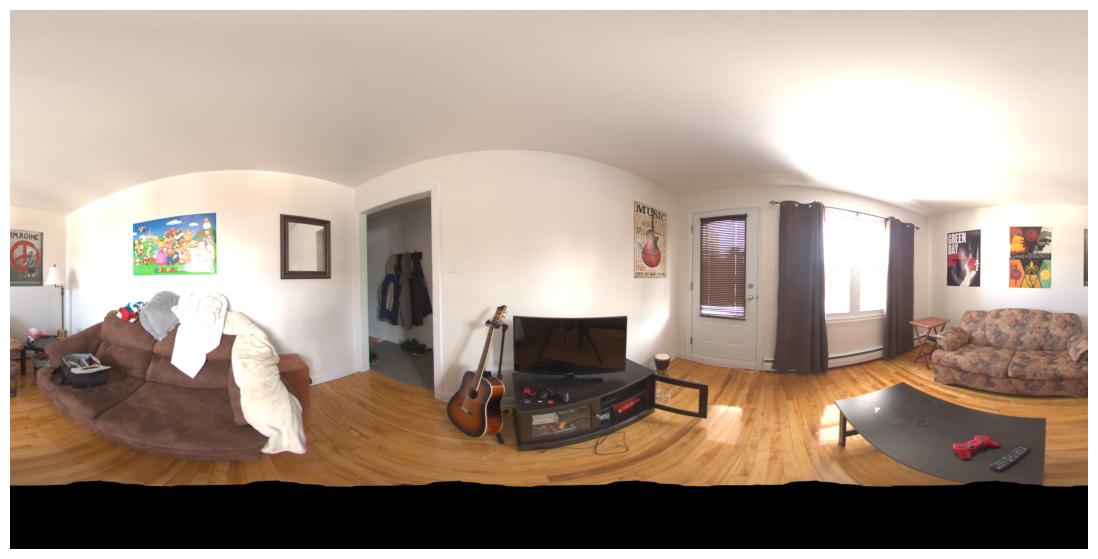}&
    \includegraphics[width=\tmplength]{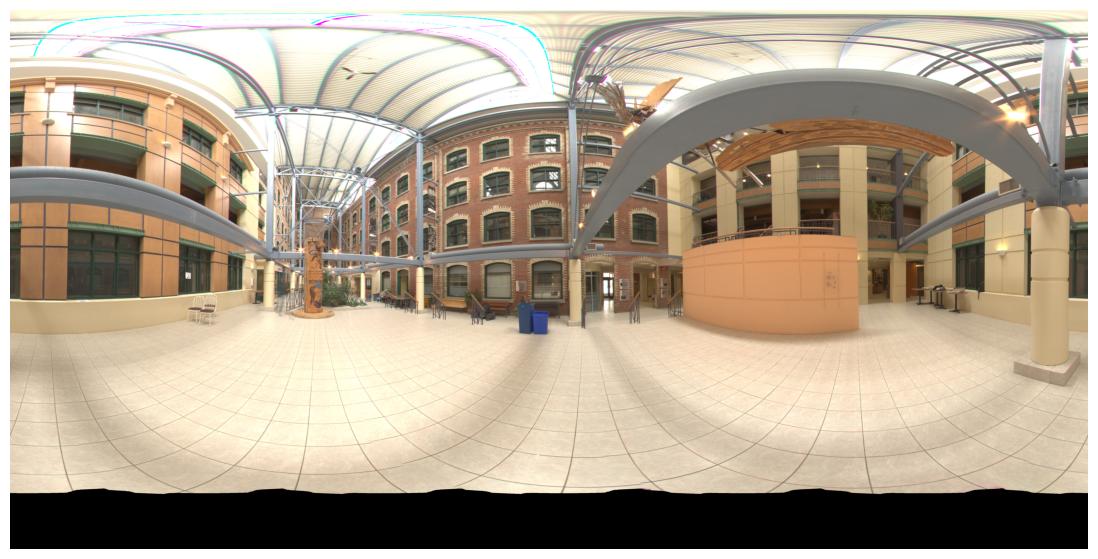}&
    \includegraphics[width=\tmplength]{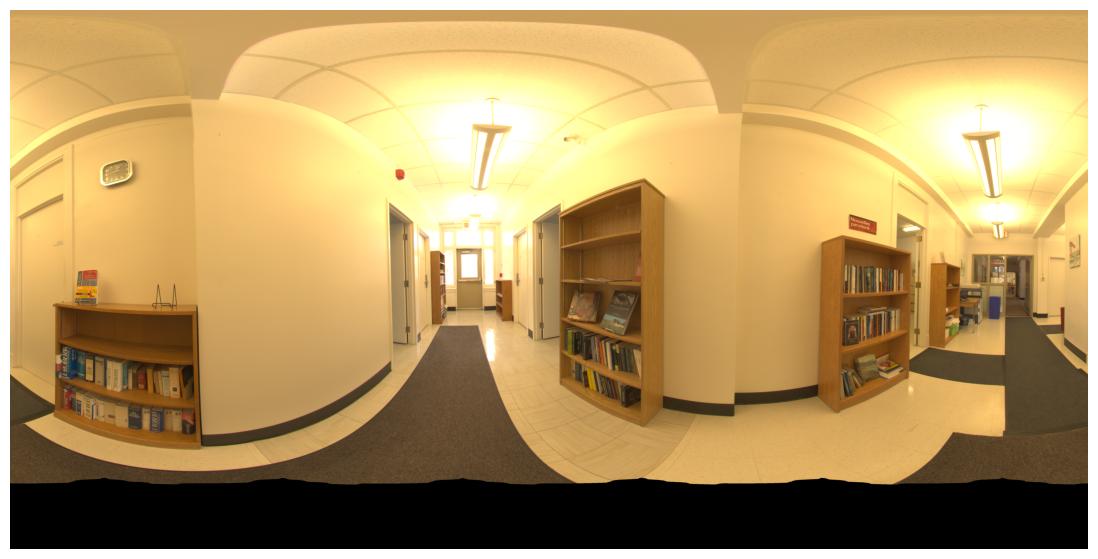}&
    \includegraphics[width=\tmplength]{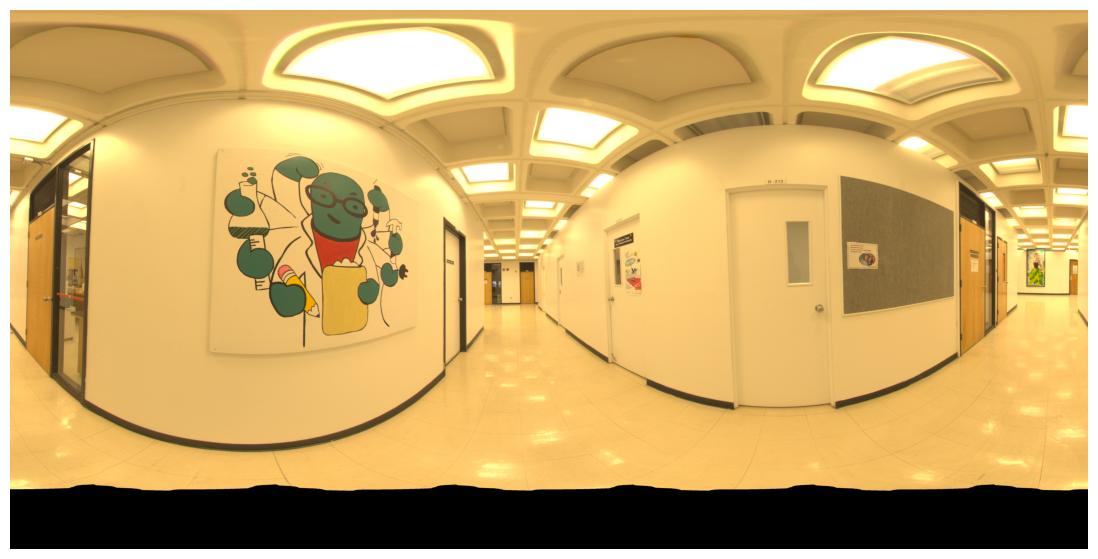}&
    \includegraphics[width=\tmplength]{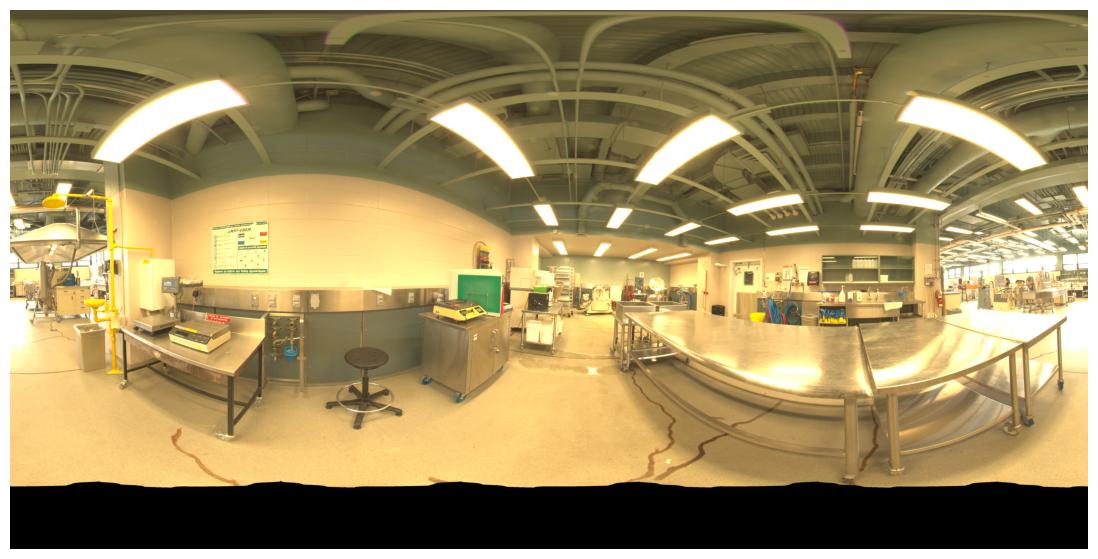}\\
    \includegraphics[width=\tmplength]{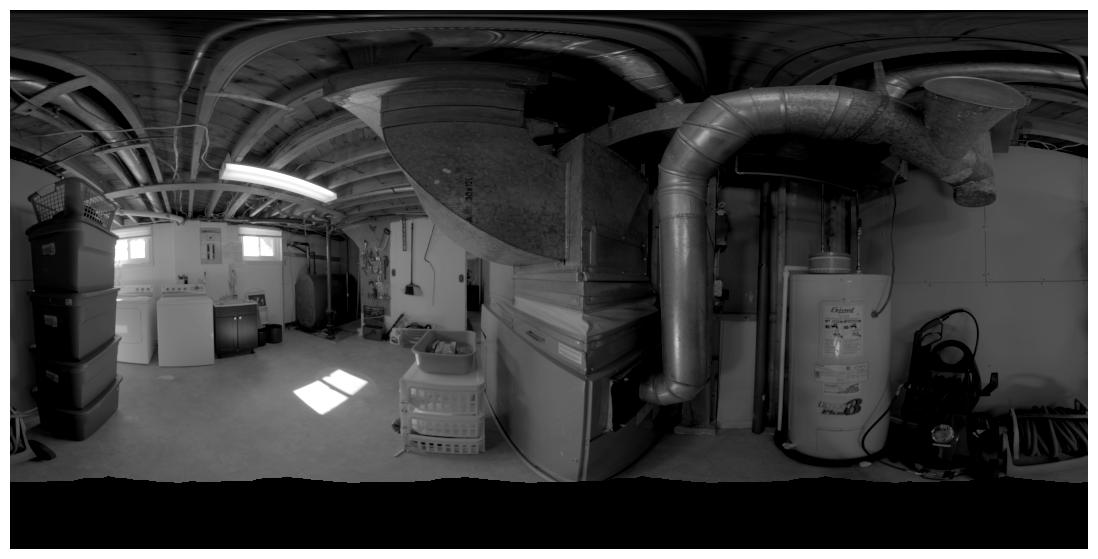}&
    \includegraphics[width=\tmplength]{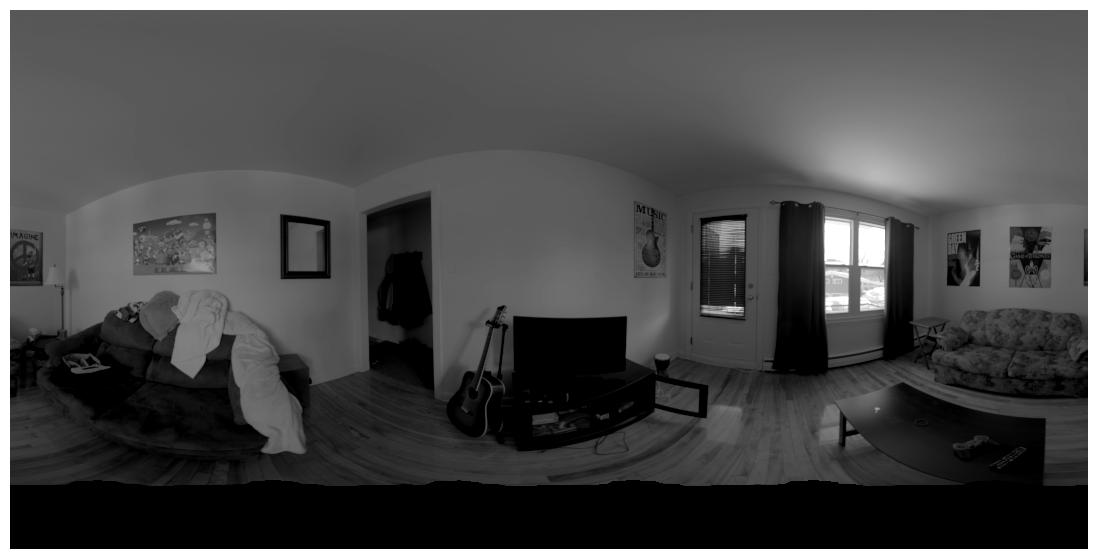}&
    \includegraphics[width=\tmplength]{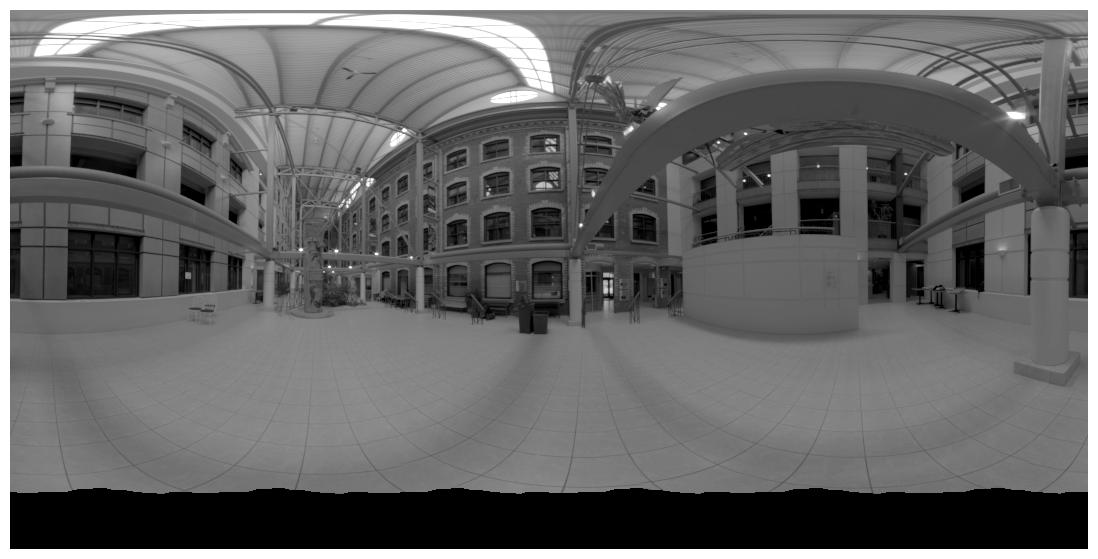}&
    \includegraphics[width=\tmplength]{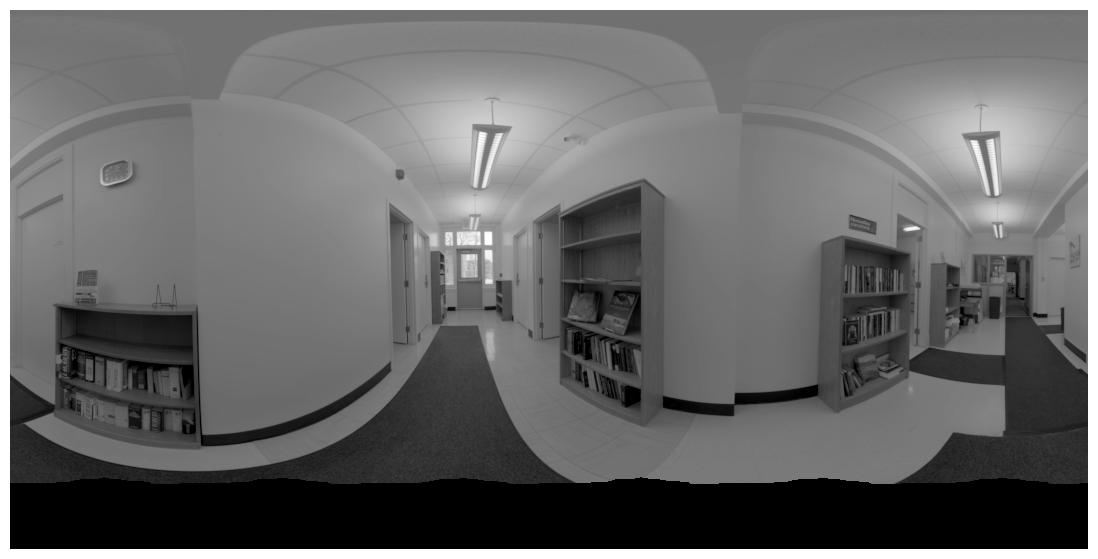}&
    \includegraphics[width=\tmplength]{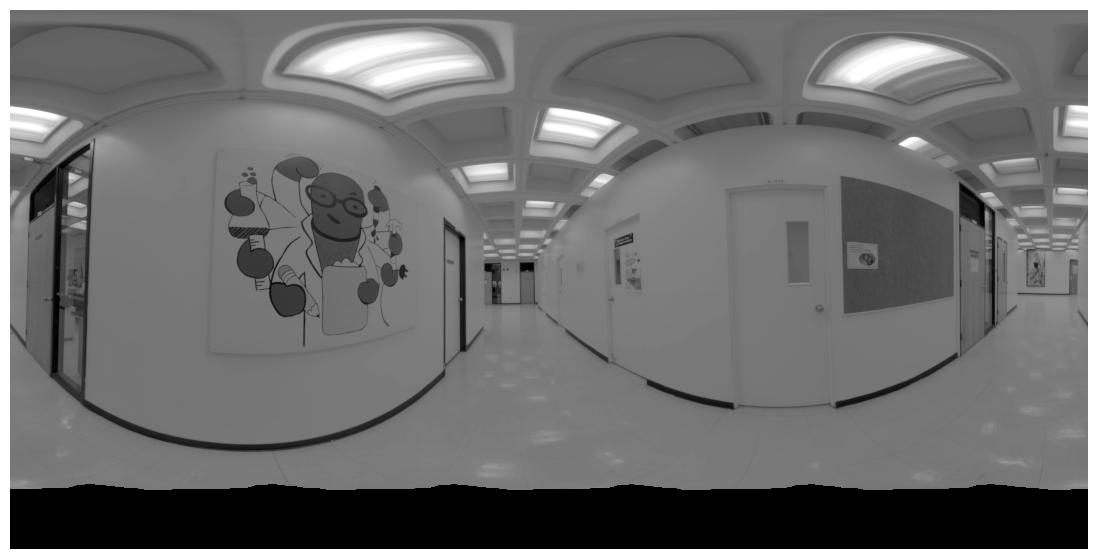}&
    \includegraphics[width=\tmplength]{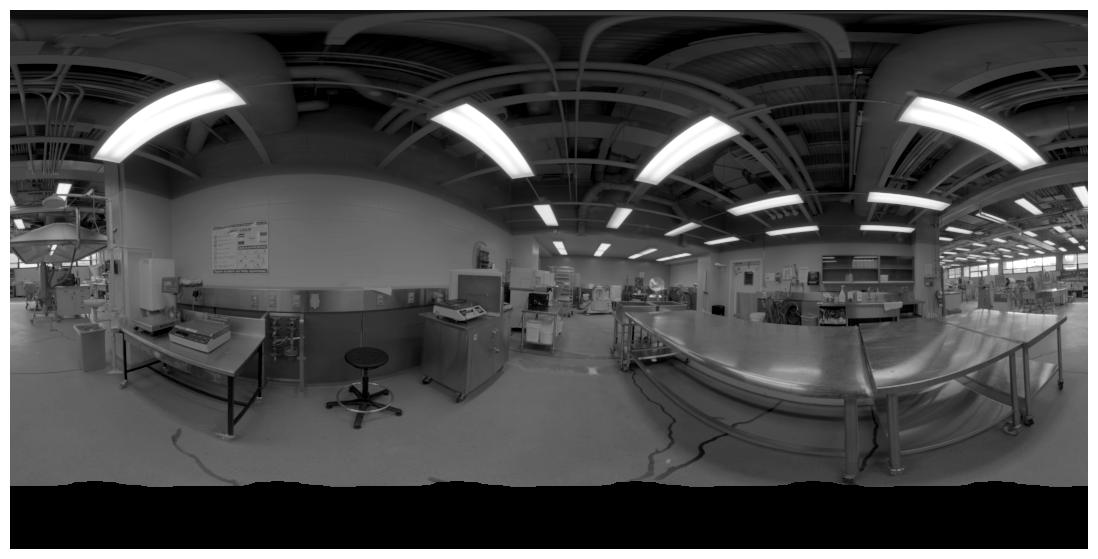}\\
    
    \valeur{92.5th} (\valeur{\SI{2099}{\lux}}) & 
    \valeur{95th} (\valeur{\SI{2723}{\lux}}) & 
    \valeur{97.5th}{\%} (\valeur{\SI{4345}{\lux}}) & 
    \valeur{99th} (\valeur{\SI{7000}{\lux}}) & 
    \valeur{99.9th} (\valeur{\SI{23411}{\lux}}) & 
    \valeur{Max} (\valeur{\SI{32431}{\lux}}) \\
    \includegraphics[width=\tmplength]{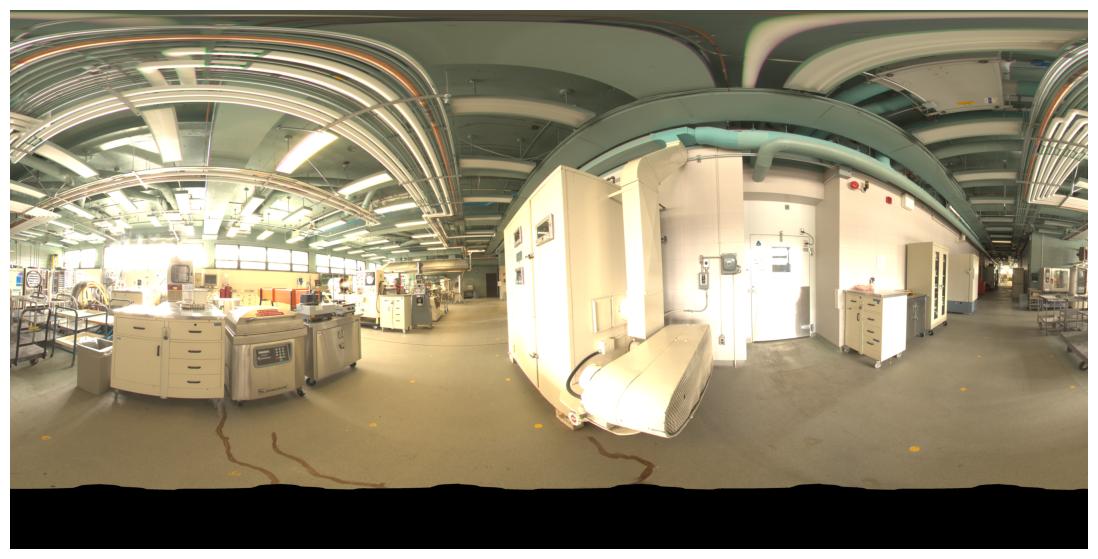}&
    \includegraphics[width=\tmplength]{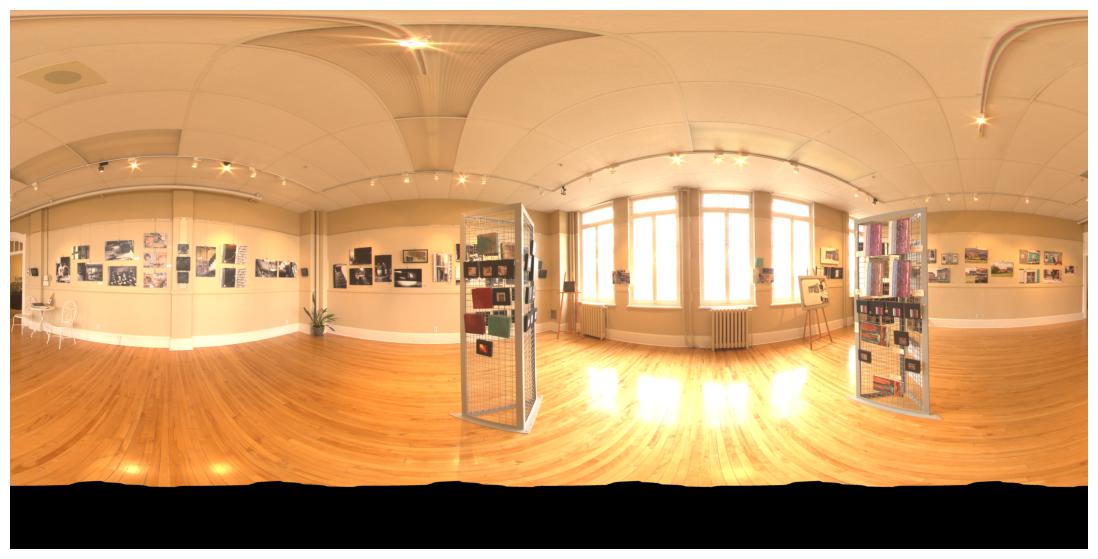}&
    \includegraphics[width=\tmplength]{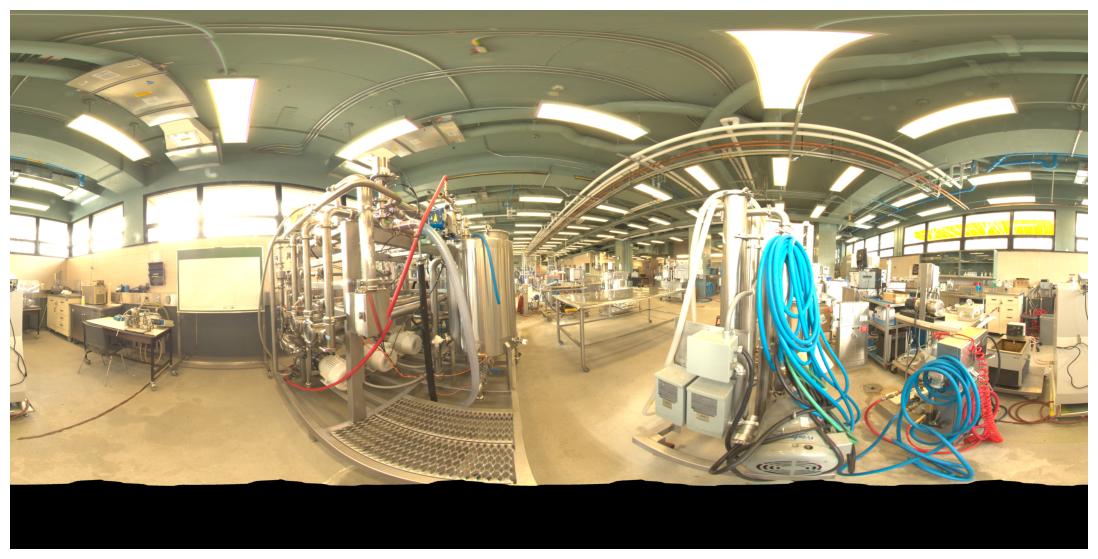}&
    \includegraphics[width=\tmplength]{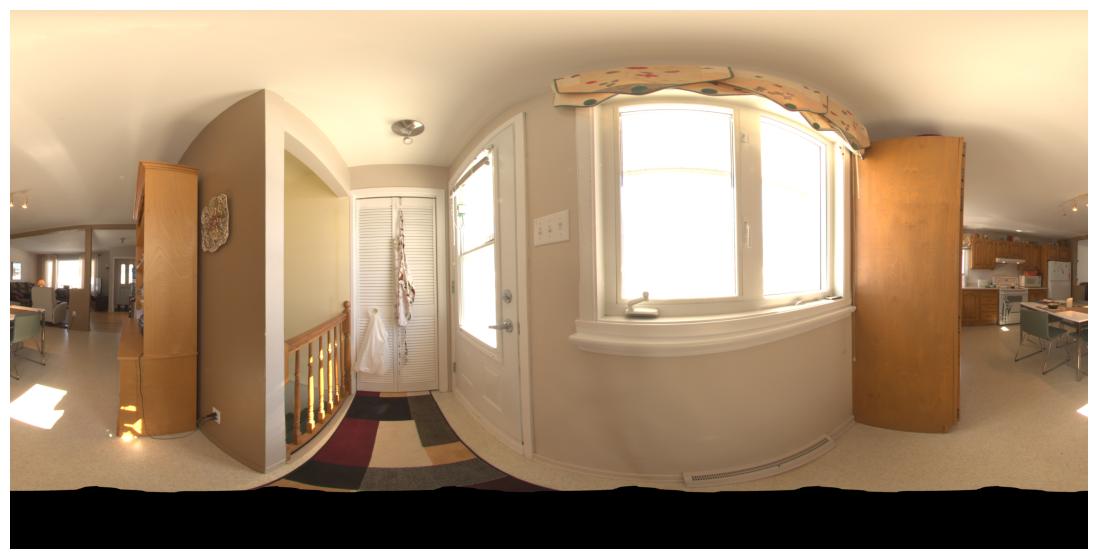}&
    \includegraphics[width=\tmplength]{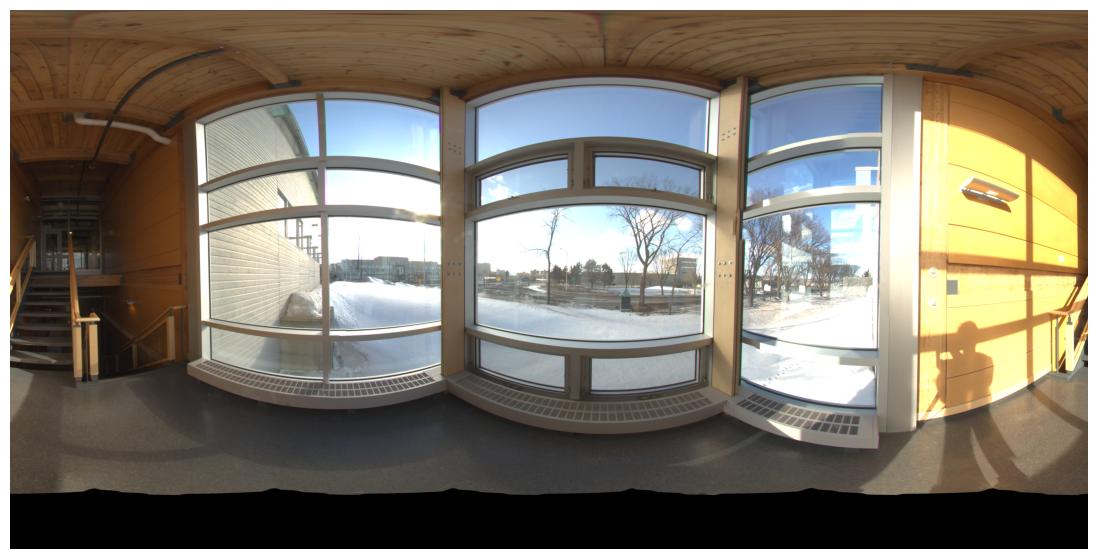}&
    \includegraphics[width=\tmplength]{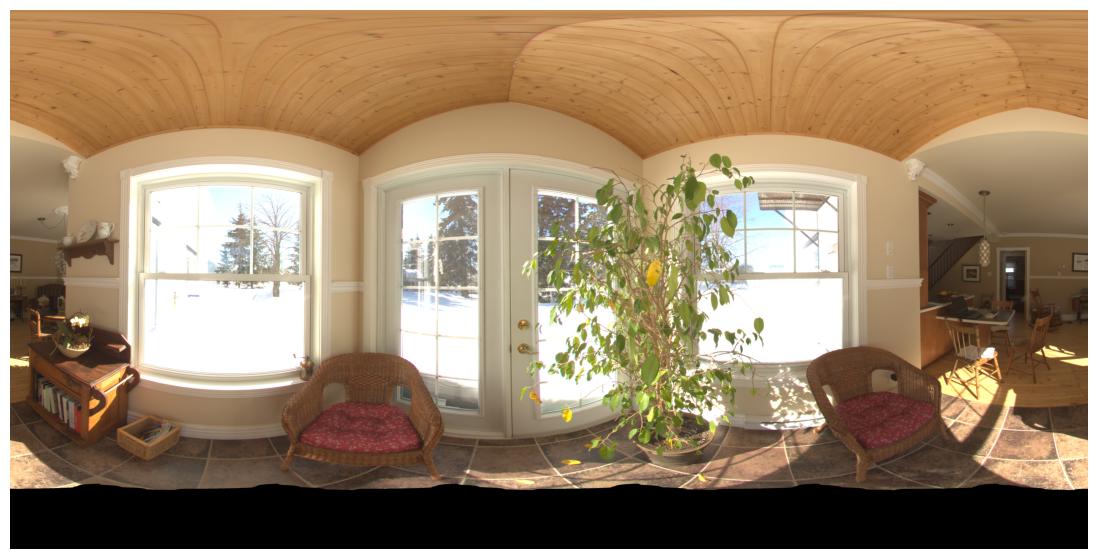}\\
    \includegraphics[width=\tmplength]{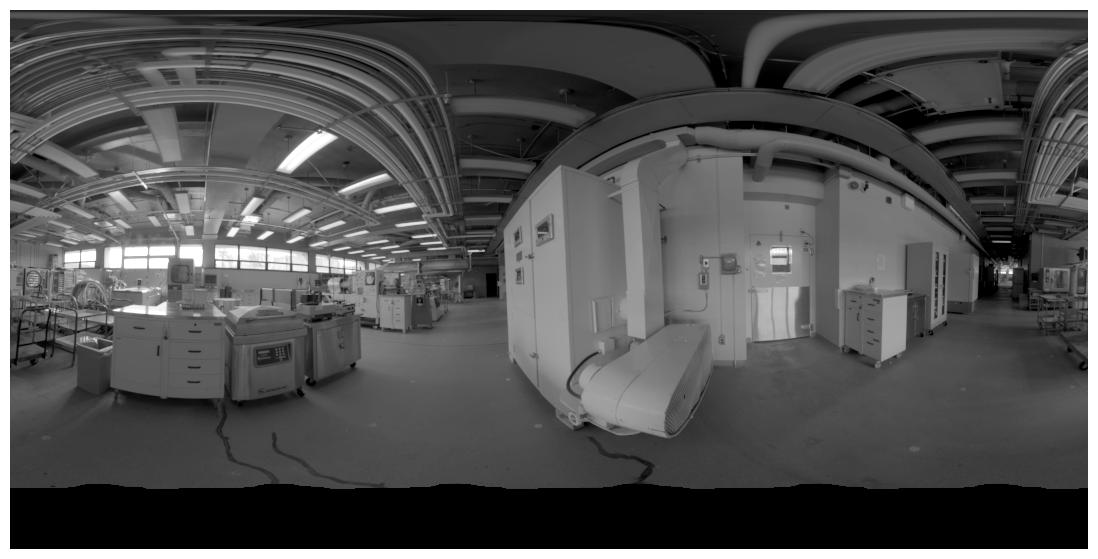}&
    \includegraphics[width=\tmplength]{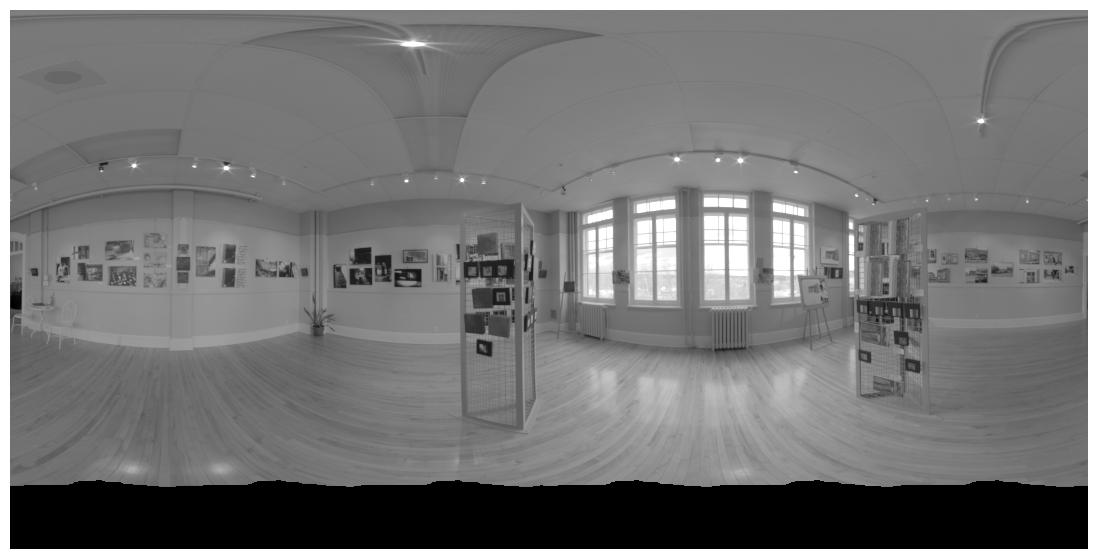}&
    \includegraphics[width=\tmplength]{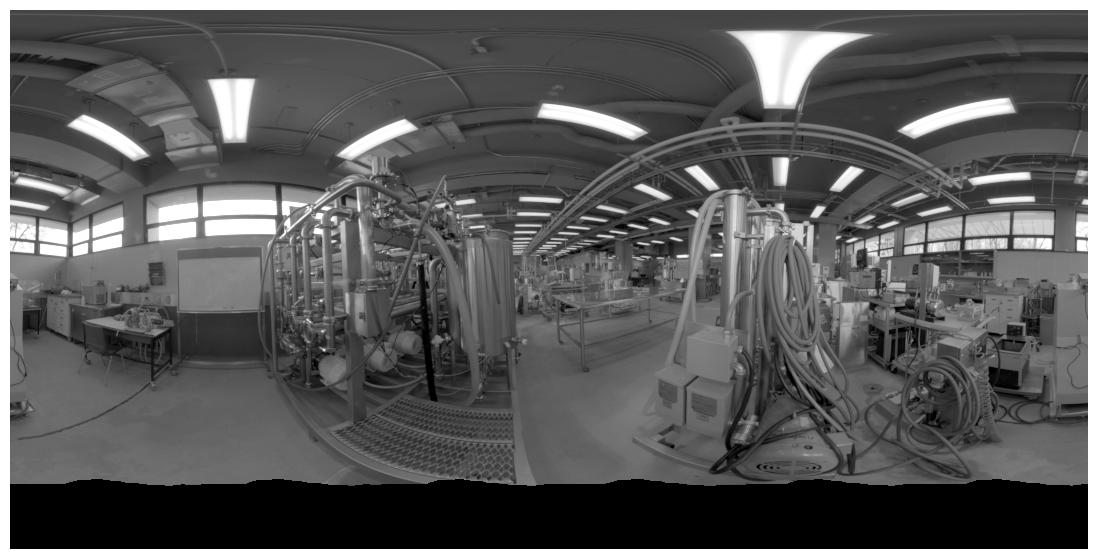}&
    \includegraphics[width=\tmplength]{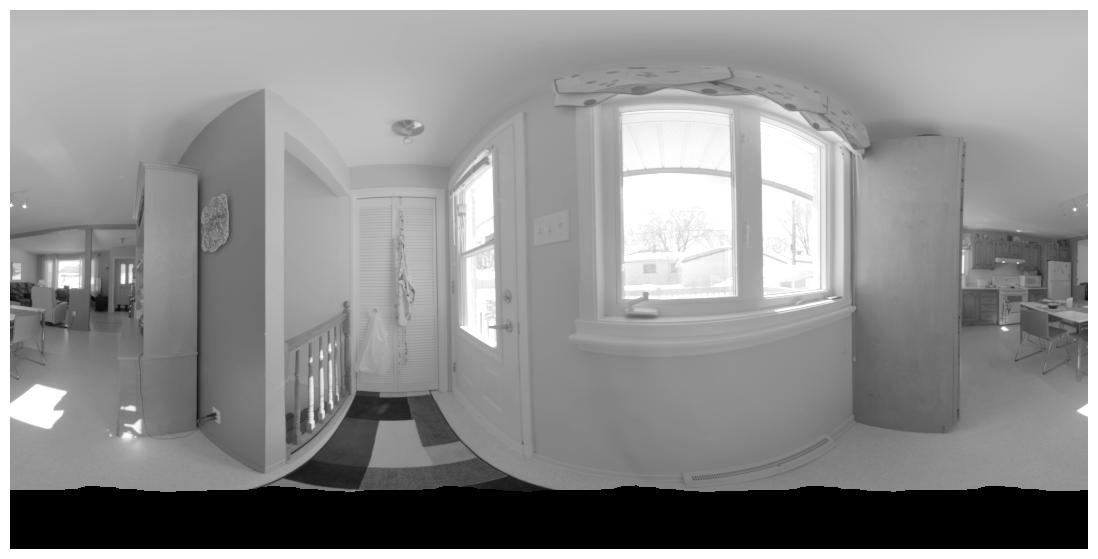}&
    \includegraphics[width=\tmplength]{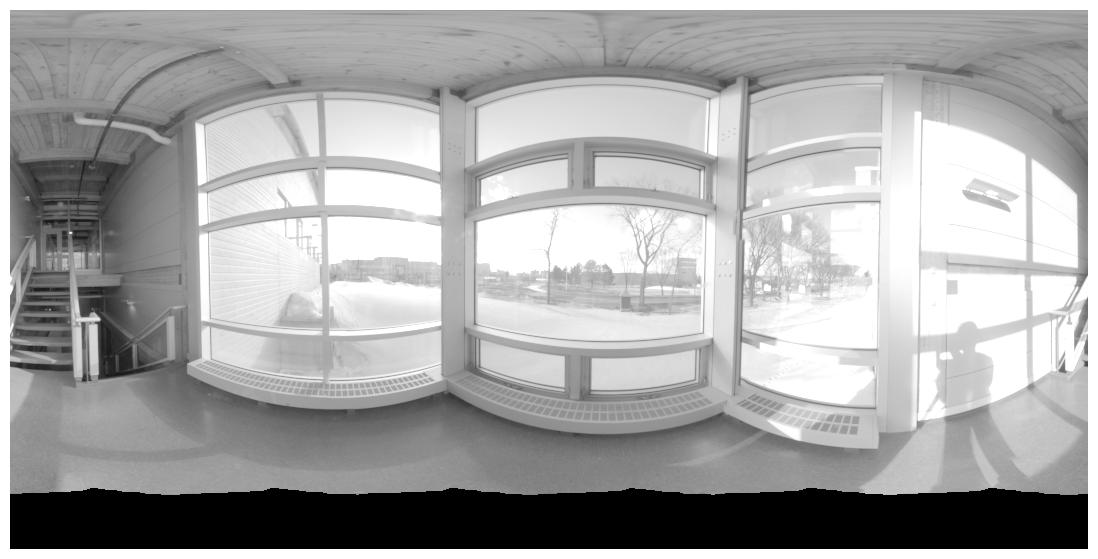}&
    \includegraphics[width=\tmplength]{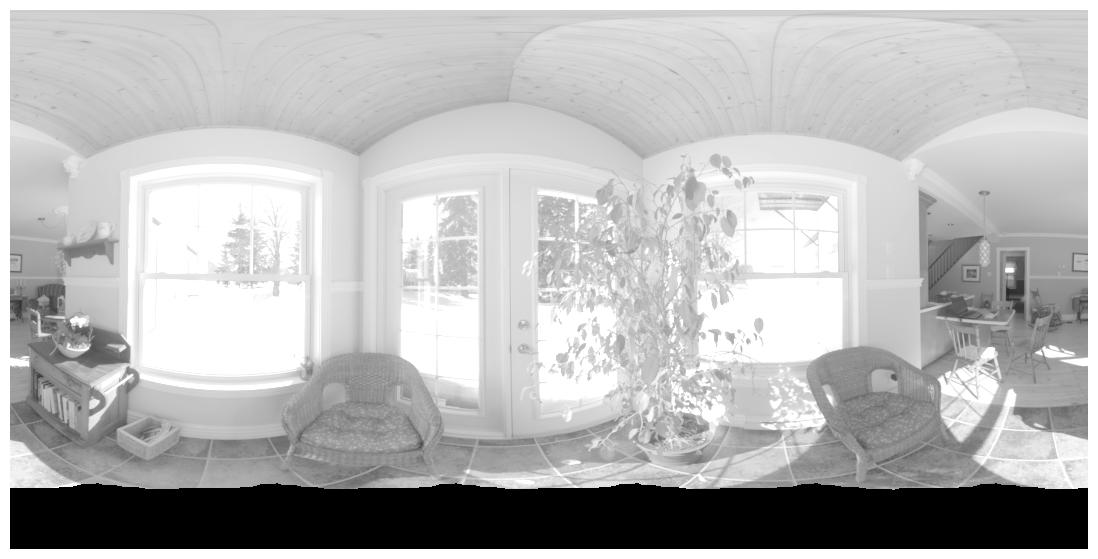}\\
    \end{tabular}
    \caption{Example scenes with mean spherical illuminance (MSI) close to the quantile values to complement fig.~3 from the main paper. Greyscale images below show the corresponding log-luminance maps.  The percentiles and corresponding measured MSI are indicated above the images.  Images are reexposed and tonemapped ($\gamma = \valeur{\num{2.2}}$) for display. Luminance color map: \includegraphics[width=3cm,trim=0.2cm 0.65cm 0.2cm 0.2cm,clip]{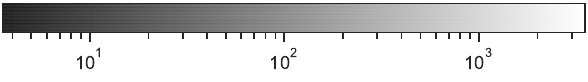}} 
    \label{fig:distribution_illuminance_examples_supp}
\end{figure*}

To complement fig. 3 of the paper, more examples of scenes contained in the dataset are presented in \cref{fig:distribution_illuminance_examples_supp}, sorted by their mean spherical illuminance (MSI), with their value close to the quantile indicated.  Below are shown the log-luminance maps associated to the scene.

\begin{figure*}
   \centering
   \scriptsize
   \setlength{\tabcolsep}{1pt} 
   \setlength{\tmplength}{0.15\linewidth}
    \begin{tabular}{cccccc}
    \valeur{Min} (\valeur{\SI{1619}{\K}}) & 
    \valeur{0.01th} (\valeur{\SI{1716}{\K}}) & 
    \valeur{1st} (\valeur{\SI{2199}{\K}}) & 
    \valeur{10th} (\valeur{\SI{2805}{\K}}) & 
    \valeur{12.5th} (\valeur{\SI{2910}{\K}}) & 
    \valeur{15th} (\valeur{\SI{3012}{\K}}) \\
    \includegraphics[width=\tmplength]{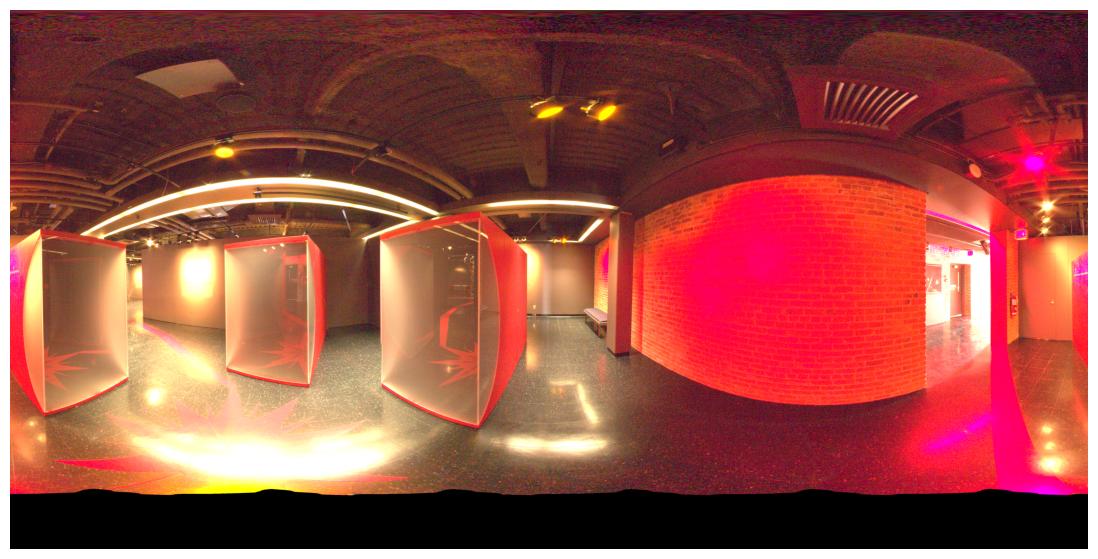}&
    \includegraphics[width=\tmplength]{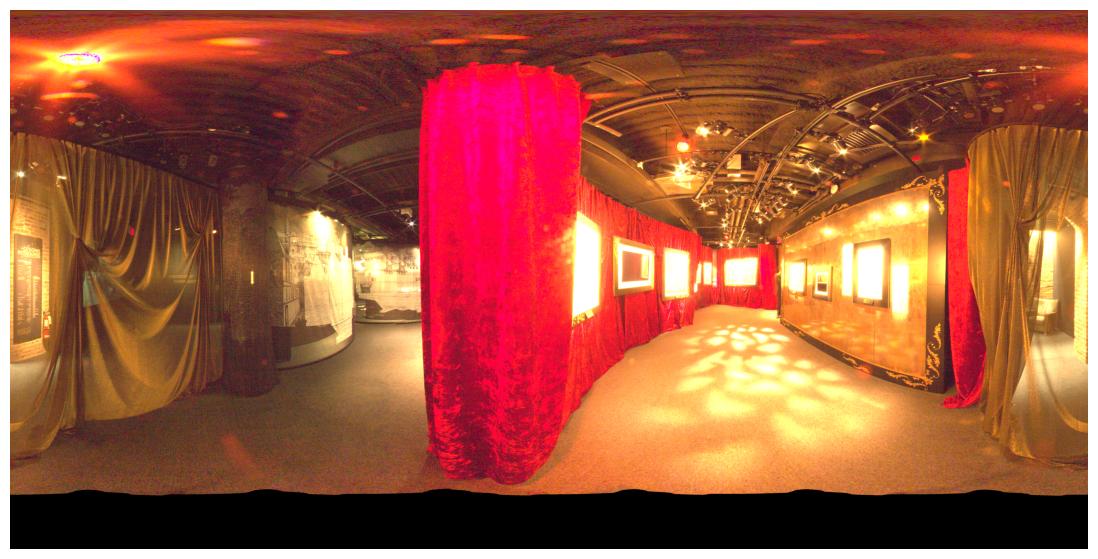}&
    \includegraphics[width=\tmplength]{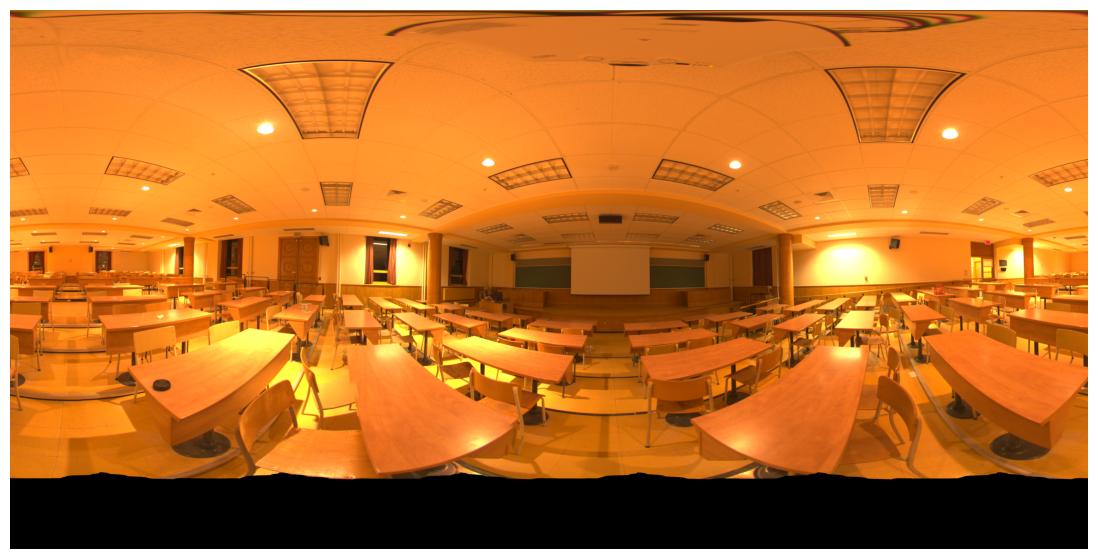}&
    \includegraphics[width=\tmplength]{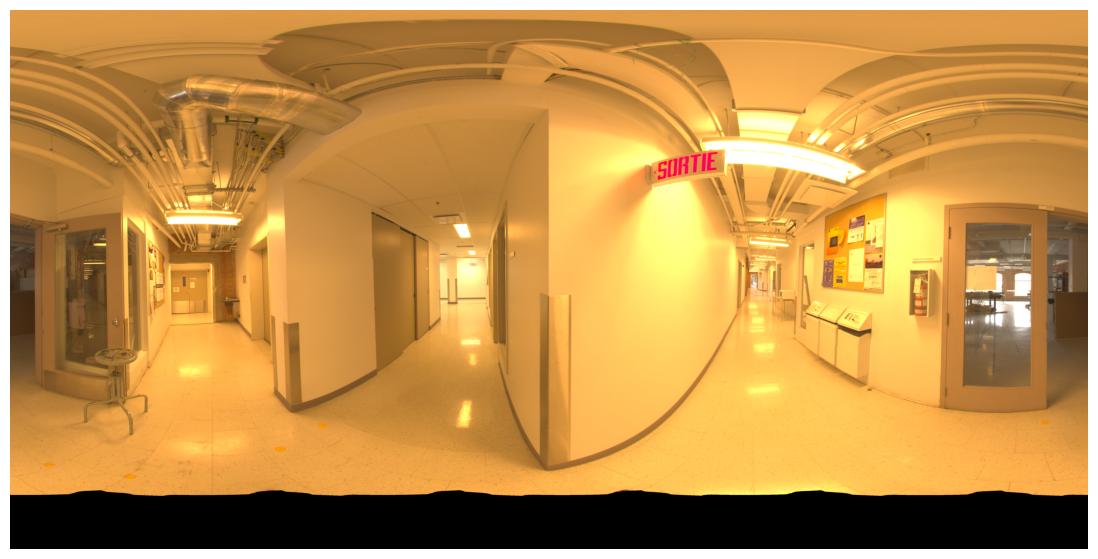}&
    \includegraphics[width=\tmplength]{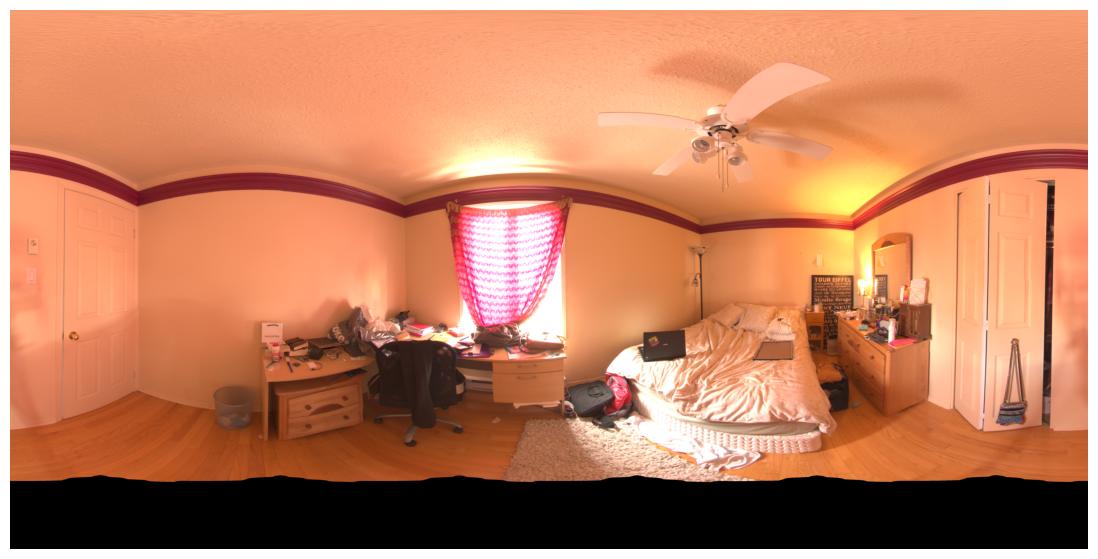}&
    \includegraphics[width=\tmplength]{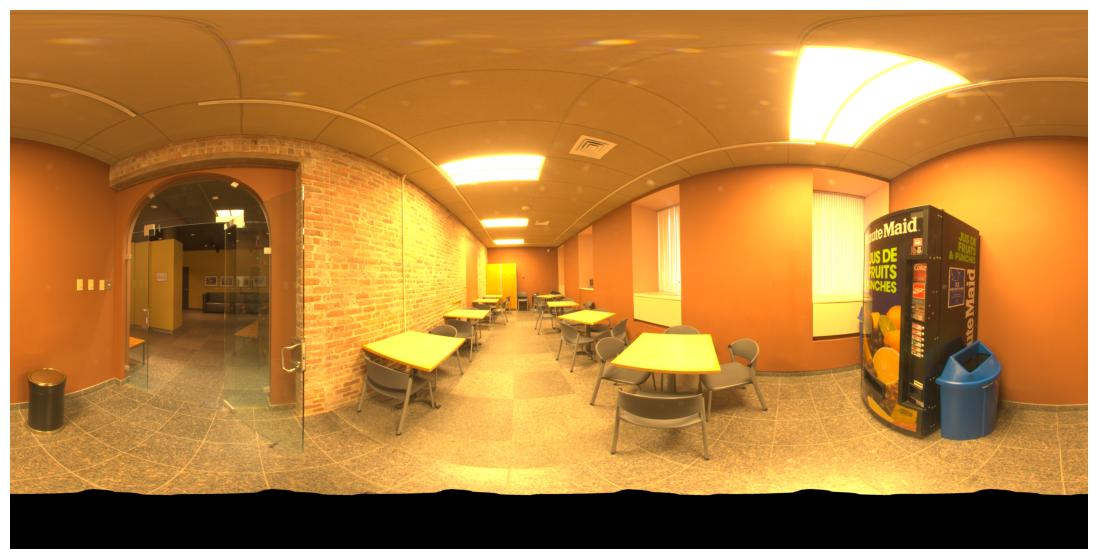}  \\
    \includegraphics[width=\tmplength]{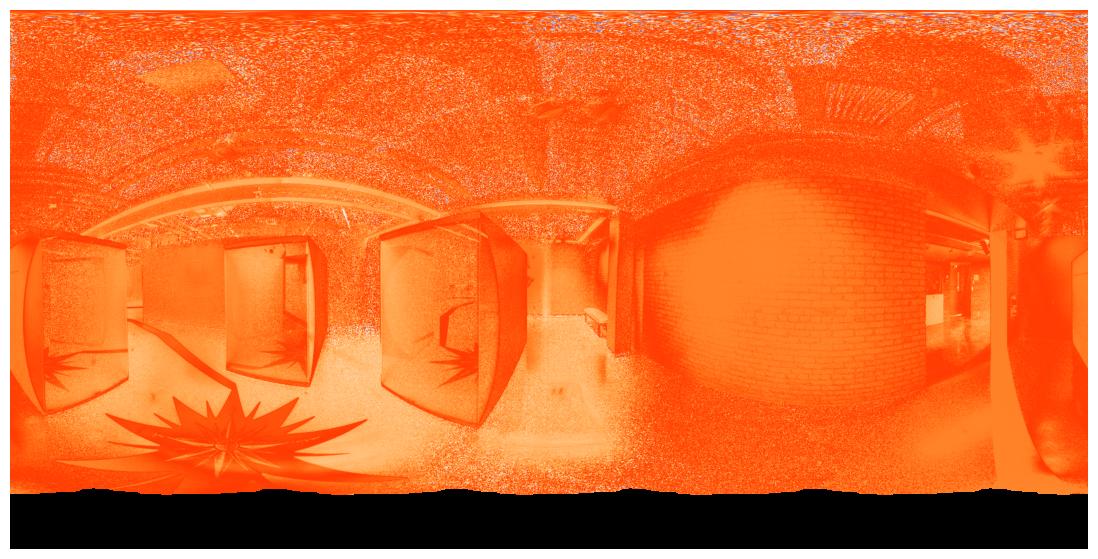}&
    \includegraphics[width=\tmplength]{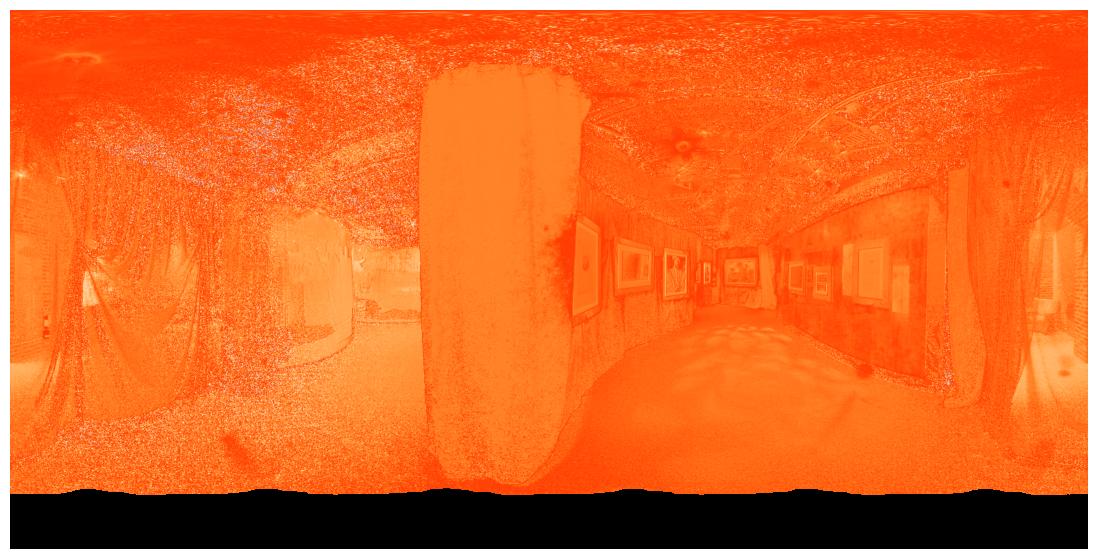}&
    \includegraphics[width=\tmplength]{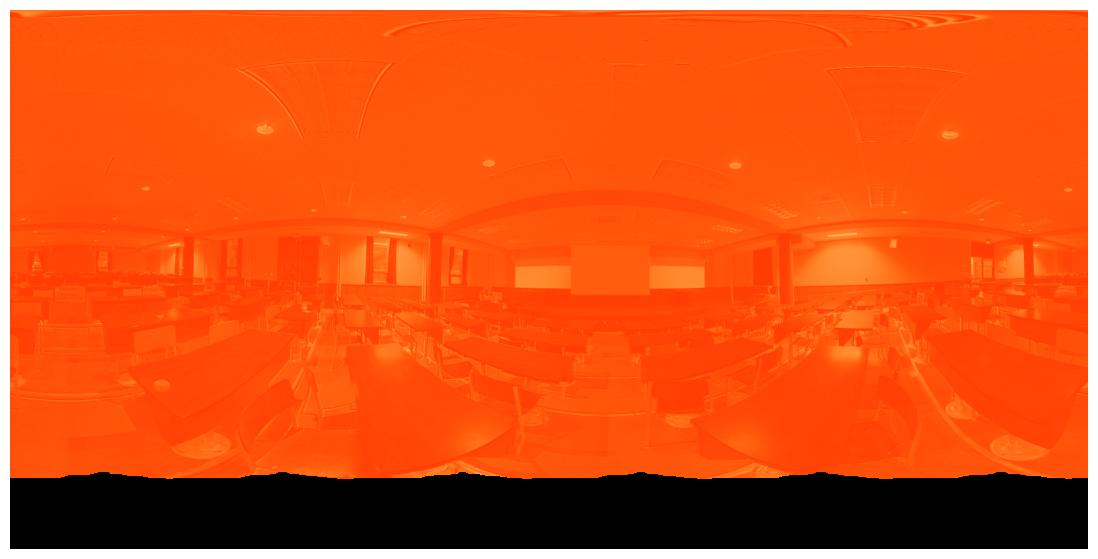}&
    \includegraphics[width=\tmplength]{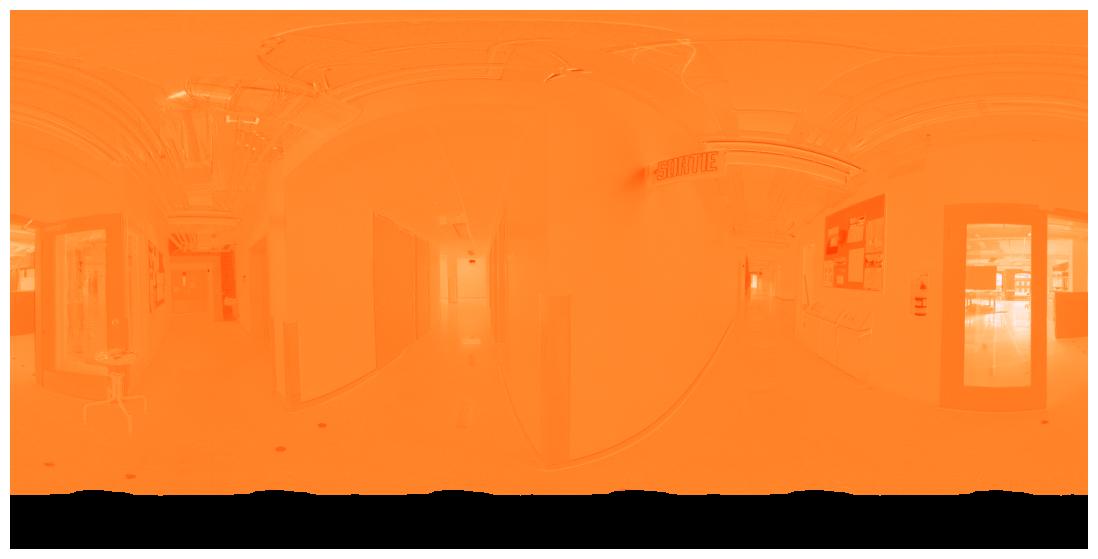}&
    \includegraphics[width=\tmplength]{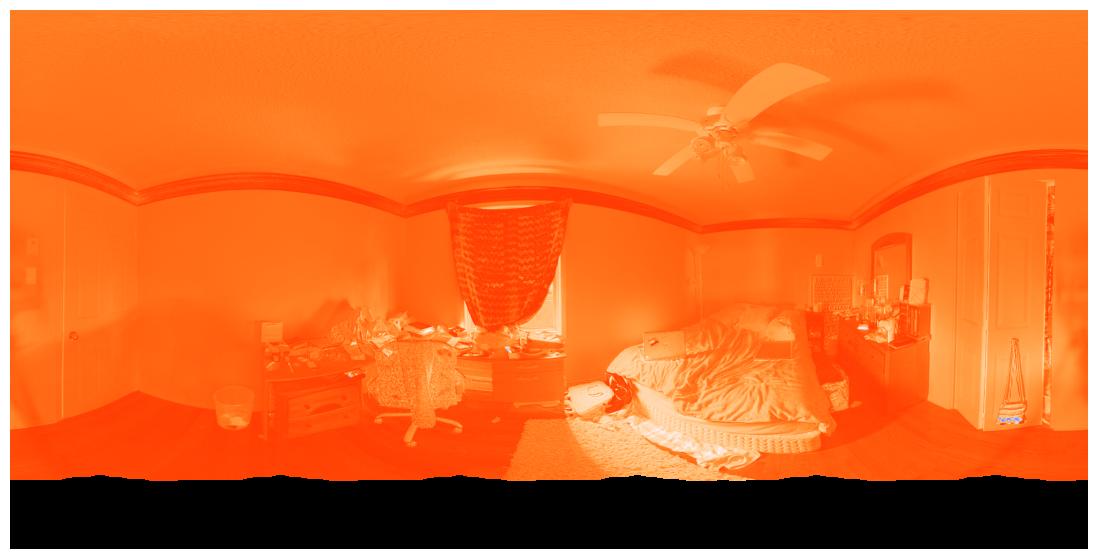}&
    \includegraphics[width=\tmplength]{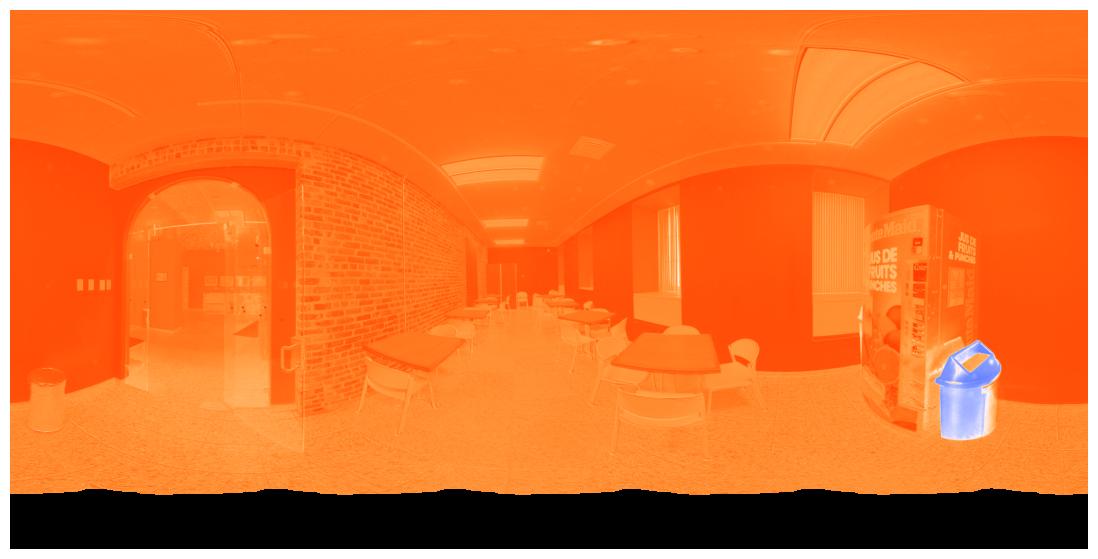}\\
    
    \valeur{17.5th} (\valeur{\SI{3113}{\K}}) & 
    \valeur{20th} (\valeur{\SI{3200}{\K}}) & 
    \valeur{22.5th} (\valeur{\SI{3253}{\K}}) & 
    \valeur{25th} (\valeur{\SI{3320}{\K}}) & 
    \valeur{27.5th} (\valeur{\SI{3360}{\K}}) & 
    \valeur{30th} (\valeur{\SI{3398}{\K}}) \\
    \includegraphics[width=\tmplength]{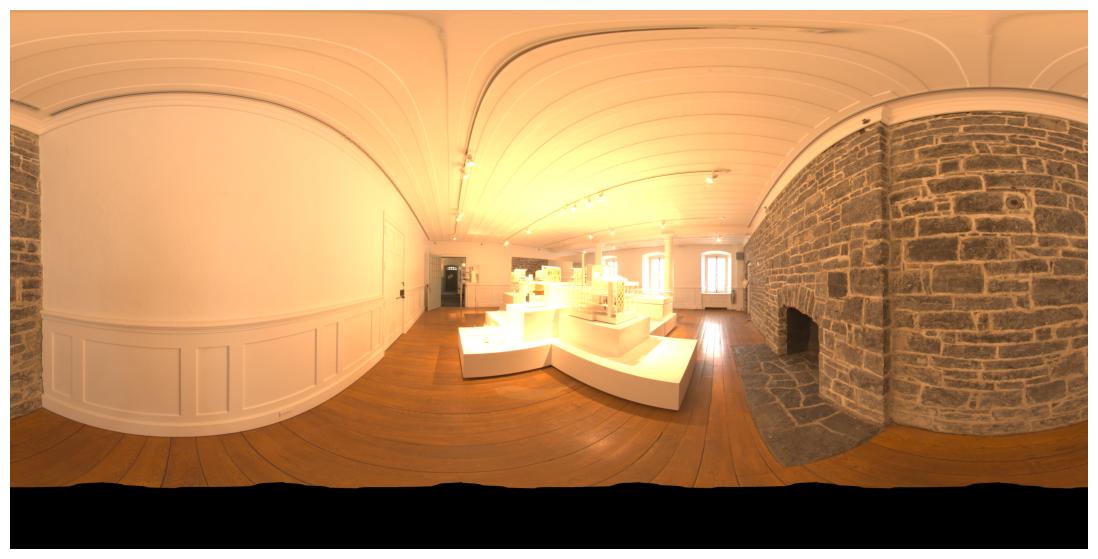}&
    \includegraphics[width=\tmplength]{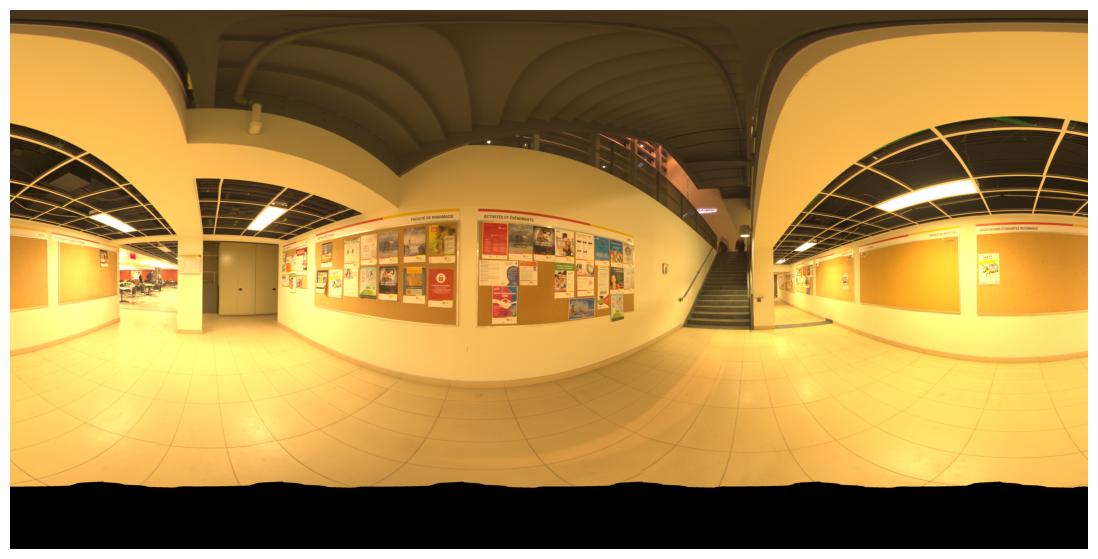}&
    \includegraphics[width=\tmplength]{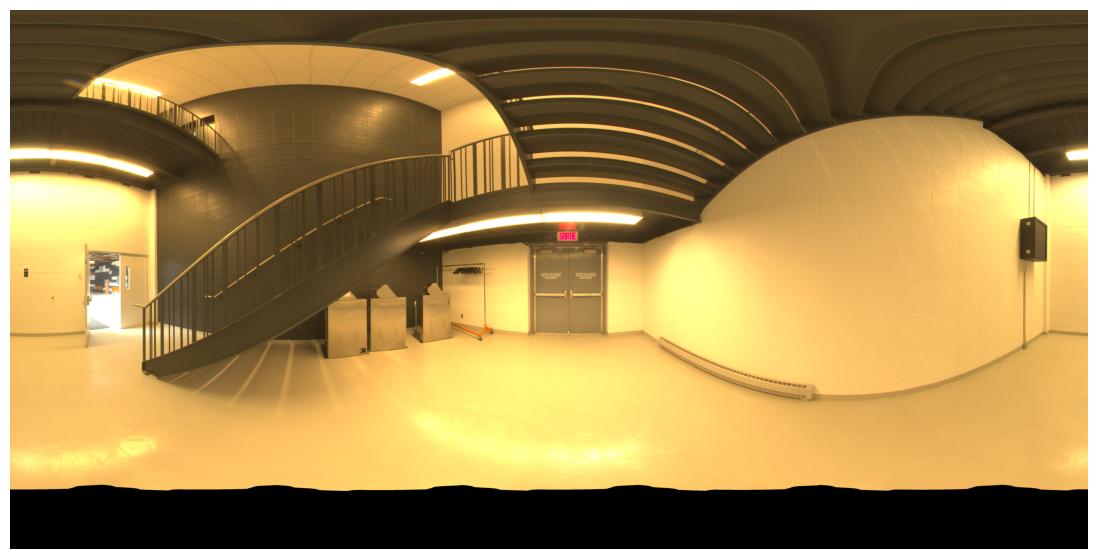}&
    \includegraphics[width=\tmplength]{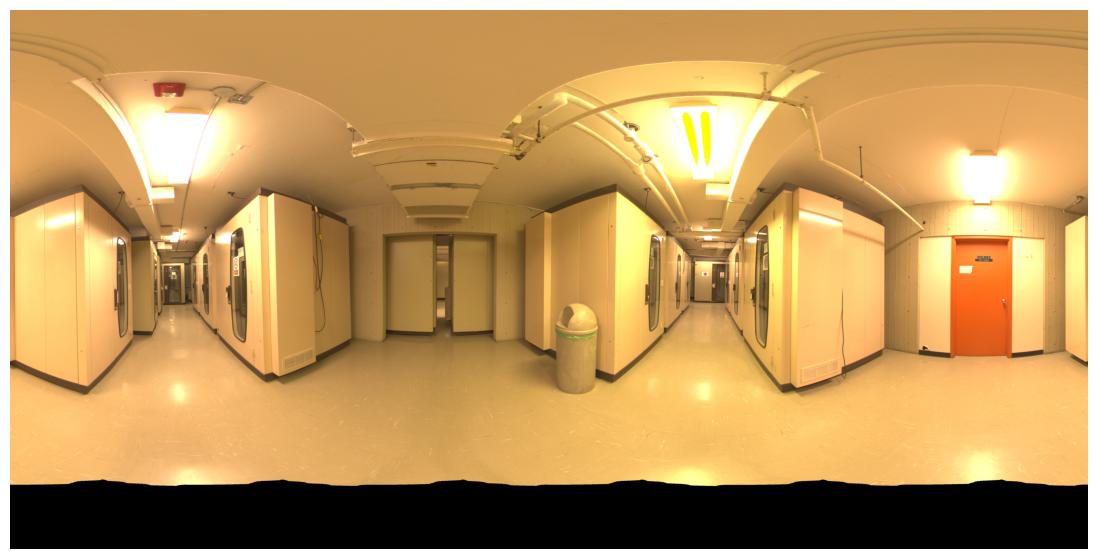}&
    \includegraphics[width=\tmplength]{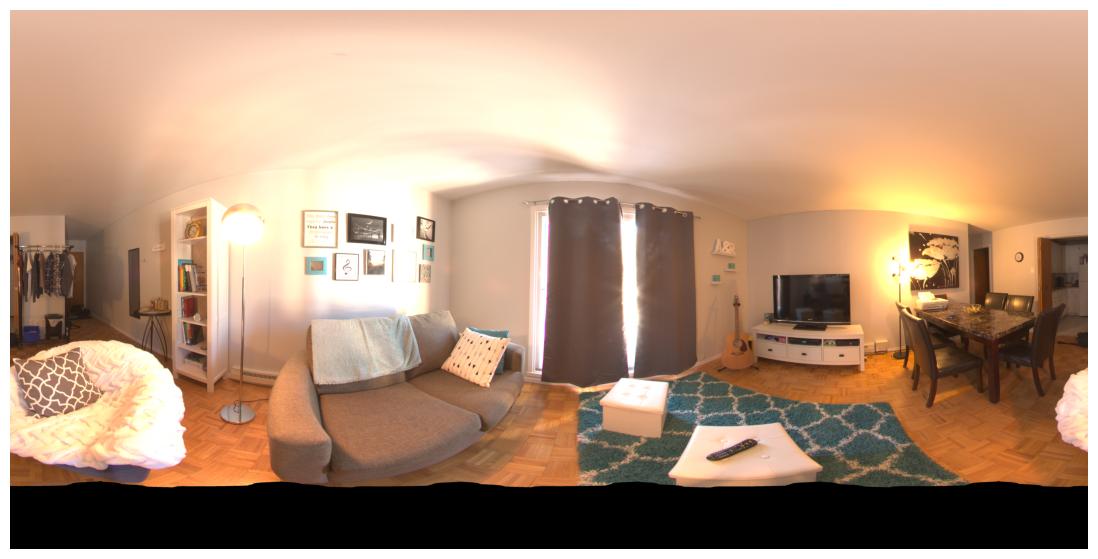}&
    \includegraphics[width=\tmplength]{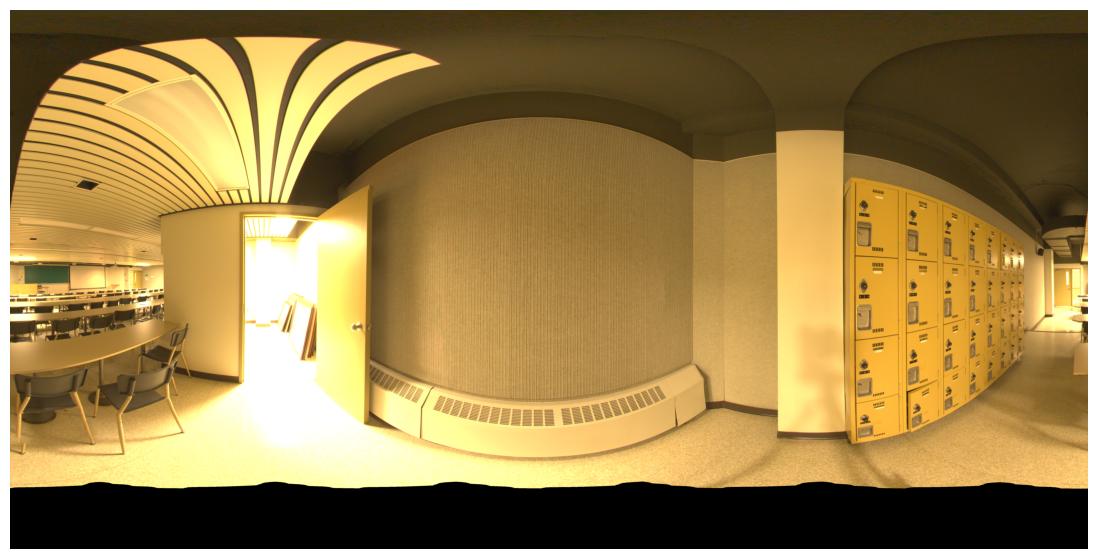}\\
    \includegraphics[width=\tmplength]{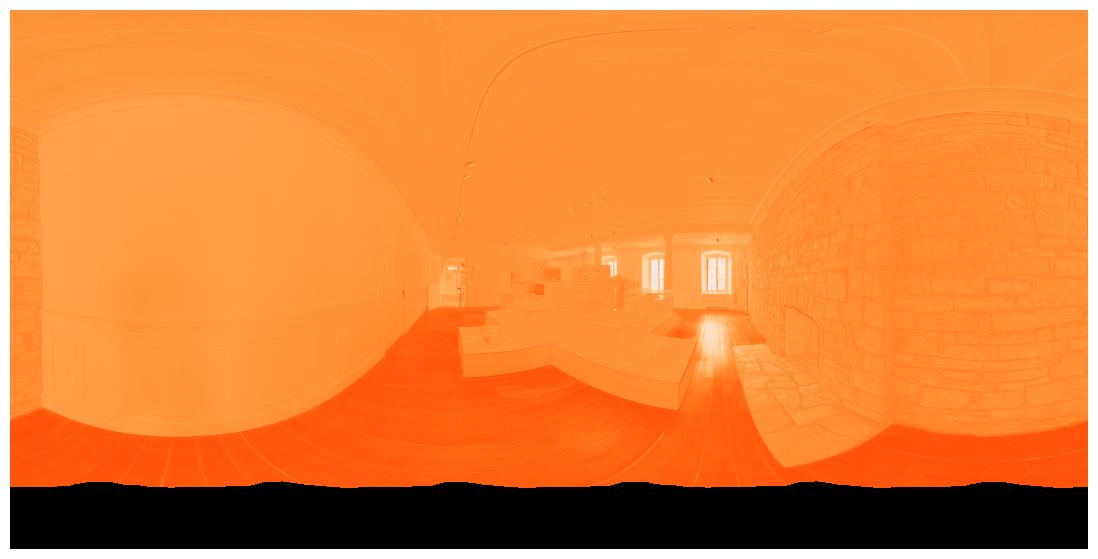}&
    \includegraphics[width=\tmplength]{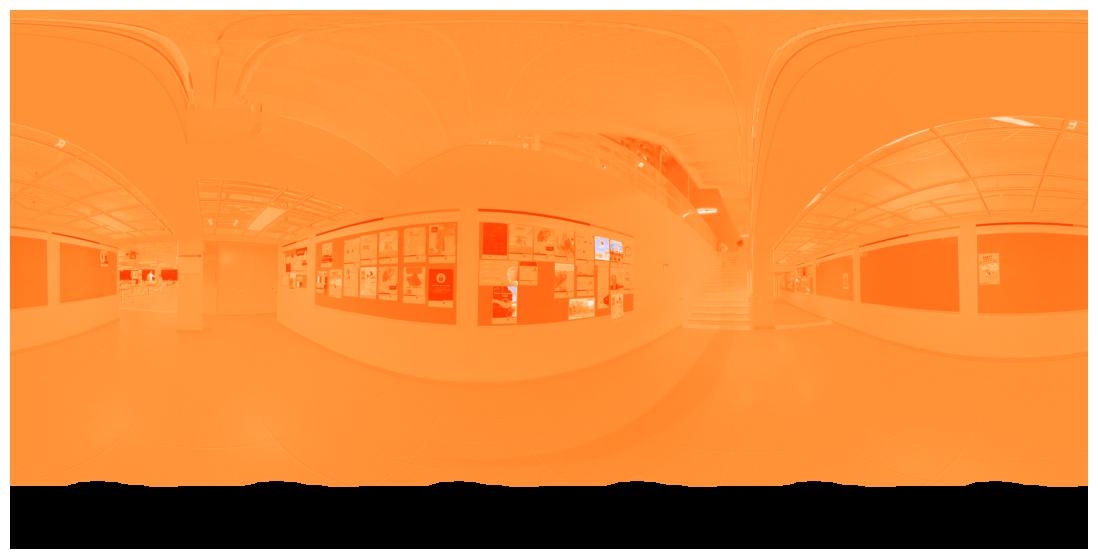}&
    \includegraphics[width=\tmplength]{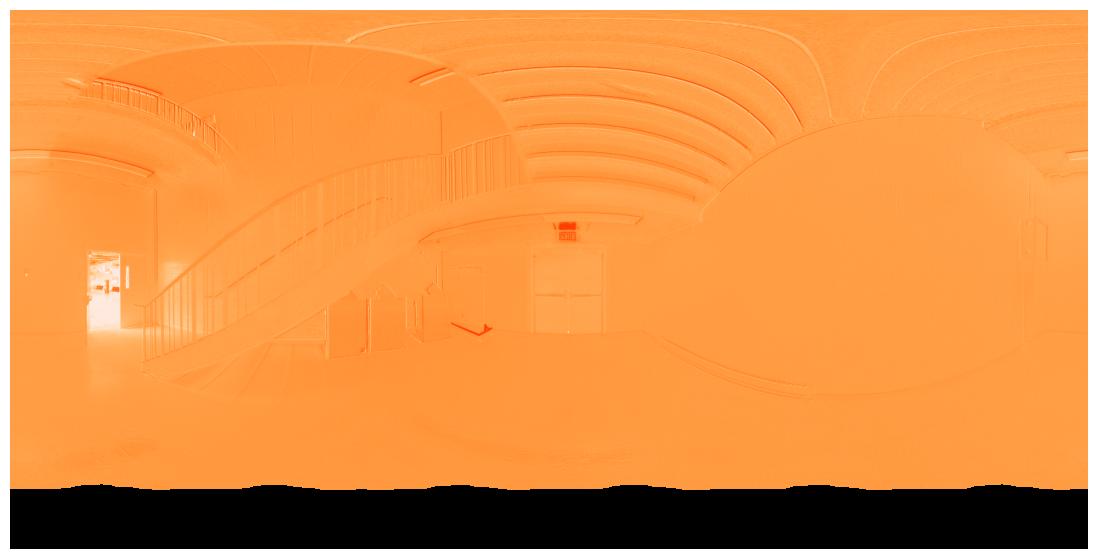}&
    \includegraphics[width=\tmplength]{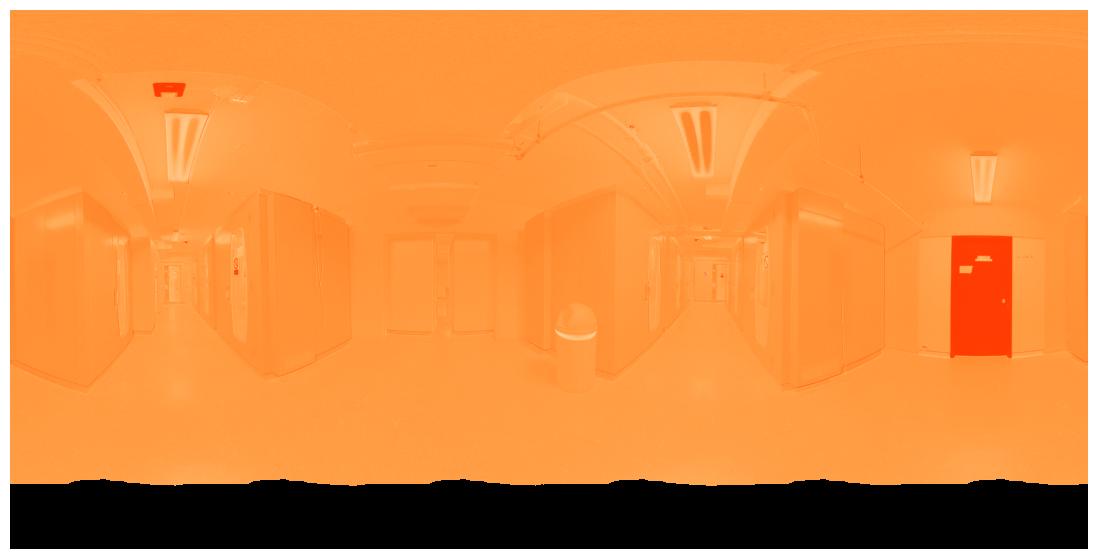}&
    \includegraphics[width=\tmplength]{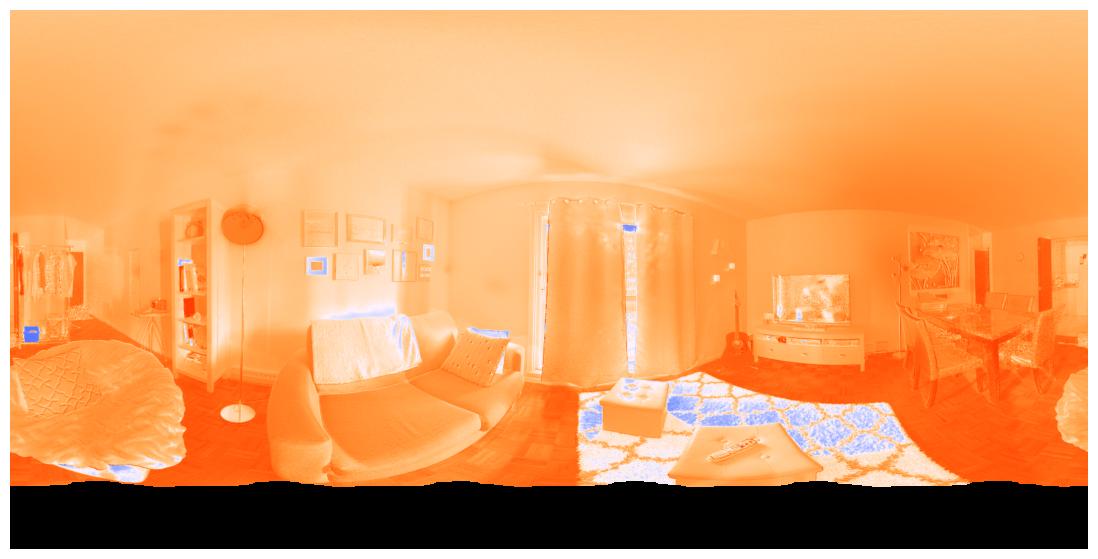}&
    \includegraphics[width=\tmplength]{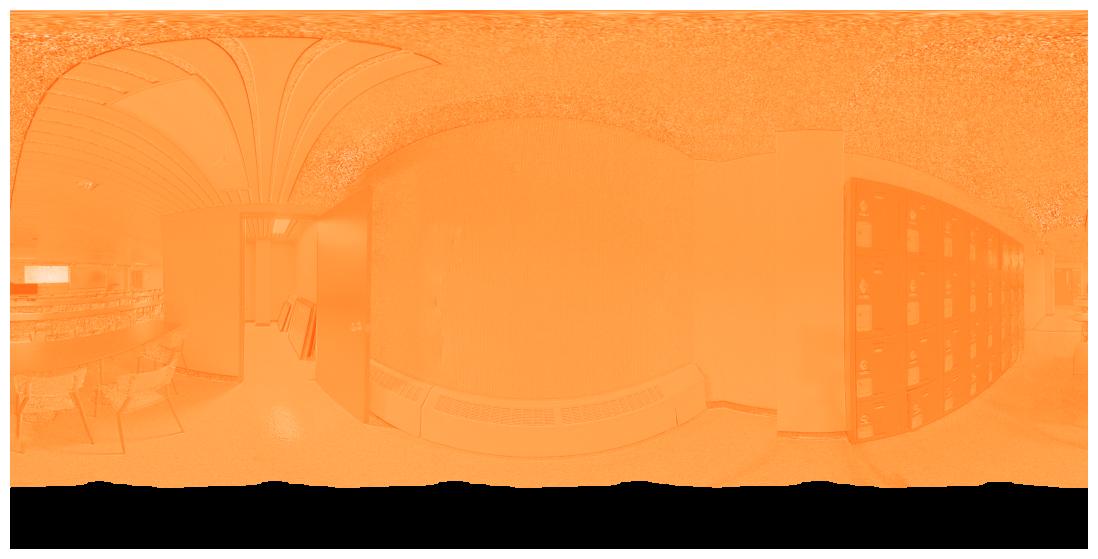}\\
    
    \valeur{32.5th} (\valeur{\SI{3435}{\K}}) & 
    \valeur{35th} (\valeur{\SI{3470}{\K}}) & 
    \valeur{37.5th} (\valeur{\SI{3507}{\K}}) & 
    \valeur{40th} (\valeur{\SI{3532}{\K}}) & 
    \valeur{42.5th} (\valeur{\SI{3558}{\K}}) & 
    \valeur{45th} (\valeur{\SI{3582}{\K}}) \\
    \includegraphics[width=\tmplength]{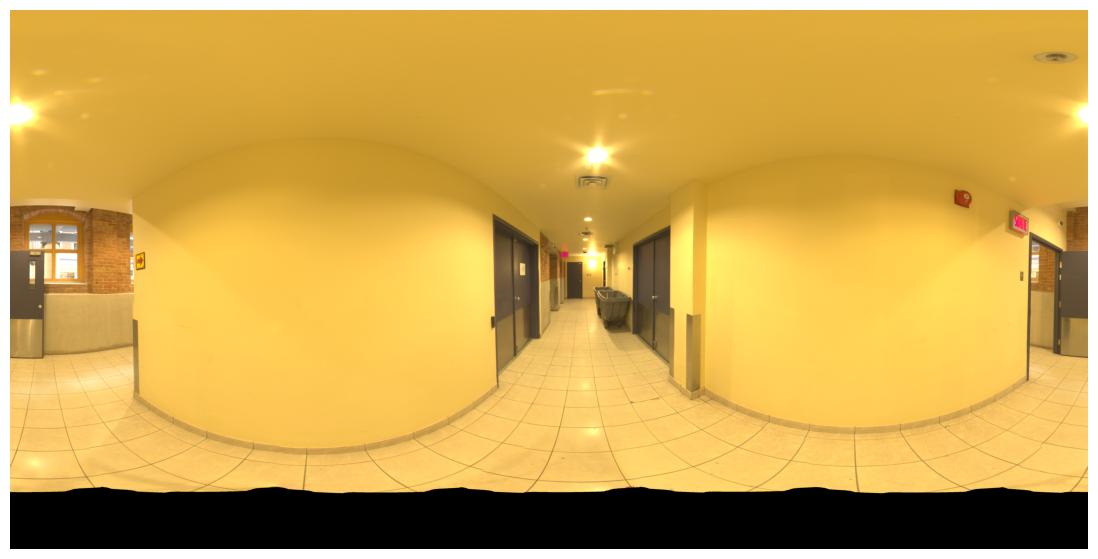}&
    \includegraphics[width=\tmplength]{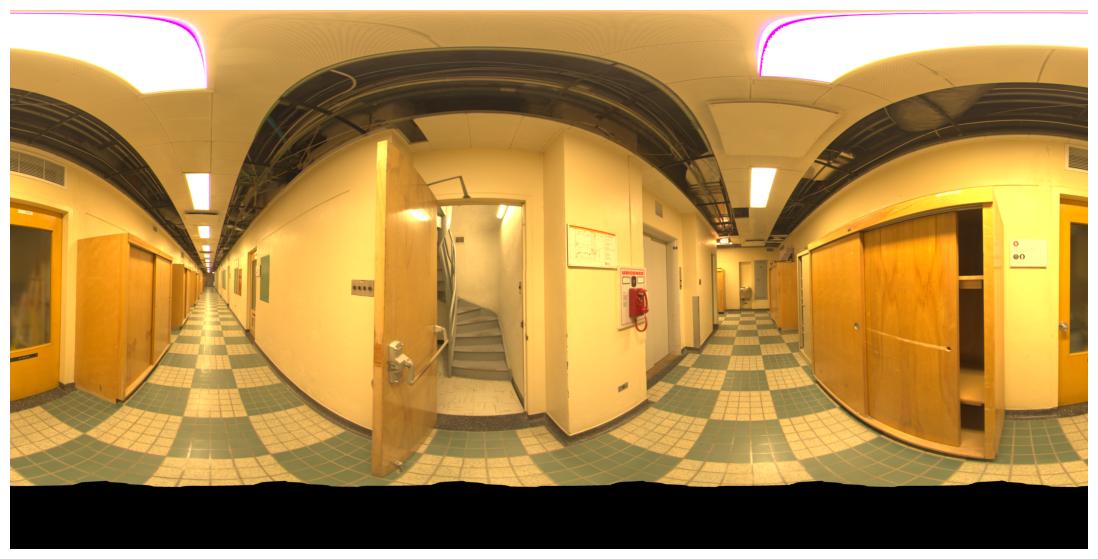}&
    \includegraphics[width=\tmplength]{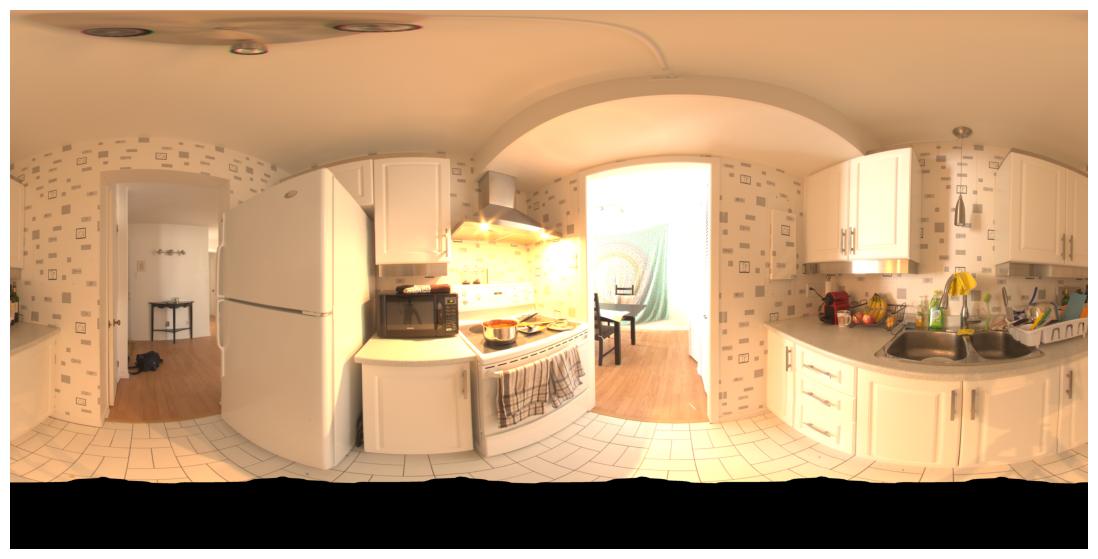}&
    \includegraphics[width=\tmplength]{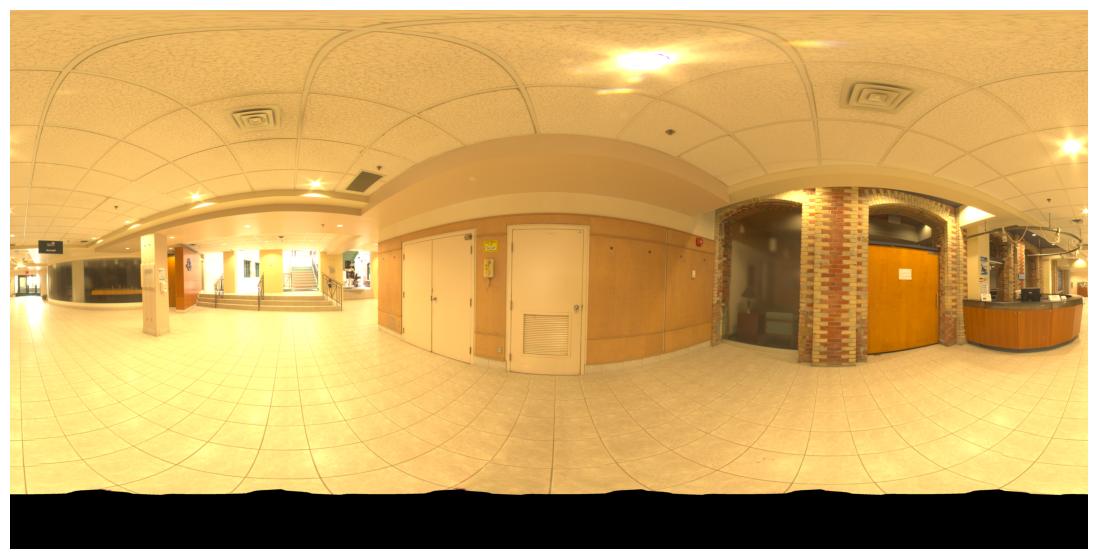}&
    \includegraphics[width=\tmplength]{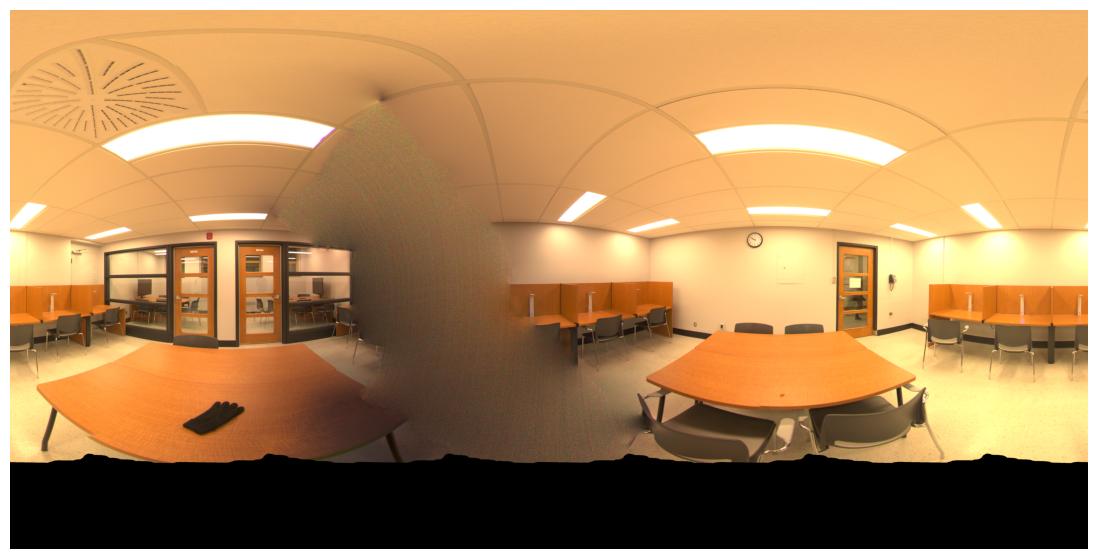}&
    \includegraphics[width=\tmplength]{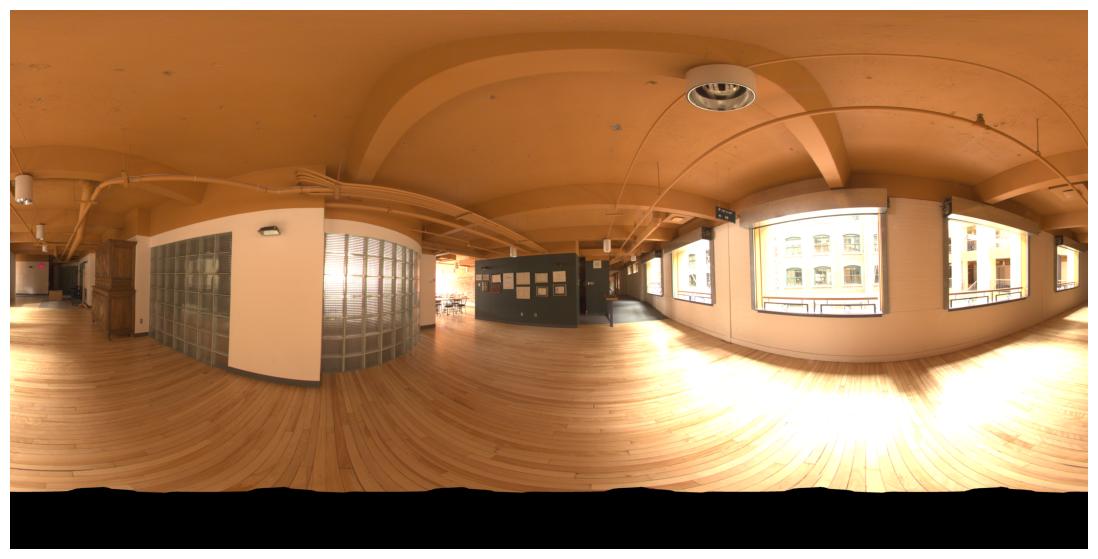}\\
    \includegraphics[width=\tmplength]{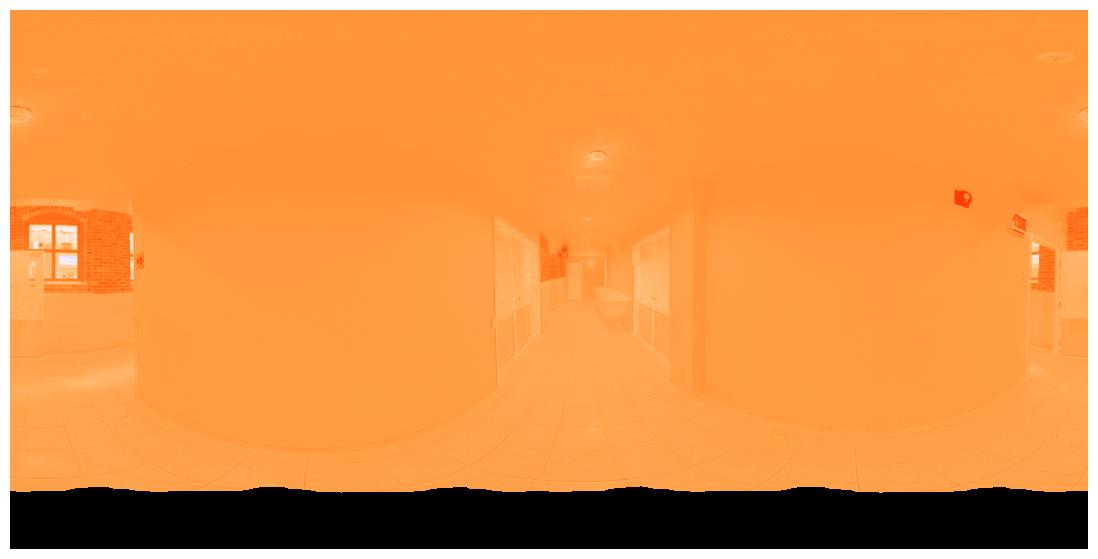}&
    \includegraphics[width=\tmplength]{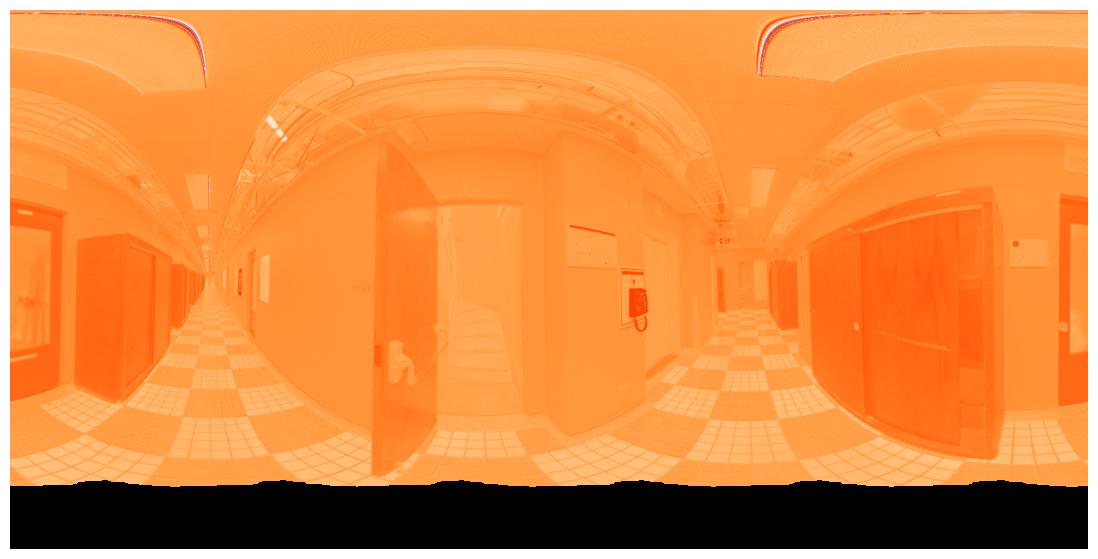}&
    \includegraphics[width=\tmplength]{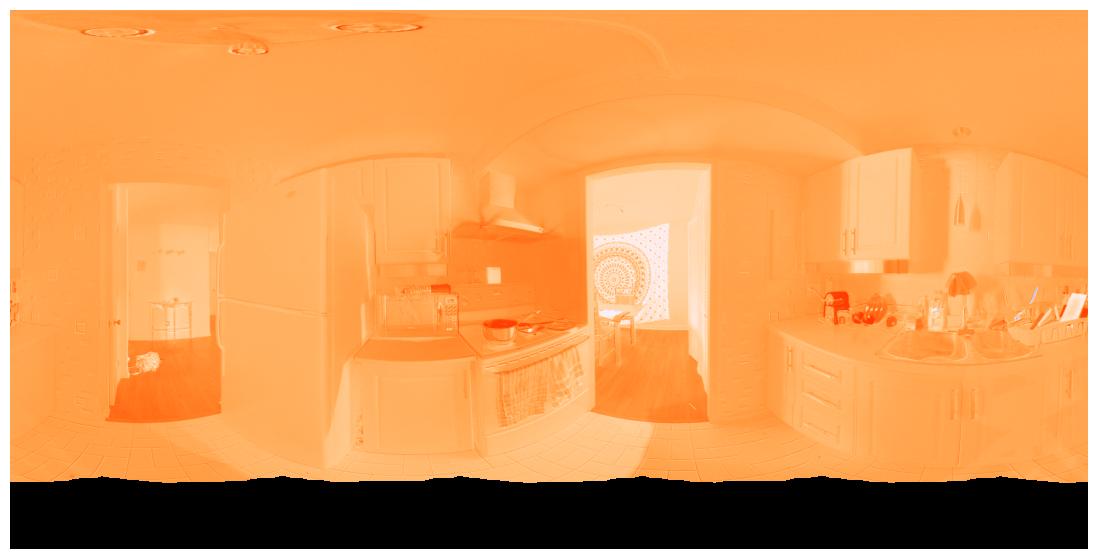}&
    \includegraphics[width=\tmplength]{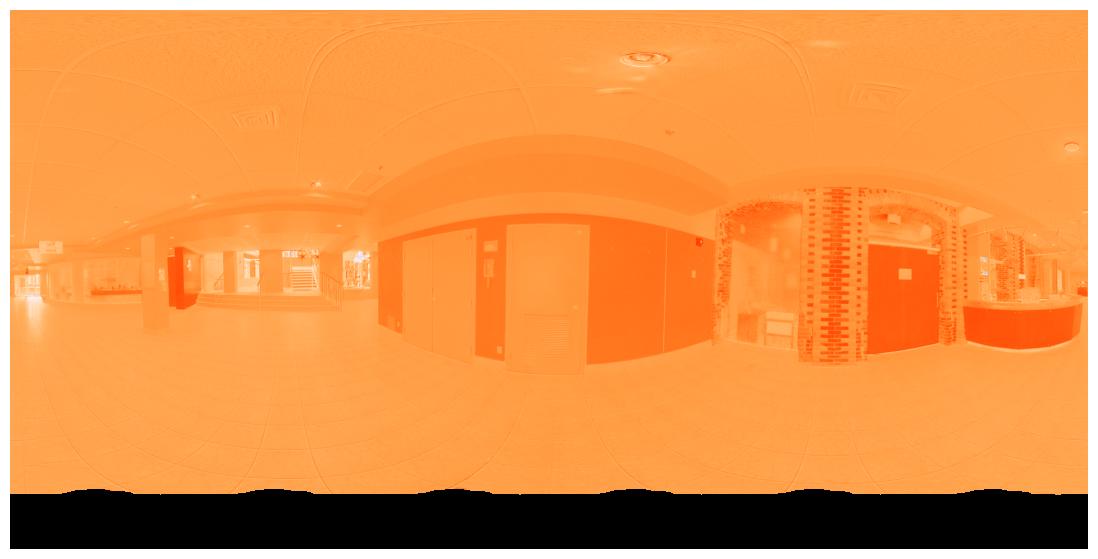}&
    \includegraphics[width=\tmplength]{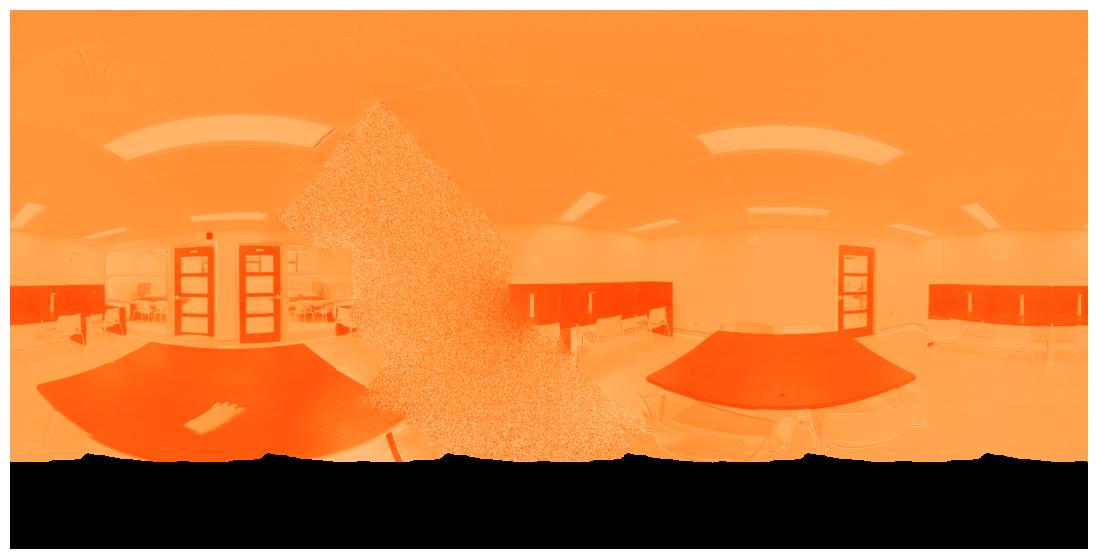}&
    \includegraphics[width=\tmplength]{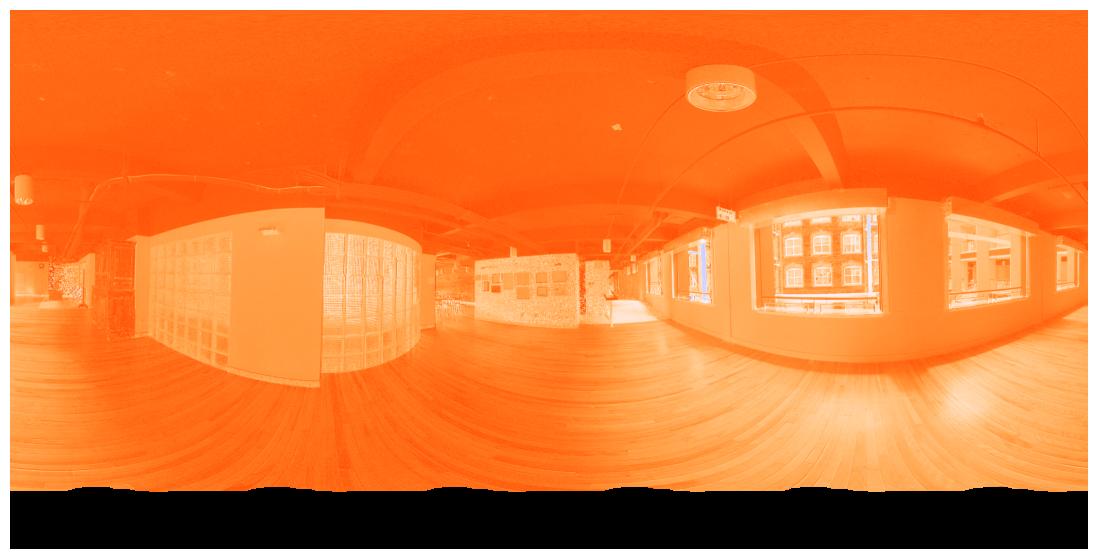}\\
    
    \valeur{47.5th} (\valeur{\SI{3618}{\K}}) & 
    \valeur{50th} (\valeur{\SI{3654}{\K}}) & 
    \valeur{52.5th} (\valeur{\SI{3688}{\K}}) & 
    \valeur{55th} (\valeur{\SI{3730}{\K}}) & 
    \valeur{57.5th} (\valeur{\SI{3784}{\K}}) & 
    \valeur{60th} (\valeur{\SI{3847}{\K}}) \\
    \includegraphics[width=\tmplength]{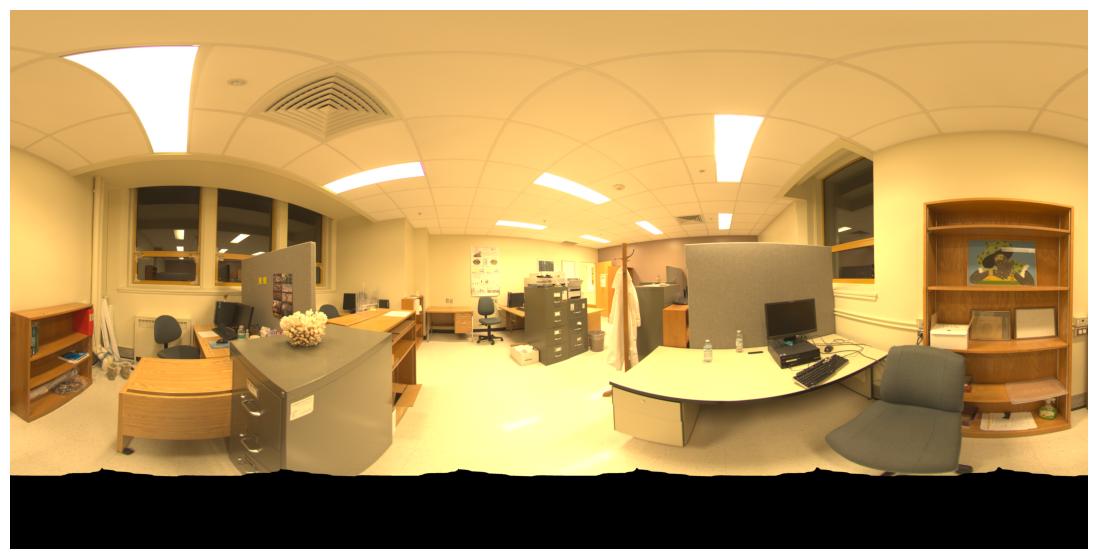}&
    \includegraphics[width=\tmplength]{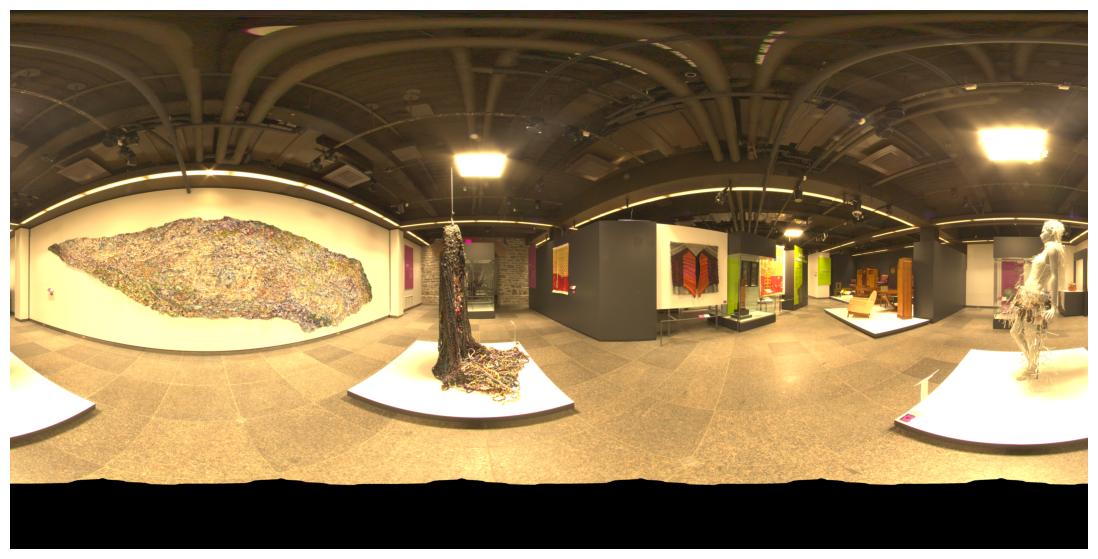}&
    \includegraphics[width=\tmplength]{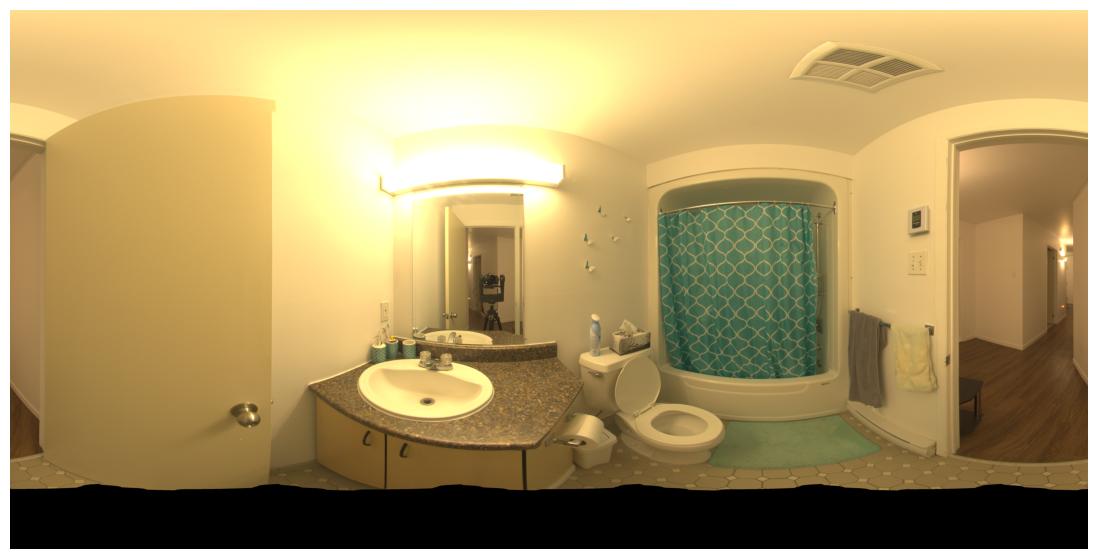}&
    \includegraphics[width=\tmplength]{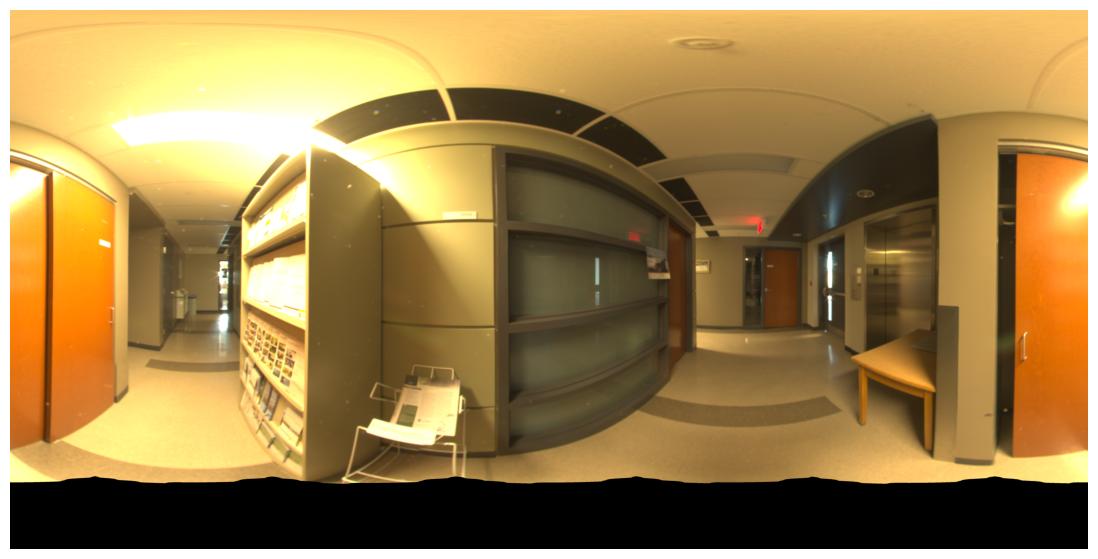}&
    \includegraphics[width=\tmplength]{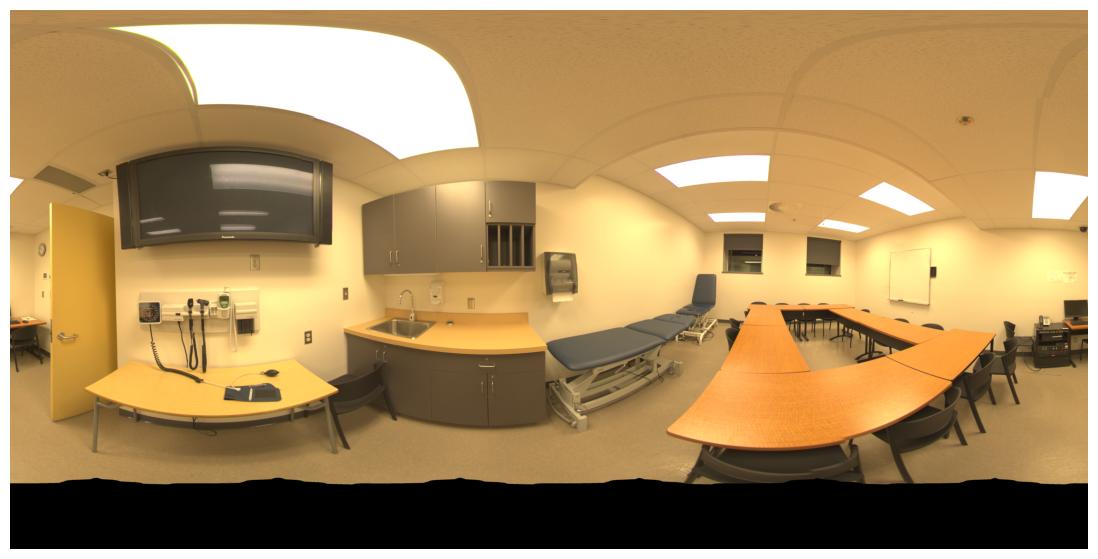}&
    \includegraphics[width=\tmplength]{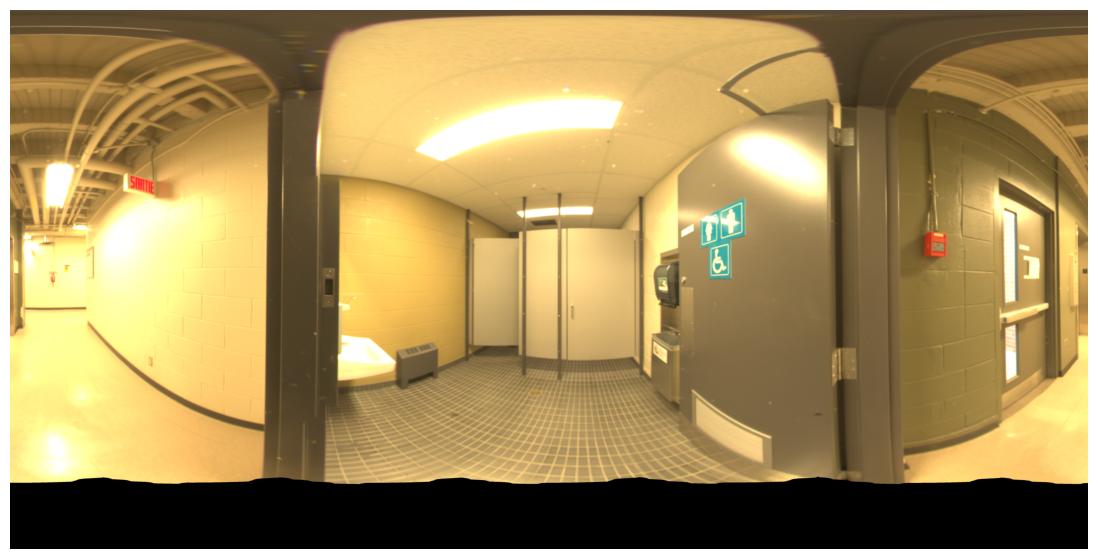}\\
    \includegraphics[width=\tmplength]{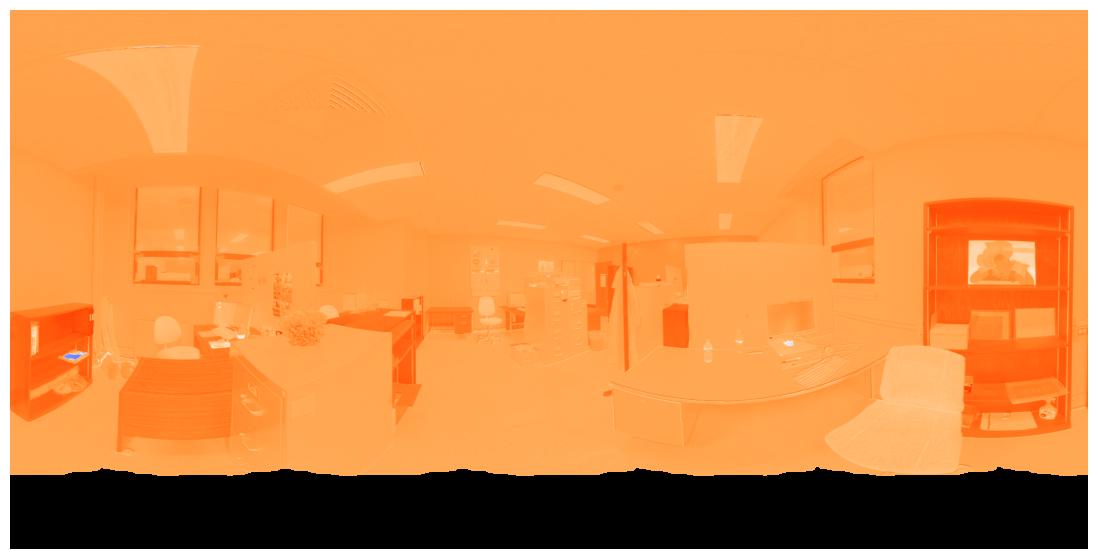}&
    \includegraphics[width=\tmplength]{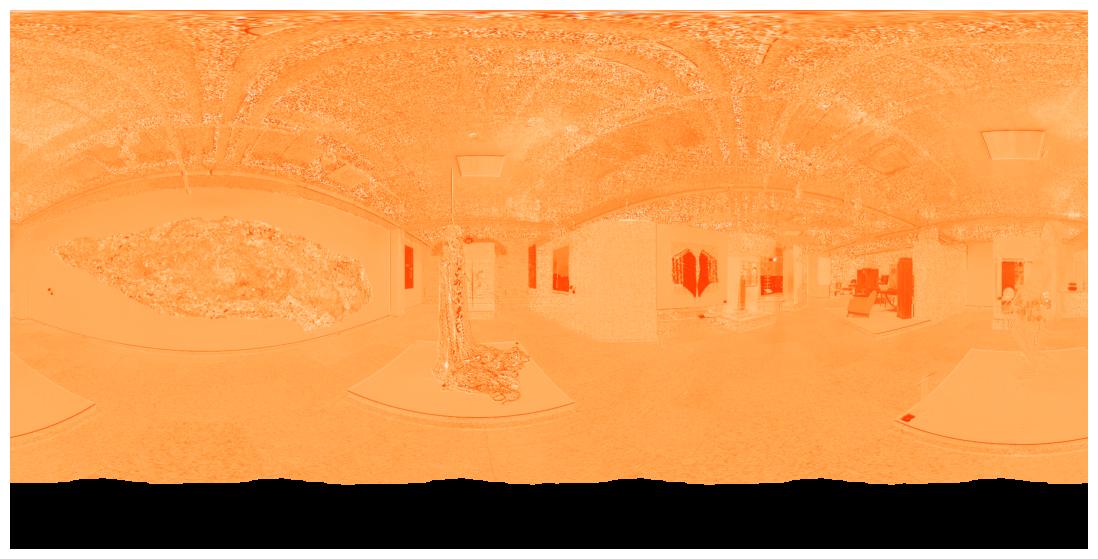}&
    \includegraphics[width=\tmplength]{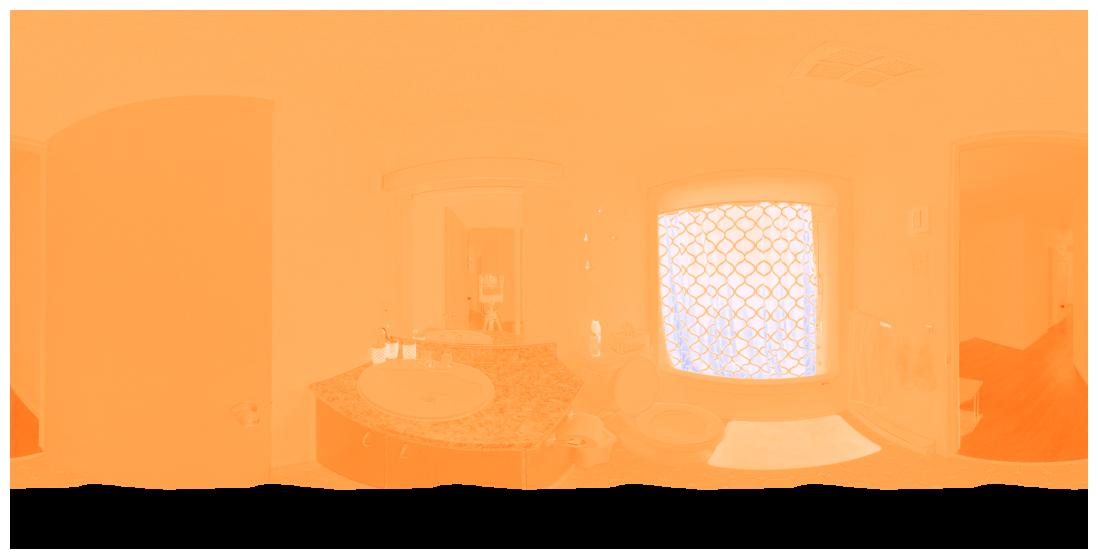}&
    \includegraphics[width=\tmplength]{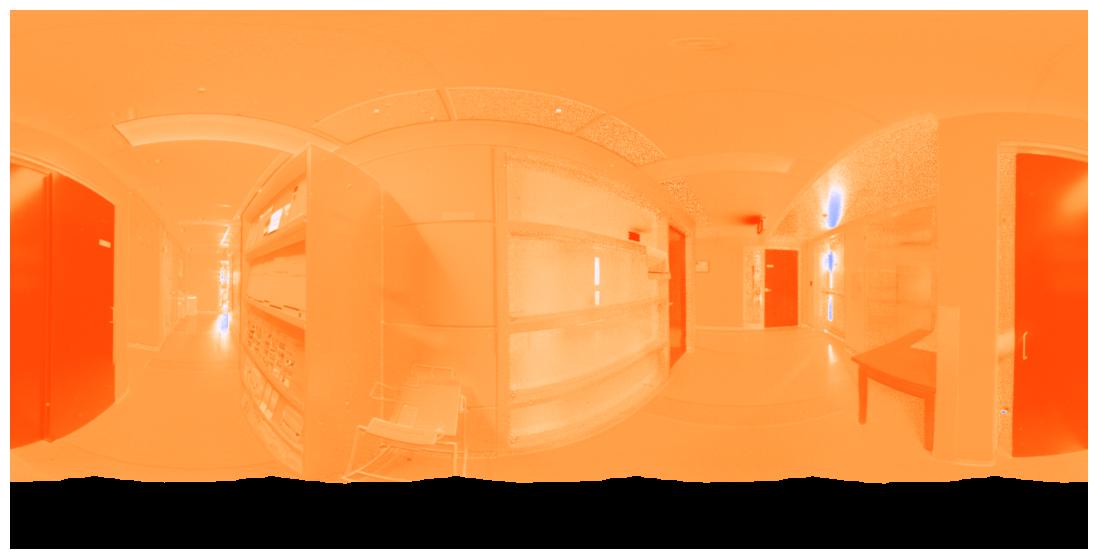}&
    \includegraphics[width=\tmplength]{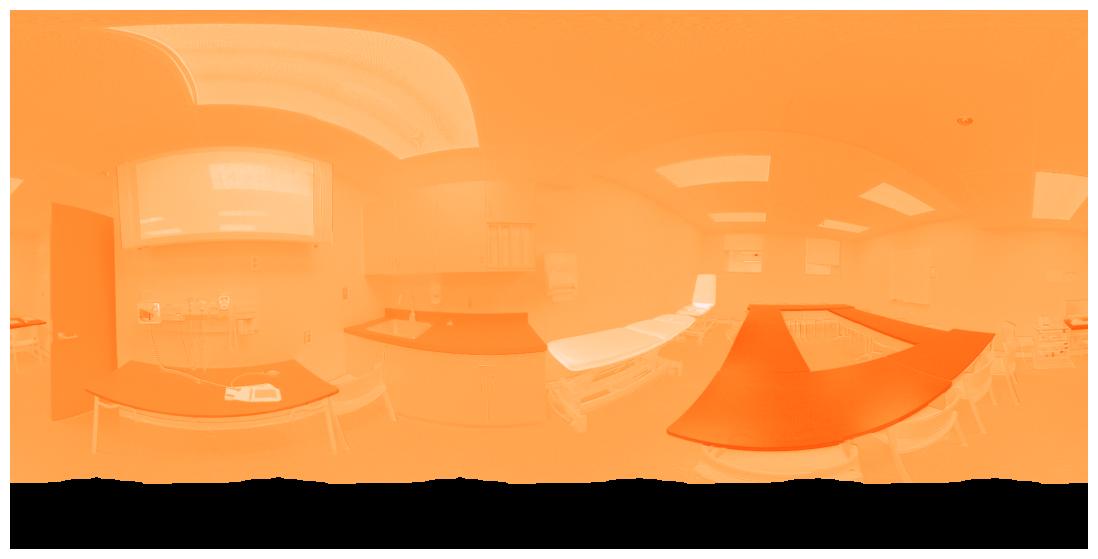}&
    \includegraphics[width=\tmplength]{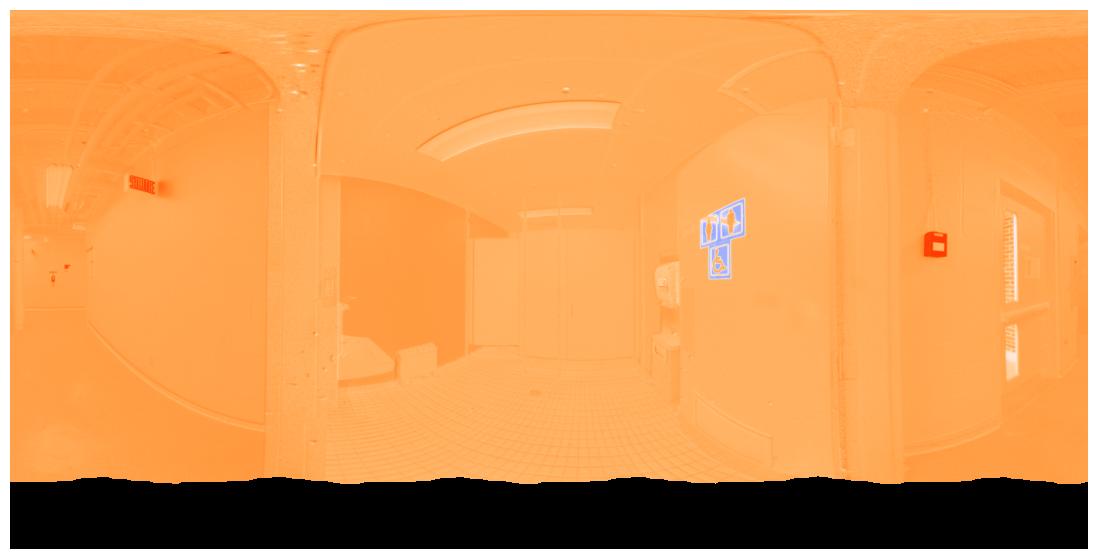}\\
    
    \valeur{62.5th} (\valeur{\SI{3910}{\K}}) & 
    \valeur{65th} (\valeur{\SI{3987}{\K}}) & 
    \valeur{67.5th} (\valeur{\SI{4104}{\K}}) & 
    \valeur{70th} (\valeur{\SI{4261}{\K}}) & 
    \valeur{72.5th} (\valeur{\SI{4428}{\K}}) & 
    \valeur{75th} (\valeur{\SI{4577}{\K}}) \\
    \includegraphics[width=\tmplength]{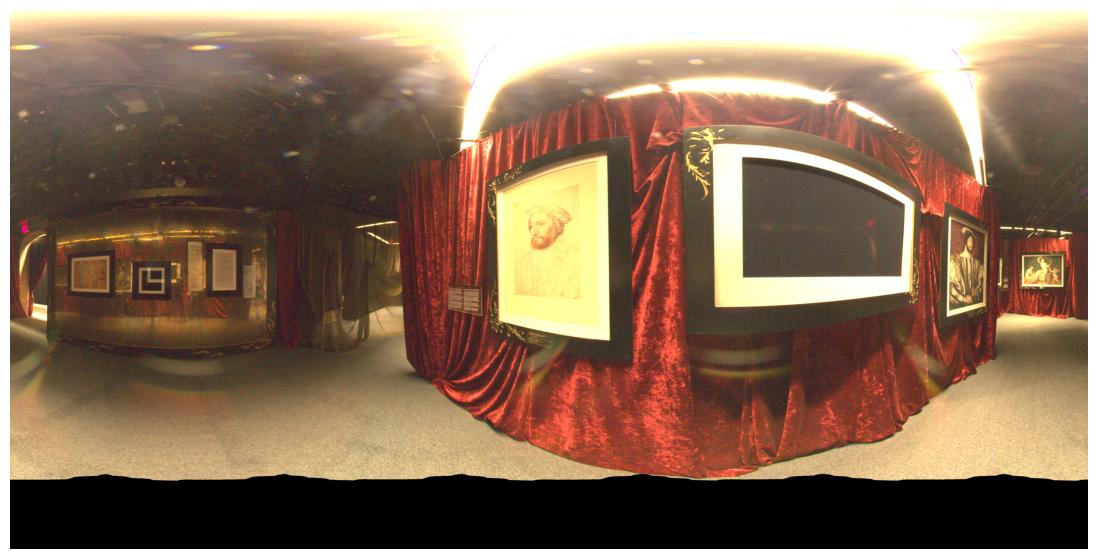}&
    \includegraphics[width=\tmplength]{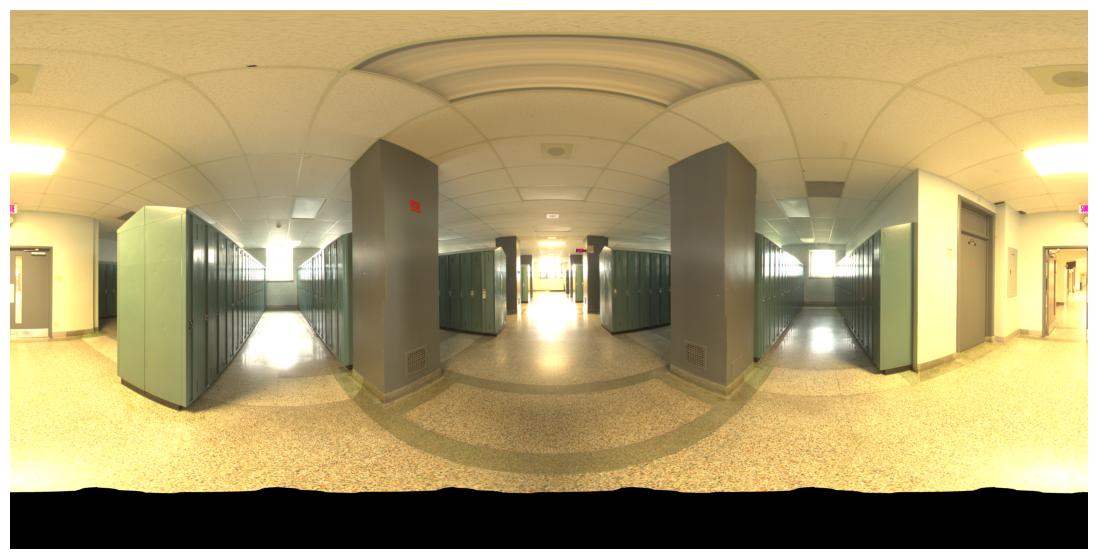}&
    \includegraphics[width=\tmplength]{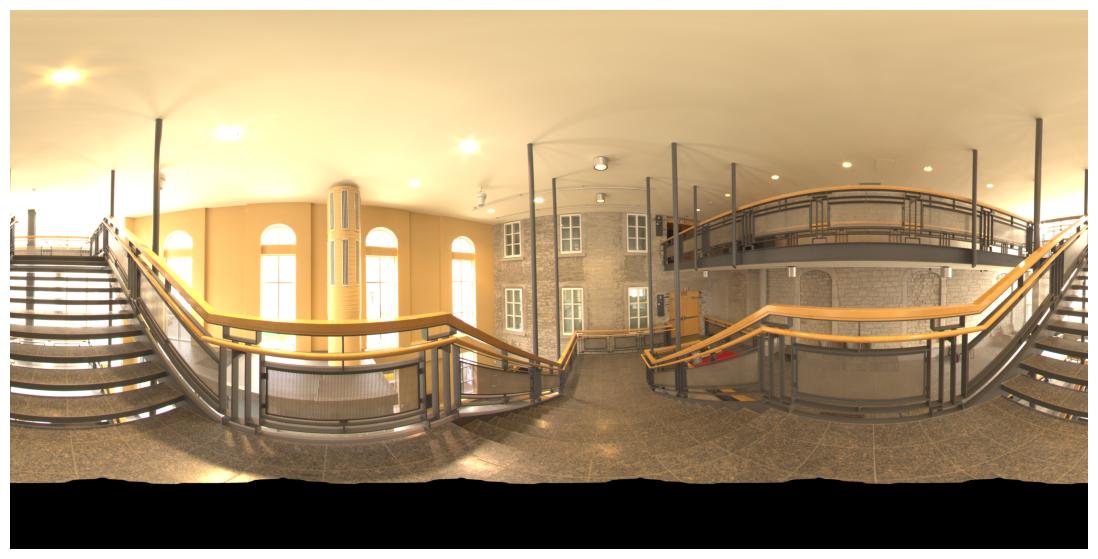}&
    \includegraphics[width=\tmplength]{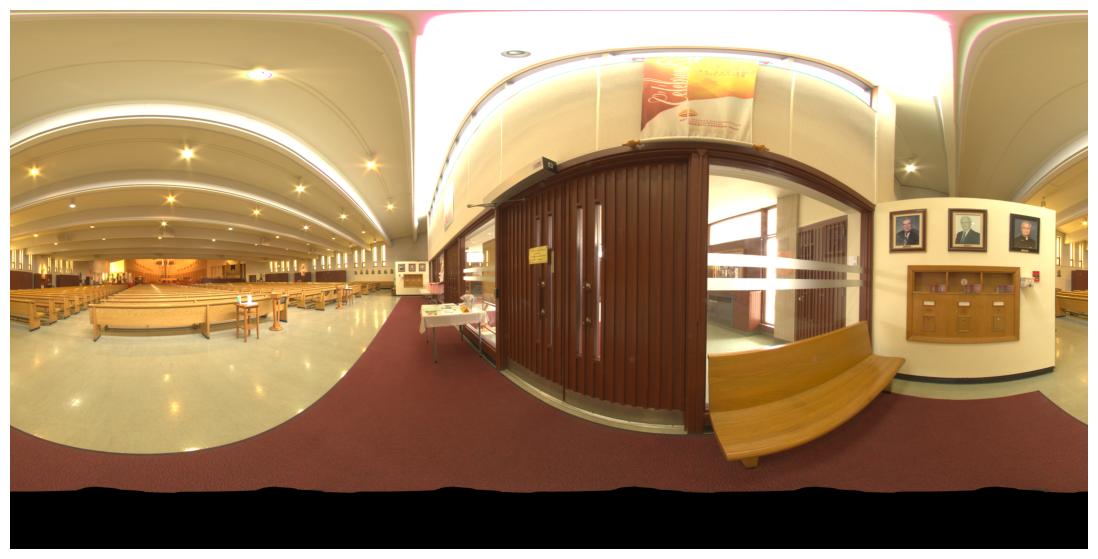}&
    \includegraphics[width=\tmplength]{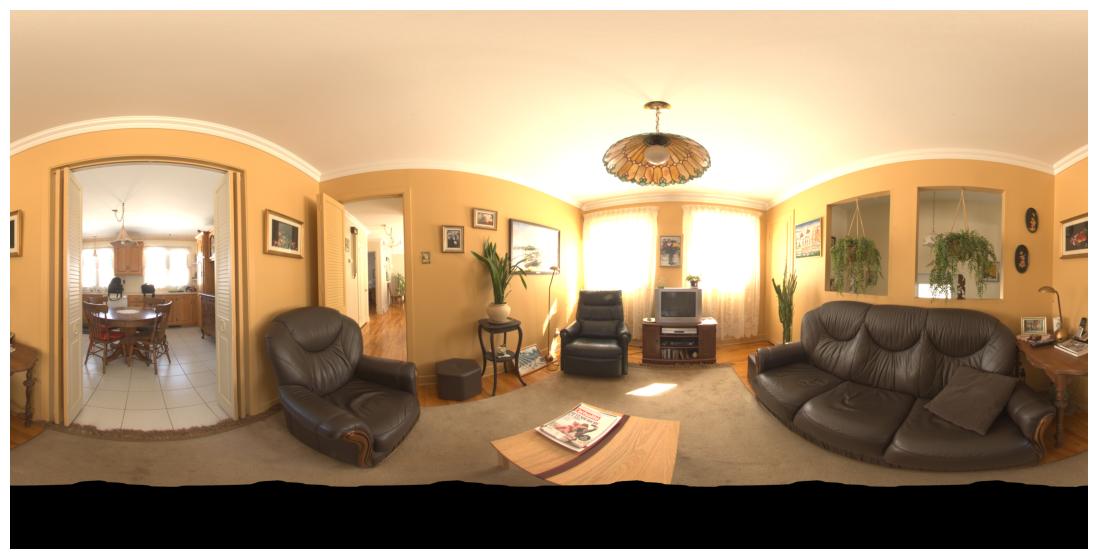}&
    \includegraphics[width=\tmplength]{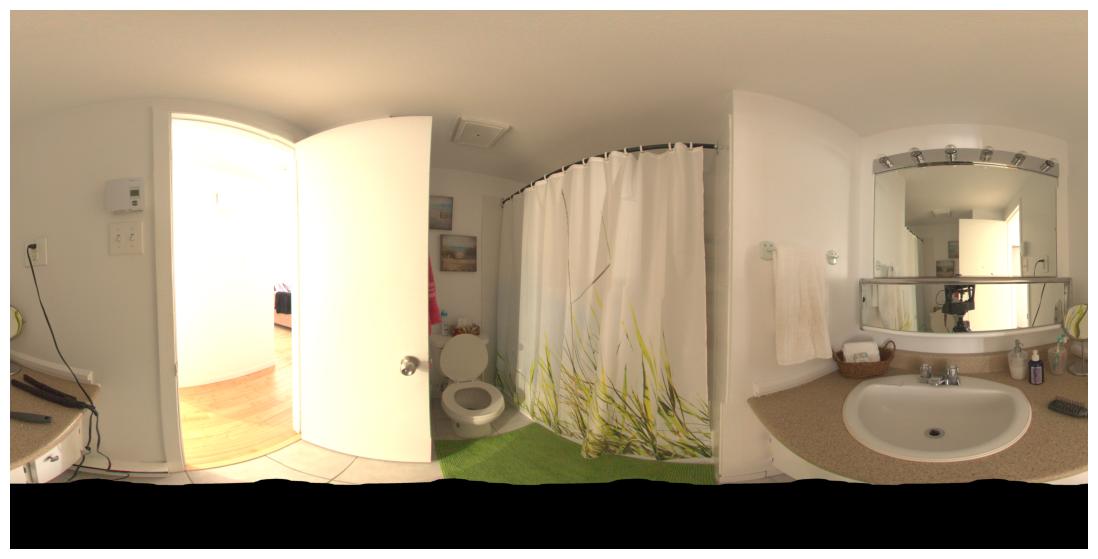}\\
    \includegraphics[width=\tmplength]{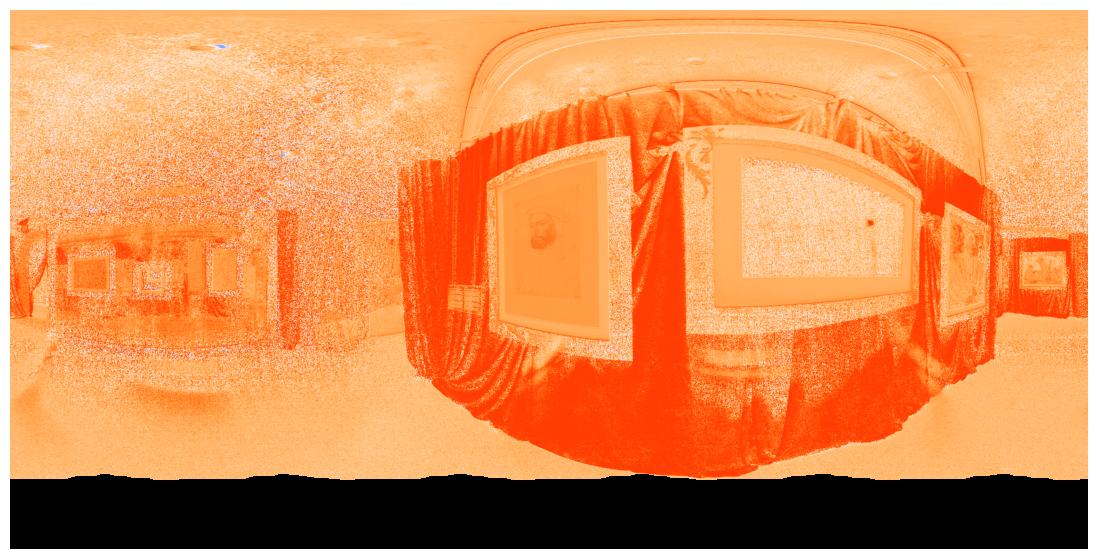}&
    \includegraphics[width=\tmplength]{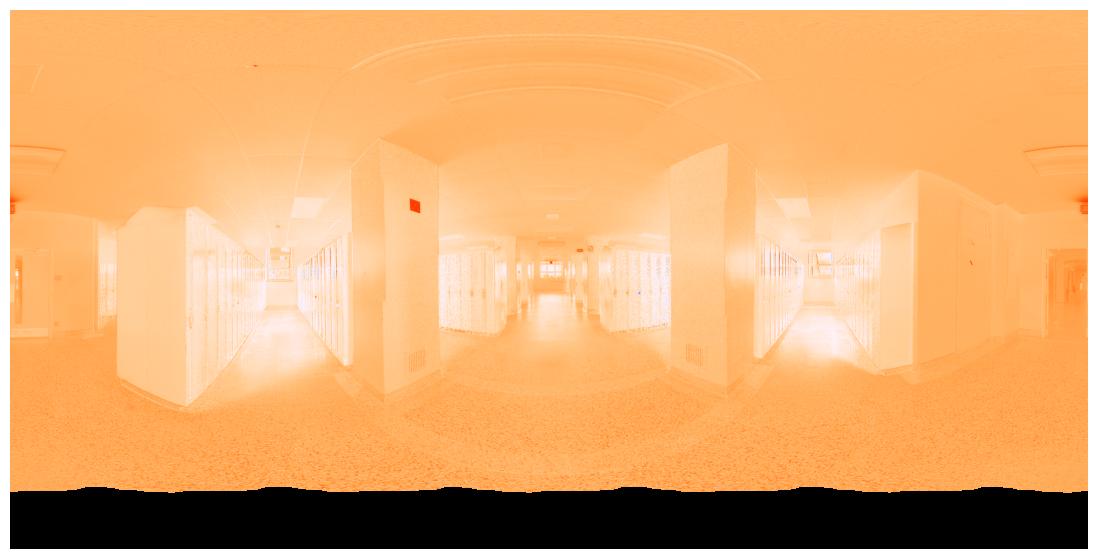}&
    \includegraphics[width=\tmplength]{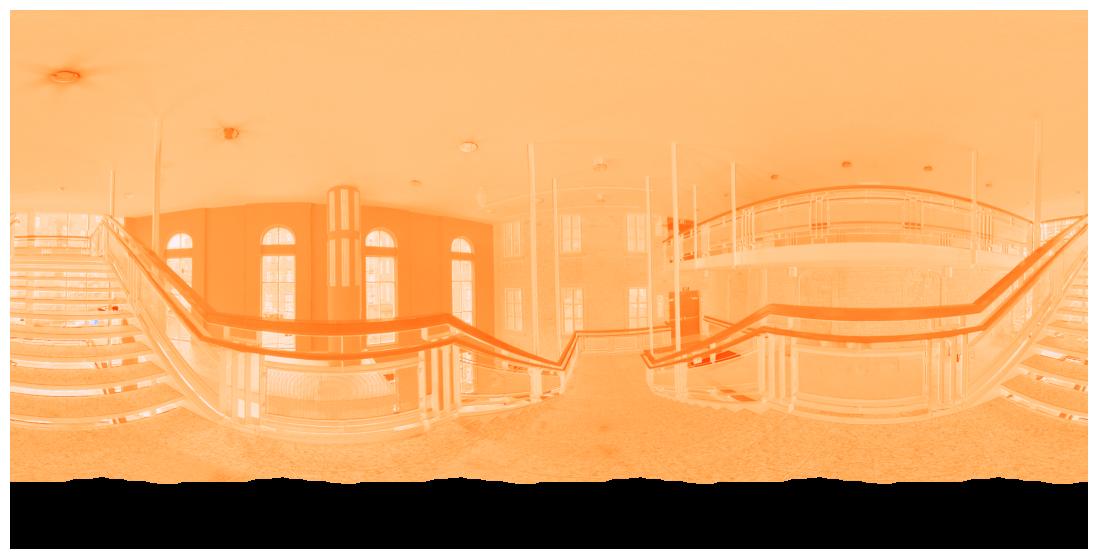}&
    \includegraphics[width=\tmplength]{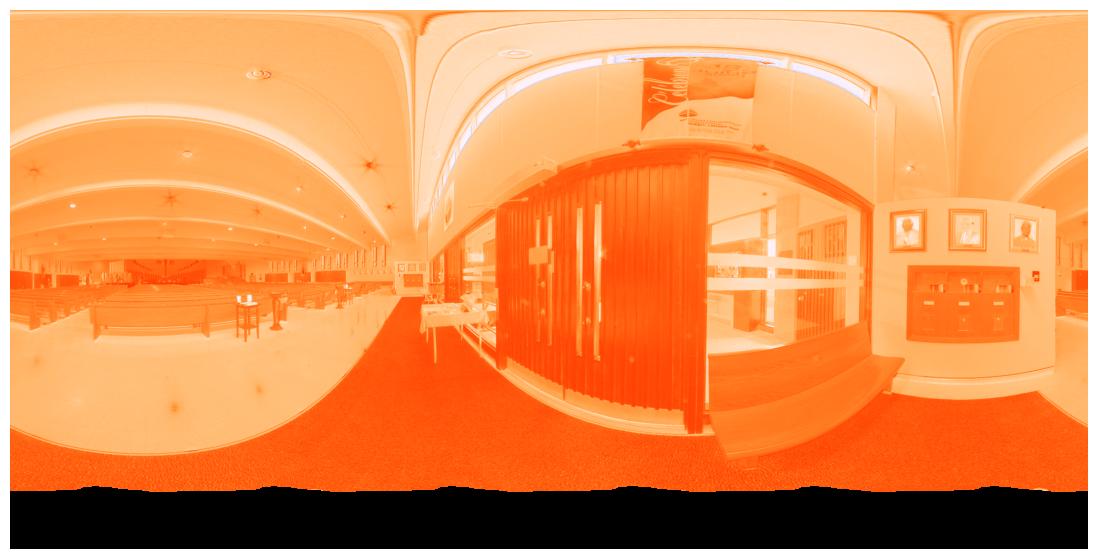}&
    \includegraphics[width=\tmplength]{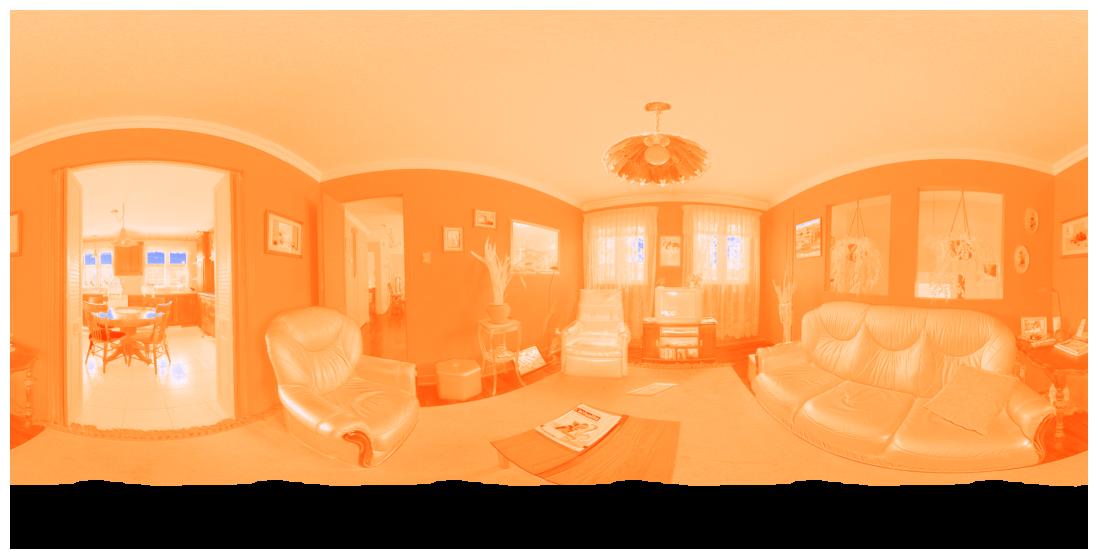}&
    \includegraphics[width=\tmplength]{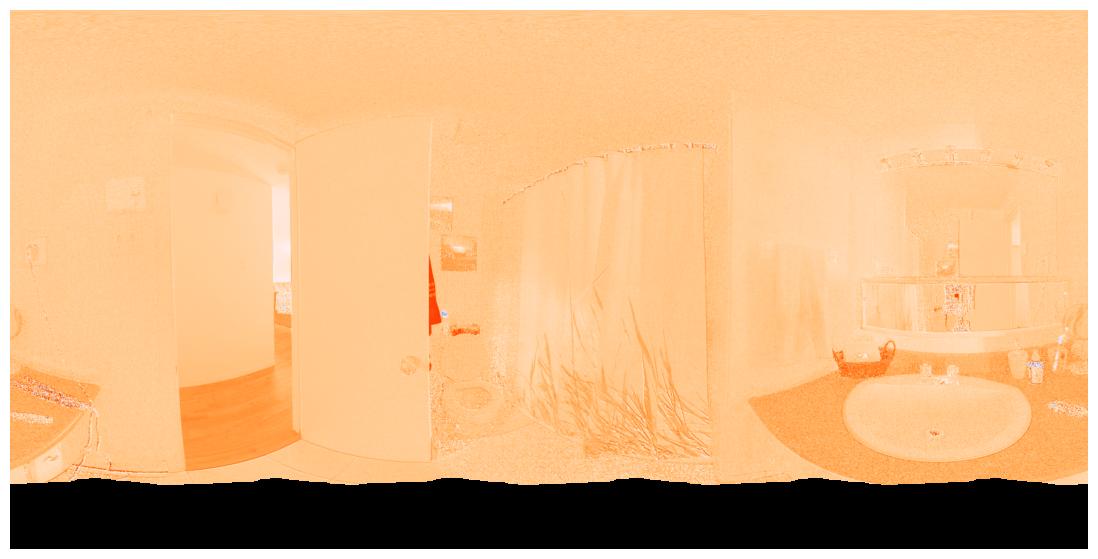}\\
    
    \valeur{77.5th} (\valeur{\SI{4707}{\K}}) & 
    \valeur{80th} (\valeur{\SI{4876}{\K}}) & 
    \valeur{82.5th} (\valeur{\SI{5043}{\K}}) & 
    \valeur{85th} (\valeur{\SI{5188}{\K}}) & 
    \valeur{87.5th} (\valeur{\SI{5358}{\K}}) & 
    \valeur{90th} (\valeur{\SI{5656}{\K}}) \\
    \includegraphics[width=\tmplength]{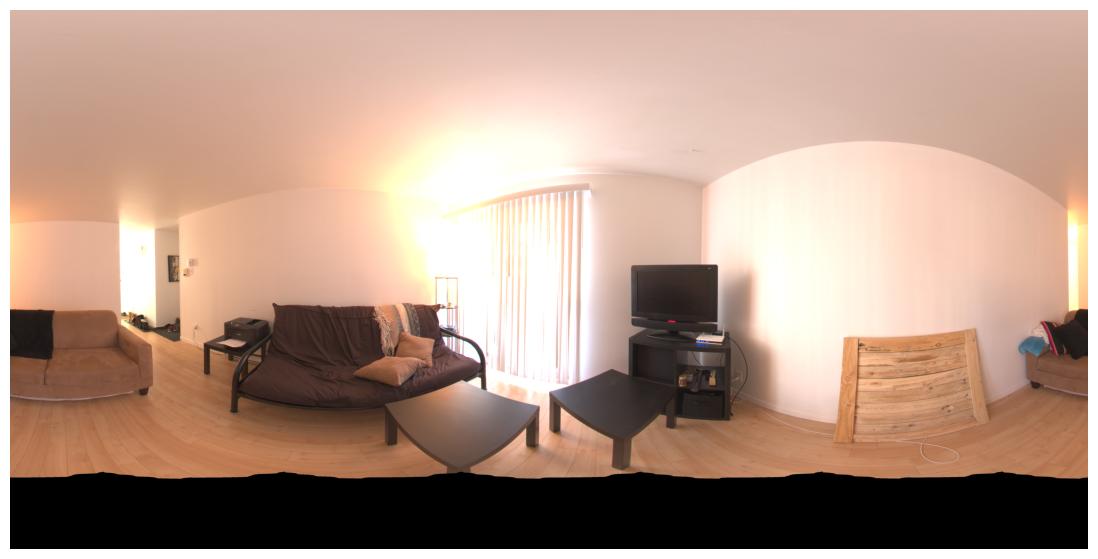}&
    \includegraphics[width=\tmplength]{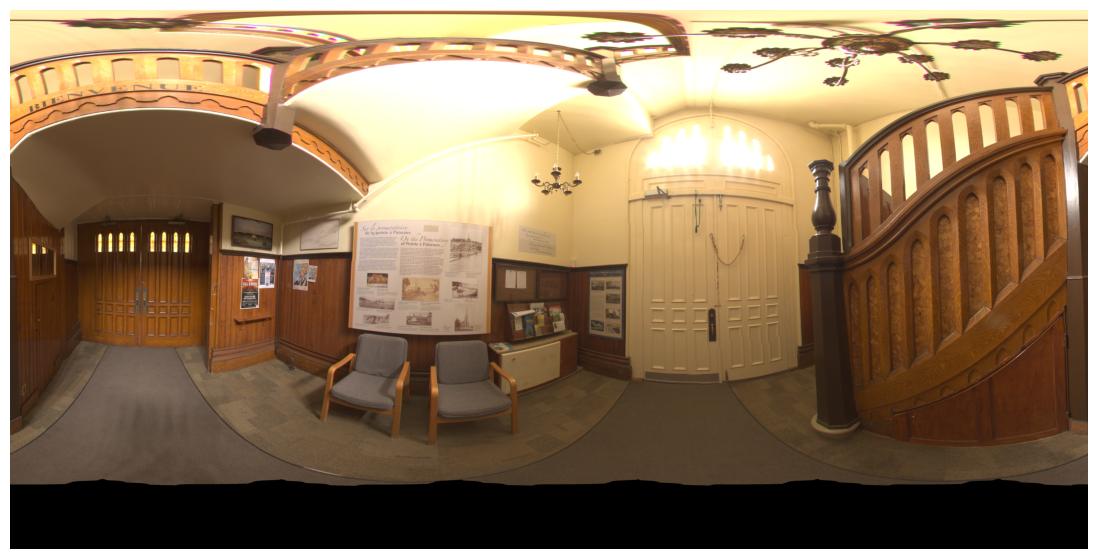}&
    \includegraphics[width=\tmplength]{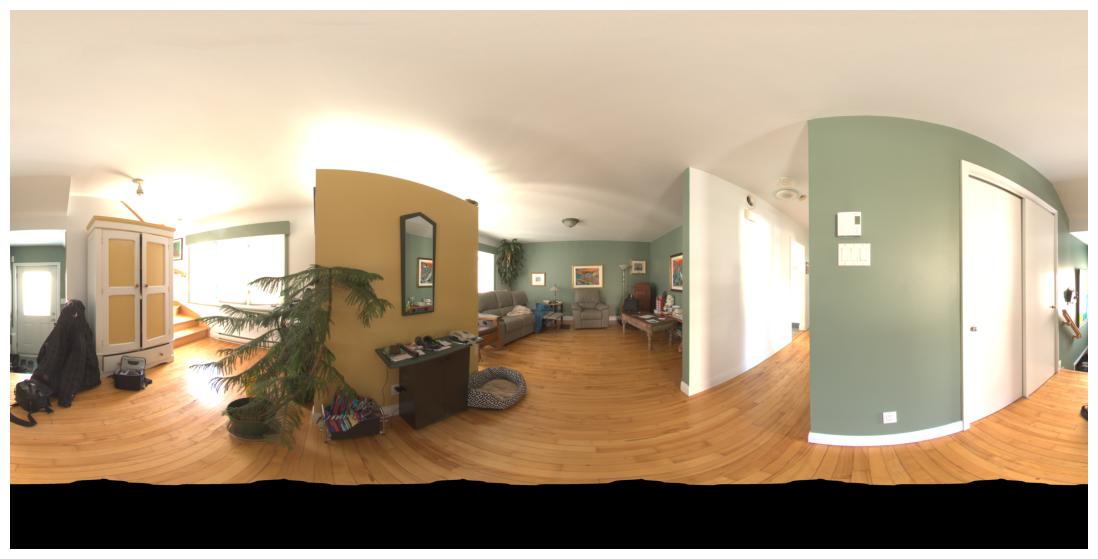}&
    \includegraphics[width=\tmplength]{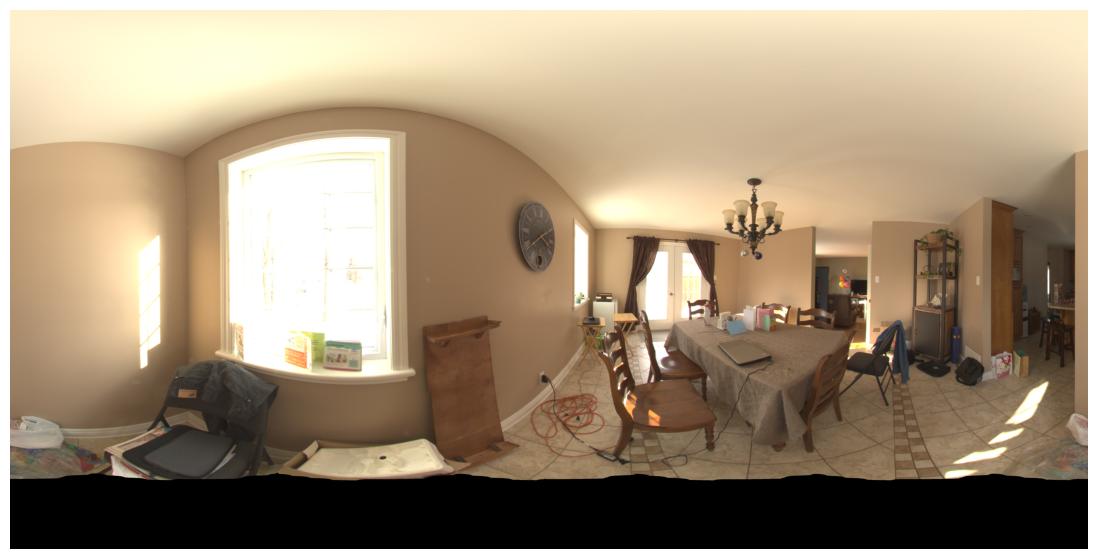}&
    \includegraphics[width=\tmplength]{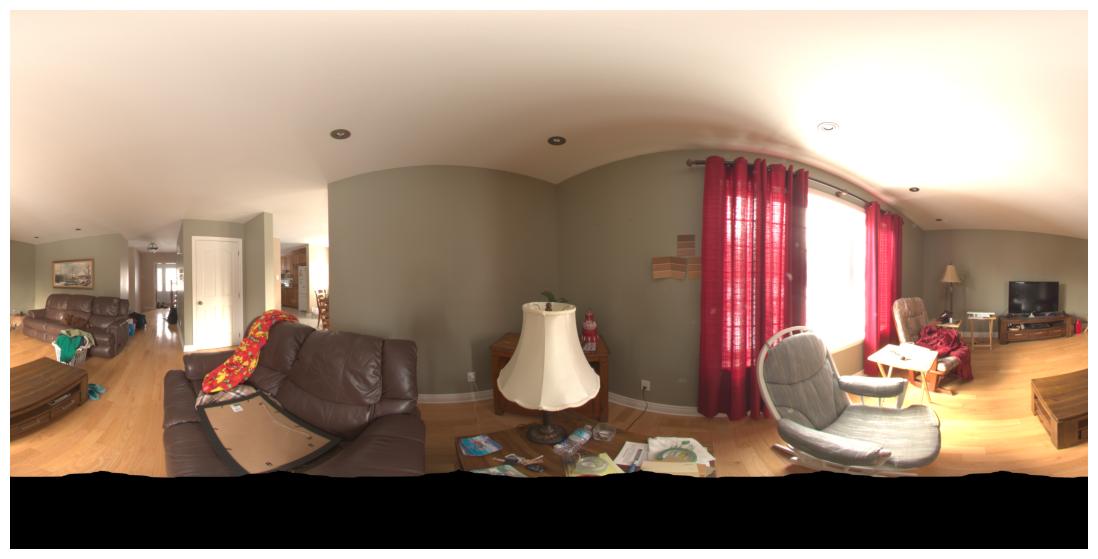}&
    \includegraphics[width=\tmplength]{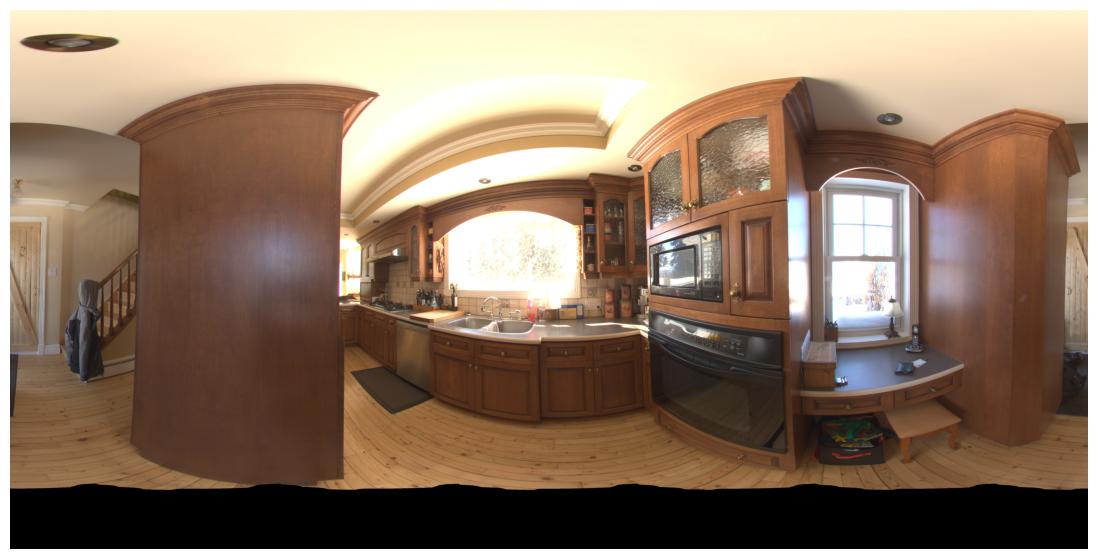}\\
    \includegraphics[width=\tmplength]{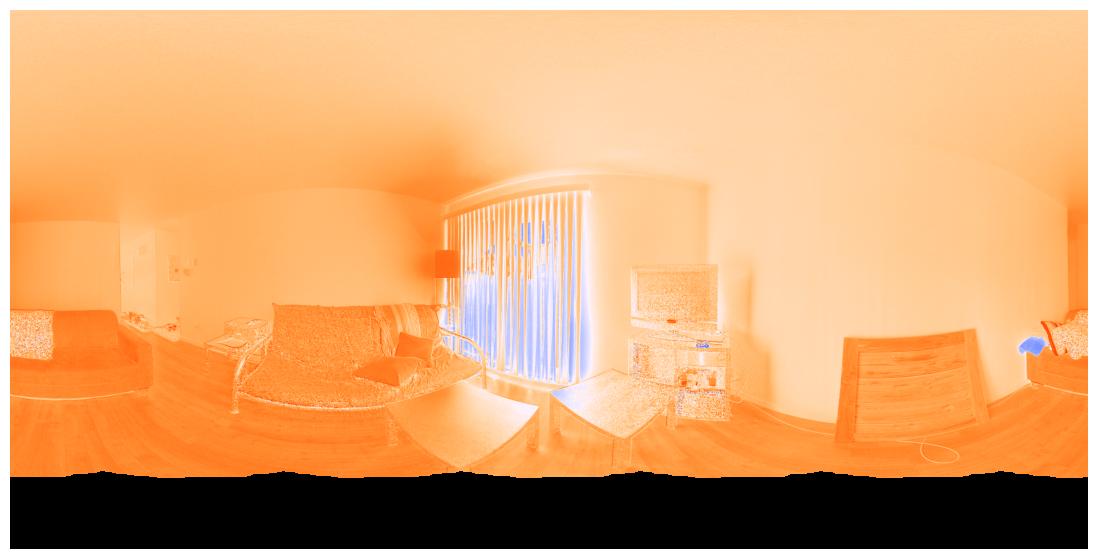}&
    \includegraphics[width=\tmplength]{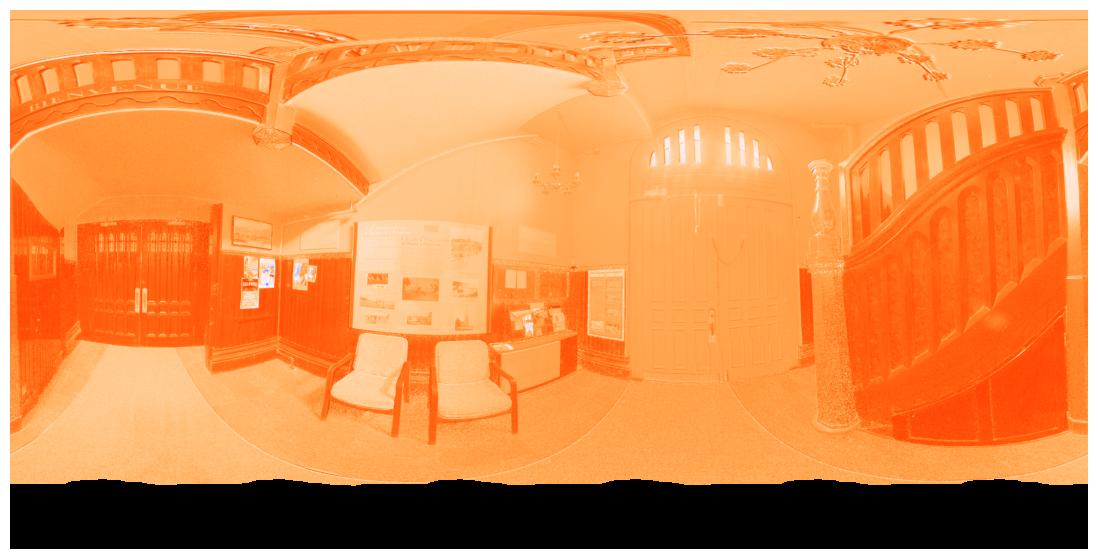}&
    \includegraphics[width=\tmplength]{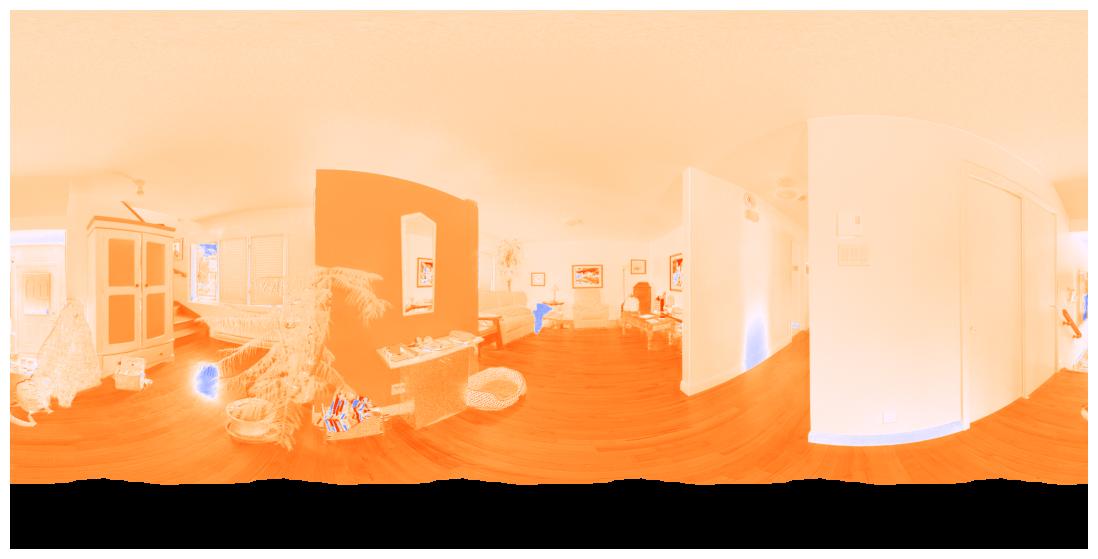}&
    \includegraphics[width=\tmplength]{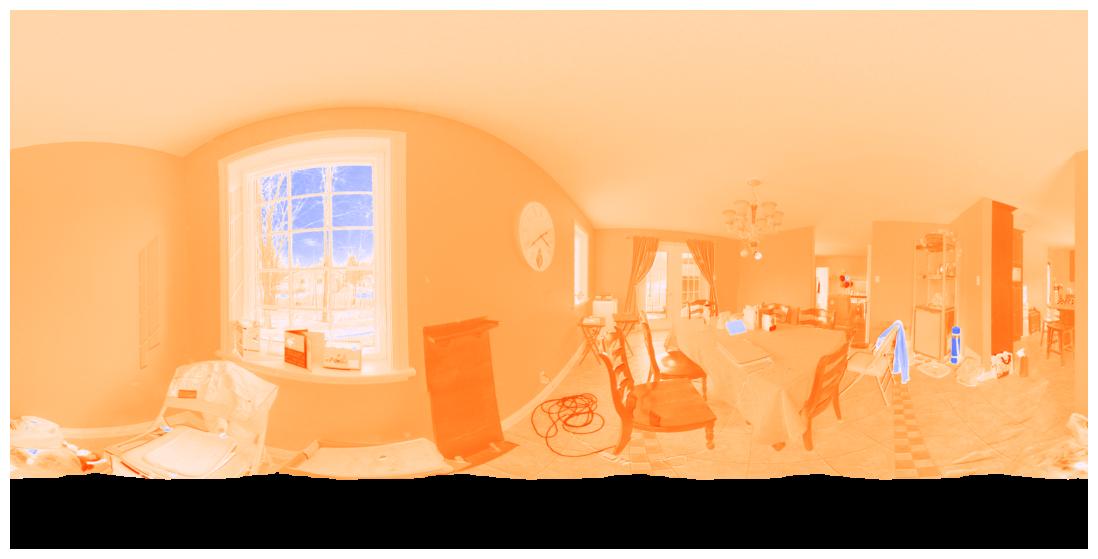}&
    \includegraphics[width=\tmplength]{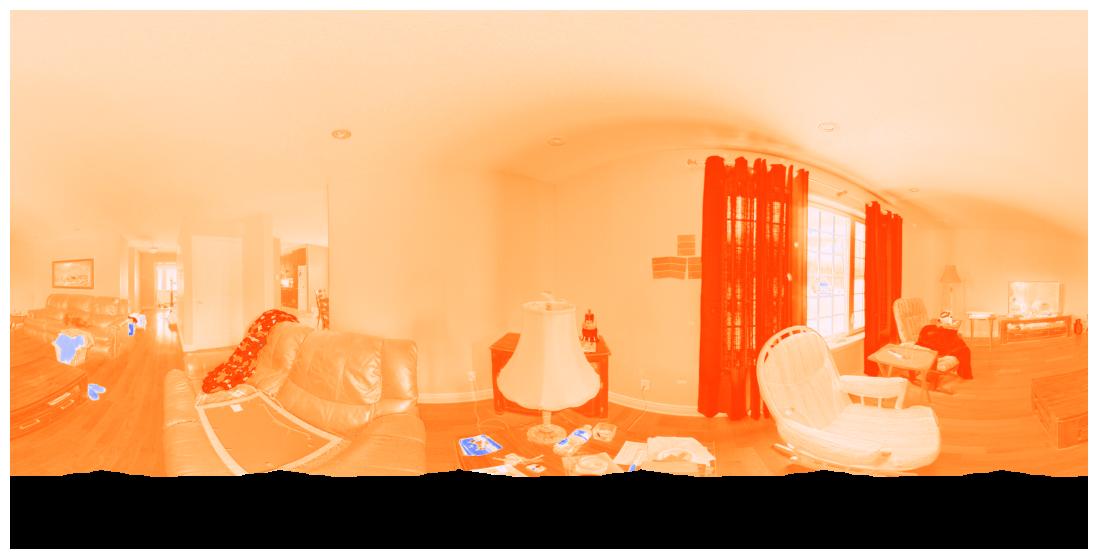}&
    \includegraphics[width=\tmplength]{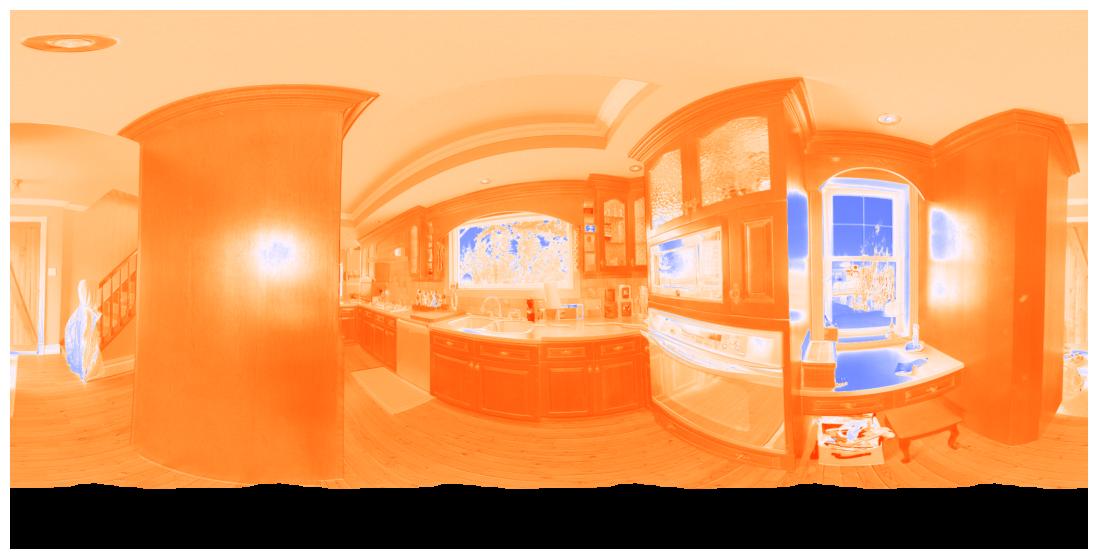}\\
    
    \valeur{92.5th} (\valeur{\SI{5919}{\K}}) & 
    \valeur{95th} (\valeur{\SI{6304}{\K}}) & 
    \valeur{97.5th} (\valeur{\SI{6856}{\K}}) & 
    \valeur{99th} (\valeur{\SI{8103}{\K}}) & 
    \valeur{99.9th} (\valeur{\SI{11565}{\K}}) & 
    \valeur{Max} (\valeur{\SI{291802}{\K}}) \\
    \includegraphics[width=\tmplength]{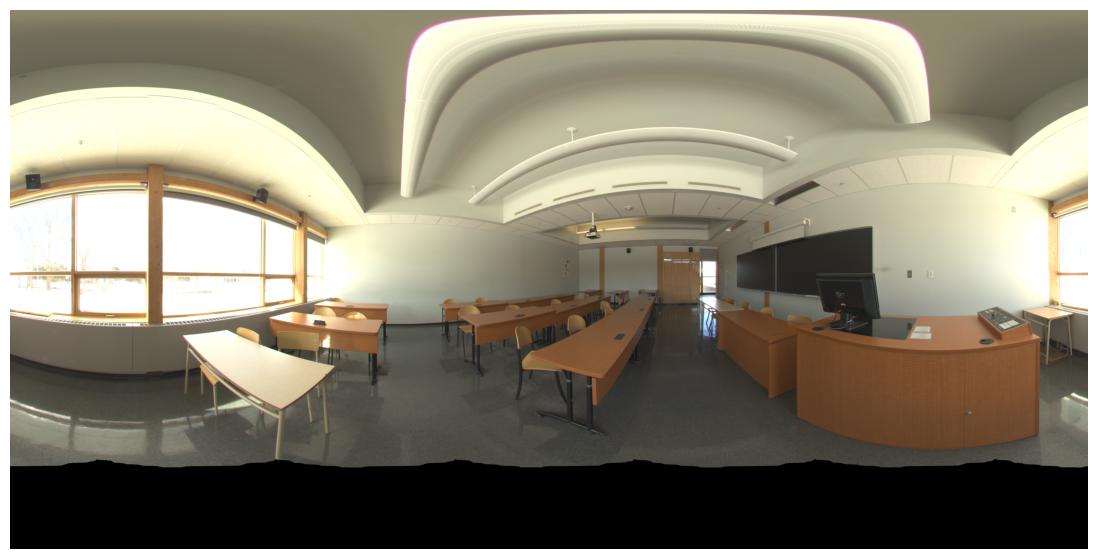}&
    \includegraphics[width=\tmplength]{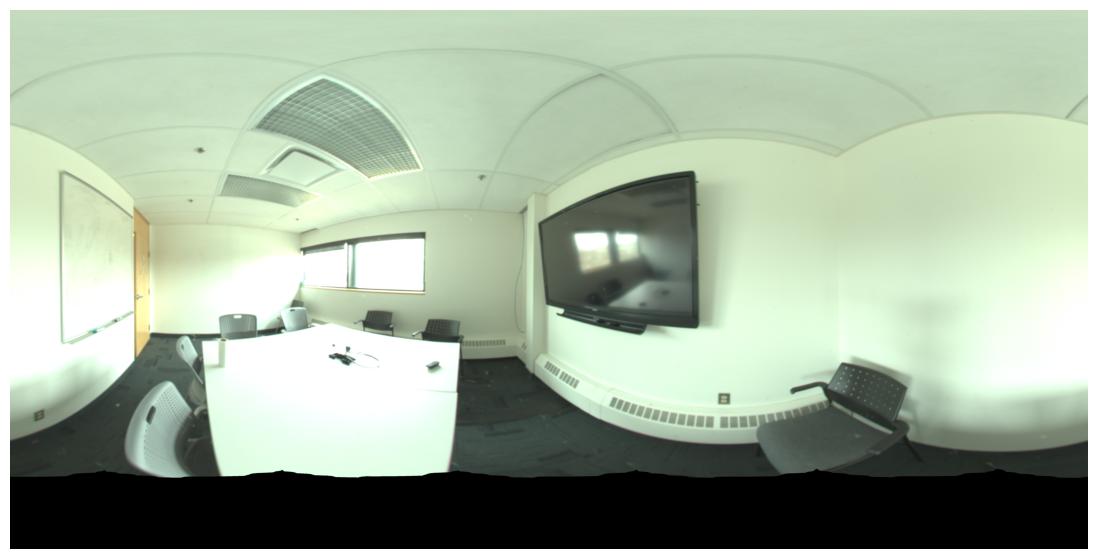}&
    \includegraphics[width=\tmplength]{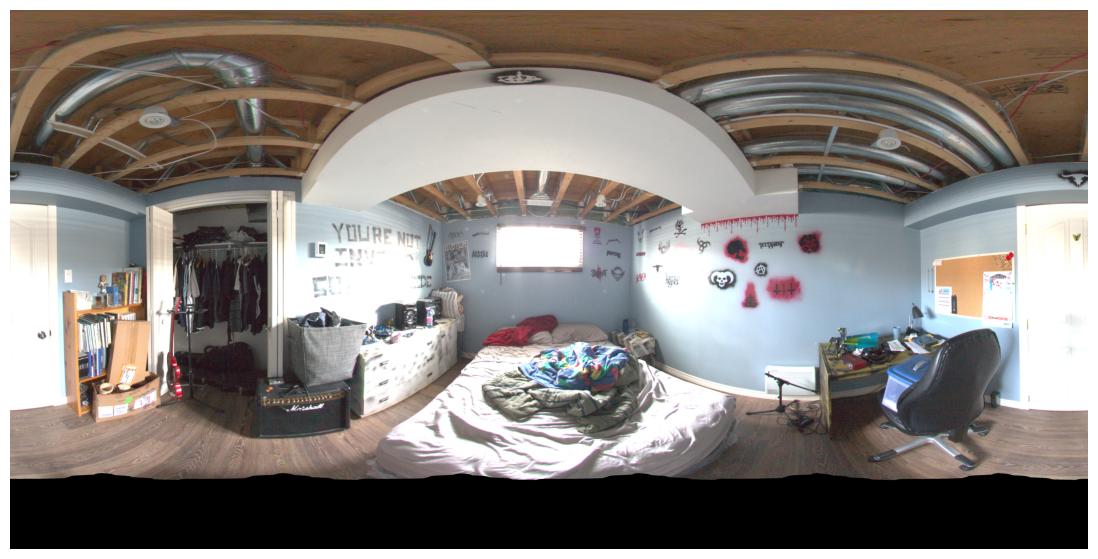}&
    \includegraphics[width=\tmplength]{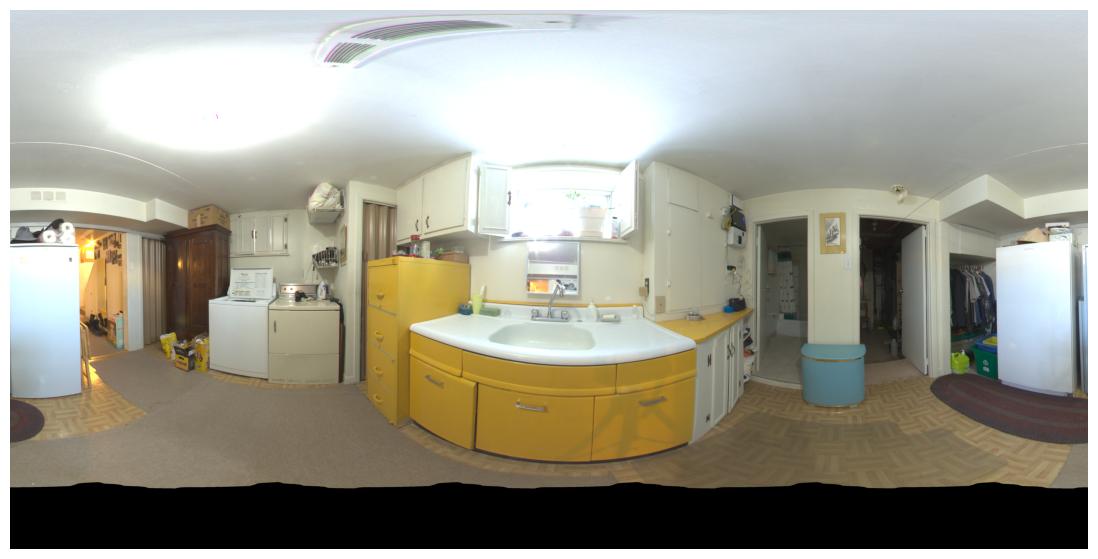}&
    \includegraphics[width=\tmplength]{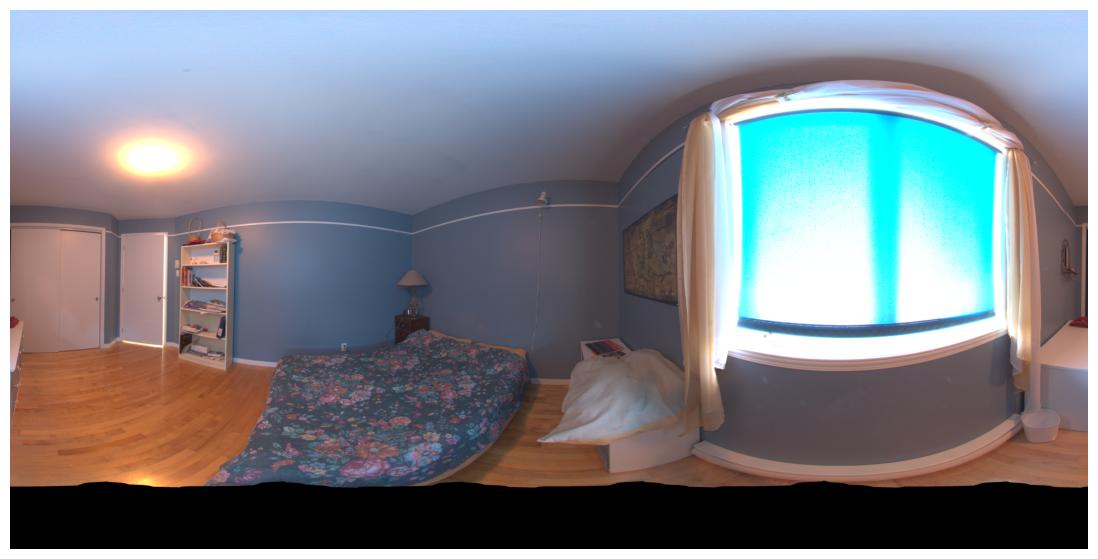}&
    \includegraphics[width=\tmplength]{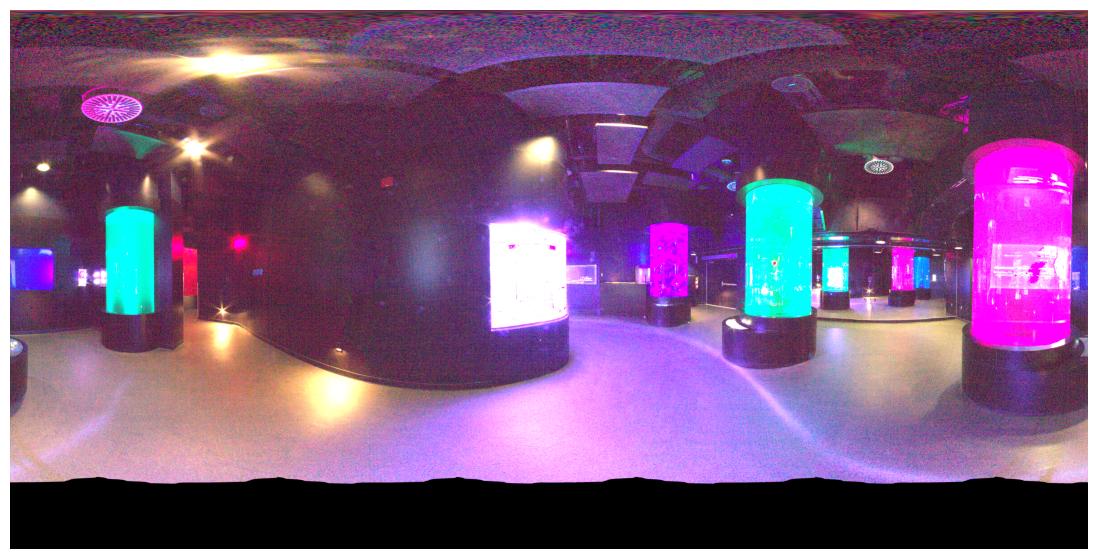}\\
    \includegraphics[width=\tmplength]{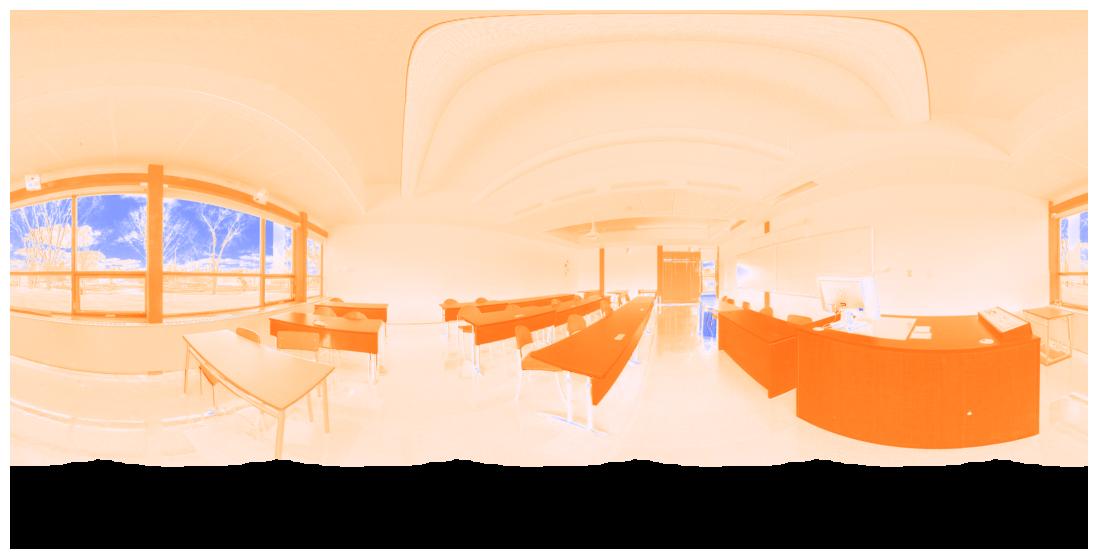}&
    \includegraphics[width=\tmplength]{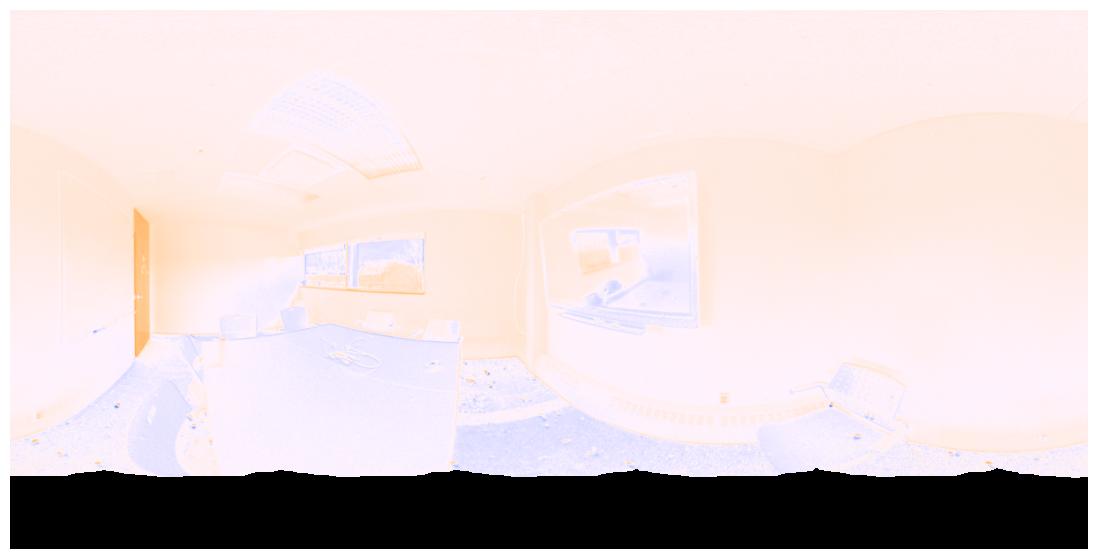}&
    \includegraphics[width=\tmplength]{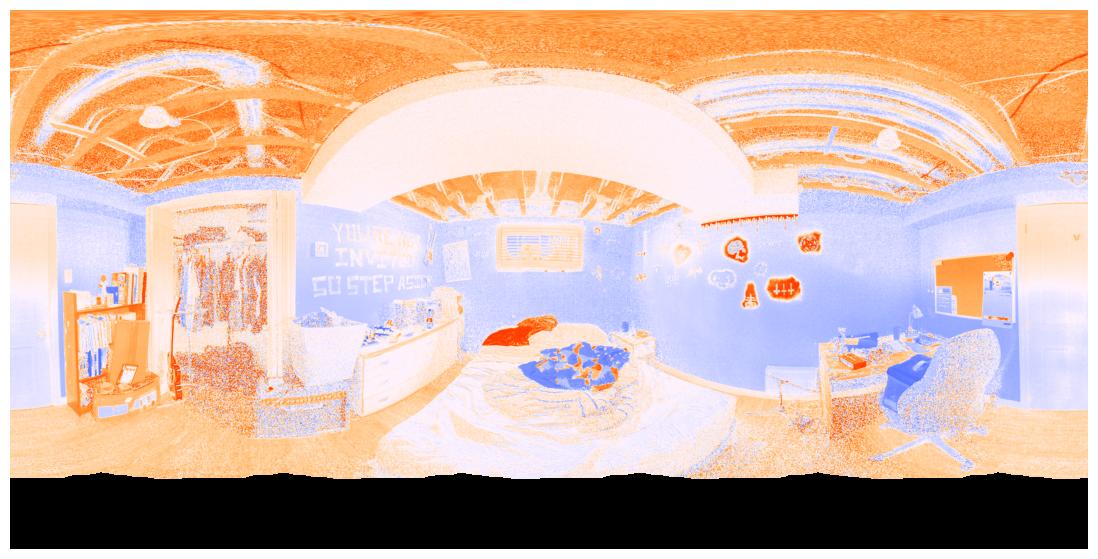}&
    \includegraphics[width=\tmplength]{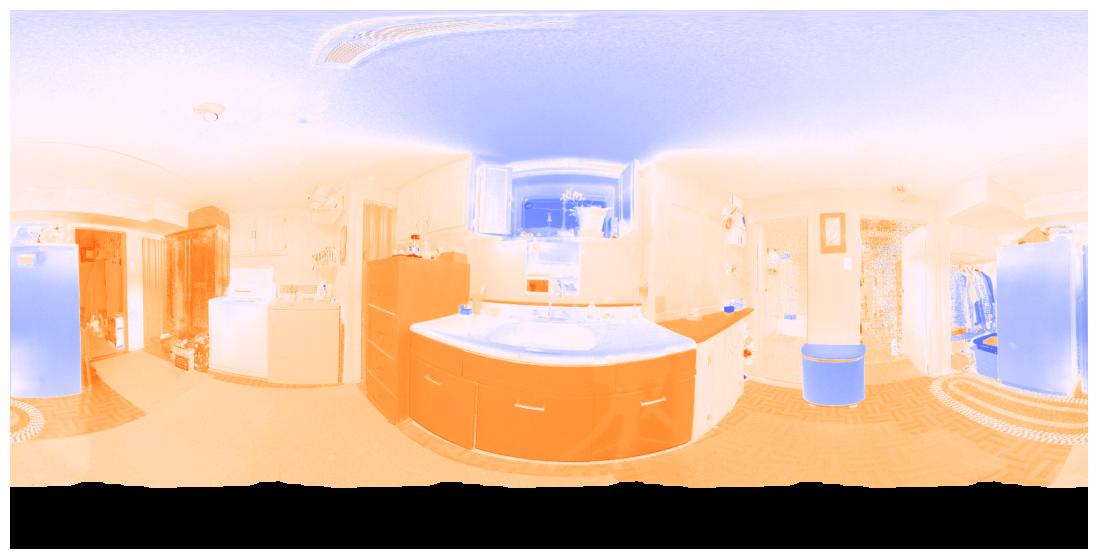}&
    \includegraphics[width=\tmplength]{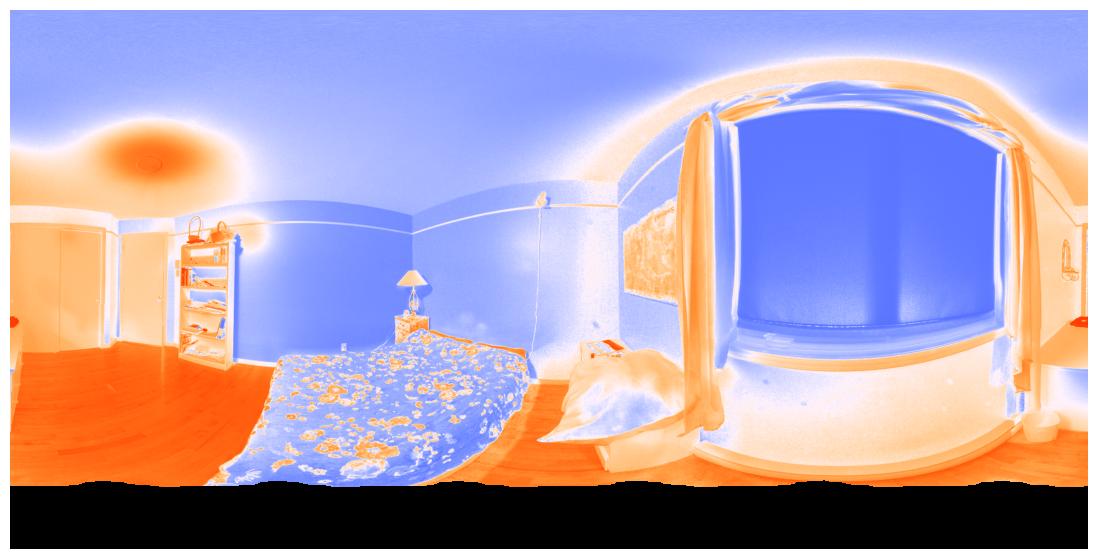}&
    \includegraphics[width=\tmplength]{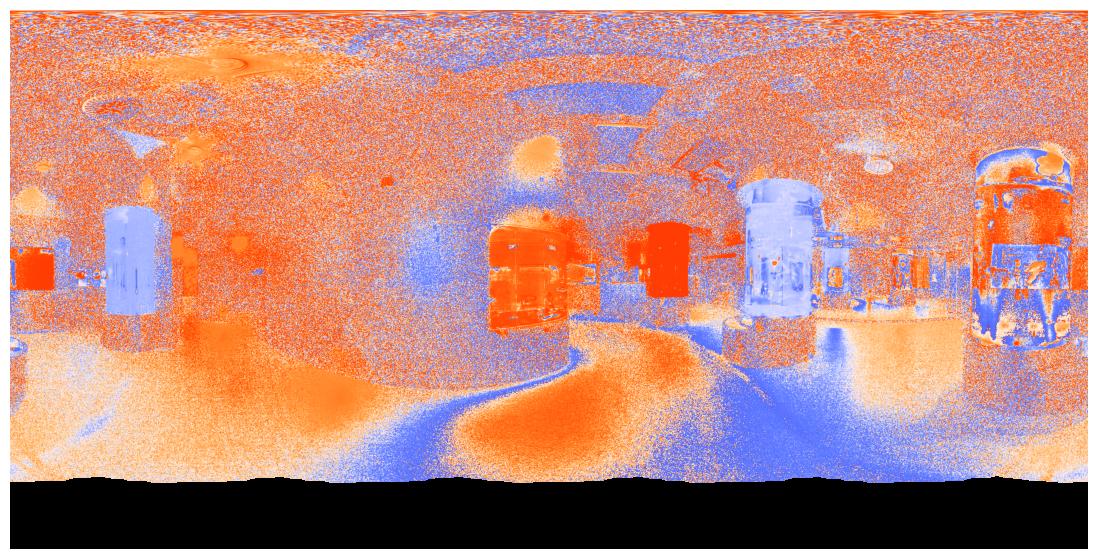}\\
    \end{tabular}
    \caption{Example scenes with CCT close to the quantile values to complement fig.~4 from the main paper.  Colored images below show the CCT map of the scenes.  The percentiles and corresponding measured scene CCT are indicated above the images.  Images are reexposed and tonemapped ($\gamma = \valeur{\num{2.2}}$) for display. CCT map: \includegraphics[width=3cm,trim=0.2cm 0.65cm 0.2cm 0.2cm,clip]{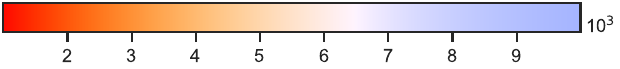}} 
    \label{fig:distribution_temp_examples_supp}
\end{figure*}

To add to fig. 4 of the paper, \cref{fig:distribution_temp_examples_supp} shows more examples of scenes, this time sorted by their CCT value.  The map below corresponds their associated CCT.

\begin{figure}
    \includegraphics[width=\linewidth]{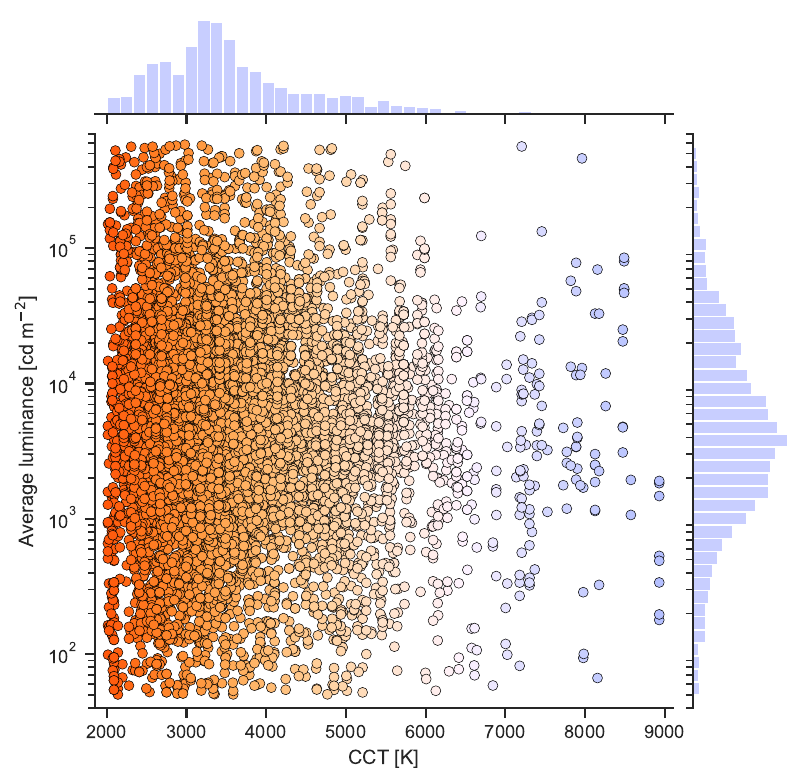}
    \caption{Correlation between the CCT and the average luminance for each light source in our calibrated dataset.  The distributions of the CCT (top) and average luminance (right) of the light sources are also displayed.  Only the light sources with a CCT in $[\valeur{\SI{2000}{\K}}, \valeur{\SI{9000}{\K}}]$ and an average luminance in $[\valeur{\SI{50}{\candela\per\m\squared}}, \valeur{\SI{600000}{\candela\per\m\squared}}]$ are included to better see the trends (\valeur{\num{10289}} out of \valeur{\num{11060}} sources).}
    \label{fig:jointplot_temperature_luminance_moyenne_sources_pano}
\end{figure}

\Cref{fig:jointplot_temperature_luminance_moyenne_sources_pano} shows the correlation between the CCT and the average luminance for the individual light sources detected discussed in sec. 4.2 of the paper (\valeur{\num{10289}} out of \valeur{\num{11060}} sources are included in the figure).  The distribution of the values are also shown on the edge of the figure.  It is possible to see that cooler sources are more frequent than the warmer (which tend to correspond to windows).  However, the distribution in average luminance is quite symmetrical.

The average value of the average luminance for all the light sources included in \cref{fig:jointplot_temperature_luminance_moyenne_sources_pano} (values in $[\valeur{\SI{50}{\candela\per\m\squared}}, \valeur{\SI{600000}{\candela\per\m\squared}}]$) is \valeur{\SI{18029}{\candela\per\m\squared}}, with a median of \valeur{\SI{3991}{\candela\per\m\squared}}.  The average value of the average luminance for all the light sources included in the dataset is \valeur{\SI{27874}{\candela\per\m\squared}}, with a median of \valeur{\SI{3854}{\candela\per\m\squared}}.  
The average value of the CCT for all the light sources included in \cref{fig:jointplot_temperature_luminance_moyenne_sources_pano} (values in in $[\valeur{\SI{2000}{\K}}, \valeur{\SI{9000}{\K}}]$) is \valeur{\SI{3633}{\K}}, with a median of \valeur{\SI{3404}{\K}}.  The average value of the CCT for all the light sources included in the dataset is \valeur{\SI{3648}{\K}}, with a median of \valeur{\SI{3380}{\K}}.

\section{Learning tasks}
\label{sec:learning}

\subsection{Input data}
\label{subsec:data}

\begin{figure}
   \centering
   \setlength{\tmplength}{\linewidth}
    Linear \quad \quad Gamma \quad \: Noise \quad \: Quantization \quad WB
    \includegraphics[width=\tmplength]{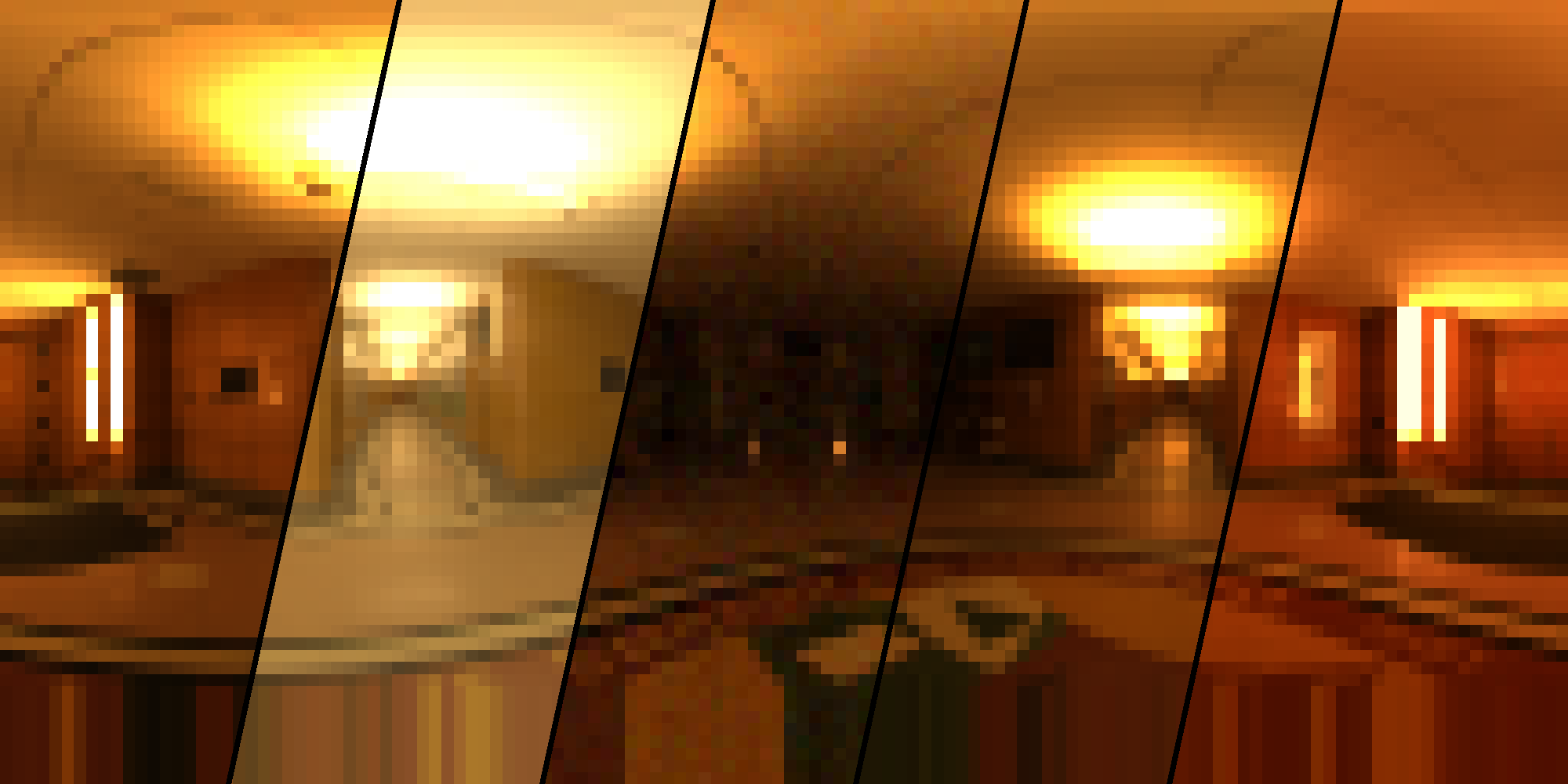} \\
    \includegraphics[width=\tmplength]{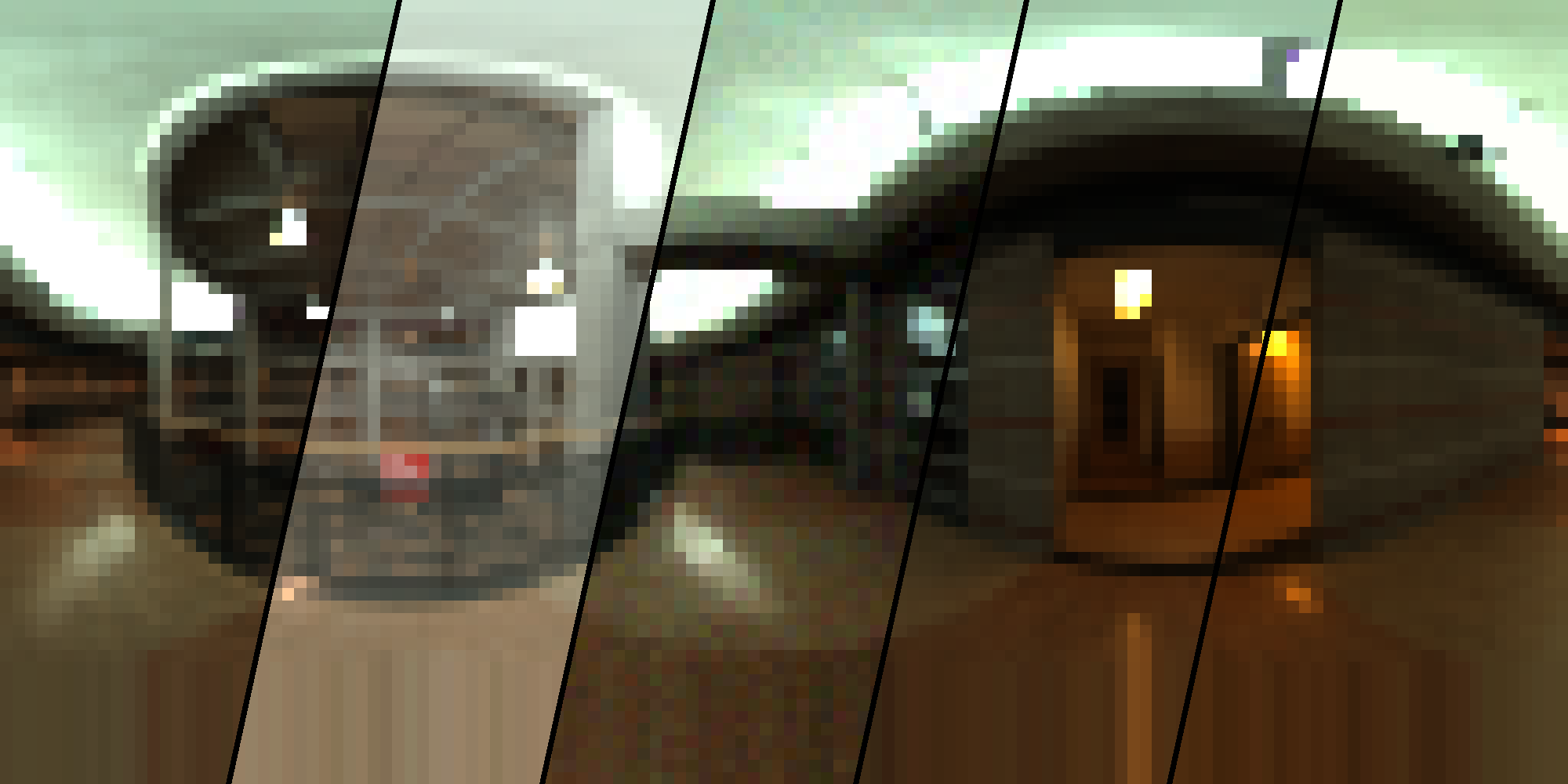} \\
    \includegraphics[width=\tmplength]{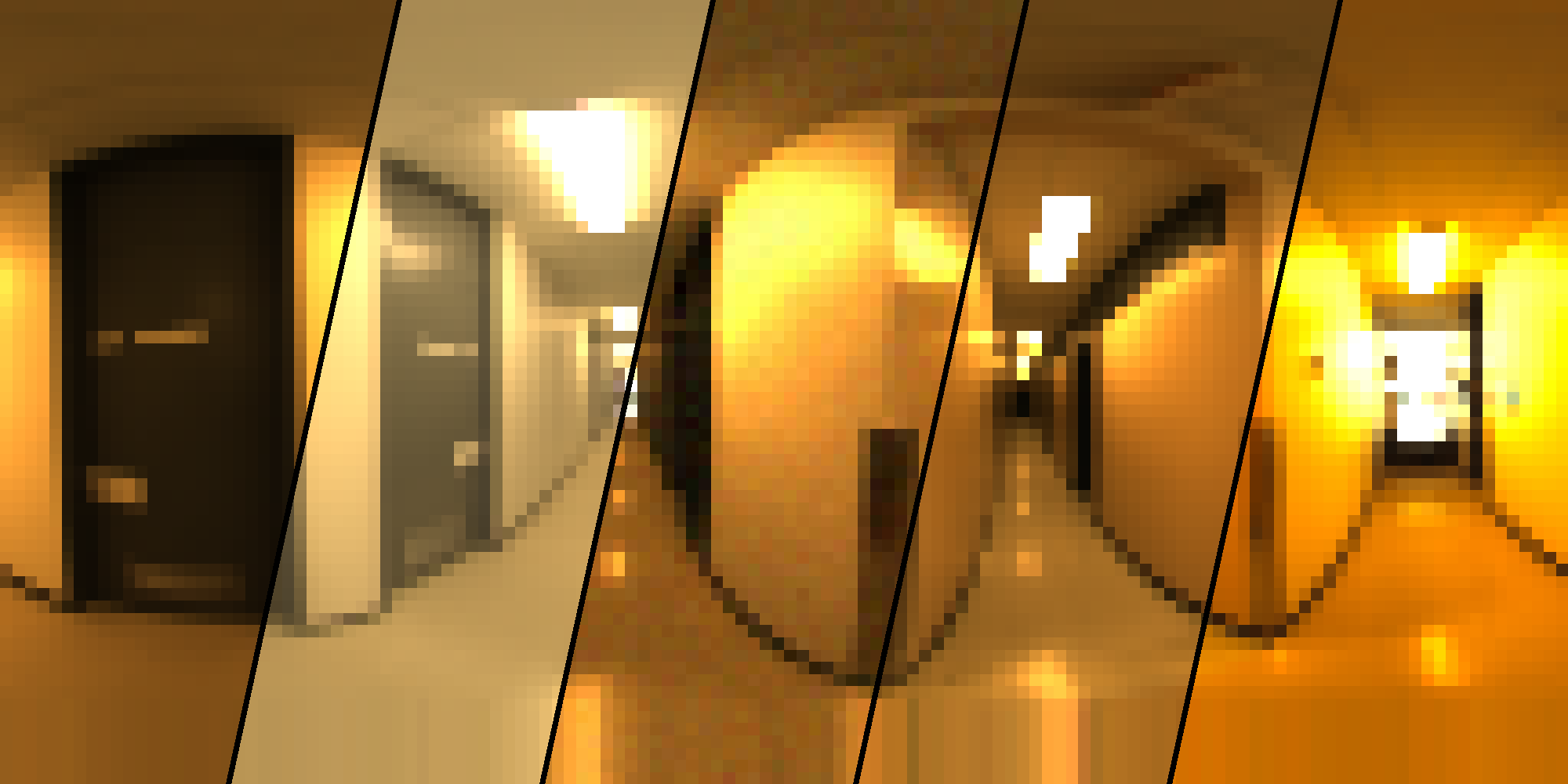} \\
    \includegraphics[width=\tmplength]{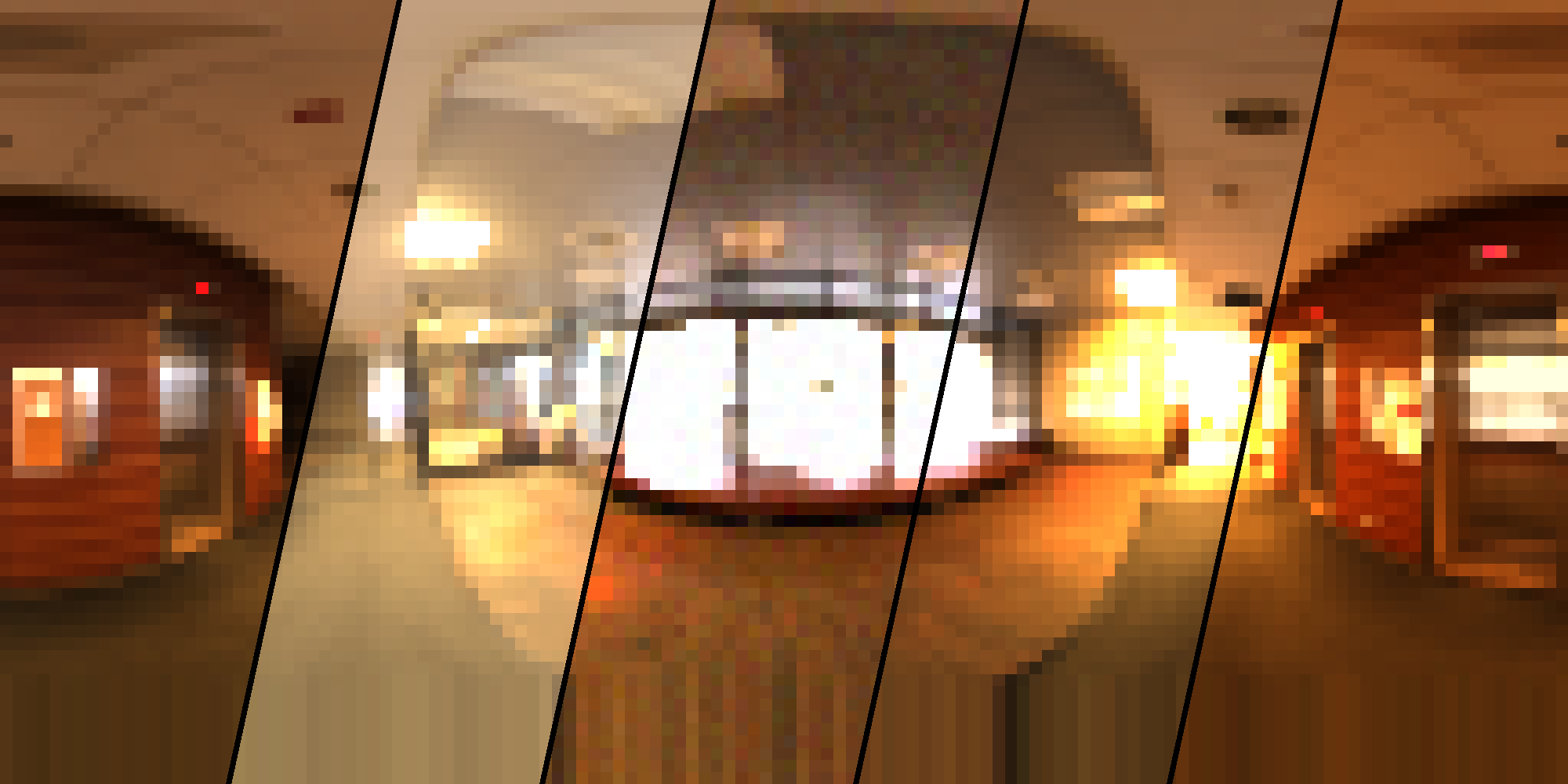} \\
    \caption{Examples of panorama inputs given to the networks. For visualization, each image is split to show 6 transformations from left to right: Linear, Gamma, Noise, Quantization, Hue. Each input is reexposed and clipped in the range [0, 1]. ``Linear'' applies no further modification. ``Gamma'' applies a gamma of $\gamma=2.2$. ``Noise'' applies additive Gaussian noise of variance uniformly drawn in the range [0, 0.03]. ``Quantization'' constraints the input in 255 values. ``WB'' applies a random WB augmenter~\cite{Afifi2019}.}
    \label{fig:learning_LDR}
\end{figure}

We apply different transformations to the input given to the networks in sec. 5 and sec. 6 of the paper. For per-pixel luminance prediction, random noise is added to the input and gamma and quantization are applied to the image. For per-pixel color prediction, a random WB augmenter~\cite{Afifi2019} is applied the input. Those transformations are visualized in \cref{fig:learning_LDR}.

\subsection{Experiments}
\label{subsec:experiments}
\begin{figure*}
   \centering
   \tiny
   \setlength{\tabcolsep}{0.5pt} 
   \setlength{\tmplength}{0.155\linewidth}
    \begin{tabular}{cccccccc}
    &1st: \valeur{3.0} (\valeur{0.7}, \valeur{\SI{97.1}{\%}}) & 
    20th: \valeur{19.1} (\valeur{18.4}, \valeur{\SI{98.2}{\%}}) & 
    40th: \valeur{42.5} (\valeur{1.9}, \valeur{\SI{98.0}{\%}}) & 
    60th: \valeur{81.1} (\valeur{78.2}, \valeur{\SI{97.1}{\%}}) & 
    80th: \valeur{193.5} (\valeur{149.2}, \valeur{\SI{95.8}{\%}}) & 
    99th: \valeur{1036.1} (\valeur{909.6}, \valeur{\SI{95.4}{\%}}) &  \multirow{5}{*}{\includegraphics[trim={{.013\linewidth} 0 0 {.010\linewidth}},clip, height=0.255\linewidth]{figures/luminanceColorBar_v.pdf}} \\
    \rotatebox{90}{\hspace{.45cm} Input} & \includegraphics[width=\tmplength]{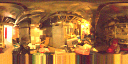}&
    \includegraphics[width=\tmplength]{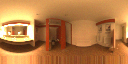}&
    \includegraphics[width=\tmplength]{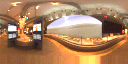}&
    \includegraphics[width=\tmplength]{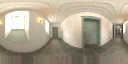}&
    \includegraphics[width=\tmplength]{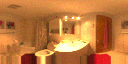}&
    \includegraphics[width=\tmplength]{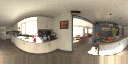}  \\
    \rotatebox{90}{\hspace{.5cm} GT} & \includegraphics[width=\tmplength]{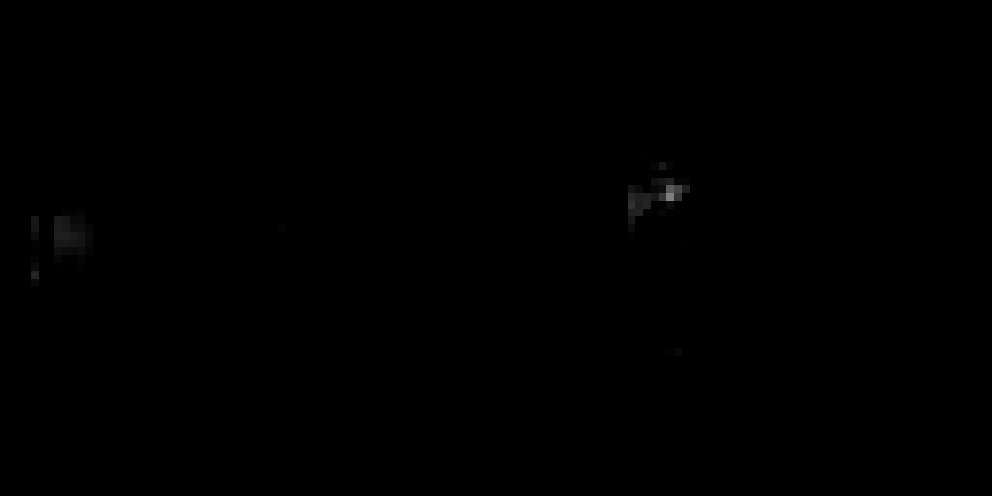}&
    \includegraphics[width=\tmplength]{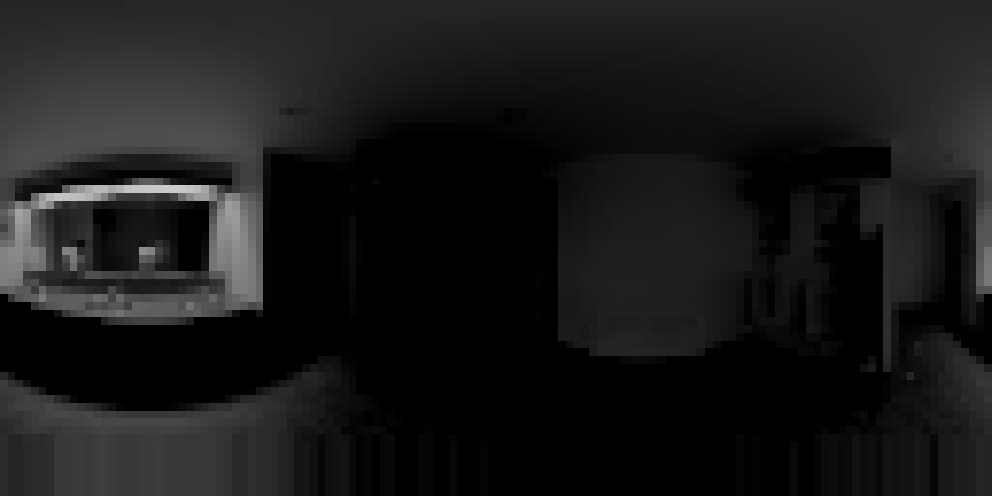}&
    \includegraphics[width=\tmplength]{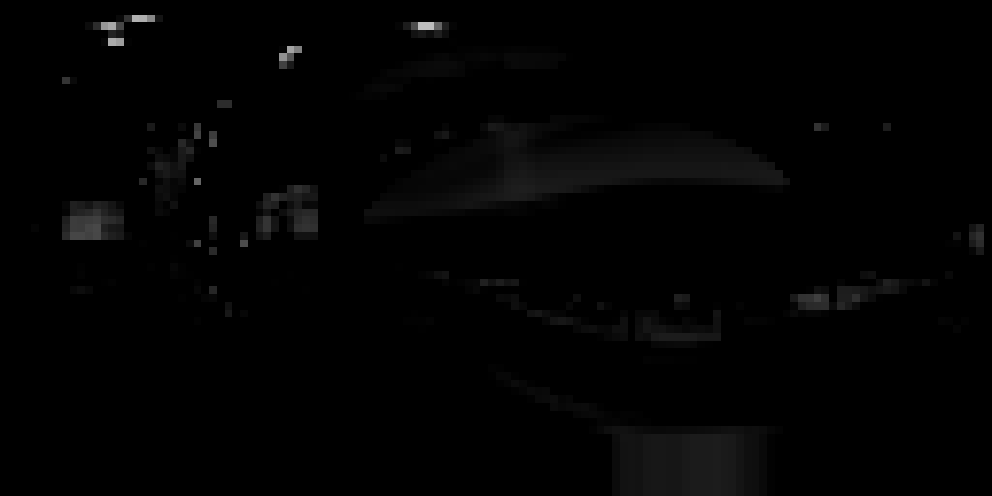}&
    \includegraphics[width=\tmplength]{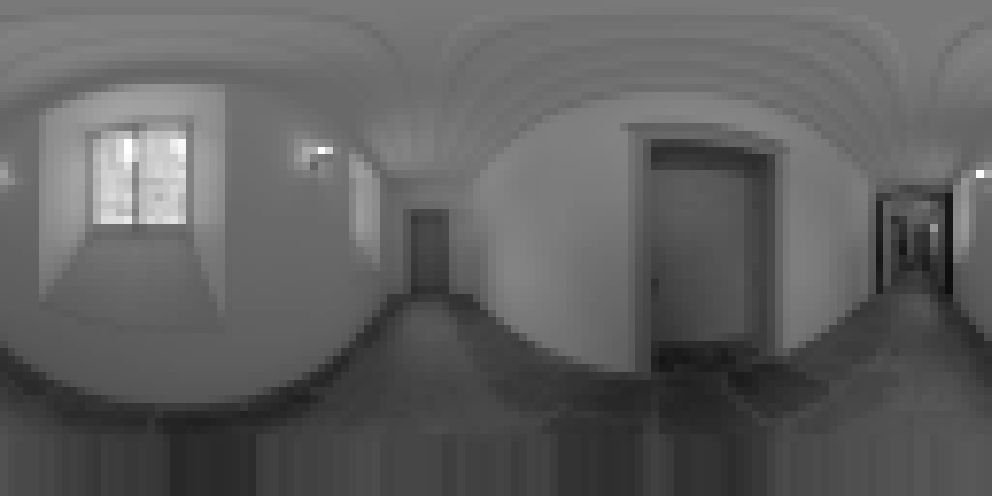}&
    \includegraphics[width=\tmplength]{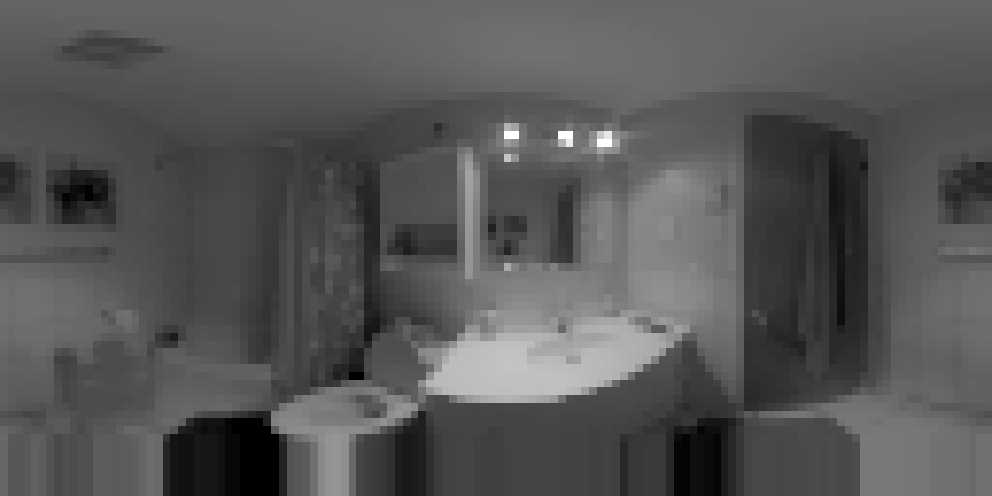}&
    \includegraphics[width=\tmplength]{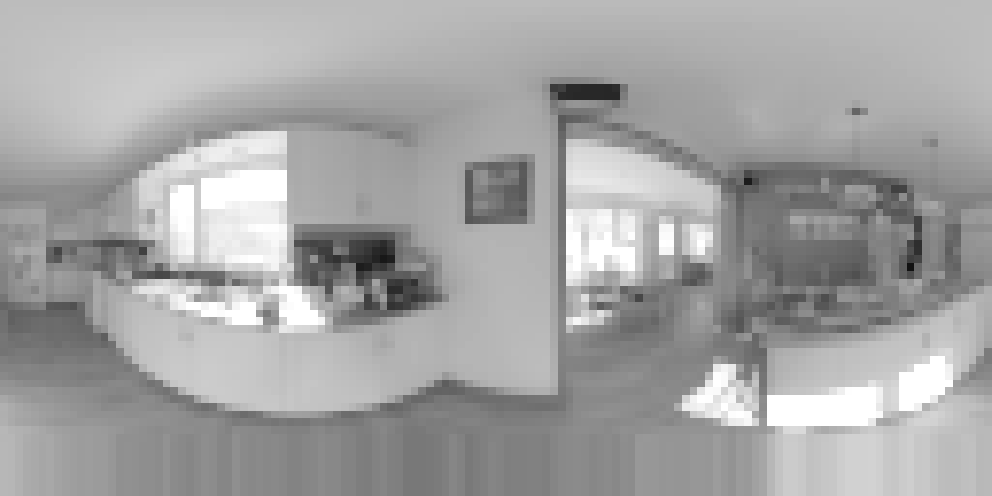}\\
    \rotatebox{90}{\hspace{.3cm} Prediction} & \includegraphics[width=\tmplength]{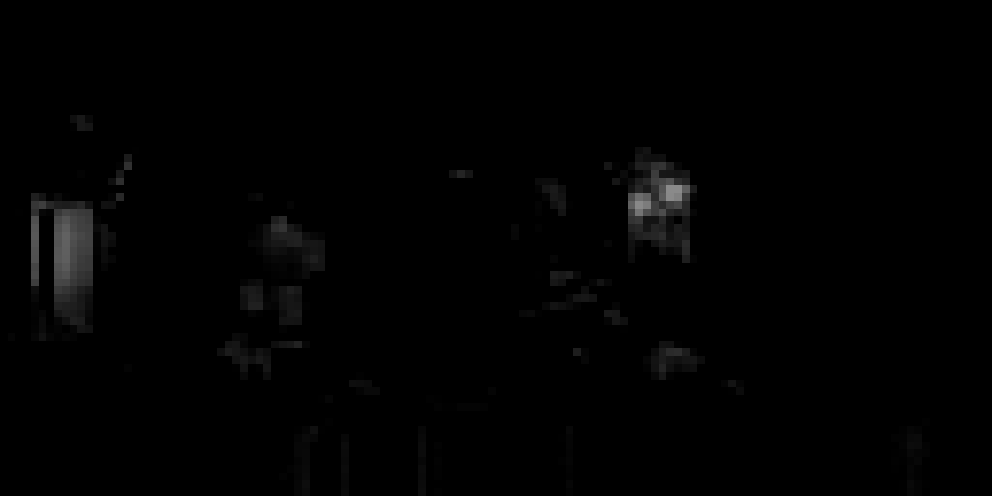}&
    \includegraphics[width=\tmplength]{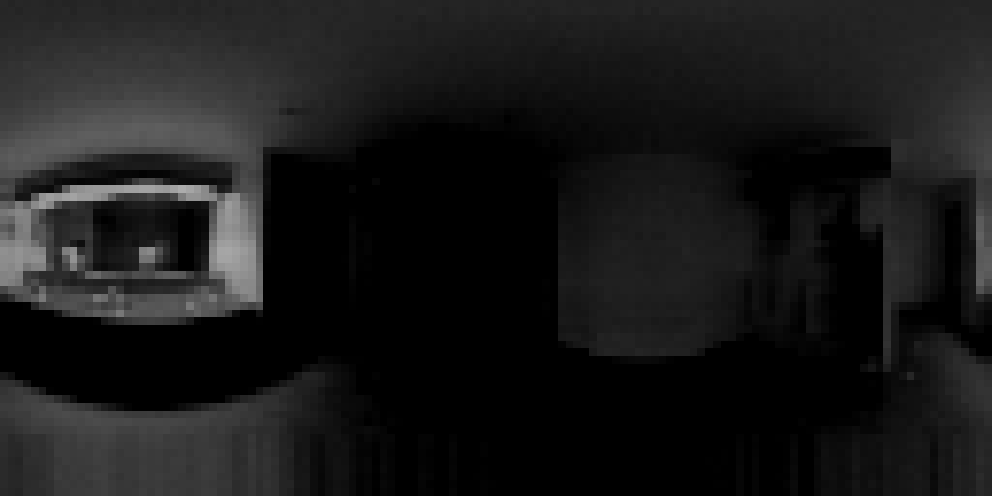}&
    \includegraphics[width=\tmplength]{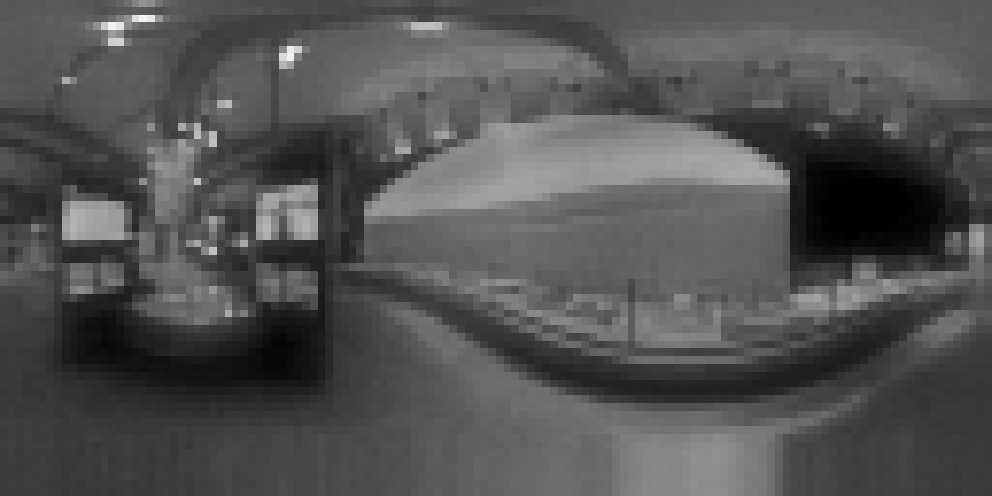}&
    \includegraphics[width=\tmplength]{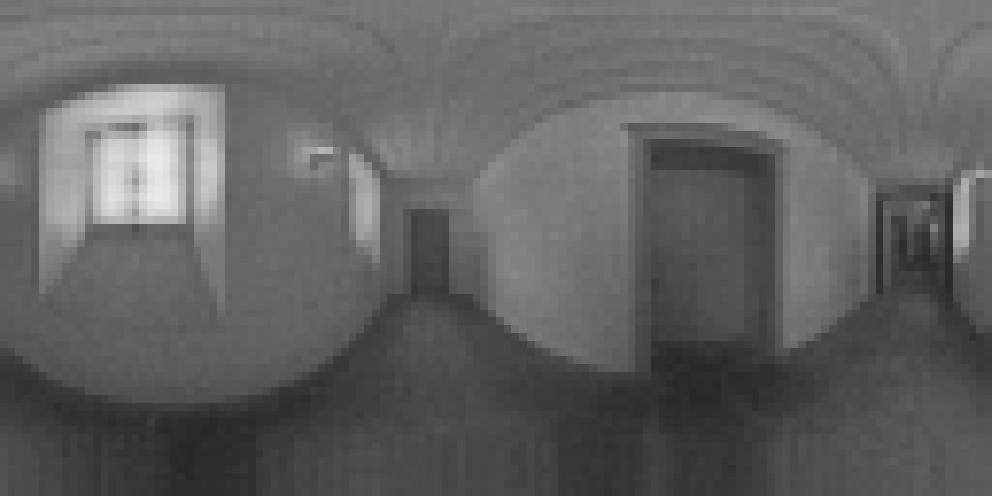}&
    \includegraphics[width=\tmplength]{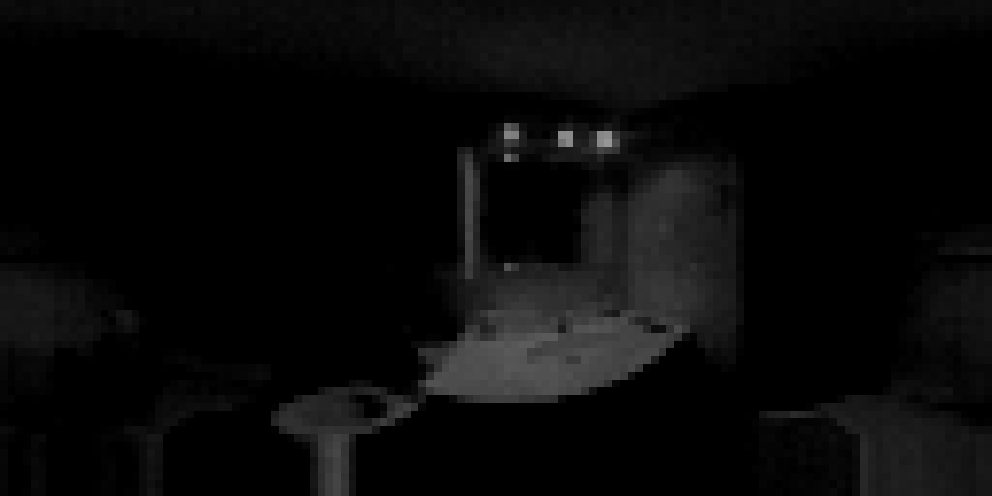}&
    \includegraphics[width=\tmplength]{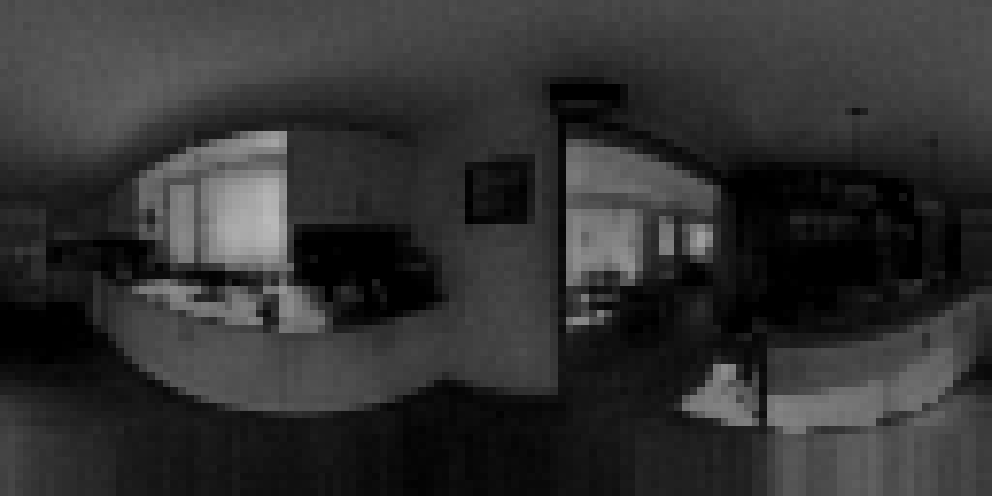}\\
    \end{tabular}
    \caption{Examples of per-pixel luminance prediction. The first row indicates the RMSE percentile: the RMSE (relative error). The ``input'' is the calibrated HDR reexposed and clipped. Other rows show the ground truth and predicted photopic luminance maps. The colormap for the luminance is shown at the right.}
    \label{fig:learning_luminance}
\end{figure*}

The following results complement the experiments of sec. 5.3 of the paper.
\paragraph{Per-pixel luminance}
\Cref{fig:learning_luminance} shows qualitative predictions, comparing the ground truth luminance to the predicted luminance. Observe that most of the errors are due to incorrect scale prediction.

\begin{figure}
    \includegraphics[width=0.8\linewidth]{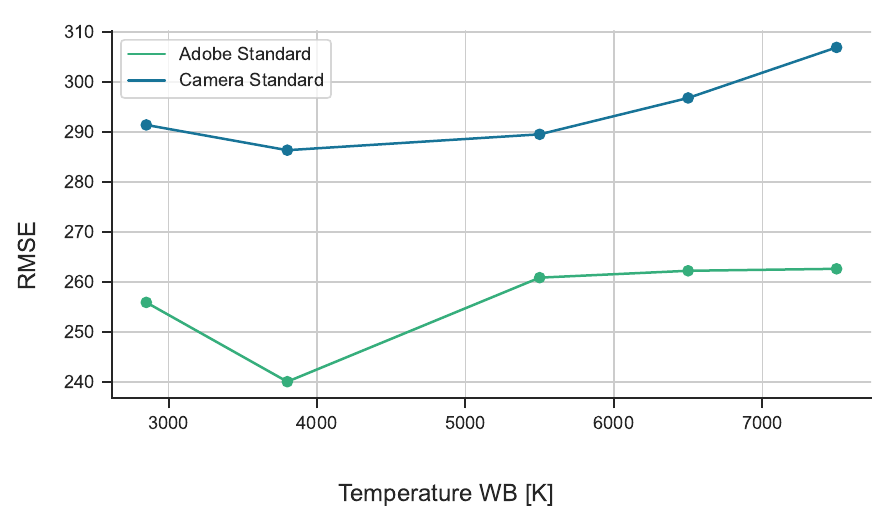}
    \caption{Test scores of color prediction with inputs at different white balance corrections with two different photofinishing profiles. The network is trained on all input corrections.}
    \label{fig:RMSE_temp}
\end{figure}

\paragraph{Per-pixel color}
\Cref{fig:RMSE_temp} shows the effect of different white balance augmentations. The network is trained and tested on all augmentation settings independently.

\paragraph{Planar illuminance}

\begin{figure*}
   \centering
   \footnotesize
   \setlength{\tabcolsep}{0.5pt} 
   \setlength{\tmplength}{0.16\linewidth}
    \begin{tabular}{cccccc}
    1st: \valeur{0.9} & 
    20th: \valeur{13.3} & 
    40th: \valeur{25.9} & 
    60th: \valeur{62.3} & 
    80th: \valeur{130.5} & 
    99th: \valeur{1539.2} \\
    \includegraphics[width=\tmplength]{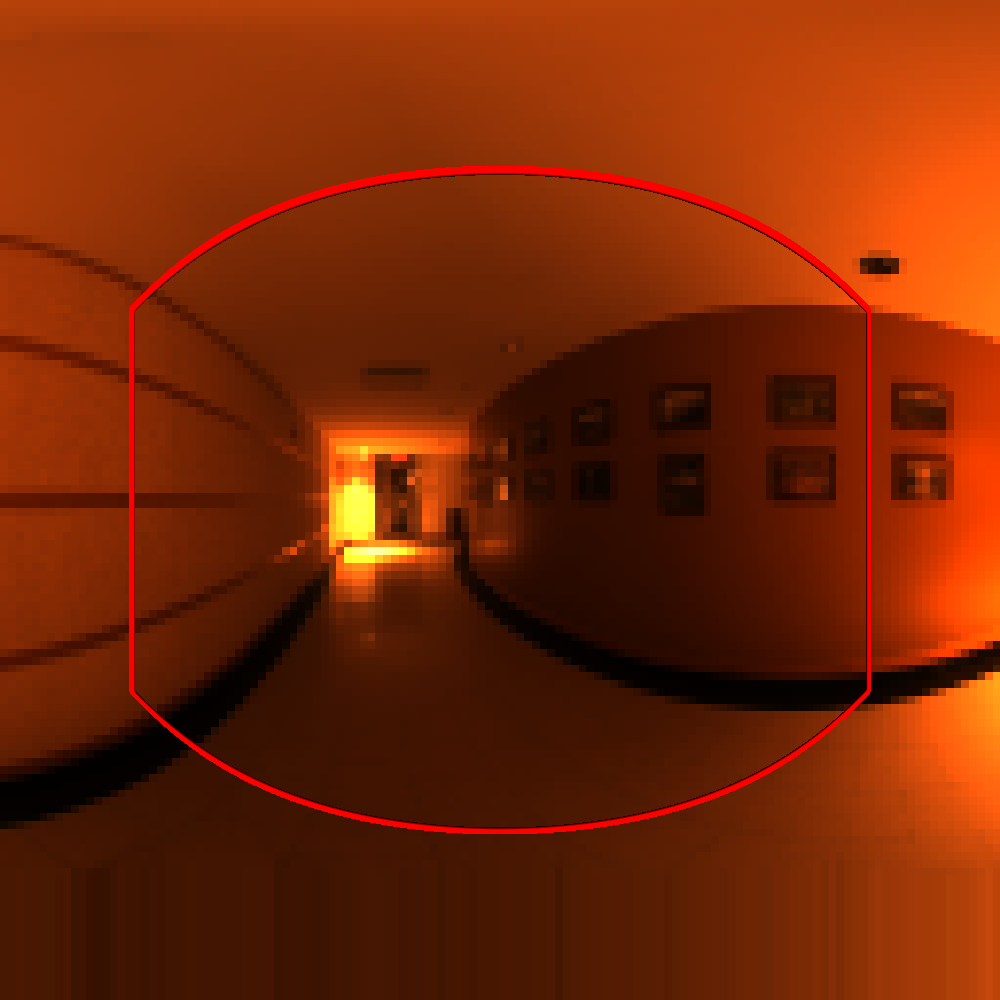}&
    \includegraphics[width=\tmplength]{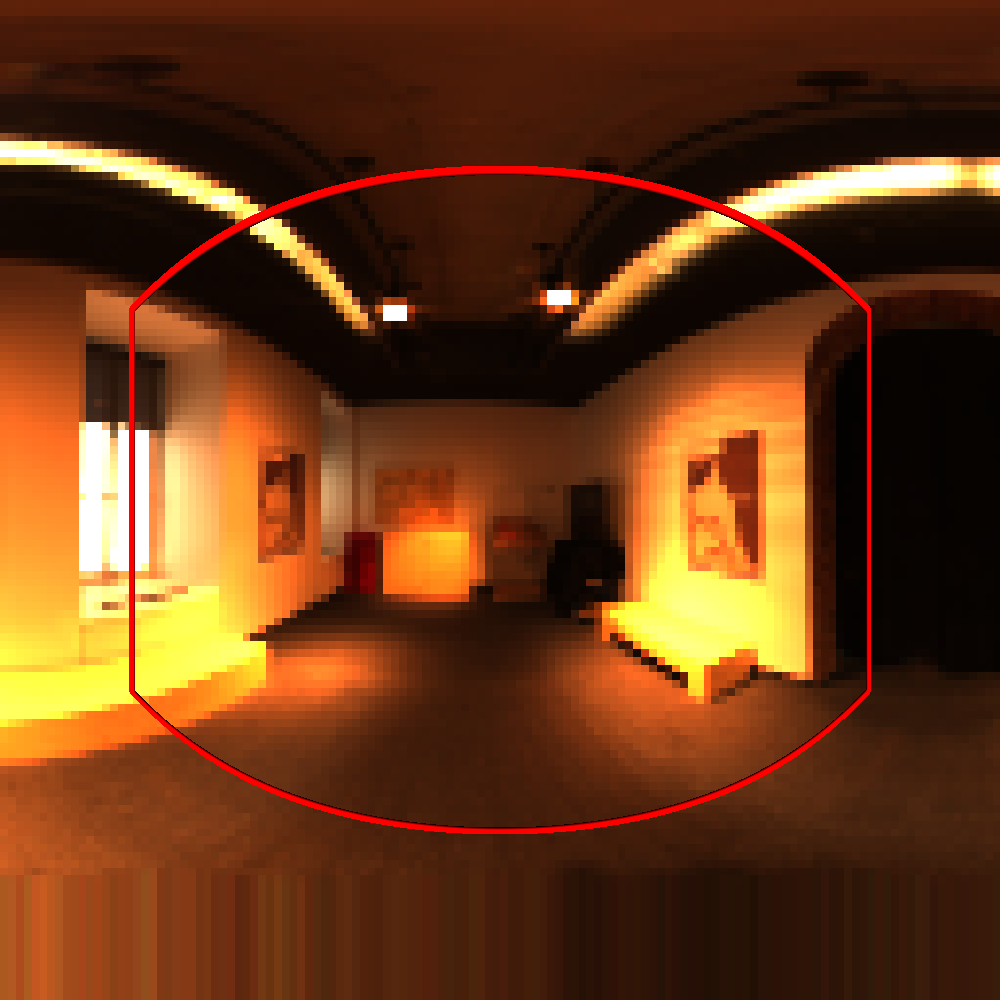}&
    \includegraphics[width=\tmplength]{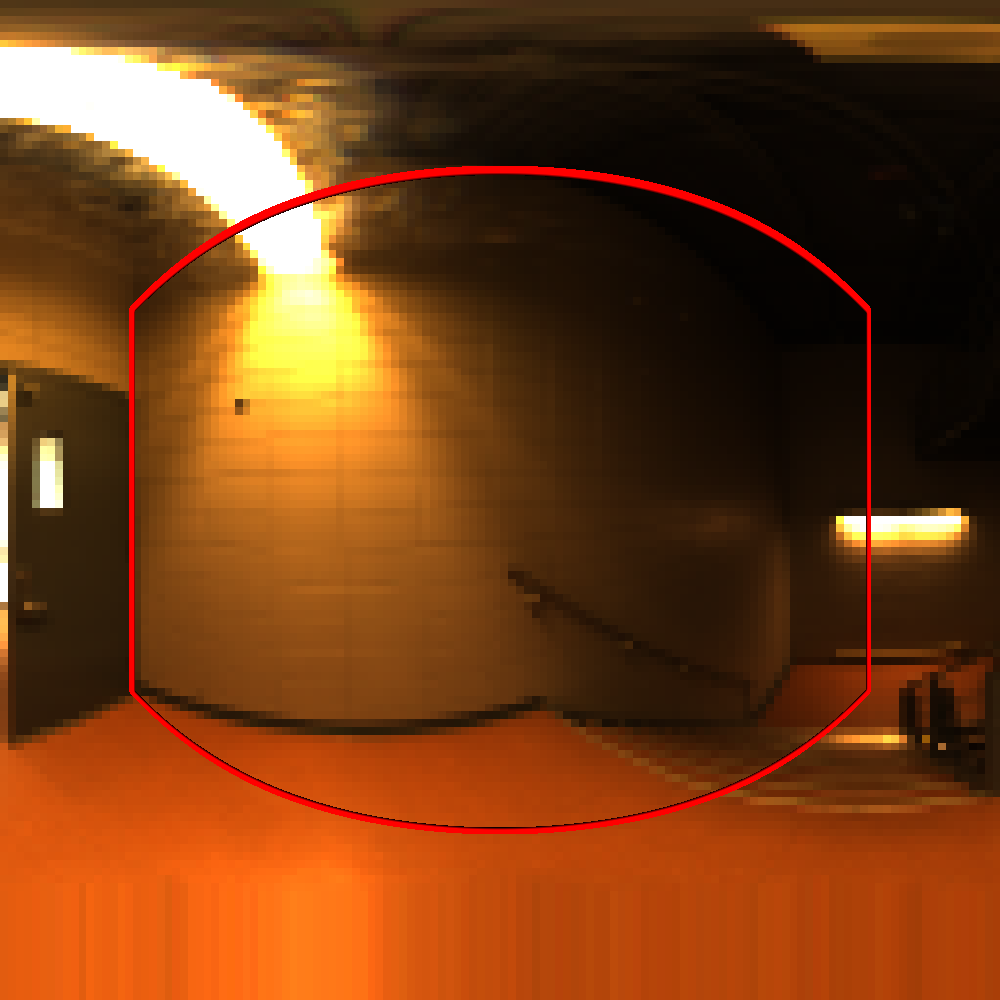}&
    \includegraphics[width=\tmplength]{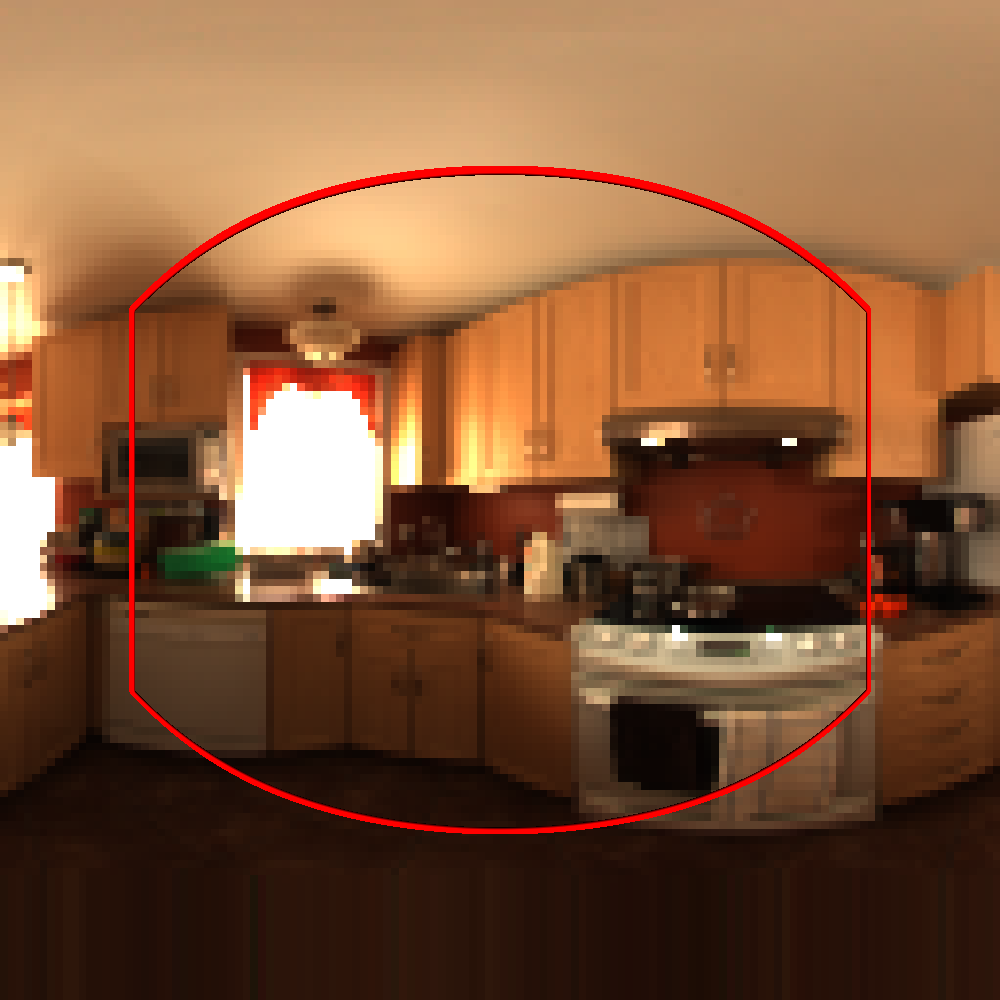}&
    \includegraphics[width=\tmplength]{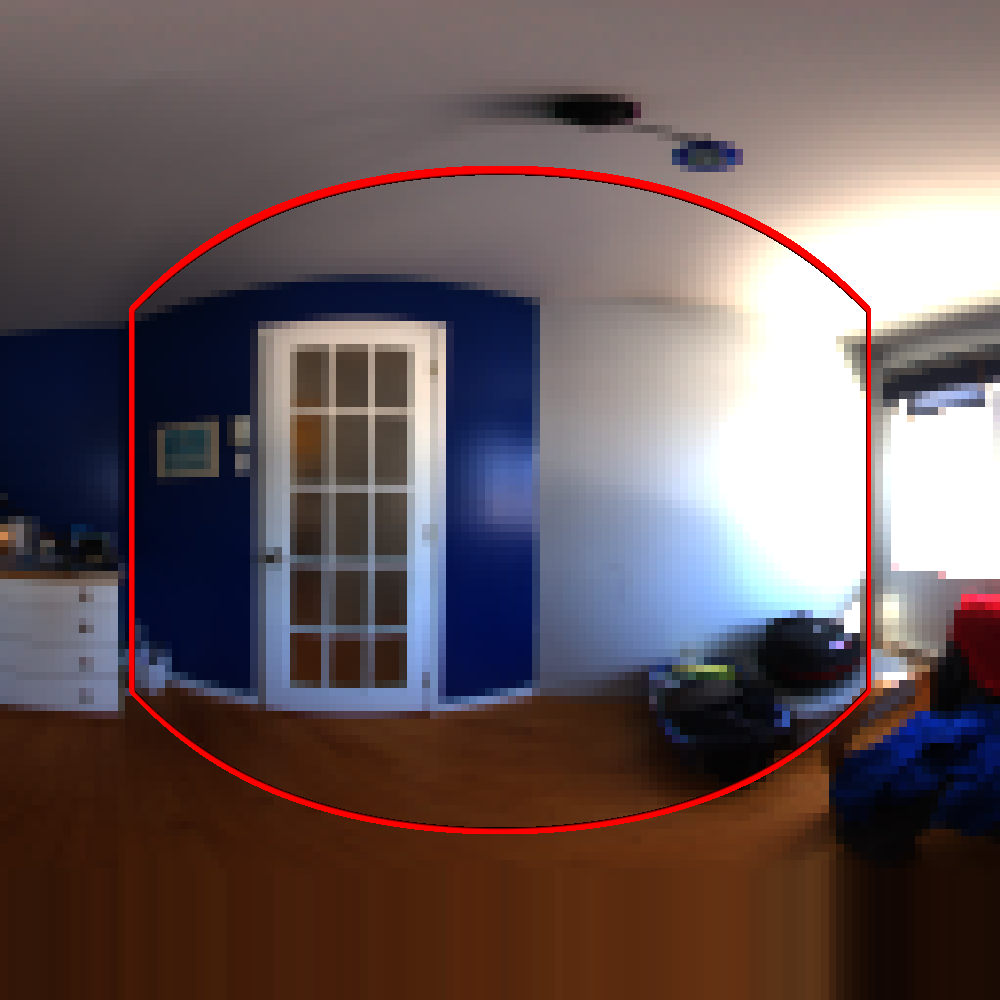}&
    \includegraphics[width=\tmplength]{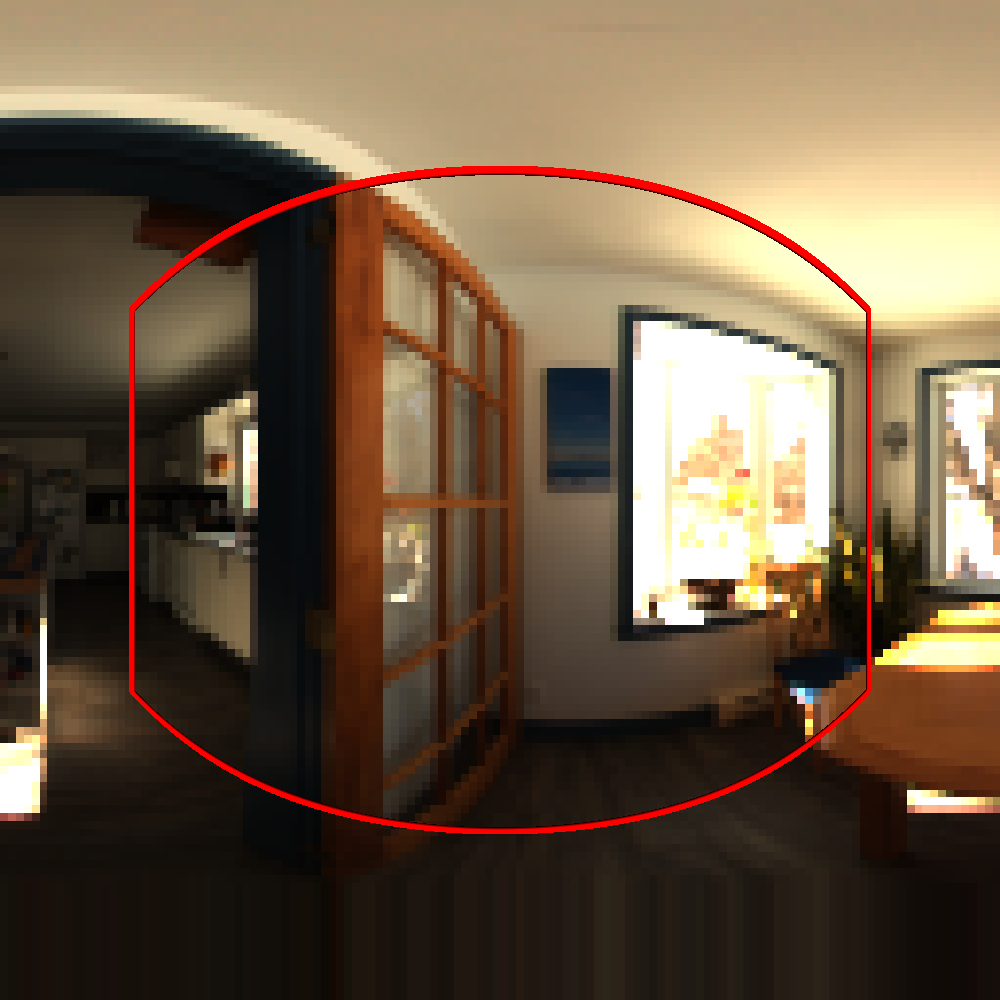}  \\
    \includegraphics[width=\tmplength]{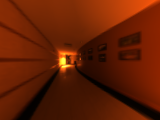}&
    \includegraphics[width=\tmplength]{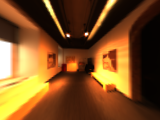}&
    \includegraphics[width=\tmplength]{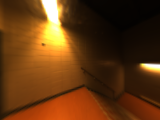}&
    \includegraphics[width=\tmplength]{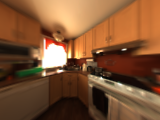}&
    \includegraphics[width=\tmplength]{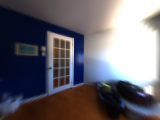}&
    \includegraphics[width=\tmplength]{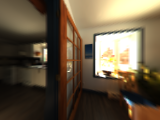}\\
    \valeur{\SI{22.4}{\lux}} / \valeur{\SI{21.5}{\lux}}&
    \valeur{\SI{22.6}{\lux}} /\valeur{\SI{9.3}{\lux}} &
    \valeur{\SI{126.6}{\lux}} / \valeur{\SI{100.7}{\lux}}&
    \valeur{\SI{117.1}{\lux}} / \valeur{\SI{179.3}{\lux}}&
    \valeur{\SI{418.7}{\lux}} / \valeur{\SI{288.3}{\lux}}&
    \valeur{\SI{1800.2}{\lux}} /\valeur{\SI{261.1}{\lux}} \\
    \end{tabular}
    \caption{Examples planar illuminance prediction with FOV of $120^\circ$. The first row indicates the RMSE percentile: the RMSE. Below are the calibrated HDR hemispheres reexposed and clipped, with the field of view of the image below outlined in red. Below is the projected HDR hemisphere reexposed and clipped given to the network. The last row shows the ground truth and predicted scalars planar illuminance respectively.}
    \label{fig:learning_illuminance}
\end{figure*}

\Cref{fig:learning_illuminance} shows illuminance predictions, along with their given inputs and the full hemispheres used for ground truth illuminance. Light sources being slightly outside the FOV and the unknown camera exposure make illuminance prediction from a single image a very difficult task. We hope our dataset will provide a useful resource to the community to tackle these challenging new problems.

\begin{figure}[t]
\centering
    \includegraphics[width=\linewidth]{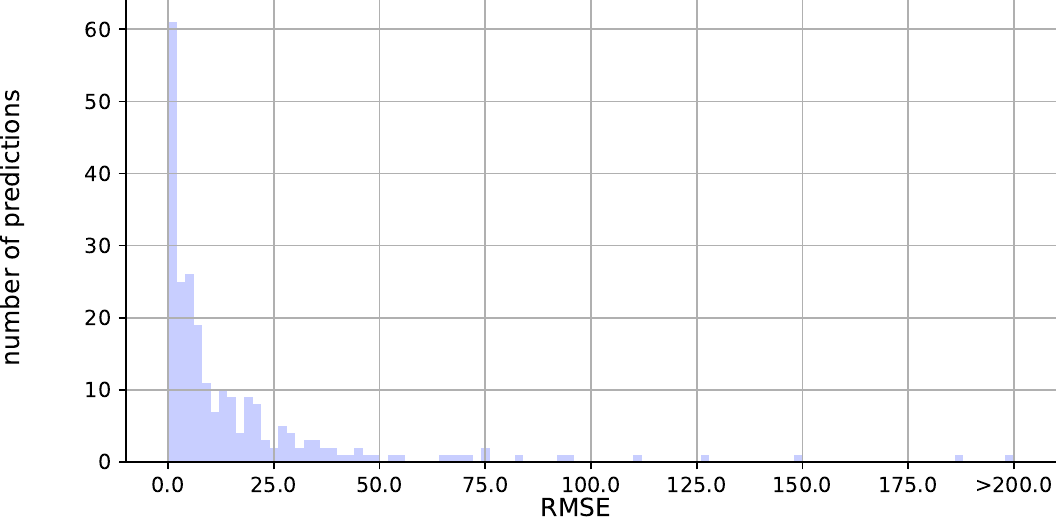} \\
    (a) HDR \\
    \includegraphics[width=\linewidth]{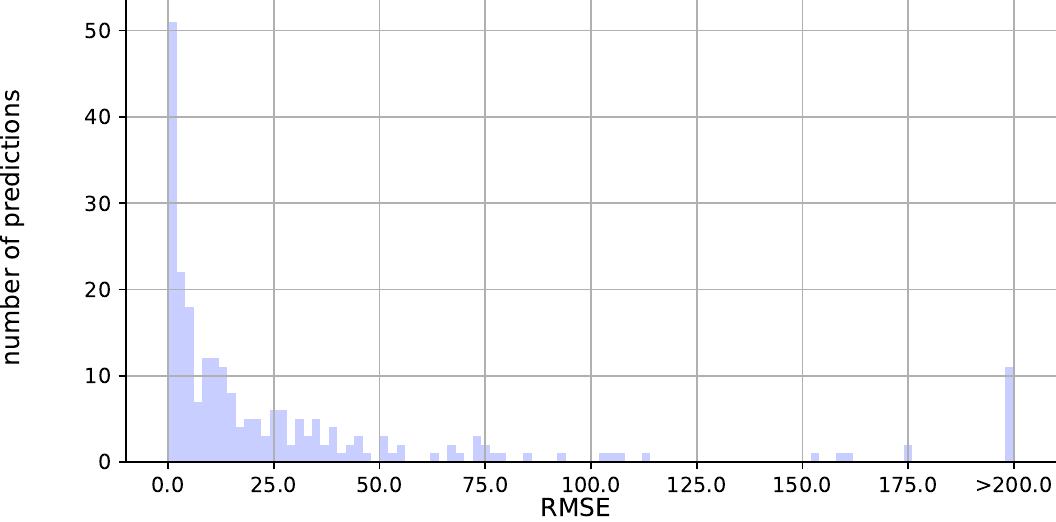} \\
    (b) LDR+scale \\
    \includegraphics[width=\linewidth]{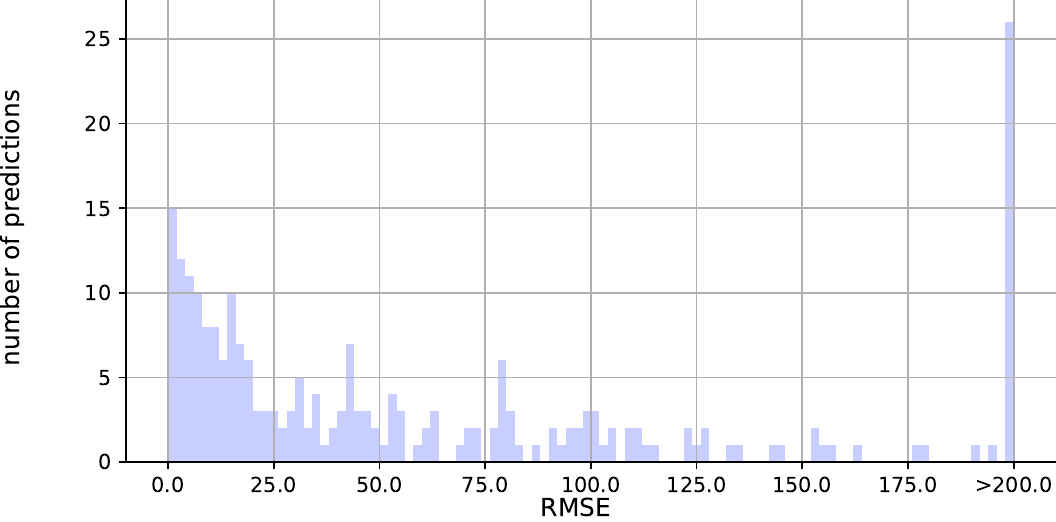} \\
    (c) LDR
    \caption{The distribution of RMSE scores with different levels of photometric information in the input. The \ang{180} hemisphere image is given as (a) HDR, (b) LDR with photometric scale, and (c) LDR without photometric scale.}
    \label{fig:learning_illumInfo}
\end{figure}

Additionally, the effect of modifying the photometric information of the input is visualized in \cref{fig:learning_illumInfo}

{\small
}




\end{document}